\documentclass[10pt,journal,compsoc]{IEEEtran}
\ifCLASSOPTIONcompsoc
  \usepackage[nocompress]{cite}
\else
  \usepackage{cite}
\fi
\ifCLASSINFOpdf
   \usepackage[pdftex]{graphicx}
   \graphicspath{{images_new_arxiv/}}
   \DeclareGraphicsExtensions{.pdf,.jpeg,.jpg,.png}
\else
\fi
\usepackage{amsmath}
\usepackage{soul}

\usepackage{array}
\usepackage[hyphens]{url}

\usepackage{color}
\usepackage{hyperref}
\usepackage{amsmath}
\usepackage{amsfonts} 
\usepackage{amssymb}
\usepackage[utf8]{inputenc} 
\usepackage{import}  
\usepackage[capitalise]{cleveref} 
\usepackage{bm} 
\usepackage{mathtools} 
\usepackage{graphicx}
\usepackage{subcaption}
\usepackage{stackengine}
\usepackage{array} 
\usepackage{multirow}
\usepackage{tabularx}

\makeatletter
\newcommand{\printfnsymbol}[1]{%
  \textsuperscript{\@fnsymbol{#1}}%
}
\makeatother

\DeclareMathOperator*{\argmin}{{argmin}}

\newcommand{\zdz}{{z,\nabla z}}
\newcommand{\dA}{{\mathrm{d}\mathcal{A}}}
\newcommand{\zz}{{z^{0}}}
\newcommand{\Norm}[1]{\left\Vert #1 \right\Vert}
\newcommand{\norm}[1]{\left\vert #1 \right\vert}
\newcommand{\diag}{\text{diag}}
\newcommand{\mean}{\text{mean}}
\newcommand{\R}{\mathbb{R}}
\newcommand{\OHR}{\Omega_{HR}}
\newcommand{\OLR}{\Omega_{LR}}

\newcommand{\comment}[1]{\ignorespaces}

\newcommand{\p}{\mathbf{p}}

\newcommand{\I}{\mathbf{I}}
\newcommand{\Ii}{{\{\I_i\}}}
\newcommand{\li}{{\{\l_i\}}}
\newcommand{\zzi}{{\{z^{0}_i\}}}
\newcommand{\mrho}{{\bm{\rho}}}
\newcommand{\n}{\mathbf{n}}
\newcommand{\m}{\mathbf{m}}
\newcommand{\mtheta}{{\bm{\theta}}}
\newcommand{\metai}{{\bm{\eta}_\I}}
\renewcommand{\u}{\mathbf{u}}
\renewcommand{\l}{\mathbf{l}}

\renewcommand{\P}{\mathcal{P}}

\renewcommand{\[}{\right]}

\begin{document}
%
\title{Photometric Depth Super-Resolution}
%
%
%
%

\author{Bjoern~Haefner\printfnsymbol{1},
        Songyou~Peng\printfnsymbol{1},
        Alok~Verma\printfnsymbol{1},
        Yvain~Qu\'eau,
        and Daniel~Cremers
\thanks{\printfnsymbol{1} Equal contribution} 
\IEEEcompsocitemizethanks{\IEEEcompsocthanksitem B. Haefner, A. Verma, and D. Cremers are with the Department
of Computer Science, Technical University of Munich, 80333, Germany.\protect\\
E-mail: $\{$bjoern.haefner,alok.verma,cremers$\}$@tum.de
\IEEEcompsocthanksitem {S. Peng is with Advanced Digital Sciences Center, University of Illinois at Urbana-Champaign, Singapore, 138602.\protect\\
E-mail: songyou.peng@adsc-create.edu.sg}
\IEEEcompsocthanksitem {Y. Qu\'eau is with the GREYC laboratory, UMR CNRS 6072, Caen, France.\protect\\
E-mail: yvain.queau@ensicaen.fr}}
}

%
%

\markboth{}%
{}
%



\IEEEtitleabstractindextext{%
\begin{abstract}
This study explores the use of photometric techniques (shape-from-shading and uncalibrated photometric stereo) for upsampling the low-resolution depth map from an RGB-D sensor to the higher resolution of the companion RGB image. A single-shot variational approach is first put forward, which is effective as long as the target's reflectance is piecewise-constant. It is then shown that this dependency upon a specific reflectance model can be relaxed by focusing on a specific class of objects (e.g., faces), and delegate reflectance estimation to a deep neural network. A multi-shot strategy based on randomly varying lighting conditions is eventually discussed. It requires no training or prior on the reflectance, yet this comes at the price of a dedicated acquisition setup. Both quantitative and qualitative evaluations illustrate the effectiveness of the proposed methods on synthetic and real-world scenarios.
\end{abstract}

}

\maketitle

\IEEEdisplaynontitleabstractindextext

%
\IEEEpeerreviewmaketitle

\IEEEraisesectionheading{\section{Introduction}\label{sec:introduction}}

%
%
%
%

\begin{figure*}[]
  \footnotesize
  \centering
  \newcommand{\mywidth}{0.12\textwidth}
  \newcommand{\mywidthlr}{0.08\textwidth}
  \newcolumntype{C}{ >{\centering\arraybackslash} m{0.02\textwidth} }
  \newcolumntype{Y}{ >{\centering\arraybackslash} m{\mywidthlr} }  
  \newcolumntype{X}{ >{\centering\arraybackslash} m{\mywidth} }
  \newcolumntype{Z}{ >{\centering\arraybackslash} m{\mywidth} }
  \setlength\tabcolsep{1pt} 
  \def\arraystretch{1}
  \begin{tabular}{C|XY||XX|XX|XX|}
\cline{2-9}
&\multicolumn{2}{|r||}{Approach} & \multicolumn{2}{c|}{SfS (Section \ref{sec:depthsr_sfs})} & \multicolumn{2}{c|}{SfS + reflectance learning (Section \ref{sec:depthsr_sfs_deep})} & \multicolumn{2}{c|}{UPS (Section \ref{sec:depthsr_ups})} \\ \cline{2-9}
&\multicolumn{2}{|r||}{Required data} & \multicolumn{2}{c|}{1 RGB-D frame} & \multicolumn{2}{c|}{1 RGB-D frame + training dataset} & \multicolumn{2}{c|}{$n\geq4$ RGB-D frames} \\
&\multicolumn{2}{|r||}{Albedo} & \multicolumn{2}{c|}{Piecewise-constant} & \multicolumn{2}{c|}{Learned (e.g., faces)}                   & \multicolumn{2}{c|}{Arbitrary} \\\cline{2-9}
&&&&&&&&\\ [-7.5pt] 
\rotatebox{90}{Rucksack} & 
\includegraphics[width=\mywidth]{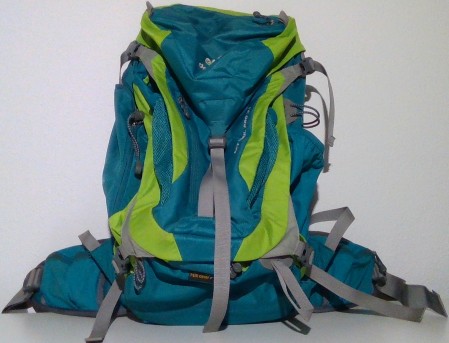} &
\includegraphics[width=\mywidthlr]{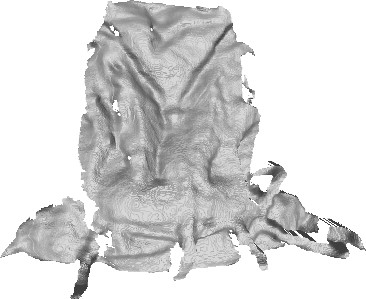} & 
\includegraphics[width=\mywidth]{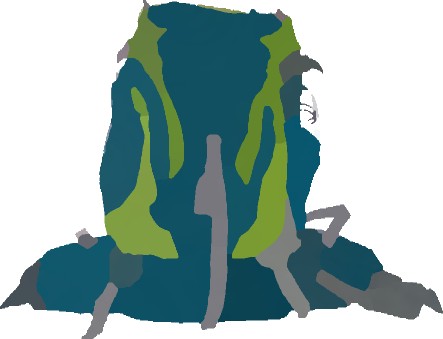} &  
\includegraphics[width=\mywidth]{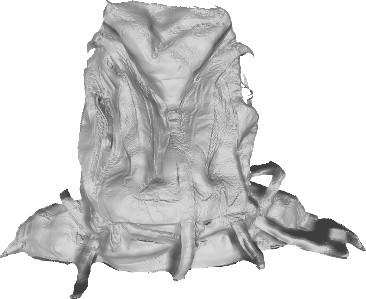} &  
\includegraphics[width=\mywidth]{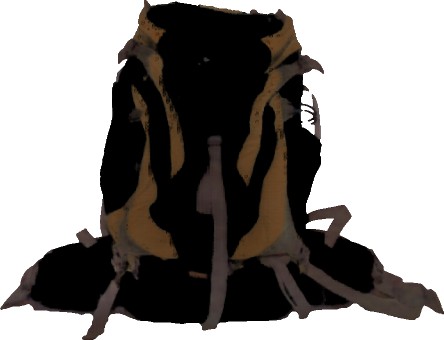} & 
\includegraphics[width=\mywidth]{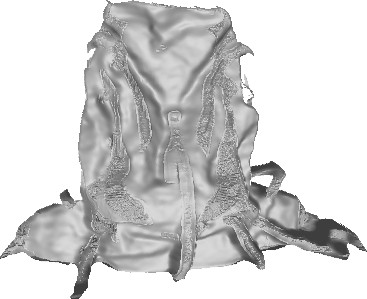} & 
\includegraphics[width=\mywidth]{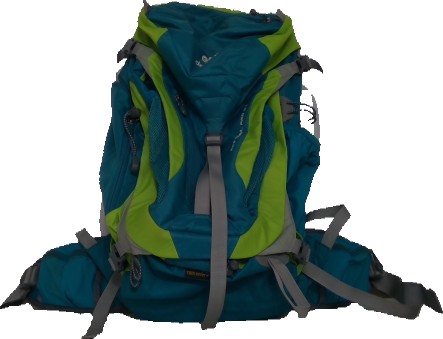} & 
\includegraphics[width=\mywidth]{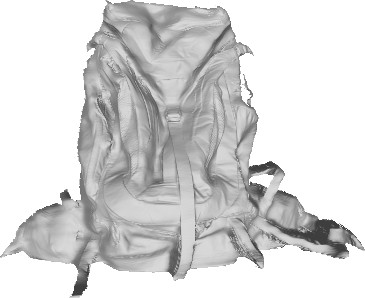} \\ [-5pt]
\rotatebox{90}{Face 1} & 
\includegraphics[width=\mywidth]{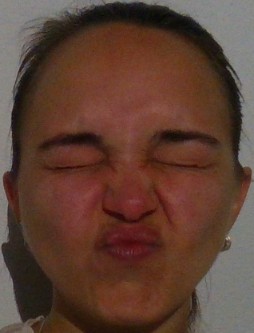} & 
\includegraphics[width=\mywidthlr]{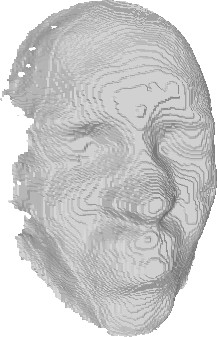} & 
\includegraphics[width=\mywidth]{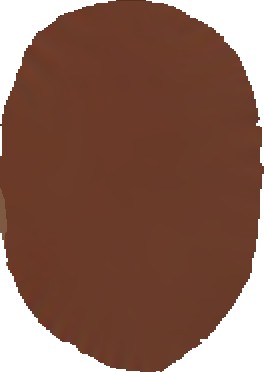} &  
\includegraphics[width=\mywidth]{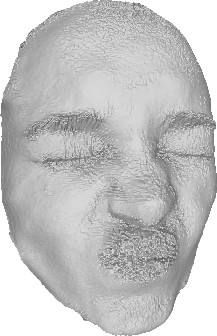} &  
\includegraphics[width=\mywidth]{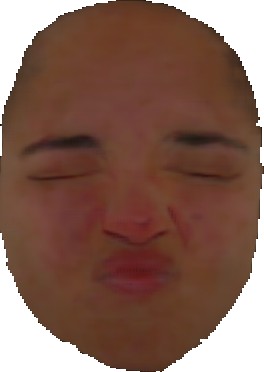} &  
\includegraphics[width=\mywidth]{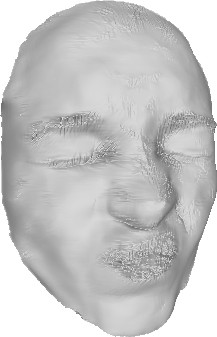} & 
\includegraphics[width=\mywidth]{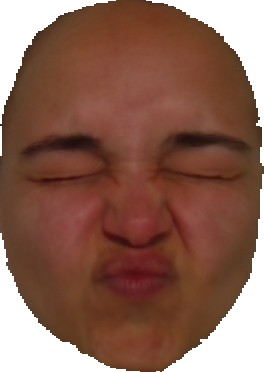} & 
\includegraphics[width=\mywidth]{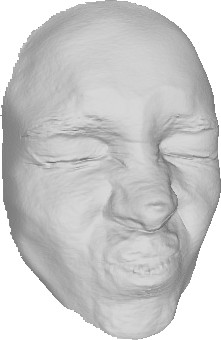} \\ [-2.5pt]
\rotatebox{90}{Tabletcase} & 
\includegraphics[width=\mywidth]{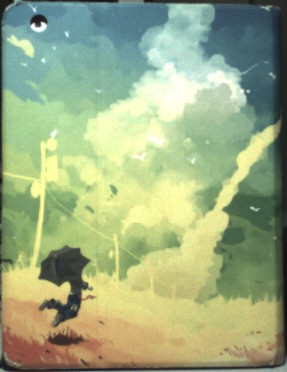} & 
\includegraphics[width=\mywidthlr]{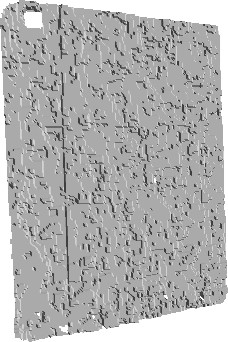} & 
\includegraphics[width=\mywidth]{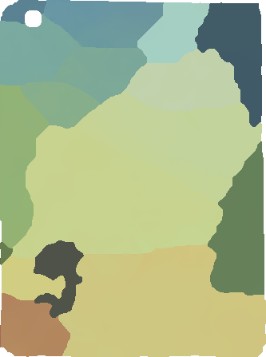} & 
\includegraphics[width=\mywidth]{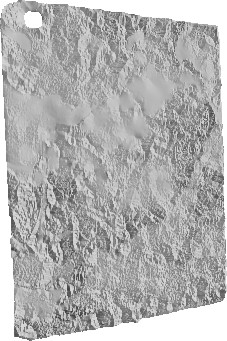} & 
\includegraphics[width=\mywidth]{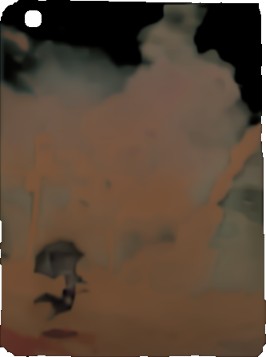} & 
\includegraphics[width=\mywidth]{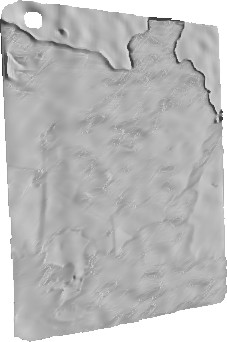} & 
\includegraphics[width=\mywidth]{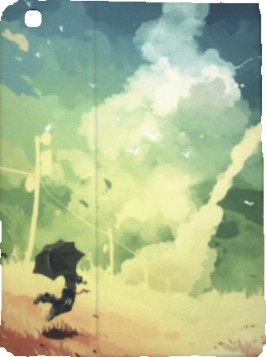} & 
\includegraphics[width=\mywidth]{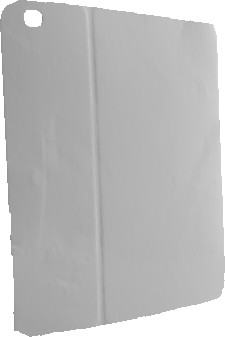} \\ \cline{2-9}
&\multicolumn{1}{|c}{$\I$} & \multicolumn{1}{c||}{$\zz$} & $\mrho$ & \multicolumn{1}{c|}{$z$} & $\mrho$ & \multicolumn{1}{c|}{$z$} & $\mrho$ & \multicolumn{1}{c|}{$z$} \\ \cline{2-9}
\end{tabular}
\caption{Photometric depth super-resolution of a low-resolution depth map $z^0$ to the higher resolution of the companion image $\I$ (first column, Rucksack and Face 1 
datasets were acquired using an Intel Realsense D415, and Tabletcase
using an Asus Xtion Pro Live). Second column: shape-from-shading (SfS) recovers high-resolution albedo~($\mrho$) and depth ($z$) from a single RGB-D frame, assuming piecewise-constant albedo. If this assumption is not satisfied (e.g., Face 1 and Tabletcase), shape estimation deteriorates. Third column: this can be circumvented by learning reflectance, an approach which is efficient as long as the target resembles the training data (here, training was carried out on human faces). Fourth column: uncalibrated photometric stereo (UPS) requires no training and handles arbitrary albedo, but it requires $n\geq4$ input frames acquired under varying illumination. See Section 6 in the supplementary material for additional comparisons.}
  \label{fig:dasteaser}
\end{figure*}

\IEEEPARstart{R}{GB-D} sensors have become very popular for 3D-reconstruction, in view of their low cost and ease of use. They deliver a colored point cloud in a single shot, but the resulting shape often misses thin geometric structures. This is due to noise, quantisation and, more importantly, the coarse resolution of the depth map. In comparison, the quality and resolution of the companion RGB image are substantially better. For instance, the Asus Xtion Pro Live device delivers $1280 \times 1024$ RGB images, but only up to $640 \times 480$ depth maps. The depth map thus needs to be up-sampled to the same resolution of the RGB image, and the latter could be analysed photometrically to reveal fine-scale details. 

However, super-resolution of a solitary depth map without additional contraints is an ill-posed problem, and retrieving geometry from either a single color image (shape-from-shading) or from a sequence of color images acquired under unknown, varying lighting (uncalibrated photometric stereo) is another ill-posed problem. The present study explores the resolution of both these ill-posedness issues by jointly performing depth super-resolution and photometric 3D-reconstruction. We call this combined approach \textit{photometric depth super-resolution}. 

The choice of jointly solving both these classic inverse problems is motivated by the observation that ill-posedness in depth super-resolution and in photometric 3D-reconstruction have different peculiarities and origins. In depth super-resolution, constraints on high-frequency shape variations are missing (there exist infinitely many ways to interpolate between two measurements), while low-frequency (e.g., concave-convex or bas-relief) ambiguities arise in photometric 3D-reconstruction. Therefore, the low-frequency geometric information necessary to disambiguate photometric 3D-reconstruction should be extracted from the low-resolution depth measurements and, symmetrically, the high-resolution photometric clues in the RGB data should provide the high-frequency information required to disambiguate depth super-resolution. One hand thus washes the other: ill-posedness in depth super-resolution is fought using photometric 3D-reconstruction, and vice-versa. 

As we shall see in Section~\ref{sec:related}, the photometric depth super-resolution problem comes down to simultaneously inferring high-resolution depth and reflectance maps, given the low-resolution depth and the high-resolution RGB images. As depicted in Figure~\ref{fig:dasteaser}, this study explores three different strategies for such a task\footnote{Codes and data can be found in \url{
https://vision.in.tum.de/data/datasets/photometricdepthsr}.}. The rest of this paper discusses them by increasing order of efficiency which, unfortunately, is inversely proportional to the amount of required resources. 1) If the available resources consist of a single RGB-D frame, then a variational approach to shape-from-shading can be followed. This approach, presented in Section \ref{sec:depthsr_sfs}, has no particular requirement in terms of acquisition setup or offline processing, yet it is effective only as long as the surface's reflectance is piecewise-constant. 2) Section~\ref{sec:depthsr_sfs_deep} then discusses a solution for eliminating this dependency upon a specific reflectance model. Pre-training a neural network for reflectance estimation allows to handle surfaces with more complex reflectance within the same variational framework. Yet, additional resources are required for offline training and the target has to resemble the objects used in the training phase (we thus focus in this section on human faces). 3) If multiple pairs of images can be acquired from the same viewing angle but under varying lighting, then one can resort to uncalibrated photometric stereo. This last strategy, discussed in Section~\ref{sec:depthsr_ups}, requires neither an assumption on the reflectance, nor offline training for a specific class of objects. However, it requires capturing more data online. Section~\ref{sec:conclusion} eventually recalls the main conclusions of this study and suggests future research directions.

\section{Problem Statement}\label{sec:related}

\begin{figure}[!ht]
\centering
\def\svgwidth{.61\linewidth}
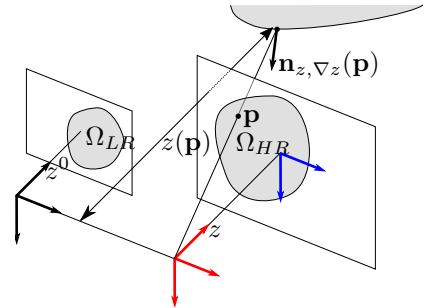
\caption{Geometric setup. Depth measurements $\zz$ are available over a low-resolution set $\OLR$, and color measurements $\I$ over a high-resolution set $\OHR$. Photometric depth super-resolution consists in estimating a high-resolution depth map $z$ out of these geometric and photometric measurements, which are connected through the surface normals $\n_{\zdz}$, see Equations \eqref{eq:1} to \eqref{eq:4}.}
\label{fig:geometry}
\end{figure}

A generic RGB-D sensor is considered, which consists of a depth sensor and an RGB camera with parallel optical axes and optical centers lying on a plane orthogonal to these axes (see Figure \ref{fig:geometry}). The images of the surface on the focal planes of the depth and the color cameras are denoted respectively by $\OLR \subset \R^2$ and $\OHR \subset \R^2$. In a single shot, the RGB-D sensor provides two 2D-representations of the surface: 
\begin{itemize}
\item A geometric one, taking the form of a mapping $\zz:\,\OLR \to \R$ between pixels in $\OLR$ and the depth of their conjugate 3D-points on the surface;
\item A photometric one, taking the form of a mapping $\I:\,\OHR \to \R^3$ between pixels in $\OHR$ and the radiance (relatively to the red, green and blue channels of the color camera) of their conjugate 3D-point. 
\end{itemize}
In real-world scenarios, the sets $\OLR$ and $\OHR$ are discrete, and the cardinality $\norm{\OLR}$ of $\OLR$ is lower than that $\norm{\OHR}$ of $\OHR$. To obtain the richest surface representation, one should thus project the depth measurements $\zz$ from $\OLR$ to $\OHR$, i.e. estimate a new, high-resolution depth map $z:\,\OHR \to \mathbb{R}$. 
To this end, we next introduce constraints arising from depth super-resolution and from photometric 3D-reconstruction.

\subsection{Geometric and Photometric Constraints}

Given the assumptions above on the alignment of the sensors, and neglecting occlusions, the low-resolution depth map $z^0$ can be considered as a downsampled version of the sought high-resolution one $z$, after warping and averaging:
\begin{equation}
\zz = K z + \eta_z,
\label{eq:1}
\end{equation}
with $\eta_z$ the realisation of a stochastic process representing measurement errors and quantisation, and $K$ a non-invertible injective linear operator combining warping, blurring and downsampling~\cite{Unger2010}, which can be calibrated beforehand~\cite{Park2011}. Solving \eqref{eq:1} in terms of the high-resolution depth map $z$ constitutes the \textit{depth super-resolution} problem, which requires additional assumptions on the smoothness of the observed surface. In this work, the latter is assumed regular, i.e. the normal to the surface exists in every visible point. Denoting by $f>0$ the focal length of the color camera, and by $\p:\, \OHR \to \R^2$ the field of pixel coordinates with respect to its principal point (blue reference coordinates system in Figure~\ref{fig:geometry}), the surface normal is defined as the following $\OHR \to \mathbb{S}^2 \subset \R^3$ field of unit-length vectors (see e.g.,~\cite{Queau_Survey}):
\begin{equation}
\n_\zdz = \frac{1}{\sqrt{\norm{f  \, \nabla z}^2 + \left( -z - \p^\top  \nabla z  \right)^2}} \begin{bmatrix}
f \, \nabla z \\ 
-z - \p^\top \nabla z 
\end{bmatrix}.
\label{eq:2}
\end{equation}

We further assume that the surface is Lambertian and lit by a collection of infinitely-distant point light sources. Lighting can then be represented in a compact manner using first-order spherical harmonics, see~\cite{Basri2003,Ramamoorthi2001} and Section 2.1 in the supplementary material. The irradiance in channel $\star \in \{R,G,B\}$ then writes
\begin{equation}
\I = \l^\top \underbrace{\begin{bmatrix}
\n_\zdz \\ 
1
\end{bmatrix}}_{:= \m_\zdz}
\mrho
 + \metai,
\label{eq:4}
\end{equation}
with $\metai:\,\OHR \to \R^3$ the realisation of a stochastic process standing for noise, quantisation and outliers, $\l \in \R^4$ the ``light vector'', $\mrho:\,\OHR \to \R^3$ the albedo (Lambertian reflectance) map and $\m_\zdz:\,\OHR \to \R^4$ a normal-dependent vector field. Solving \eqref{eq:4} in terms of the high-resolution depth map $z$ constitutes the \textit{photometric 3D-reconstruction} problem, where reflectance $\mrho$ and lighting $\l$ represent hidden variables to estimate.  

\textit{Photometric depth super-resolution} aims at inferring $z$ out of $\zz$ and $\I$, while ensuring consistency with the super-resolution constraint in~\eqref{eq:1} and with the photometric one in~\eqref{eq:4}. Before elaborating on three strategies for solving this problem, let us first review related works. 

\subsection{Related Works}

Single depth image super-resolution requires solving Equation~\eqref{eq:1} in terms of the high-resolution depth map $z$. Since $K$ is not invertible, this is an ill-posed problem: there exist infinitely many choices for interpolating between observations, cf. Section 2.2 in the supplementary material. Disambiguation can be carried out by adding observations obtained from different viewing angles~\cite{Goldlucke2014,Maier2015,Schuon2009}. In the more challenging case of a single viewing angle, a smoothness prior on the high-resolution depth map can be added and a variational approach can be followed~\cite{Unger2010}. One may also resort to machine learning techniques relying on a dictionary of low- and high-resolution depth or edge patches~\cite{Macaodha2012,Xie2016}. Such a dictionary can even be constructed from a single depth image by looking for self-similarities~\cite{Hornacek2013,Li2014}. Nevertheless, learning-based depth super-resolution methods remain prone to over-fitting~\cite{Xie2015}, which can be avoided by combining the respective merits of machine learning and variational approaches~\cite{Ferstl2015,Riegler2016}. 

Shape-from-shading~\cite{Horn1970,Breuss2012,Durou2008,Zhang1999} is another classic inverse problem which aims at inferring shape from a single image of a scene, by inverting an image formation model such as \eqref{eq:4}. Common numerical strategies for this task include variational~\cite{Horn1986,Ikeuchi1981} and PDE methods~\cite{Cristiani2007,Falcone1997,Lions1993,Rouy1992}. However, even when reflectance and lighting are known, shape-from-shading is still ill-posed due to the underlying concave~$/$~convex ambiguity, cf. Section 2.2 in the supplementary material. Obviously, even more ambiguities arise under more realistic lighting and reflectance assumptions: any image can be explained by a flat shape illuminated uniformly but painted in a complex manner, by a white and frontally-lit surface with a complex geometry, or by a white planar surface illuminated in a complex manner~\cite{Adelson1996}. Shape-from-shading under uniform reflectance but natural lighting has been studied~\cite{Huang2011,Johnson2011,Richter2015,Queau2017}, but the case with unknown reflectance requires the introduction of additional priors~\cite{Barron2015}. This can be avoided by actively controlling the lighting, a variant of shape-from-shading known as photometric stereo which allows to estimate both shape and reflectance~\cite{Woodham1980}. The problem with uncalibrated lighting is however ill-posed: it can be solved only up to a linear ambiguity~\cite{Hayakawa1994} which, assuming integrability of the normals, reduces to a generalised bas-relief (GBR) one under directional lighting~\cite{Belhumeur1999}, and to a Lorentz one under natural lighting~\cite{Basri2007}. Resolution of such ambiguities by resorting to additional priors~\cite{Alldrin2007,Papadhimitri2014b,JMIV2015}, extensions to non-Lambertian reflectance~\cite{Lu2018} and natural illumination~\cite{Mo2018} remain active research topics for which public benchmarks exist~\cite{Shi2018}. Recent developments in this field include PDE-based variational methods~\cite{CVPR2017} and machine learning solutions~\cite{Ikehata2018,Chen2019}.

Shape-from-shading has recently gained new life with the emergence of RGB-D sensors. Indeed, the rough depth map can be used as prior to ``guide'' shape-from-shading and thus circumvent its ambiguities. This has been achieved in both the multi-view~\cite{Choe2017,Maier2017,Zollhoefer2015} and the single-shot~\cite{Han2013,Kim2015,Or-El2016,Or-El2015,Wu2014,Yu2013} cases. Still, the resolutions of the input image and depth map are assumed equal, and the same holds for approaches resorting to photometric stereo instead of shape-from-shading~\cite{Anderson2011,Chatterjee2015,Xie2018,Zhang2018}. In fact, depth super-resolution and photometric 3D-reconstruction have been widely studied, but rarely together. Several methods were proposed to coalign the depth edges in the super-resolved depth map with edges in the high-resolution color image~\cite{Diebel2006,Ferstl2013,Park2011,Yang2007,Li2018,Hui2016}, but such approaches only consider sparse color features and may thus miss thin geometric structures. Some authors super-resolve the photometric stereo results~\cite{Tan2008}, and others generate high-resolution images using photometric stereo~\cite{Chaudhuri2005}, but none employ low-resolution depth clues except those of~\cite{Lu2013}, who combine calibrated photometric stereo with structured light sensing. However, this involves a non-standard setup and careful lighting calibration, and reflectance is assumed to be uniform. Such issues are circumvented in the building blocks  \cite{Haefner2018} and \cite{Peng2017} of this study, which deal with photometric depth super-resolution based on, respectively, shape-from-shading and photometric stereo. Let us present the former approach, which is a single-shot solution to photometric depth super-resolution based on a variational approach to shape-from-shading.

\section{Single-shot Depth Super-resolution using Shape-from-shading}\label{sec:depthsr_sfs}

In this section, the input data consists of a single RGB-D frame, i.e. a high-resolution image $\I$ and a low-resolution depth map $\zz$. To obtain a high-resolution depth map $z$ consistent with both the geometric constraint \eqref{eq:1} and the photometric one~\eqref{eq:4}, we consider a variational approach which comes down to solving the optimization problem~\eqref{eq:variational}. Following \cite{Mumford1994}, such a variational formulation can be derived from a Bayesian rationale. 

\subsection{Bayesian-to-Variational Rationale}

Besides the high-resolution depth map $z$, neither the reflectance $\mrho$ nor the lighting vector $\l$ is known. We treat the joint recovery of these three quantities as a maximum a posteriori (MAP) estimation problem. To this end we aim at maximising the posterior distribution of $\I$ and $\zz$ which, according to Bayes rule, writes
\begin{equation}\label{eq:map}
\P(z,\mrho,\l | \zz,\I) = \frac{\P(\zz,\I|z,\mrho,\l) \, \P(z,\mrho,\l)}{\P(\zz,\I)}.
\end{equation}
In~\eqref{eq:map}, the denominator is the evidence, which is a constant with respect to the variables $z,\mrho$ and $\l$ and can thus be neglected during optimisation. The numerator is the product of the likelihood $\P(\zz,\I|z,\mrho,\l)$ and the prior distribution $\P(z,\mrho,\l)$, which both need to be further discussed.

The measurements of depth and image observations being done using separate sensors, $\zz$ and $\I$ are statistically independent and thus the likelihood factors out as $\P(\zz,\I|z,\mrho,\l) = \P(\zz|z,\mrho,\l) \, \P(\I|z,\rho,\l)$. Furthermore, we assume that the process of how the depth map $\zz$ is acquired is depending neither on lighting $\l$ nor on reflectance~$\mrho$. Given this, the marginal likelihood for the depth map~$\zz$ can be written as $\P(\zz|z,\mrho,\l) = \P(\zz|z)$. Assuming that noise $\eta_z$ in \eqref{eq:1} is homoskedastic, zero-mean and Gaussian-distributed with variance $\sigma_z^2$, we further have $\P(\zz|z) \propto \exp\left\{-\frac{\Norm{Kz-\zz}_{2}^2}{2\sigma_z^2}\right\}$ (here $\Norm{\cdot}_{2}$ is the $\ell^2$-norm over $\OLR$). Concerning the marginal likelihood of $\I$, we assume the random variable $\metai$ in \eqref{eq:4} follows a homoskedastic Gaussian distribution with zero mean and covariance matrix $\diag(\sigma_I^2,\sigma_I^2,\sigma_I^2)\in\R^{3\times3}$, thus $\P(\I|z,\mrho,\l)\propto \exp\left\{-\frac{\Norm{\l^\top \m_\zdz \, \mrho-\I}_{2}^2}{2\sigma_I^2}\right\}$ (this time, $\Norm{\cdot}_{2}$ is the $\ell^2$-norm over $\OHR$). Therefore, the likelihood in \eqref{eq:map} is given by
\begin{equation}\label{eq:likelihood}
\P(\zz\!\!,\I|z,\mrho,\l) \!\propto\! \exp\!\left\{\!-\frac{\Norm{\!K\!z\!-\!\zz\!}_{2}^2}{2\sigma_z^2} \!-\!\frac{\Norm{\!\l^\top \! \m_\zdz \, \mrho\!-\!\I\!}_{2}^2}{2\sigma_I^2}\!\right\}\!.
\end{equation}

The prior distribution $\P(z,\mrho,\l)$ in \eqref{eq:map} can be derived in a similar manner. The Lambertian assumption implies independence of reflectance from geometry and lighting, and the distant-light assumption implies independence of geometry and lighting. Therefore, $z$, $\mrho$ and $\l$ are statistically independent and the prior distribution factors out as 
\begin{equation}\label{eq:priors}
\P(z,\mrho,\l) = \P(z)\P(\mrho)\P(\l).
\end{equation}
Regarding lighting, we do not want to favor any particular situation and thus we opt for an improper prior:
\begin{equation}\label{eq:improper}
\P(\l) = \text{constant}.
\end{equation}
The prior on $z$ is slightly more evolved. As we want to prevent oversmoothing (Sobolev regularisation) and/or staircasing artefacts (total variation regularisation), we make use of a minimal surface prior~\cite{Graber2015}. To this end, a parametrisation $\dA_\zdz:\OHR\to\R$ mapping each pixel to the corresponding area of the surface element is required. This writes $\dA_\zdz = \frac{z}{f^2}\sqrt{\norm{f  \, \nabla z}^2 + \left( -z - \p^\top \nabla z  \right)^2}$, and the total surface area is then given by $\Norm{\dA_\zdz}_{1}$ (here $\Norm{\cdot}_{1}$ is the $\ell^1$-norm over $\OHR$). Introducing a free parameter $\alpha>0$ to control the surface smoothness, the minimal surface prior can then be stated as
\begin{equation}\label{eq:depth_prior}
\P(z)\propto\exp\left\{ -\frac{\Norm{\dA_\zdz}_{1}}{\alpha} \right\}.
\end{equation}
Following the Retinex theory \cite{Land1977}, reflectance $\mrho$ can be assumed piecewise-constant, resulting in a Potts prior
\begin{equation}\label{eq:albedo_prior}
\P(\mrho)\propto\exp\left\{ -\frac{\Norm{\nabla\mrho}_{0}}{\beta} \right\},
\end{equation}
with $\beta >0$ controlling the degree of discontinuities in the reflectance $\mrho$. Note that $\mrho$ is a vector field, thus for each pixel~$\p$, $\nabla\mrho(\p) = \left[\nabla\rho_R(\p), \nabla\rho_G(\p), \nabla\rho_B(\p)\right]^\top\in\R^{3\times 2}$, and we use the following definition of the $\ell^0$-``norm'' over~$\OHR$: $
\Norm{\nabla \mrho}_{0} := \sum_{\p\in\OHR} \begin{cases} 
      0 & \text{if} \norm{\nabla \mrho(\p)}_F=0, \\
      1 & \text{else}
   \end{cases}
$, with~$\norm{\cdot}_F$ the Frobenius norm over $\R^{3\times 2}$.

The MAP estimate for depth, reflectance and lighting is eventually attained by maximising the posterior distribution \eqref{eq:map} or, equivalently, minimising its negative logarithm. Plugging Equations \eqref{eq:likelihood} to \eqref{eq:albedo_prior} into \eqref{eq:map}, and discarding all additive constants, this comes down to solving the following variational problem: 
\begin{equation}\label{eq:variational}
\min_{z, \mrho, \l }  \Norm{\l^{\!\top}\! \m_\zdz \, \mrho -\!\I}_{2}^2 \!\!+ \mu\!\Norm{\!K\!z\!-\!\zz\!}_{2}^2  \!+ \nu\!\Norm{\dA_\zdz\!}_{1} \!+ \lambda\!\Norm{\nabla\!\mrho}_{0},
\end{equation}
where the trade-off parameters $(\mu,\nu,\lambda)$ are given by
\begin{align}\label{eq:parameter_ratios_sfs}
\mu&=\frac{\sigma_I^2}{\sigma_z^2}, & \nu&=\frac{\sigma_I^2}{\alpha}, & \lambda&=\frac{\sigma_I^2}{\beta}.
\end{align}

\subsection{Numerical Solving of \eqref{eq:variational}}\label{sec:depthsr_sfs:numerics}
The variational problem in \eqref{eq:variational} is not only nonconvex, but also inherits a nonlinear dependency upon the gradient of $z$, see \eqref{eq:4} along with \eqref{eq:2}. Compared to other methods, which overcome this issue by either following a two-step approach via optimising over the normals and then fitting an integrable surface to it \cite{Han2013} (a strategy which may fail if the estimated normals are non-integrable), or by freezing the nonlinearity~\cite{Or-El2015} (which may yield convergence issues, in view of the nonconvexity of the optimisation problem), we solve for the depth directly and without any approximation. To this end we follow~\cite{Queau2017} and turn the global-and-nonlinear problem~\eqref{eq:variational} into a sequence of global-yet-linear and nonlinear-yet-local ones. This can be achieved by introducing an auxiliary vector field $\mtheta:\OHR\to\R^3$ with $\mtheta:=(\zdz)$ and rewriting~\eqref{eq:variational} as the following equivalent constrained optimisation problem:
\begin{align}\label{eq:numerics}
\min_{z,\mrho,\l,\mtheta} & \Norm{\l^\top\m_{\mtheta}\, \mrho-\I}_{2}^2 \! + \mu\!\Norm{Kz-\zz}_{2}^2 \!+ \nu\!\Norm{\dA_\mtheta}_{1} \!+ \lambda\!\Norm{\nabla \mrho}_{0} \nonumber \\
\text{s.t.}\,\,&\mtheta = (\zdz).
\end{align}

To solve the nonconvex, non-smooth and constrained optimisation problem \eqref{eq:numerics} we make use of a multi-block ADMM scheme \cite{Boyd2011,Eckstein1992,Glowinski1975}. This comes down to iterating a sequence consisting of minimisations of the augmented Lagrangian
\begin{align}\label{eq:lagrangian}
\mathcal{L}(z,\mrho,\l,\mtheta,\u)= & \Norm{\l^\top \m_\mtheta \, \mrho -\I}_{2}^2 \!+\! \mu\!\Norm{K\!z\!-\!\zz}_{2}^2 \!+\!\nu\!\Norm{\dA_\mtheta}_{1} \nonumber\\
 & \!\!\!\!\!\!\!\!\!\!\!\!\!\!\!\!\!\!\!\!\!\!\!\!\!\!\!\!\!\!\!\!\!\!\!\! + \lambda\Norm{\nabla\mrho}_{0} + \left(\mtheta-(\zdz)\right)^\top\!\u + \frac{\kappa}{2}\Norm{\mtheta-(\zdz)}_{2}^2
\end{align}
over the primal variables $z,~\mrho,~\l$ and $\mtheta$, and one gradient ascent step over the dual variable $\u:\OHR\to\R^3$ ($\kappa>0$ can be viewed as a step size).

At iteration $(k)$, one sweep of this scheme writes as:
\begin{align}
& \mrho^{(k+1)} =  \argmin_{\mrho}  \Norm{\l^{(k)^\top } \m_{\mtheta^{(k)}} \, \mrho -\I}_{2}^2  + \lambda\Norm{\nabla\mrho}_{0}, \label{eq:rho_update} \\
& \l^{(k+1)} =  \argmin_{\l}  \Norm{\l^\top \m_{\mtheta^{(k)}}\,\mrho^{(k+1)}-\I}_{2}^2, \label{eq:l_update} \\
& \mtheta^{(k+1)} =  \argmin_{\mtheta} \Norm{\l^{(k+1) \top} \m_{\mtheta} \, \mrho^{(k+1)} -\I}_{2}^2 \label{eq:theta_update} \\
&  \qquad\qquad\qquad + \nu\Norm{\dA_\mtheta}_{1} + \frac{\kappa}{2}\Norm{\mtheta - (\zdz)^{(k)} + \u^{(k)}}_{2}^2,    \nonumber\\
& z^{(k+1)} = \argmin_{z}  \mu\!\Norm{K\!z\!-\!\zz}_{2}^2 \!\!+\! \frac{\kappa}{2}\!\Norm{\mtheta^{(k+1)} \!-\! (\zdz) \!+\! \u^{(k)}}_{2}^2,  \label{eq:z_update} \\
& \u^{(k+1)} = \u^{(k)} + \mtheta^{(k+1)} - (\zdz)^{(k+1)}. \label{eq:u_update}
\end{align}
The albedo subproblem \eqref{eq:rho_update} is solved using the primal-dual algorithm \cite{Strekalovskiy2014}. The lighting update step in \eqref{eq:l_update} is done using the pseudo-inverse. The $\mtheta$-update \eqref{eq:theta_update} is a nonlinear optimisation subproblem, yet free of neighboring pixel dependency thanks to the proposed splitting. It can be solved independently in each pixel using the implementation~\cite{Schmidt} of the L-BFGS method \cite{Liu1989}. Eventually, the conjugate gradient method is applied on the normal equations of \eqref{eq:z_update}, which is a sparse linear least squares problem.

Our initial values for $(k)=(0)$ are chosen to be $\mrho^{(0)} = \I$, $\l^{(0)} = [0,0,-1,0]^\top$, $z^{(0)}$ an inpainted~\cite{InpaintNans} and smoothed~\cite{He2013} version of $\zz$ followed by bicubic interpolation to upsample to the image domain $\OHR$, $\mtheta^{(0)} = (\zdz)^{(0)}$, $\u^{(0)} = 0$ and $\kappa = 10^{-4}$. Due to the problem being non-smooth and nonconvex, to date no convergence result has been established and we leave this as future work. Nevertheless, in our experiments we have never encountered any problem reaching convergence, which we consider as reached if the relative residual falls below some threshold:
\begin{equation}
r_\text{rel} := \frac{\Norm{z^{(k+1)} - z^{(k)}}_{2}}{\Norm{z^{(0)}}_{2}}<10^{-5},
\end{equation}
and if the constraint $\mtheta=(\zdz)$ is numerically satisfied, i.e.
\begin{align}
r_\text{c} :=& \left(\mtheta^{(k+1)}-(\zdz)^{(k+1)}\right)^\top \u^{(k+1)}\nonumber\\
&+ \frac{\kappa}{2}\Norm{\mtheta^{(k+1)}-(\zdz)^{(k+1)}}_{2}^2<5\cdot10^{-6}.
\end{align}
To ensure the latter, the step size $\kappa$ is multiplied by a factor of $2$ after each iteration.

The scheme is implemented in Matlab, except the albedo update \eqref{eq:rho_update} which is implemented in CUDA. Depending on the datasets, convergence is reached between $10s$ and $90s$.

\subsection{Experiments}\label{sec:depthsr_sfs:experiments}

Although the optimal value of each parameter can be deduced using \eqref{eq:parameter_ratios_sfs}, it can be difficult to estimate the noise statistics in practice, thus we consider $(\mu,\nu,\lambda)$ as tunable hyperparameters. We first carried out a series of experiments on synthetic datasets, which showed that the set of parameters $(\mu,\nu,\lambda) = (0.1, 0.7, 1)$ seems appropriate, cf. Section 3.2 in the supplementary material. Using these values, we then conducted qualitative and quantitative comparison of our results against the state-of-the-art single-shot approaches~\cite{Xie2016,Yang2007,Or-El2015}, on synthetic datasets and publicly available real-world ones from~\cite{Shi2018,Zollhoefer2015,Maier2017}. The proposed method appeared to represent the best compromise between the recovery of high- and low-frequency geometric information. These experimental results can be found in Sections 3.3 to 3.6 in the supplementary material. 

\begin{figure}[!ht]
  \centering
  \newcommand{\mywidth}{0.127\textwidth}
  \newcommand{\mywidthlr}{0.07\textwidth}
  \newcolumntype{C}{ >{\centering\arraybackslash} m{0.02\textwidth} }
  \newcolumntype{X}{ >{\centering\arraybackslash} m{\mywidth} }
  \newcolumntype{Y}{ >{\centering\arraybackslash} m{\mywidthlr} }
  \setlength\tabcolsep{1pt} 
  \def\arraystretch{0}
  \begin{tabular}{CXXYX}
    & $\I$ & $\mrho$ & $\zz$ & $z$ \\
    \rotatebox{90}{Android} &
    \includegraphics[width=\mywidth]{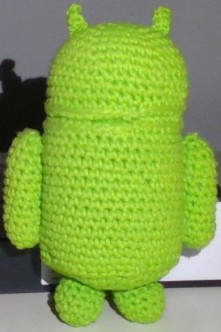}&
    \includegraphics[width=\mywidth]{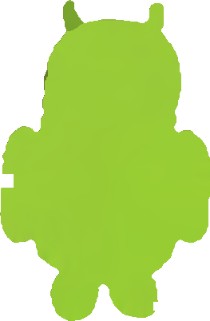}&
    \includegraphics[width=\mywidthlr]{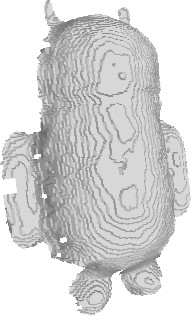}&
    \includegraphics[width=\mywidth]{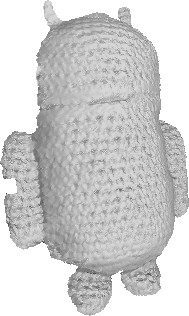}\\
    \rotatebox{90}{Basecap} &
    \includegraphics[width=\mywidth]{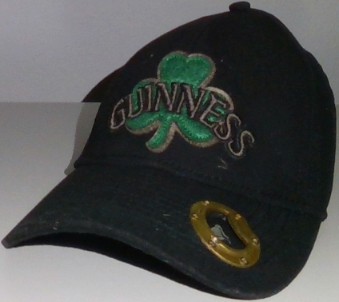}&
    \includegraphics[width=\mywidth]{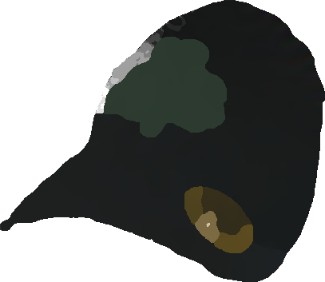}&
    \includegraphics[width=\mywidthlr]{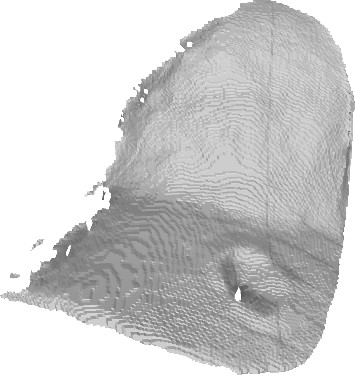}&
    \includegraphics[width=\mywidth]{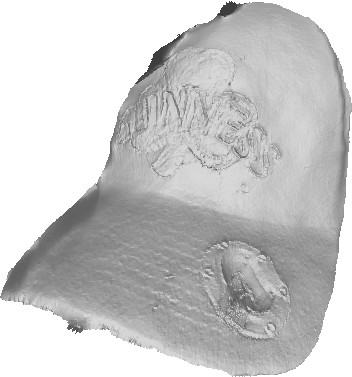}\\
    \rotatebox{90}{Minion} &
    \includegraphics[width=\mywidth]{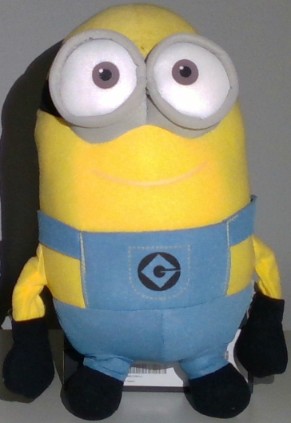}&
    \includegraphics[width=\mywidth]{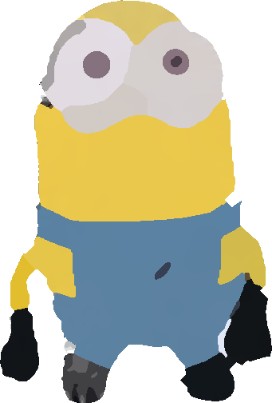}&
    \includegraphics[width=\mywidthlr]{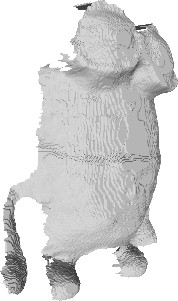}&
    \includegraphics[width=\mywidth]{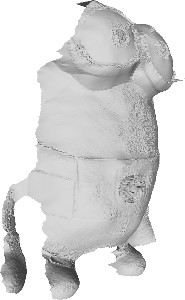}\\
  \end{tabular}
  \caption{Qualitative results obtained using the proposed single-shot approach on three real-world datasets captured with an Intel Realsense D415 camera. Even when intensity is very low (second row), or when under- or over-segmentation of reflectance happens (third row), the minimal surface prior prevents artefacts from arising while still allowing the recovery of thin geometric structures.}
  \label{fig:realworld_results_sfs}
\end{figure}

Next, we qualitatively evaluated our approach on data we captured ourselves with an Intel RealSense D415 ($1280\times720$ RGB and $320\times240$ depth) and an Asus Xtion Pro Live camera ($1280\times1024$ RGB and $320\times240$ depth). Data was captured indoor with an LED attached to the camera in order to reinforce shading in the RGB images. The objects of interest were manually segmented from background before processing. Figure \ref{fig:realworld_results_sfs} shows the resulting estimates of $\rho$ and~$z$ (1D depth profiles highlighting the recovery of thin structures can be found in Section 3.6 in the supplementary material). In the simplest ``Android'' experiment, all shading information is explained with geometry since the Potts prior prevents shading information being propagated into reflectance. The ``Basecap'' experiment is slightly more challenging due to the presence of areas with very low intensity. However, in such cases minimal surface ensures robustness, while fine details such as the stitches on the peak or the rivet of the bottle opener can still be recovered. The geometry of the 3-dimensional ``GUINNESS'' stitching is also correctly explained in terms of geometric variations and not as albedo. Although under- and over-segmentation of reflectance can be observed in the ``Minion'' experiment (cf. the eyes, the ``Gru'' logo in the center of the dungaree, or the left foot), this does not seem to affect depth estimation too much.

Another interesting qualitative result is the ``Rucksack'' experiment in Figure~\ref{fig:dasteaser}, where the very thin wrinkles are appropriately interpreted in terms of slight geometric variations. However, our method fails whenever the reflectance of the pictured object does not fit the Potts prior, see for instance the ``Face 1'' and ``Tabletcase'' experiments in Figure~\ref{fig:dasteaser}. For such objects with smoothly varying reflectance the piecewise-constant albedo assumption induces bias which propagates to the estimated depth. Indeed, the prior forbids to explain thin brightness variations in terms of reflectance, and thus the depth is forced to account for them, which results in noisy high-resolution depth maps. These failure cases illustrate the difficulty of designing a Bayesian prior which would properly split geometry and albedo information. The rest of this manuscript discusses two different strategies to circumvent this issue: by replacing the albedo estimation brick of the proposed variational framework with a deep neural network, or by acquiring additional data. The former approach is described in the next section.

\section{Depth Super-Resolution using Shape-from-Shading and Reflectance Learning}\label{sec:depthsr_sfs_deep}

The need for a strong prior on the target's reflectance is a serious bottleneck in single-shot depth super-resolution using shape-from-shading. To circumvent this issue, we investigate in this section the combination of a deep learning strategy (to estimate reflectance) with a simplified version of the proposed variational framework (to carry out depth super-resolution, with pre-estimated reflectance).


\subsection{Motivations and Construction of our Method}

If we replace the assumption of a piecewise-constant albedo by the much stronger assumption of known albedo, the variational problem from the previous section comes down to jointly achieving depth super-resolution and low-order lighting estimation, and is thus substantially simplified. Yet, the task of designing a reflectance prior which is both realistic and numerically tractable is replaced with that of designing an efficient method for estimating a reflectance map out of a high-resolution RGB image. Luckily, this problem has long been investigated in the computer vision community: it is an intrinsic image decomposition problem. Some variational solutions exist~\cite{IntrisicImageUsingOpt2011,Barron2015}, yet they rely on explicit reflectance priors and thus suffer from the same limitations as the previously proposed approach. One recent alternative is to rather resort to convolutional neural networks (CNNs), see for instance~\cite{Fan2018}.

One important issue pertaining to CNN-based albedo estimation techniques is the lack of inter-class generalisation. Nevertheless, as long as the object to be analysed resembles those used during the training stage, the albedo estimates are satisfactory (see Section 2.3 in the supplementary material). Therefore, our proposal is to replace our man-made reflectance prior (piecewise-constantness) by a less explicit prior on the class of objects that the target belongs to. In this section, we focus on the class of human faces, as e.g., in \cite{FaceDecomWithPriors2014}, in view of both the richness of geometric details to recover and the complexity of the reflectance.

Let us emphasise that we resort to CNNs only for reflectance estimation and not for geometry refinement, although several deep learning strategies are able to provide shape clues~\cite{Eigen2014,FaceNormals2017,NeuralFace2017,Pixel2mesh2018,Sfsnet18}. Indeed, such methods have shown commendable results yet they are fraught with good-to-the-eye but possibly physically-incorrect geometry estimates, probably because during testing time they are unfettered by any concrete physics-based model and prior. Given that we do already have a physics-based depth refinement framework at hand, which furthermore makes use of the available low-resolution geometric clues from the depth sensor, we believe it is more sound to pick the best from both worlds~-~deep learning and variational methods. The solution we advocate thus contains two building blocks: a deep neural network prior-lessly learns the mapping from the input RGB image to reflectance for a particular class of objects (here, human faces), and then our variational framework based on shape-from-shading provides a physically-sound numerical framework for depth super-resolution. 


\subsection{Reflectance Learning}

\begin{figure}[!ht]
  \centering
  \newcommand{\mywidth}{0.055\textwidth}
  \newcolumntype{X}{ >{\centering\arraybackslash} m{\mywidth} }
  \setlength\tabcolsep{1.5pt} 
  \def\arraystretch{0}
  \begin{tabular}{XXXXXXXX}
	\includegraphics[width=\mywidth]{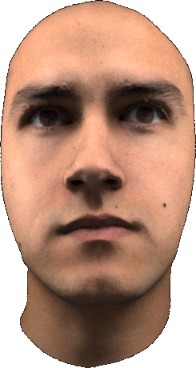}&
	\includegraphics[width=\mywidth]{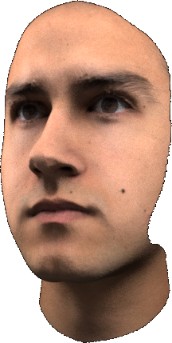}&
	\includegraphics[width=\mywidth]{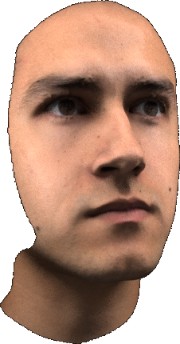}&
	\includegraphics[width=\mywidth]{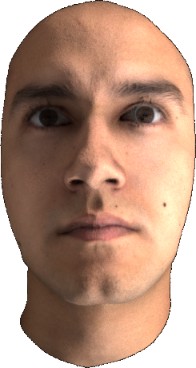}&
	\includegraphics[width=\mywidth]{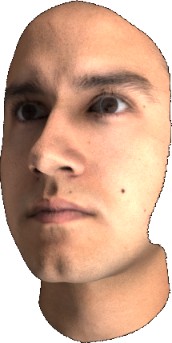}&
	\includegraphics[width=\mywidth]{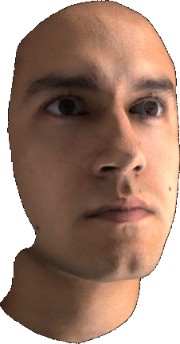}&
	\includegraphics[width=\mywidth]{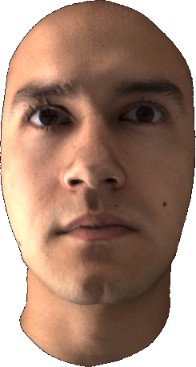}&
	\includegraphics[width=\mywidth]{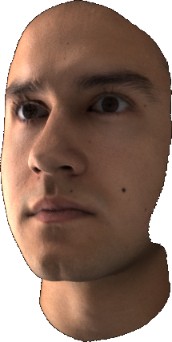}\\
	\includegraphics[width=\mywidth]{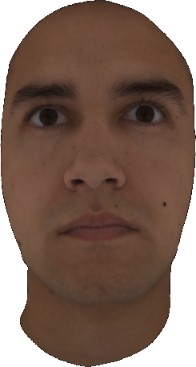}&
	\includegraphics[width=\mywidth]{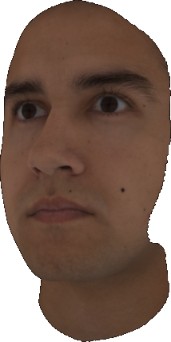}&
	\includegraphics[width=\mywidth]{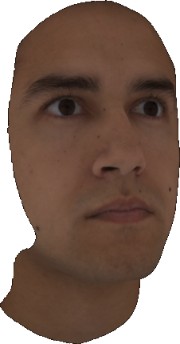}&
	\includegraphics[width=\mywidth]{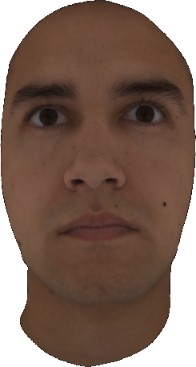}&
	\includegraphics[width=\mywidth]{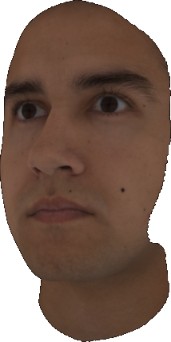}&
	\includegraphics[width=\mywidth]{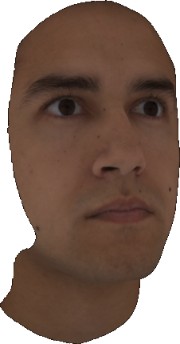}&
	\includegraphics[width=\mywidth]{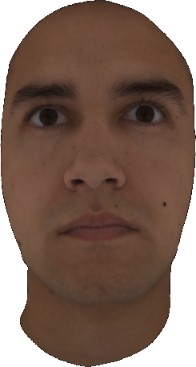}&
	\includegraphics[width=\mywidth]{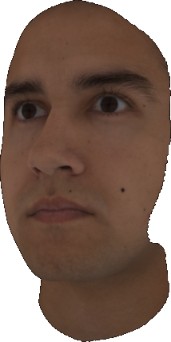}
  \end{tabular}
	\caption{Examples of human faces rendered under varying viewing and lighting conditions (top), along with the corresponding albedo maps (bottom).}
	\label{fig:renderedFaces}
\end{figure}

To train a CNN for the estimation of the human face's reflectance, one needs at his disposal hundreds of facial images in vivid lighting and viewing conditions, along with the corresponding albedo maps (see Figure~\ref{fig:renderedFaces}). This could be achieved using photometric stereo, yet the process would be very tedious. Training a neural network using synthetic images is a much simpler alternative: for instance, the approach from \cite{Jian2017} resorts to the ShapeNet 3D-model library for estimating the albedo of inanimate objects. We follow a similar approach, but dedicated to human faces.

We consider for this purpose the ICT-3DRFE database \cite{Database3drfe2007,Database3drfe2011}, which comprises of 3D meshes of human faces, reflectance maps and normal maps. These databases were captured using a Light Stage, which provides fine-detailed shape and reflectance. Using a rendering software like Blender, one can then relight the faces and change viewing angles in order to obtain hundreds of shaded RGB images along with ground-truth albedo maps. Our training dataset consists of 21 faces, each enacting 15 different expressions. For each face and each expression, several images are acquired under varying lighting conditions induced by combining ten extended light sources. In practice, eight different lighting conditions are simulated by modulating the intensity of each light source, in accordance to the usual lighting in homes and offices e.g., light sources on the ceiling, walls, windows etc. Furthermore, rendering of the faces is done from three different viewing angles, i.e. center, slight left and slight right. Eventually, the images are generated using the Lambertian reflectance model. In total, after pruning the dataset and augmenting the faces for lighting, viewpoint and specularity, the training set comprises of 5175 images. Figure \ref{fig:renderedFaces} shows some rendering examples, along with the corresponding ground-truth albedo maps.

A CNN is then trained to learn the mapping from the rendered face images to the corresponding ground-truth reflectance. Our network architecture is based on U-Net~\cite{unet2015}. Generally, U-Net comprises of convolution and nonlinear layers which downsample the input to a 1D array and then upsample to the same input size using transpose convolution and nonlinear layers. Apart from these layers, an important architectural nuance of U-Net is the skip connections between downsampling and upsampling layers. This allows U-Net to produce sharp results, which is crucial for albedo estimation. Let us emphasise that the architecture of this network is remarkably simple, cf. Section 2.3 in the supplementary material. Once reflectance estimation is dropped out, the variational problem~\eqref{eq:variational} for joint depth super-resolution and lighting estimation also becomes rather simple. Still, the appropriate combination of such simple frameworks does provide state-of-the-art results, as we shall see in the following.

\subsection{Experiments}\label{sec:depthsr_deep:experiments}

Since the numerical framework for estimating lighting and high-resolution depth is the same as the one discussed in Section~\ref{sec:depthsr_sfs}, we use exactly the same parameters as in this section. Using these parameters, we carried out qualitative and quantitative comparison of our results against state-of-the-art methods which perform deep neural network-based depth super-resolution with the same kind of inputs as our method~\cite{Hui2016}, and deep neural network-based shape-from-shading on low-resolution RGB data (without depth super-resolution)~\cite{Sfsnet18}. Our method appears to outperform the state-of-the-art both qualitatively and quantitatively on synthetic and publicly available real-world data from~\cite{Shi2018}. We also compared our reflectance learning-based approach with the previously discussed variational approach, and the learning-based method better refines the geometry of faces, which illustrates the benefit of dropping a handcrafted prior in favor of a more general learning framework (see Section~4.2 in the supplementary material).

\begin{figure}[!ht]
	\centering
	\newcommand{\mywidth}{0.13\textwidth}
	\newcommand{\mywidthlr}{0.07\textwidth}
	\newcolumntype{C}{ >{\centering\arraybackslash} m{0.02\textwidth} }
	\newcolumntype{X}{ >{\centering\arraybackslash} m{\mywidth} }
	\newcolumntype{Y}{ >{\centering\arraybackslash} m{\mywidthlr} }
	\setlength\tabcolsep{0.2pt} 
	\def\arraystretch{0.1}
	\begin{tabular}{CXXYX}
		&$\I$ & $\mrho$ & $\zz$ & $z$ \\		 
    \rotatebox{90}{Face 2}&		
		\includegraphics[width=\mywidth]{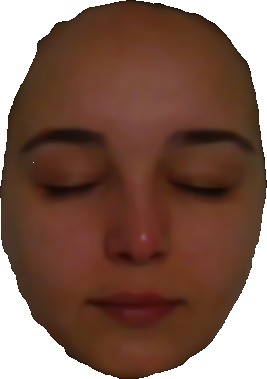}&
		\includegraphics[width=\mywidth]{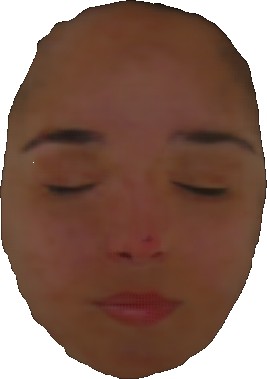}&
		\includegraphics[width=\mywidthlr]{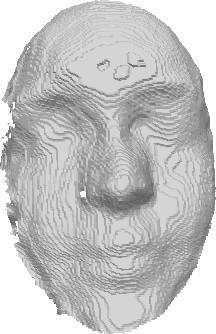}&
		\includegraphics[width=\mywidth]{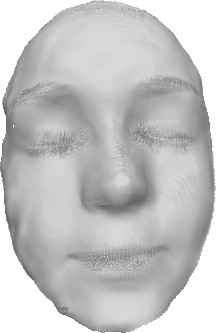}\\
    \rotatebox{90}{Face 3}&		 
		\includegraphics[width=\mywidth]{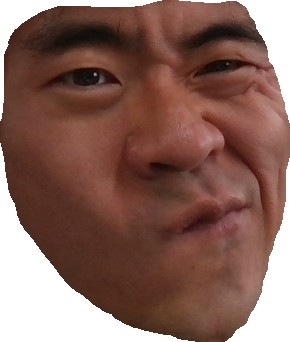}&
		\includegraphics[width=\mywidth]{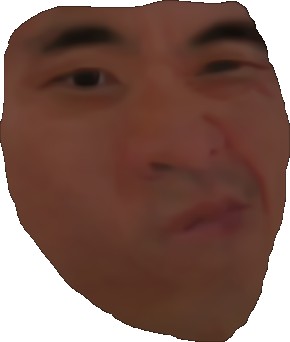}&
		\includegraphics[width=\mywidthlr]{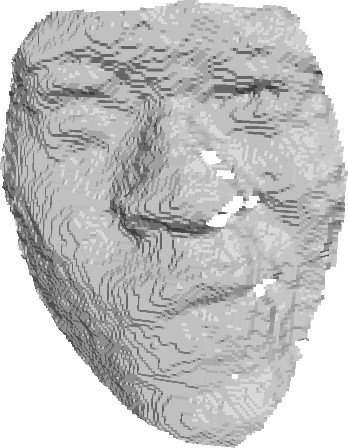}&
		\includegraphics[width=\mywidth]{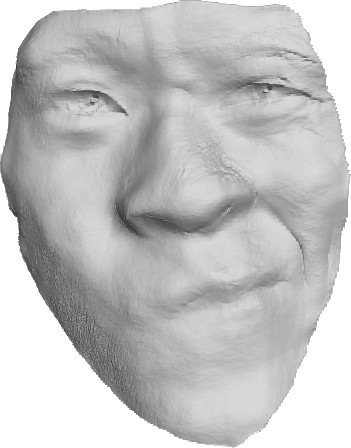}\\
    \rotatebox{90}{Face 4}&		
		\includegraphics[width=\mywidth]{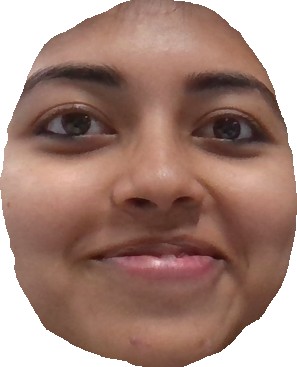}&
		\includegraphics[width=\mywidth]{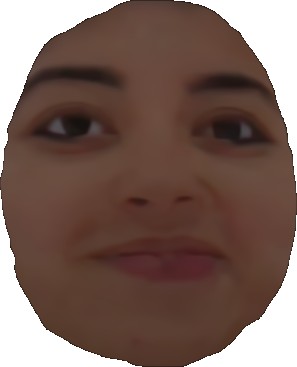}&
		\includegraphics[width=\mywidthlr]{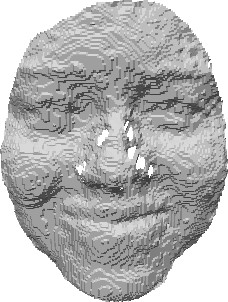}&
		\includegraphics[width=\mywidth]{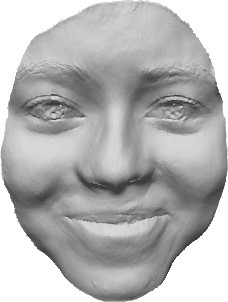}\\
    \rotatebox{90}{Face 5}&		 
		\includegraphics[width=\mywidth]{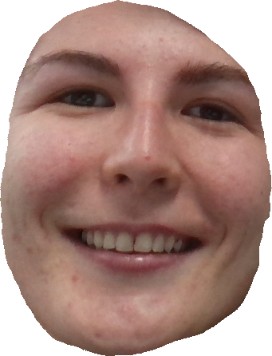}&
		\includegraphics[width=\mywidth]{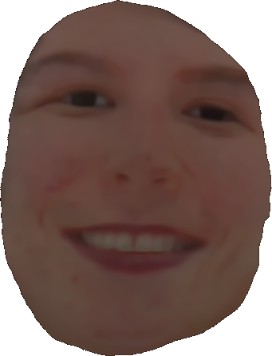}&
		\includegraphics[width=\mywidthlr]{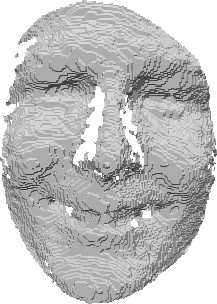}&
		\includegraphics[width=\mywidth]{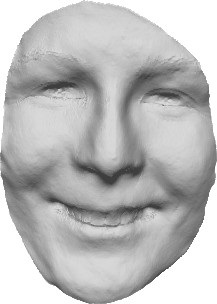}\\
	\end{tabular}
	\caption{Results of the proposed variational approach to photometric depth super-resolution, using deep learning to estimate reflectance. Data was captured with an Intel Realsense D415 camera. }
	\label{fig:realworld_results_deepnet}
\end{figure}

Next, we qualitatively evaluated our method on data we captured ourselves using an Intel RealSense D415 ($1280\times720$ RGB and $640\times360$ depth). The results in Figure~\ref{fig:realworld_results_deepnet} illustrate the ability of the proposed approach to recover detail-preserving geometry with subtle wrinkles and teeth details, in contrast with pure deep learning methods which are less accurate (see Section 4.3 in the supplementary material). Eventually, comparing the result on the ``Face 1'' dataset (Figure~\ref{fig:dasteaser}) against the shape-from-shading result from Section~\ref{sec:depthsr_sfs} also confirms the interest of replacing a model-based prior by a learning framework. However, the ``Rucksack''' and ``Tabletcase'' experiments of Figure~\ref{fig:dasteaser} also highlight the limitation of the proposed learning-based solution: whenever the object significantly departs from usual facial appearance, the reflectance fails and artifacts arise in the depth map. This can also be observed on objects from the DiLiGenT dataset~\cite{Shi2018} (see Section 4.4 in the supplementary material), although our approach still outperforms other learning-based ones. The only way to circumvent such an issue is to acquire more data in a photometric stereo manner, as discussed in the next section.

\section{Multi-Shot Depth Super-Resolution using Photometric Stereo}\label{sec:depthsr_ups}

Single-shot depth super-resolution requires some prior knowledge of the surface reflectance, either in terms of a piecewise-constant prior or of adequation to a learning database. The only way to get rid of such priors consists in acquiring multiple observations under varying lighting, i.e. performing uncalibrated photometric stereo.

Let us consider from now on a sequence of images $\Ii$, $i \in \{1,\dots,n\}$ and $n \geq 4$, captured under varying lighting conditions denoted by $\li$. The image formation model~\eqref{eq:4} is then turned into the following system of $n$ equations:
\begin{equation}
\I_i = \l_i^\top \m_\zdz \, \mrho + \metai_i,\quad i \in \{1,\dots,n\}.
\label{eq:PS_model}
\end{equation}

In~\eqref{eq:PS_model}, neither the depth $z$ nor the reflectance map $\mrho$ depends on $i$. Hence, their estimation is much more constrained in comparison with shape-from-shading. Nevertheless, nescience of the lighting vectors $\li$ makes the joint estimation of shape, reflectance and lighting an ill-posed problem: as discussed in Section~\ref{sec:related}, the arising ambiguities cannot be resolved without the introduction of additional priors. As we shall see now, in the context of RGB-D sensing the need for such priors can be circumvented and a purely data-driven approach can be followed. In other words, the low-resolution depth information act as a natural disambiguation prior for uncalibrated photometric stereo and, equally, the tailored photometric based-prior implicitly ensures surface regularity for depth map super-resolution.

\subsection{Maximum Likelihood-Based Solution}

Let us recall that the single-shot approach discussed in Section~\ref{sec:depthsr_sfs} required priors on the regularity of both the depth and the reflectance maps. By considering \textit{multiple} RGB-D frames $\{\I_i,z^{0}_i\}$, $i \in \{1,\dots,n\}$  of a static scene obtained under varying (though unknown) lighting, we hope to end up with a variational framework free of such man-made priors. To this end, we consider a maximum likelihood framework instead of a Bayesian one. 

Considering again the independence of depth and image observations as well as the independence of shape from reflectance and lighting, the joint likelihood of the observations ${\{\I_i,z^{0}_i\}}$ can be factored out as follows: 
\begin{equation}\label{eq:likelihood_sr_ups}
\P({\{\I_i,z^{0}_i\}} | z,\mrho,\li) = \P(\Ii | z,\mrho,\li) \, \P(\zzi | z).
\end{equation}
Under the assumption that the random variables $\metai_i$ in~\eqref{eq:PS_model} are homoskedastically distributed according to zero-mean Gaussian laws with the same covariance matrix $\diag(\sigma_{I}^2,\sigma_{I}^2,\sigma_{I}^2)$, the marginal likelihood for $\Ii$ can be explicitly written as
\begin{equation}\label{eq:likelihood_Ii}
\P(\Ii | z,\mrho,\li) \!\propto\! \exp\!\left\{\!-\frac{\sum_{i}\Norm{\l_i^\top\! \m_\zdz \, \mrho-\I_i}_{2}^2}{2\sigma_I^2}\!\right\}.
\end{equation}
Assuming that the $n$ low-resolution depth maps $z^0_i$ are consistent with the super-resolution model~\eqref{eq:1}, and that the $n$ corresponding random variables ${\eta_z}_i$ follow a zero-mean Gaussian distribution with same variance $\sigma_{z}^2$, the marginal likelihood for $\zzi$ writes as
\begin{equation}\label{eq:likelihood_zzi}
\P(\zzi | z) \propto \exp\left\{-\frac{\sum_{i}\Norm{Kz-z^0_i}_{2}^2}{2\sigma_z^2}\right\}.
\end{equation}

Maximum likelihood estimation of depth, reflectance and lighting consists in maximising the joint likelihood \eqref{eq:likelihood_sr_ups} or, equivalently, minimising its negative logarithm. Neglecting all additive constants and plugging \eqref{eq:likelihood_Ii} and \eqref{eq:likelihood_zzi} into \eqref{eq:likelihood_sr_ups}, this writes as the following variational problem: 
\begin{equation}\label{eq:variational_ups}
\min_{z,\mrho,\li} \sum_{i} \Norm{Kz-z^0_i}_{2}^2 + \gamma\Norm{\l_i^\top \m_\zdz \, \mrho -\I_i}_{2}^2,
\end{equation}
with the trade-off parameter $\gamma$ given by the ratio $\gamma=\frac{\sigma_z^2}{\sigma_I^2}$.
Let us emphasise the simplicity of the photometric stereo-based variational model~\eqref{eq:variational_ups}, in comparison with the one obtained using shape-from-shading, cf. \eqref{eq:variational}. Although one may think that more data introduces more complexity to such problems, we can clearly see here that in fact Problem~\eqref{eq:variational_ups} is naturally easier by itself as it does not include non-smooth prior terms on the albedo and the depth, but only two data terms. As discussed next, this allows a much simpler numerical strategy to be followed. 

\subsection{Numerical Solving of \eqref{eq:variational_ups}}\label{sec:depthsr_ups:numerics}

Contrarily to the shape-from-shading problem~\eqref{eq:variational}, in~\eqref{eq:variational_ups} the nonlinearity arises only from the unit-length constraint on the normals. Therefore, we opt for a simpler numerical solution based on fixed point iterations. Considering \eqref{eq:2} and~\eqref{eq:4},~\eqref{eq:variational_ups} can be rewritten as 
\begin{equation}\label{eq:variational_ups2}
\min_{z,\mrho,\li} \  \sum_{i} \Norm{Kz-z^0_i}_{2}^2  + \gamma\Norm{ 
\l_i^\top 
\begin{bmatrix}
\tilde{\n}_\zdz / d_\zdz \\
1
\end{bmatrix}
  \mrho-\I_i}_{2}^2,
\end{equation}
with $\n_\zdz = \tilde{\n}_\zdz /  d_\zdz$, where $d_\zdz$ is a scalar field ensuring the unit-length constraint of the normals:
\begin{equation}
d_\zdz =\sqrt{\norm{f  \, \nabla z}^2 + \left( -z - \p^\top \nabla z  \right)^2},
\end{equation}
and $\tilde{\n}_\zdz$ is a vector field encoding the normal direction: 
\begin{equation}
\tilde{\n}_\zdz = \begin{bmatrix}
f \, \nabla z \\ 
-z - \p^\top \nabla z
\end{bmatrix}. 
\end{equation}
In~\eqref{eq:variational_ups2}, only $d_\zdz$ depends in a nonlinear way on the unknown depth $z$. Therefore, it seems natural to solve \eqref{eq:variational_ups2} iteratively, while freezing the nonlinearity (contrarily to the shape-from-shading case, in photometric stereo we experimentally found this fixed point strategy to be convergent, though we leave the convergence proof for future work). At iteration $(k)$ and with the current estimates $(\mrho^{(k)}, \{\l_i^{(k)}\}, z^{(k)})$, one sweep of this scheme reads:
\begin{align}
& \mrho^{(k+1)} =  \argmin_{\mrho}  \!\sum_{i} \! \Norm{
\l^{(k)\top}_i \!\!
\begin{bmatrix}
\tilde{\n}_{z^{(k)}\!,\nabla z^{(k)}} / d_{z^{(k)}\!,\nabla z^{(k)}} \\
1
\end{bmatrix}
\!\mrho \!-\!\I_i}_{2}^2\!, \label{eq:rho_update_ups}\\
& \l^{(k+1)}_i  = \argmin_{\l_i}  \Norm{
\l^{\top}_i \!\!
\begin{bmatrix}
\tilde{\n}_{z^{(k)}\!,\nabla z^{(k)}} / d_{z^{(k)}\!,\nabla z^{(k)}} \\
1
\end{bmatrix}
\!\mrho^{(k+1)}
-\I_i
}_{2}^2  \forall i, \label{eq:l_update_ups} \\
& z^{(k+1)} = \argmin_{z}  \sum_{i}\Norm{Kz-z^0_i}_{2}^2 \label{eq:z_update_ups} \\[-0.5em]
& \qquad\qquad\qquad + \gamma\Norm{
\l^{(k+1)\top}_i \!\!
\begin{bmatrix}
\tilde{\n}_{z\!,\nabla z} / d_{z^{(k)}\!,\nabla z^{(k)}} \\
1
\end{bmatrix}
\!\mrho^{(k+1)}
-\I_i
}_{2}^2. \nonumber
\end{align}
All three problems \eqref{eq:rho_update_ups}, \eqref{eq:l_update_ups} and \eqref{eq:z_update_ups} are linear least-squares problems which we solve using the conjugate gradient method on the normal equations.

Our initial values for $(k)=(0)$ are chosen to be $\mrho^{(0)} = \mean(\Ii)$, $\l^{(0)}_i = [0,0,-1,0]^\top$ $\forall i$, and $z^{(0)}$ a smoothed version of $\mean(\zzi)$ using the guided filter~\cite{He2013} followed by bicubic interpolation to upsample to the image domain $\OHR$. As in Section \ref{sec:depthsr_sfs:numerics}, to verify convergence we check if the relative residual $r_\text{rel}$ falls below some threshold. In our experiments convergence was reached within at most 15 iterations, which corresponds to a few minutes in our Matlab implementation. 

\subsection{Experiments}\label{sec:depthsr_ups:experiments}

We first considered synthetic datasets in order to experimentally determine appropriate values for the hyper-parameter $\gamma$ and the number $n$ of images. The values $\gamma = 0.01$ and $n \in [10,30]$ were found to represent an appropriate compromise between accuracy and runtime (see Section 5.2 in the supplementary material). We then carried out qualitative and quantitative comparisons of our results against state-of-the-art uncalibrated photometric stereo~\cite{Papadhimitri2014b}, shading-based depth refinement using a low-resolution RGB image \cite{Or-El2015} and image-driven depth super-resolution using an anisotropic Huber-loss as regularisation term \cite{Unger2010,Werlberger2009}. Our approach was found to be the most effective on both synthetic and publicly available real-world datasets~\cite{Shi2018}. These experiments can be found in Sections 5.3 to 5.5 in the supplementary material. 

\begin{figure}[!ht]
\centering
  \newcommand{\mywidth}{0.1269\textwidth}
  \newcommand{\mywidthlr}{0.07\textwidth}
  \newcolumntype{C}{ >{\centering\arraybackslash} m{0.02\textwidth} }
  \newcolumntype{Y}{ >{\centering\arraybackslash} m{\mywidthlr} }
  \newcolumntype{X}{ >{\centering\arraybackslash} m{\mywidth} }
  \setlength\tabcolsep{1pt} 
  \def\arraystretch{3}
  \begin{tabular}{CXXYX}
      & one of $\I_i$ & $\mrho$ & $\zz$ & $z$ \\
    \rotatebox{90}{Face 2}&		
		\includegraphics[width=\mywidth]{Deep_Learning/realData/ina12Face}&
		\includegraphics[width=\mywidth]{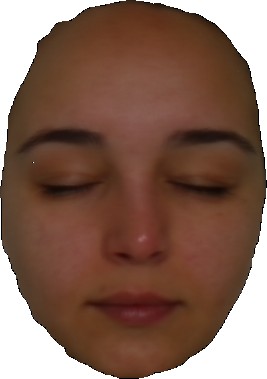}&
		\includegraphics[width=\mywidthlr]{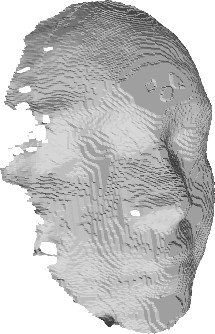}&
		\includegraphics[width=\mywidth]{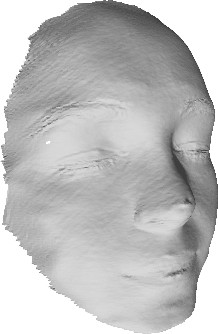}\\
		
    \rotatebox{90}{Shirt} &
    \includegraphics[width = \mywidth]{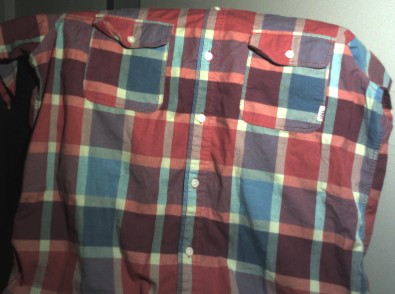}  &
    \includegraphics[width = 0.12\textwidth]{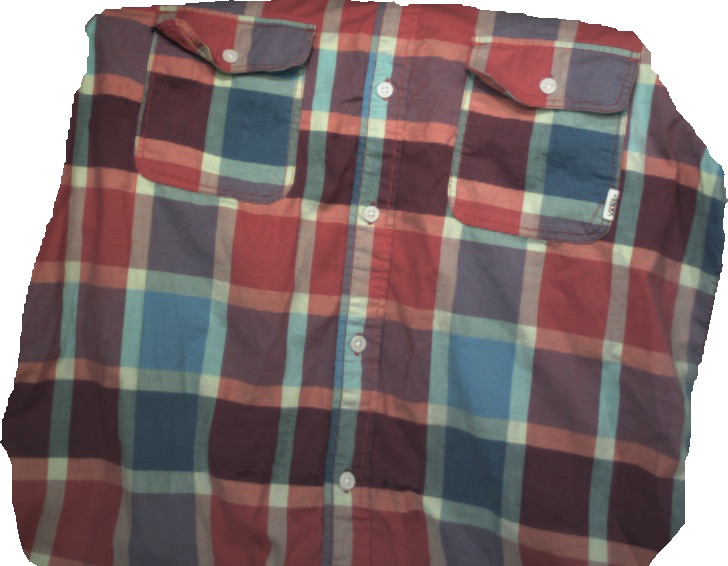}  &  
    \includegraphics[width = \mywidthlr]{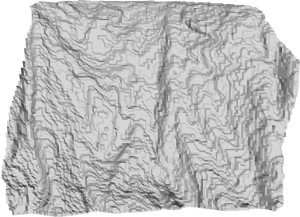} & 
    \includegraphics[width = \mywidth]{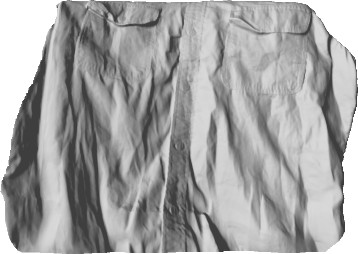} \\
          
    \rotatebox{90}{Backpack} &
    \includegraphics[width = \mywidth]{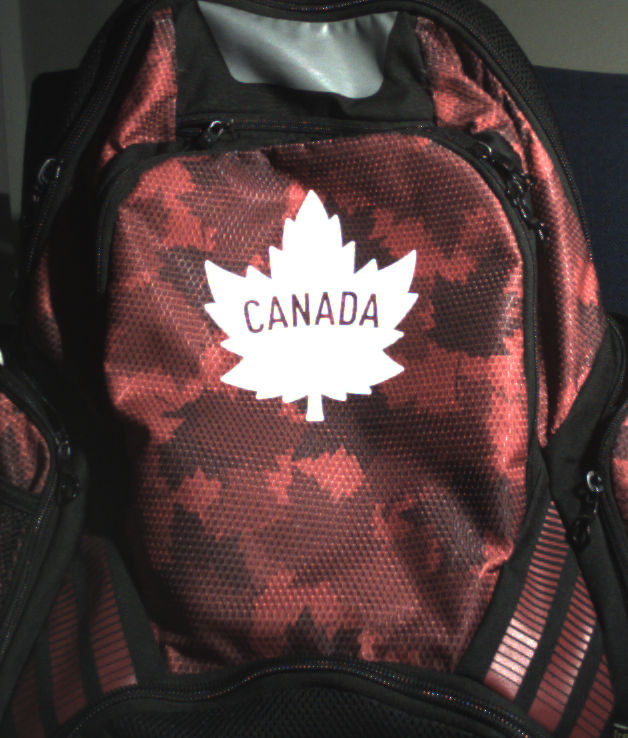}  &
    \includegraphics[width = 0.12\textwidth
    ]{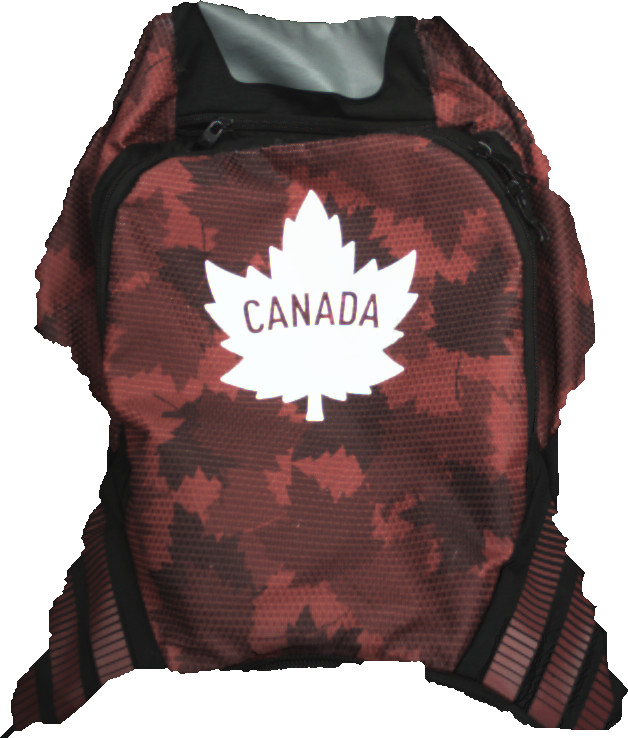}  &  
    \includegraphics[width = \mywidthlr]{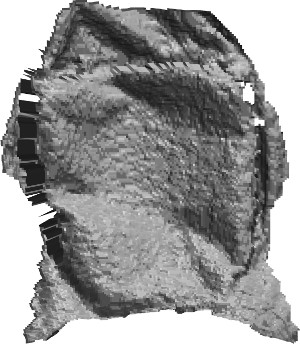} & 
    \includegraphics[width = \mywidth]{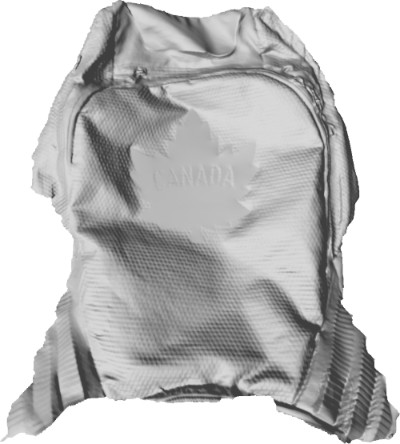} \\
    
    \rotatebox{90}{Oven mitt} &
    \includegraphics[width = \mywidth]{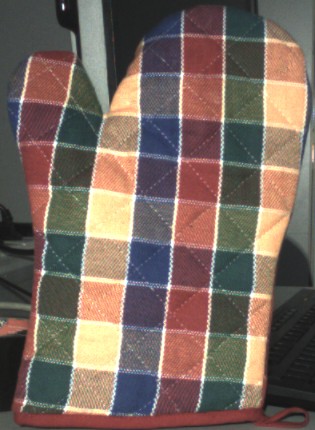}  &
    \includegraphics[width = 0.12\textwidth]{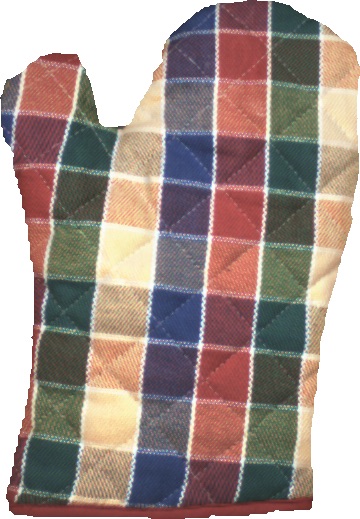}  &  
     \includegraphics[width = \mywidthlr]{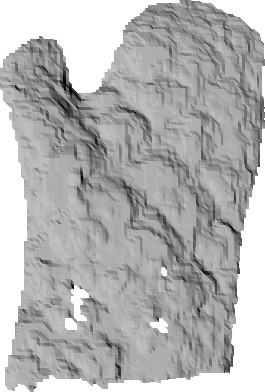} & 
   \includegraphics[width = \mywidth]{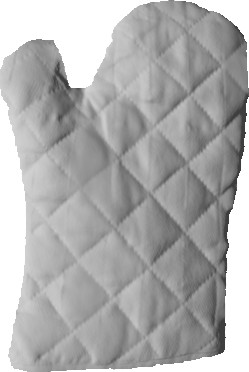}\\


    \rotatebox{90}{Hat} &
    \includegraphics[width = \mywidth]{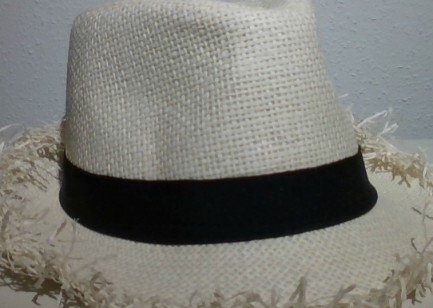}  &
    \includegraphics[width = \mywidth]{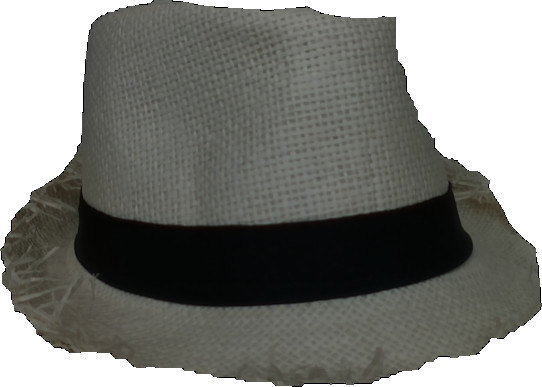}  &  
    \includegraphics[width = \mywidthlr]{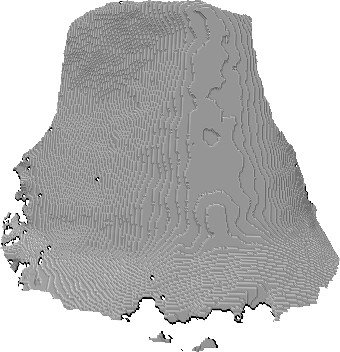} & 
    \includegraphics[width = \mywidth]{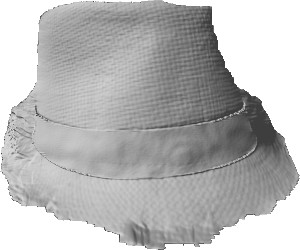}\\
    
    \rotatebox{90}{Vase} &
    \includegraphics[width = \mywidth]{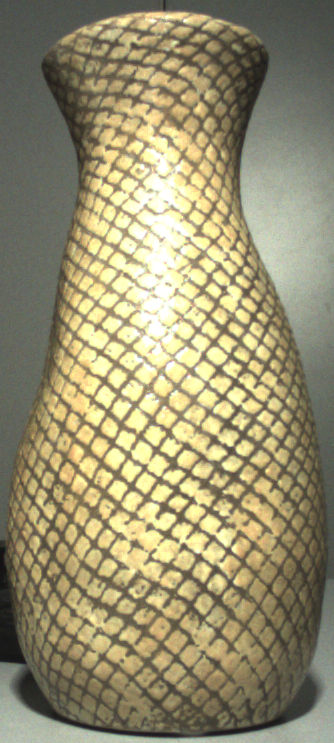}  &
    \includegraphics[width = 0.11\textwidth]{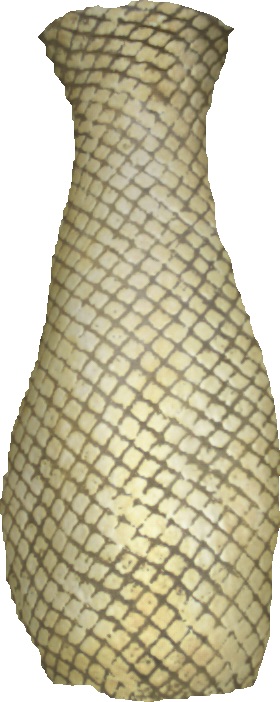}  &  
    \includegraphics[width = \mywidthlr]{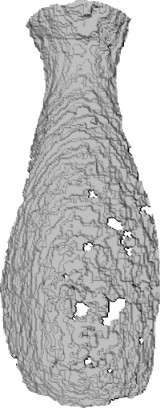} & 
    \includegraphics[width = 0.11\textwidth]{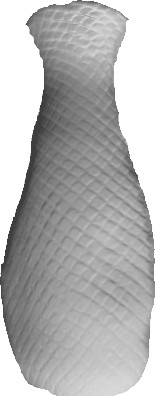}
  \end{tabular}
\caption{Qualitative results of our uncalibrated photometric stereo-based method, on real-world data captured using a RealSense D415 (``Hat'' and ``Face 2'') or an Xtion Pro Live (five other datasets).}
\label{fig:ups_exp_real}
\end{figure}

Then, we carried out a qualitative evaluation of our results on data we captured ourselves using an Asus Xtion Pro Live ($1280\times1024$ RGB and $320\times240$ depth) and an Intel Realsense D415 ($1280\times720$ RGB and $640\times480$ depth). The setup is the same as in Section~\ref{sec:depthsr_sfs:experiments}, just multiple images of the same static scene with static camera under varying lighting conditions are captured. Varying lighting was created by freely moving a handheld LED light source during the capturing process. From each image sequence, $n=20$ high-resolution RGB images $\I_i$ and low-resolution depth images $z^0_i$ were randomly extracted. Results are displayed in Figure~\ref{fig:ups_exp_real}. ``Face 2'' results are even more satisfactory compared to the deep learning-based approach in Figure~\ref{fig:realworld_results_deepnet}, despite a small spike on the nose due to a small specular spot being present in every input image. Even the fine wrinkles and the buttons of the ``Shirt'' are recovered. The thin structures of the ``Backpack'' are appropriately recovered and the partly very low reflectance does not seem to deteriorate the depth estimate. The ``Oven mitt`` contains fine stitching structures which are successfully separated from the estimated albedo. The very fine geometric details of ``Hat'' are appropriately recovered in the depth, although some shading information remains visible in the reflectance. Interestingly, although our method is based on the Lambertian reflectance assumption, the high-quality shape of the reflective ``Vase'' can still be reconstructed and even where color is saturated at the specular regions, fine-scale geometric details are recovered. Eventually, among the three methods proposed in this article, only the uncalibrated photometric stereo-based approach can handle all three datasets in Figure~\ref{fig:dasteaser}, since reflectance is constrained neither to be piecewise-constant (``Rucksack'') nor to be that of a face (``Face 1''): the smoothly-varying albedo of the ``Tabletcase'' is appropriately estimated, and separated from the thin geometric wrinkle.

\section{Conclusion}\label{sec:conclusion}

We investigated the use of photometric techniques for solving the depth super-resolution problem in RGB-D sensing. Three strategies were put forward: i) a shape-from-shading approach which requires a single RGB-D frame but is limited to objects exhibiting piecewise-constant reflectance, ii) a reflectance learning one which loosens this assumption by delegating reflectance estimation to a deep neural network trained on a specific class of objects such as faces, and iii) an uncalibrated photometric stereo setup which bypasses the need for albedo prior or training by acquiring additional data. These three approaches represent a continuum of solutions to photometric depth super-resolution with increasing level of accuracy, yet increasing amount of required resources. 

This work may still be completed in several manners. First, the theoretical properties (proofs of convergence, existence and uniqueness of solutions, etc.) of the proposed numerical schemes may be explored. Second, all the methods presented here explicitly use the linear Lambertian image formation model: a natural line of future research would be to improve robustness to off-Lambertian effects such as specularities and cast-shadows, by resorting either to robust estimation techniques as in~\cite{CVPR2017}, or to non-Lambertian image formation models as in~\cite{Chen2017}. Eventually, the combination of deep learning and variational techniques might be further explored, for instance by devoting not only reflectance estimation to a deep neural network, but also lighting estimation as in~\cite{Gardner2017}. Put together, these novelties could allow our approaches to handle more general surfaces as well as more general illumination conditions. 

%



\ifCLASSOPTIONcompsoc
  \section*{Acknowledgments}
\else
  \section*{Acknowledgment}
\fi

The authors wish to thank Thomas M\"ollenhoff and Robert Maier for helpful discussions and comments.

\ifCLASSOPTIONcaptionsoff
  \newpage
\fi



%
%
%

%
\newpage
\begin{IEEEbiography}[{\includegraphics[width=1in,height=1.25in,clip,keepaspectratio]{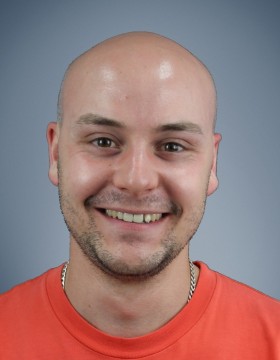}}]{Bjoern Haefner}
received his B.Sc. in Mathematics from the OTH Regensburg in 2013 and his M.Sc. in Mathematics in Science and Engineering from the Technical University of Munich in 2016. Since mid November 2016, he is a full-time PhD student in the Computer Vision and Artificial Intelligence chair at the Technical University of Munich. His research interests include RGB-D data processing for 3D reconstruction using variational methods.
\end{IEEEbiography}

\begin{IEEEbiography}
[{\includegraphics[width=1in,height=1.25in,clip,keepaspectratio]{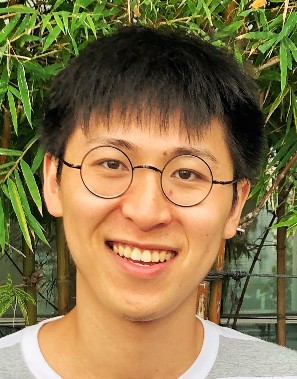}}]
{Songyou Peng}
received the Erasmus Mundus M.Sc. in Computer Vision and Robotics in 2017. Between 2016 and 2017, he spent some time doing research at INRIA Grenoble and Technical University of Munich.  Since 2018 he is a research engineer at Advanced Digital Sciences Center in Singapore. His research interests are computer vision and machine learning.
\end{IEEEbiography}

\begin{IEEEbiography}
[{\includegraphics[width=1in,height=1.25in,clip,keepaspectratio]{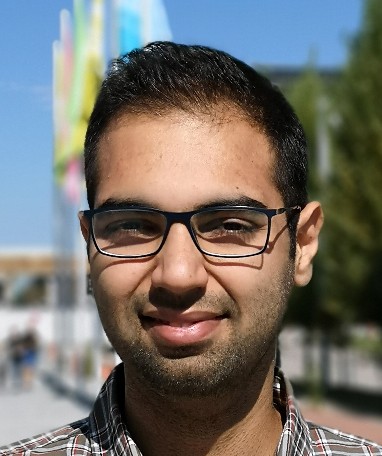}}]
{Alok Verma}
is pursuing a Master's degree in Biomedical Computing at the Technical University of Munich, Germany since 2017. Previously he worked as a senior electrical and software engineer at Philips Healthcare, Bangalore, India focusing on C-Arm X-ray Systems. His research interests are computer vision and deep learning for medical and non-medical images.   
\end{IEEEbiography}


\begin{IEEEbiography}
[{\includegraphics[width=1in,height=1.25in,clip,keepaspectratio]{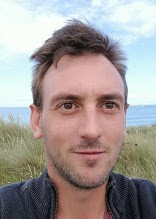}}]
{Yvain Qu\'eau}
received his Ph.D from INP-ENSEEIHT, Université de Toulouse, in 2015. From 2016 to 2018 he was a postdoctoral researcher in Technical University Munich, Germany, and then an associate processor with ISEN Brest, France. Since 2018 he is a CNRS researcher with the GREYC laboratory, Université de Caen, France. His research focuses on variational methods for solving inverse problems in computer vision.
\end{IEEEbiography}

\begin{IEEEbiography}
[{\includegraphics[width=1in,height=1.25in,clip,keepaspectratio]{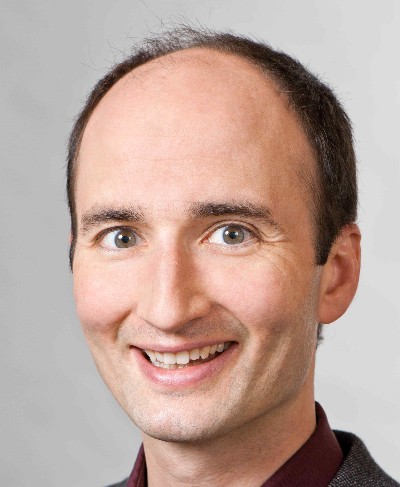}}]
{Daniel Cremers}
received the PhD degree in computer science from the University of Mannheim, Germany. Subsequently, he spent two years as a postdoctoral researcher with UCLA and one year as a permanent researcher at Siemens Corporate Research, Princeton. From 2005 until 2009, he was associate professor with the  University of Bonn. Since 2009 he holds the Chair of Computer Vision and Artificial Intelligence at the Technical University of Munich. He received numerous awards including the Gottfried-Wilhelm Leibniz Award 2016, the biggest award in German academia.
\end{IEEEbiography}




\clearpage
\appendices

\section{Organization of the document}

This document is structured as follows. Section~\ref{sec:supp_1} contains general comments on photometric 3D-reconstruction and depth super-resolution: the derivation of the RGB image formation model used through the paper, a visual description of the ambiguities arising in depth super-resolution and in shape-from-shading, and some general information regarding the reflectance learning-based approach. The rest of the document is devoted to the individual experimental evaluation of each of the proposed methods: Section~\ref{sec:supp_3} contains the shape-from-shading experiments, Section~\ref{sec:supp_4} the reflectance learning ones, and Section~\ref{sec:supp_5} evaluates the uncalibrated photometric stereo-based approach. Section~\ref{sec:supp_6} eventually concludes the document by presenting a unified comparison of the results obtained with the three proposed methods.

\section{Generalities}
\label{sec:supp_1}

\subsection{Derivation of the RGB image formation model}

This subsection is devoted to the derivation of the RGB image formation model (Eq.~(3) in the main paper), which relates the irradiance measurements and the surface normals. The following derivation is adapted from~\cite[Sect. 2.2]{LEDS}, with an extension of the model to RGB images and spherical harmonics lighting.  

We first assume that the surface is Lambertian, i.e. its appearance is independent from the viewing angle. A consequence of this assumption is that the surface's reflectance~$\rho$ at a surface point is a simple scalar quantity called the albedo, which is independent from the incident light direction. 

Next, we assume that the surface is lit by a single, infinitely distant light source represented by a direction $\omega$ on the visible hemisphere. The spectral radiance at a surface point is thus given by
\begin{equation}
L(\lambda,\omega) = \phi(\lambda,\omega) \, \dfrac{ \rho(\lambda)}{\pi} \, \max\{0,\mathbf{s}(\omega)^\top \mathbf{n}_{z,\nabla z}\},
\label{eq:1}
\end{equation}
with $\lambda$ the wavelength, $\phi(\cdot,\omega)$ the spectrum of the source associated with direction~$\omega$, $\rho(\cdot)$ the spectral reflectance of the surface point, $\mathbf{s}(\omega)$ the unit-length vector pointing towards the light source associated with direction~$\omega$, and $\n_\zdz$ the outer unit-length surface normal. 

Now, let us assume that the surface is observed under natural illumination, rather than lit by one single light source. Let us represent natural illumination by a collection of infinitely distant point light sources, each of them being represented by a direction $\omega$. The total spectral radiance of a surface point is obtained by summing the individual contributions from each source, i.e. by integrating~\eqref{eq:1} over the visible hemisphere:
\begin{equation}
L(\lambda) = \dfrac{ \rho(\lambda)}{\pi} \int_{\mathbb{S}^2} \phi(\lambda,\omega) \, \max\{0,\mathbf{s}(\omega)^\top \mathbf{n}_{z,\nabla z}\}\,\mathrm{d}\omega. 
\label{eq:2}
\end{equation}

We further assume that the sensor's response is linear, and that the RGB camera is focused on the surface. Then, the sensor's spectral irradiance, in the pixel conjugate to the surface point, is given by
\begin{equation}
E(\lambda) = \beta \cos^4 \alpha \, L(\lambda),
\label{eq:3}
\end{equation} 
where $\beta$ depends on the sensor's aperture and magnification, and where $\alpha$ is the angle between the viewing angle and the optical axis (the $\cos^4 \alpha$ factor is thus responsible for darkening at the periphery of images). 

The intensity recorded by the camera in channel $\star$, $\star \in \{R,G,B\}$,  is proportional to the sum of all spectral sensor's irradiances, weighted by the camera's transmission spectrum. Denoting by $\gamma$ this proportionality coefficient, this writes as
\begin{equation}
I_\star = \gamma \int_{\mathbb{R}^+} c_\star(\lambda) E(\lambda) \, \mathrm{d}\lambda,
\label{eq:4}
\end{equation}
with $c_\star(\lambda)$ the transmission spectrum of camera's channel~$\star$. 

We further assume that all the light sources are achromatic, i.e. that
\begin{equation}
\phi(\lambda,\omega) = \phi(\omega)
\label{eq:5}
\end{equation}
(this assumption implies that color will be interpreted in terms of surface's reflectance by our algorithms, rather than in terms of lighting). 

Plugging Equations~\eqref{eq:2},~\eqref{eq:3} and~\eqref{eq:5} into~\eqref{eq:4} yields
\begin{equation}
I_\star = \rho_\star \int_{\mathbb{S}^2} \phi(\omega) \max\{0,\mathbf{s}(\omega)^\top \mathbf{n}_{z,\nabla z}\} \,\mathrm{d}\omega,
\label{eq:6}
\end{equation}
with 
\begin{equation}
\rho_\star := \dfrac{\gamma \beta \cos^4 \alpha}{\pi} \int_{\mathbb{R}^+} c_\star(\lambda) \rho(\lambda) \, \mathrm{d}\lambda
\label{eq:7}
\end{equation}
the ``albedo'', relatively to channel $\star$ (note that $\rho_\star$ does not characterize the surface, since it depends upon the sensor's response, its aperture and magnification, etc.).

Next, we approximate the integral in~\eqref{eq:6} using spherical harmonics~\cite{Basri2003,Ramamoorthi2001}. In this work we consider the first-order case, which already captures more than $85\%$ of natural illumination~\cite{Frolova2004}, and leave the extension to second-order spherical harmonics as future work. The spherical harmonics approximation reads
\begin{equation}
 \int_{\mathbb{S}^2} \  \phi(\omega) \max\{0,\mathbf{s}(\omega)^\top \mathbf{n}_{z,\nabla z}\} \,\mathrm{d}\omega \approx \l^\top \m_\zdz 
\label{eq:8}
\end{equation} 
with $\l \in \R^4$ the achromatic ``light vector'' (which is the same for all pixels), and 
\begin{equation} 
\m_\zdz := \begin{bmatrix}
\n_\zdz \\ 
1
\end{bmatrix}
\label{eq:9}
\end{equation}
a geometric vector depending upon the surface normals.

Plugging~\eqref{eq:8} into~\eqref{eq:6}, we obtain
\begin{equation}
I_\star = \rho_\star \, \l^\top \m_\zdz,\quad \star \in \{R,G,B\}.
\label{eq:10}
\end{equation}
Denoting 
\begin{equation}
\I := \begin{bmatrix}
I_R \\
I_G \\
I_B
\end{bmatrix}
\quad\text{and}\quad
\mrho := \begin{bmatrix}
\rho_R \\
\rho_G \\
\rho_B
\end{bmatrix},
\label{eq:11}
\end{equation}
and assuming that~\eqref{eq:10} is satisfied up to additive noise, we eventually obtain the RGB image formation model (Eq. (3) in the paper)  by plugging together the three equations in~\eqref{eq:10}:
\begin{equation}
\I = 
\l^\top \m_\zdz \, \mrho + \metai,
\label{eq:12}
\end{equation}
with $\metai$ the realisation of a stochastic process.

\subsection{Ambiguities in Depth Super-resolution and Shape-from-shading}

\begin{figure}[!th]
\centering
\def\svgwidth{.9\linewidth}
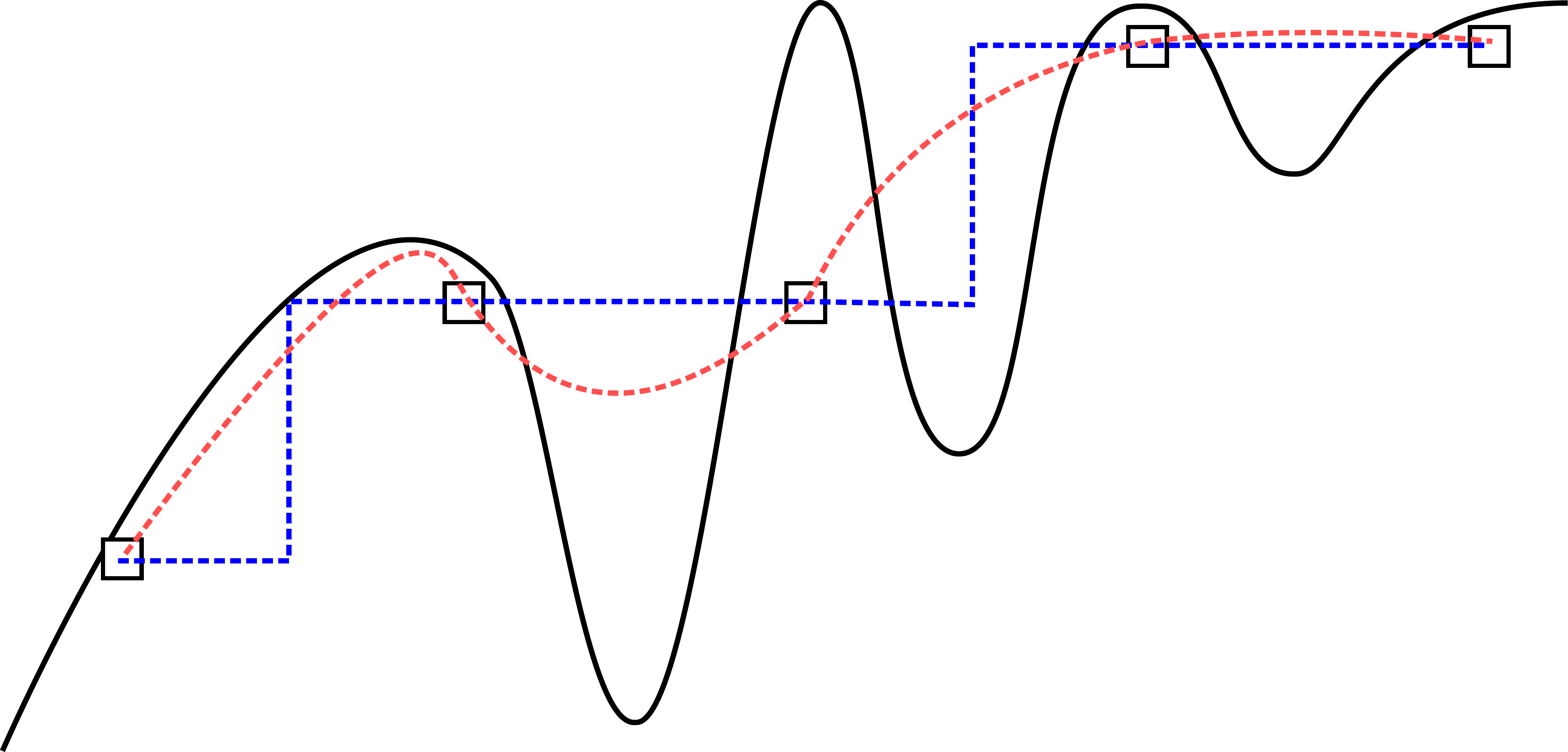
\caption{There exist infinitely many ways (dashed lines) to interpolate between low-resolution depth samples (rectangles). Our disambiguation strategy builds upon shape-from-shading applied to the companion high-resolution color image (cf. Figure~\ref{fig:2}), in order to resurrect the fine-scale geometric details of the genuine surface (solid line).}
\label{fig:1}
\end{figure}

\begin{figure}[!ht]
\centering
\def\svgwidth{.9\linewidth}
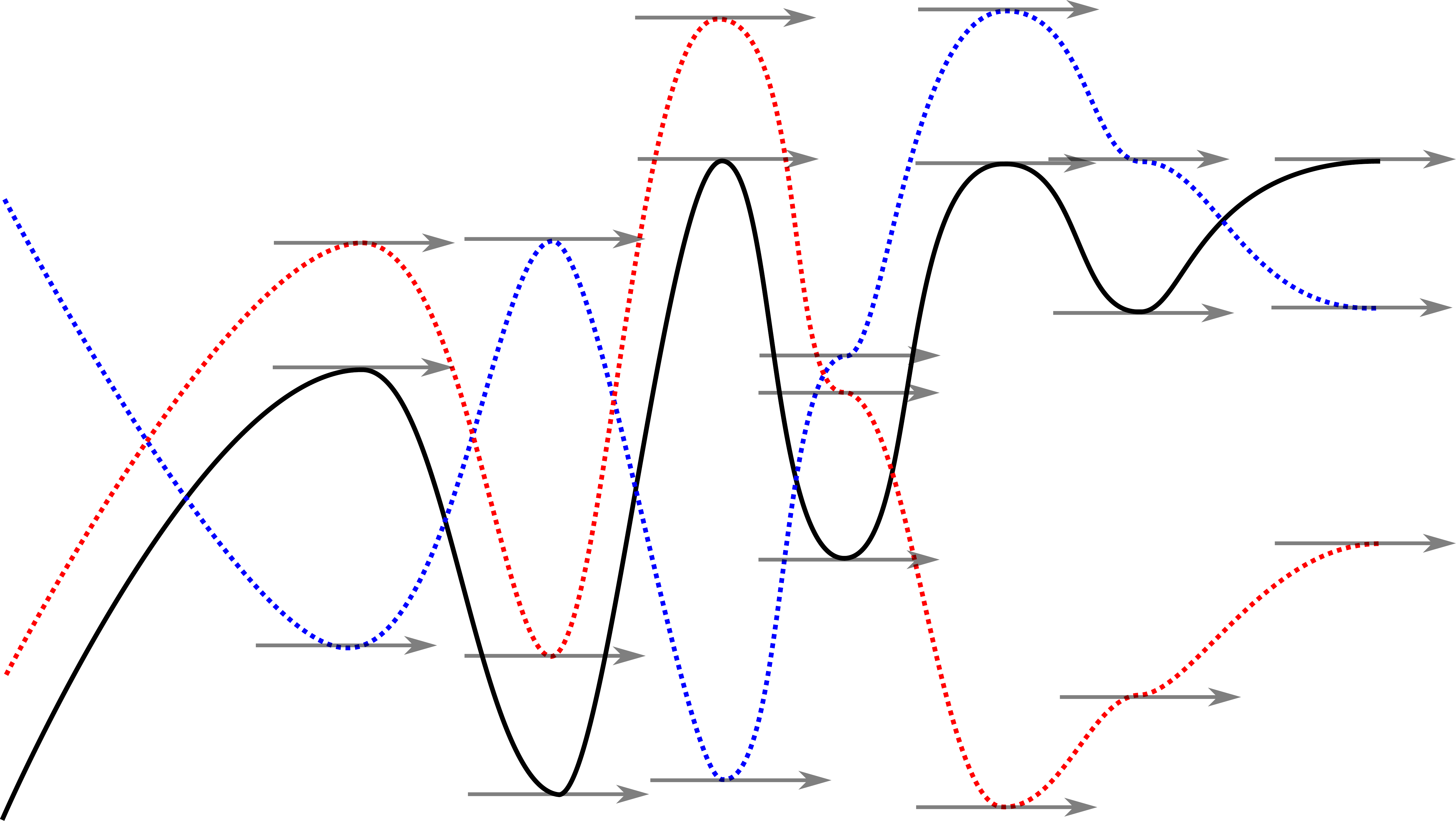
\caption{Shape-from-shading suffers from the concave $/$ convex ambiguity: the genuine surface (solid line) and both the surfaces depicted by dashed lines produce the same image, if lit and viewed from above. We put forward  low-resolution depth clues (cf. Figure~\ref{fig:1}) for disambiguation.}
\label{fig:2}
\end{figure}

This subsection illustrates the ambiguities arising in depth super-resolution and in photometric 3D-reconstruction, in order to visually motivate the choice of their joint solving. As can be seen in Figure~\ref{fig:1}, in super-resolution high-frequency geometric clues are missing and thus there exist infinitely many ways to interpolate between low-resolution samples. On the contrary, shape-from-shading suffers from the concave $/$ convex ambiguity: though the surface orientation is unambiguous in critical points (arrows in Figure~\ref{fig:2}), two such singular points may be connected either by ``going up'' or by ``going down''. Therefore, it seems reasonable to rely on high-frequency photometric clues to disambiguate depth super-resolution, and on low-frequency geometric clues to disambiguate photometric 3D-reconstruction. 

\subsection{Generalities on Reflectance Learning-based Depth Super-resolution}

We now illustrate the creation of the training dataset and the network's architecture, and justify why we focused on a particular class of objects in the learning-based approach.

%
%
%

\begin{figure}[!ht]
	\centering	
	\begin{tabular}{cc}
	\includegraphics[width=.46\linewidth]{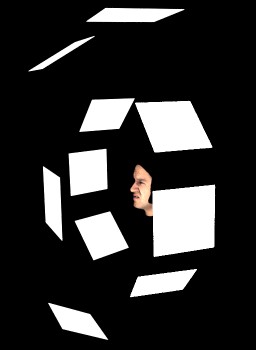} &
	\includegraphics[width=.46\linewidth]{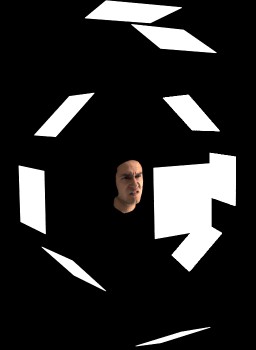}	
	\end{tabular}
	\caption{Rendering of synthetic faces for generating training data. The white planes represent switchable extended light sources, which are independently controlled to create multiple illumination conditions. Multiple images can then be captured under different illumination and viewing angles. }
	\label{fig:lightSources}
\end{figure}

Figure~\ref{fig:lightSources} illustrates the generation of training data. We consider ground truth geometry and reflectance of various human faces from the ICT-3DRFE database~\cite{Database3drfe2011}. A rendering software is used to generate multiple images of these faces under different viewing and lighting scenarios. Lighting variations are created by turning off and on several extended sources, emulating usual indoor lighting conditions. 

\begin{figure}[ht]
	\centering
	\includegraphics[width=.45\textwidth]{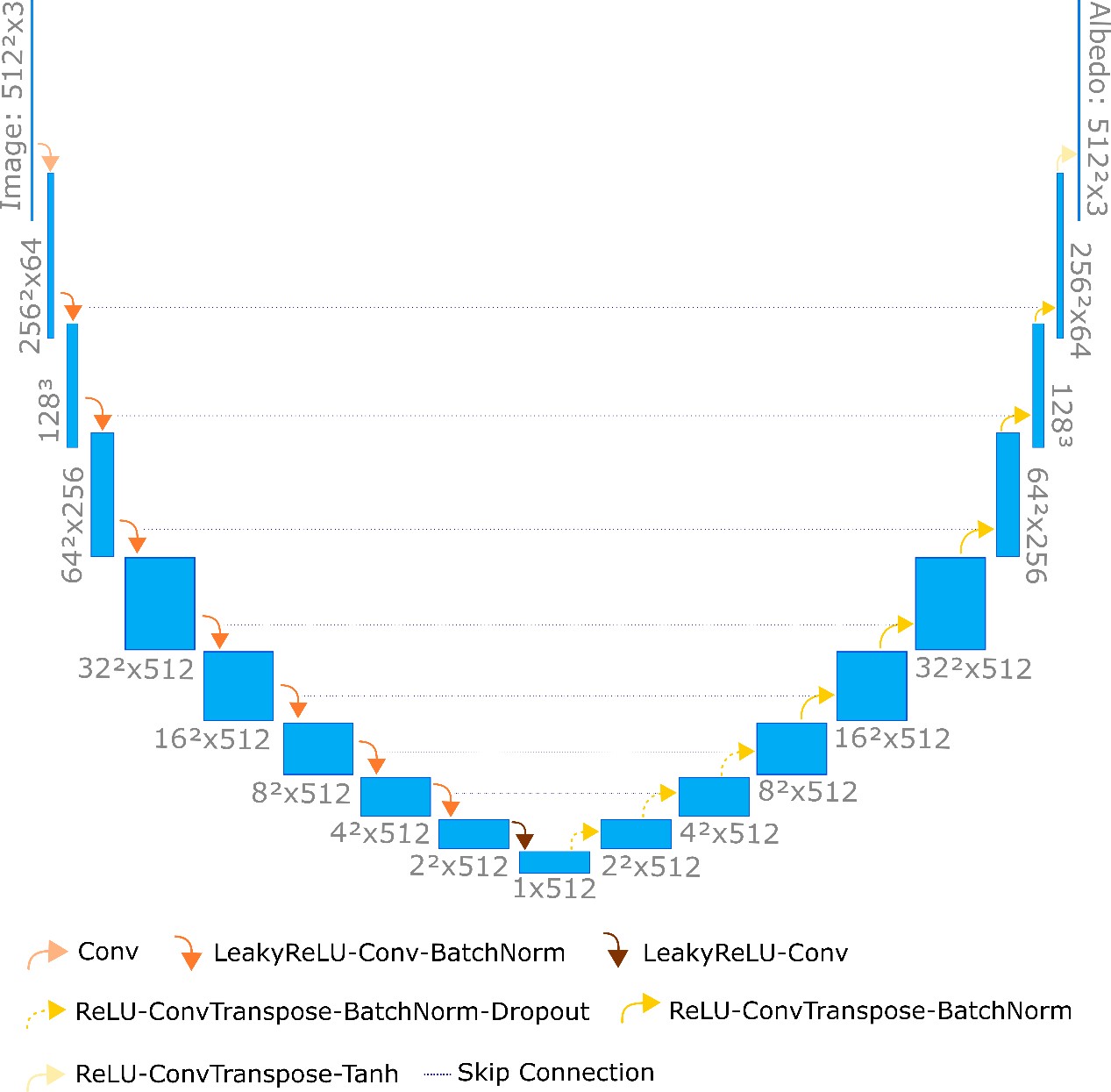}
	\caption{The U-Net Architecture used for albedo estimation. The top two layers are the input and output, respectively. The arrows' color represent the operations of the other hidden layers. Skip connections (dotted lines) concatenate the left and right layers.}
	\label{fig:network}
\end{figure}

Figure~\ref{fig:network} illustrates the architecture of the neural network. It is a U-Net architecture comprising an initial convolution layer of kernel size 4, stride 2 and padding 1; after which there are repeated blocks of 8 ReLU-Conv-BatchNorm layers. This results in downsampling of a 512x512 resolution image to a 1x512 vector at the bottleneck of the ``U''. Then, the 1D array is upsampled to input resolution with multiple ReLU-Transpose Convolution-BatchNorm layers. Dropout is also used in a few layers to allow for randomness while learning the mapping from input images to albedo maps. Finally, the L1 loss is considered, which favors sharper output compared to the L2 loss.

\begin{figure}[!ht]
  \centering
  \newcommand{\mywidth}{0.155\textwidth}
  \newcolumntype{X}{ >{\centering\arraybackslash} m{\mywidth} }
  \setlength\tabcolsep{0.1pt} 
  \def\arraystretch{1}
  \begin{tabular}{XXX}
    Input Image & \cite{Takuya2015} & {\vspace*{2em}\cite{Jian2017}} \\[2em]
    \includegraphics[width=\mywidth]{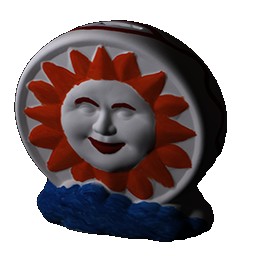}&
		\includegraphics[width=\mywidth]{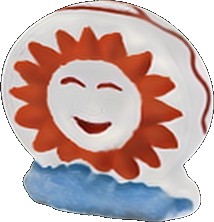}&
		\vspace*{1em}\includegraphics[width=\mywidth]{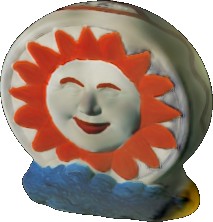}\\[1em]
		\includegraphics[width=\mywidth]{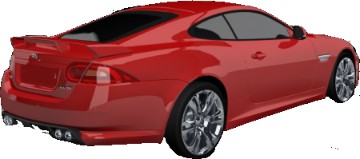}&
		\includegraphics[width=\mywidth]{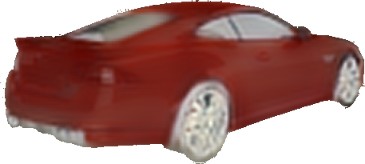}&
		\vspace*{1em}\includegraphics[width=\mywidth]{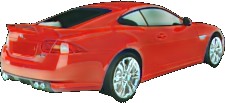}\\[1em]
		\includegraphics[width=\mywidth]{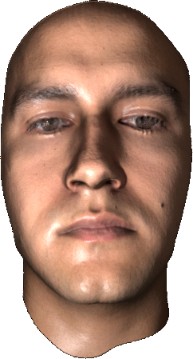}&
		\includegraphics[width=\mywidth]{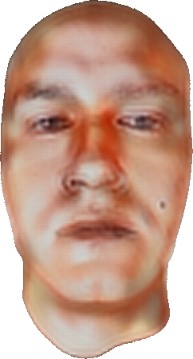}&
		\vspace*{1em}\includegraphics[width=\mywidth]{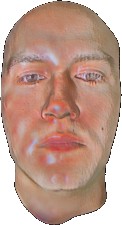}\\[1em]
    \includegraphics[width=\mywidth]{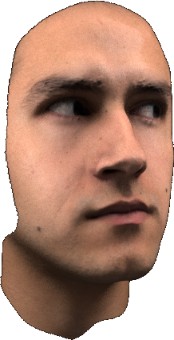}&
		\includegraphics[width=\mywidth]{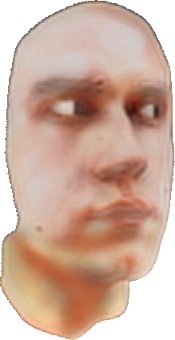}&
		\includegraphics[width=\mywidth]{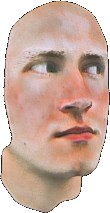}
  \end{tabular}
	\caption{Learning-based albedo estimation applied to an object from the MIT database (first row), a car from the ShapeNet dataset (second row), and two images of human faces we generated with a renderer using the ICT-3DRFE database~\cite{Database3drfe2011}. This illustrates the lack of inter-class generalisation inherent to learning-based techniques: the approach from \cite{Takuya2015}, trained on the MIT dataset, fails on the ShapeNet car and on faces, and the one from~\cite{Jian2017}, trained on the ShapeNet dataset, fails on the MIT object and on faces: in both cases albedo estimation is not satisfactory since the objects do not resemble the training data.}
	\label{fig:compareIntrinsicDecom1}
\end{figure}

Eventually, Figure~\ref{fig:compareIntrinsicDecom1} illustrates the lack of inter-class generalisation which is inherent to learning-based methods. For instance, the approach of \cite{Takuya2015} (trained on Sintel~\cite{Sintel2012} and MIT~\cite{MIT2009} datasets) performs well on the MIT object but poorly on the ShapeNet car image, because such an object was not present in the learning database. For the same reason, the alternative approach of \cite{Jian2017} (trained on ShapeNet objects~\cite{Shapenet2015}) performs well on the ShapeNet car but fails on the MIT object, and both approaches fail on the face image since the latter resembles none of the training data. Due to this lack of inter-class generalisation, we choose to focus in our approach on the specific class of human faces. 

\begin{figure*}[!ht]
  \centering
  \newcommand{\mywidth}{0.22\textwidth}
  \newcommand{\mywidthlr}{0.13\textwidth}
  \newcolumntype{C}{ >{\centering\arraybackslash} m{0.02\textwidth} }
  \newcolumntype{X}{ >{\centering\arraybackslash} m{\mywidth} }
  \begin{tabular}{CXXXX}
    & Lucy\cite{Levoy2005data} & Thai Statue\cite{Levoy2005data} & Armadillo\cite{Levoy2005data} & Joyful Yell\cite{Bendansie2015} \\
    $\I$ &
    \includegraphics[width=\mywidth]{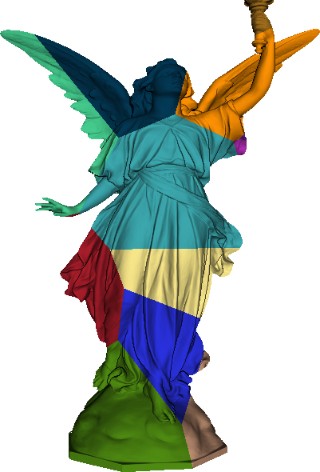}&
    \includegraphics[width=\mywidth]{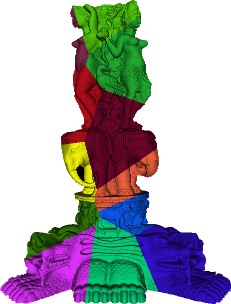}&
    \includegraphics[width=\mywidth]{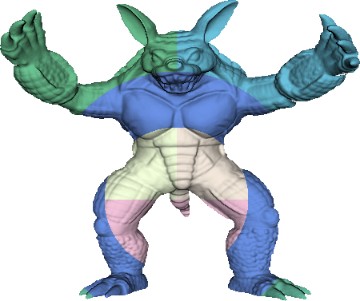}&
    \includegraphics[width=\mywidth]{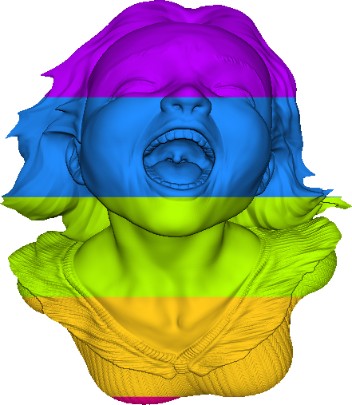} \\
    & ``voronoi'' albedo & ``voronoi'' albedo & ``rectcircle'' albedo & ``bar'' albedo\\\\
    $\zz$ &
    \includegraphics[width=\mywidthlr]{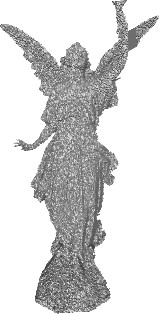}&
    \includegraphics[width=\mywidthlr]{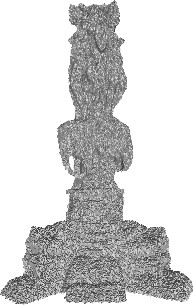}&
    \includegraphics[width=\mywidthlr]{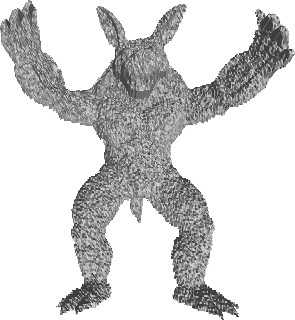}&
    \includegraphics[width=\mywidthlr]{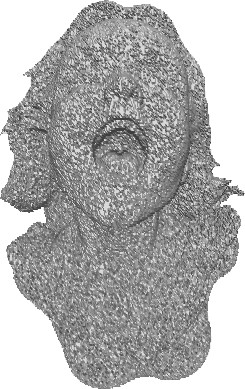}
  \end{tabular}
  \caption{Illustration of synthetic data used for evaluation of the single-shot approach based on shape-from-shading. High-resolution RGB images $\I$, of size $480\times640$, are generated using high-resolution ground truth depth and reflectance maps, and adding noise. Low-resolution depth maps $\zz$ are created by downsampling the ground truth depth maps with scaling factors of $8$, $4$ and $2$ (the second row shows the low-resolution depth maps with a scaling factor of $2$), and adding noise.}
  \label{fig:synthetic_data_sfs}
\end{figure*}

\section{Evaluation of the Single-shot Approach based on Shape-from-shading}
\label{sec:supp_3}

\subsection{Creation of the Synthetic Data}

Figure~\ref{fig:synthetic_data_sfs} illustrates the synthetic data used for evaluation, which is generated using four different 3D-shapes (``Lucy'', ``Thai Statue'', ``Armadillo'' and ``Joyful Yell''), each of them rendered using three different albedo maps (``voronoi'', ``rectcircle'' and ``bar'') and three different scaling factors (2, 4 and 8) for the low-resolution depth image.  To this end, 3D-meshes are rendered into high-resolution ground truth depth maps of size $480\times640$, which are then downsampled. Then, additive zero-mean Gaussian noise with standard deviation $10^{-4}$ times the squared original depth value (consistently with real-world measurements from \cite{Khoshelham2012}) is added to the low-resolution depth maps, which are eventually quantised. High-resolution RGB images are rendered from the ground truth depth map using the first-order spherical harmonics model with $\l=[0,0,-1,0.2]^\top$ using the three different high-resolution reflectance maps, and an additive zero-mean Gaussian noise with standard deviation $1\%$ the maximum intensity is eventually added to the RGB images.

\subsection{Tuning the Hyper-parameters}

\begin{figure*}[!ht]
  \centering
  \newcolumntype{C}{>{\centering\arraybackslash} m{0.005\textwidth} }
  \newcolumntype{X}{>{\centering\arraybackslash} m{0.31\textwidth} }
  \begin{tabular}{CXXX}
    & $\mu$ & $\nu$ & $\lambda$ \\
    \rotatebox{90}{RMSE (depth)} &
    \includegraphics[width=0.225\textwidth]{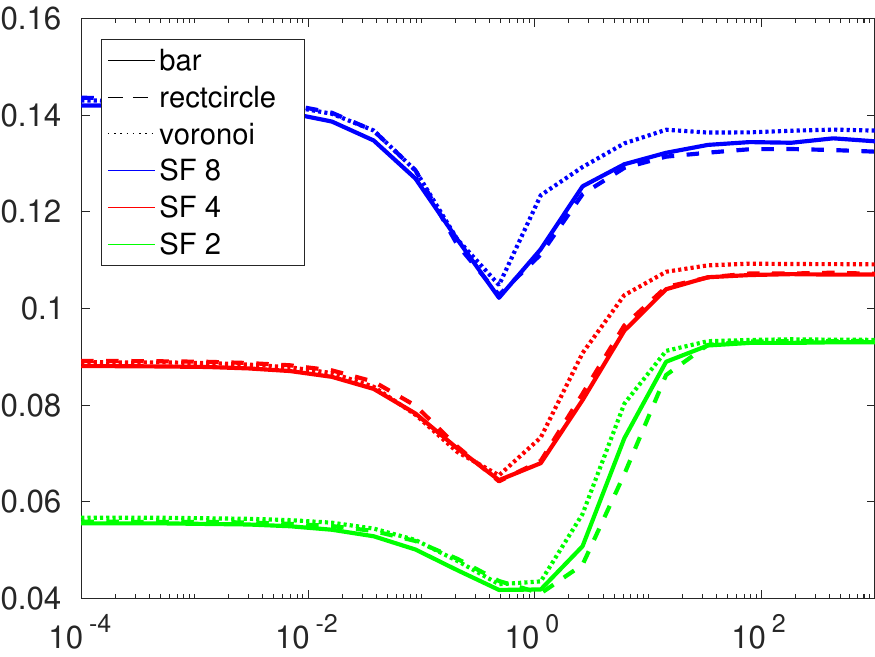}&
    \includegraphics[width=0.225\textwidth]{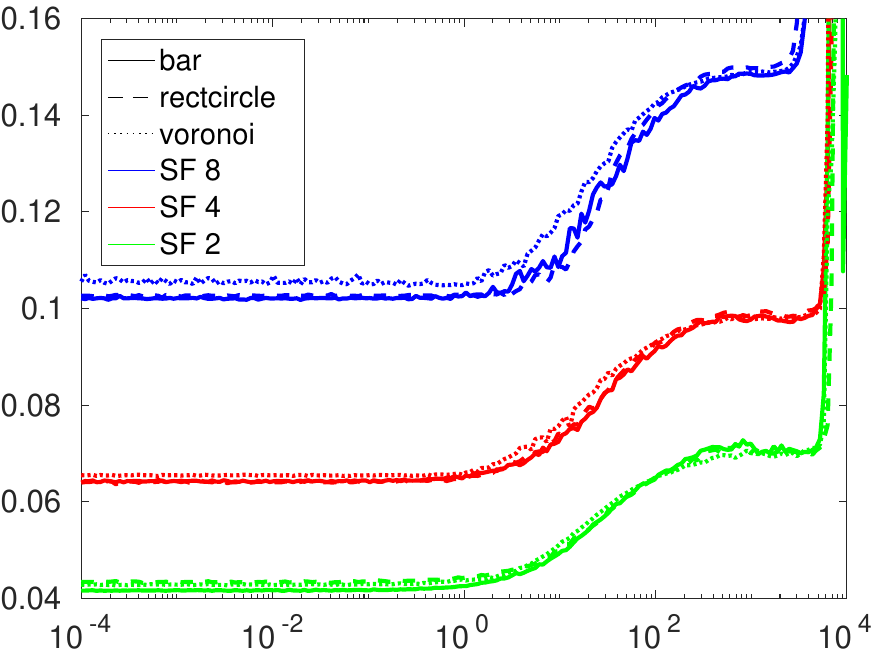}&
    \includegraphics[width=0.225\textwidth]{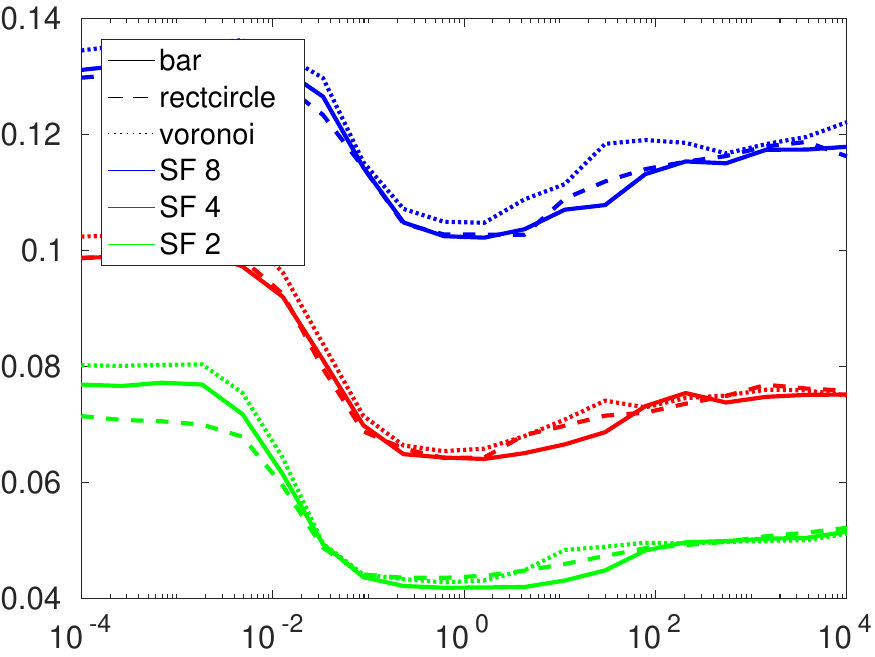}\\
    \rotatebox{90}{MAE (depth)} &
    \includegraphics[width=0.225\textwidth]{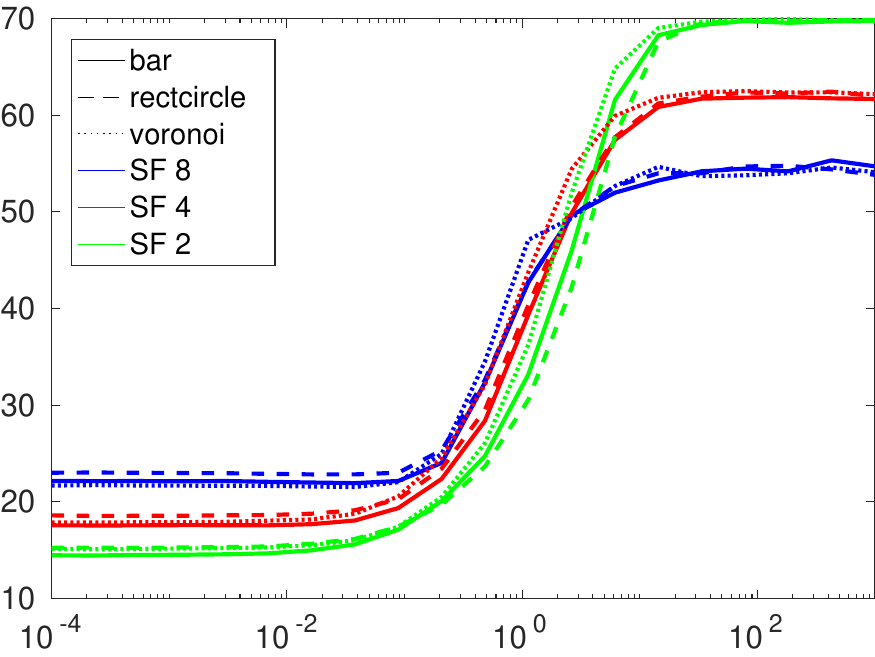}&
    \includegraphics[width=0.225\textwidth]{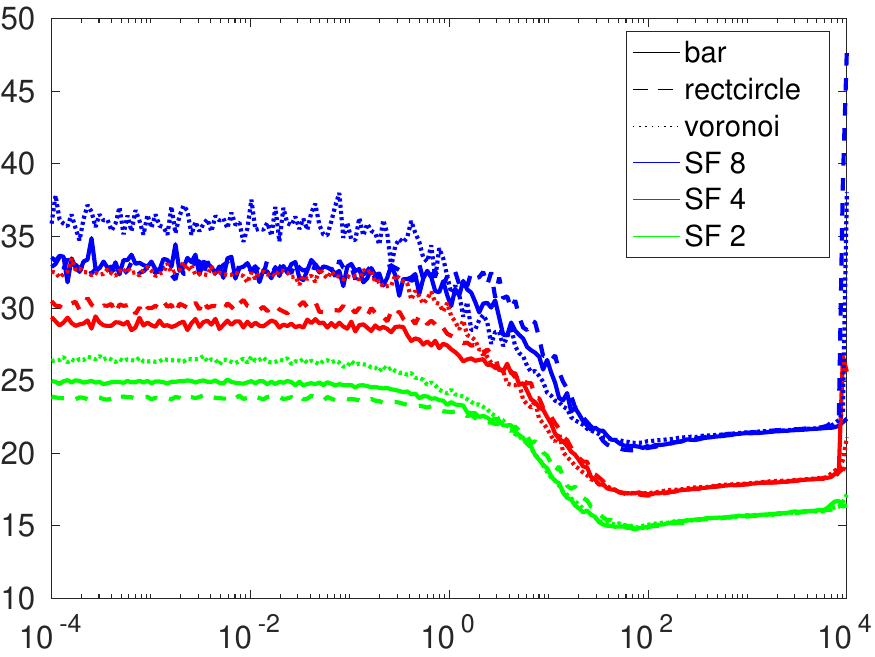}&
    \includegraphics[width=0.225\textwidth]{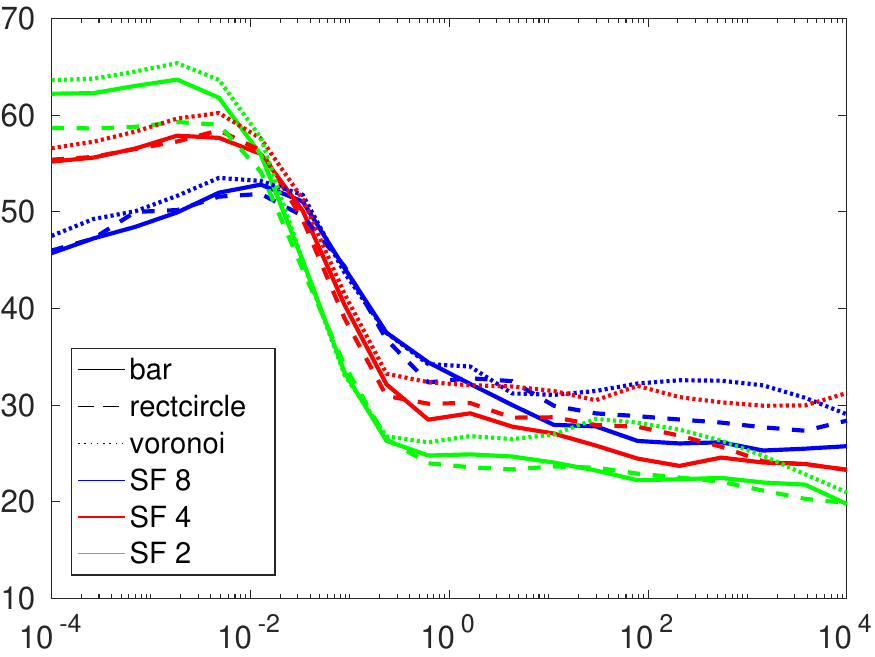}\\
    \rotatebox{90}{RMSE (albedo)} &
    \includegraphics[width=0.225\textwidth]{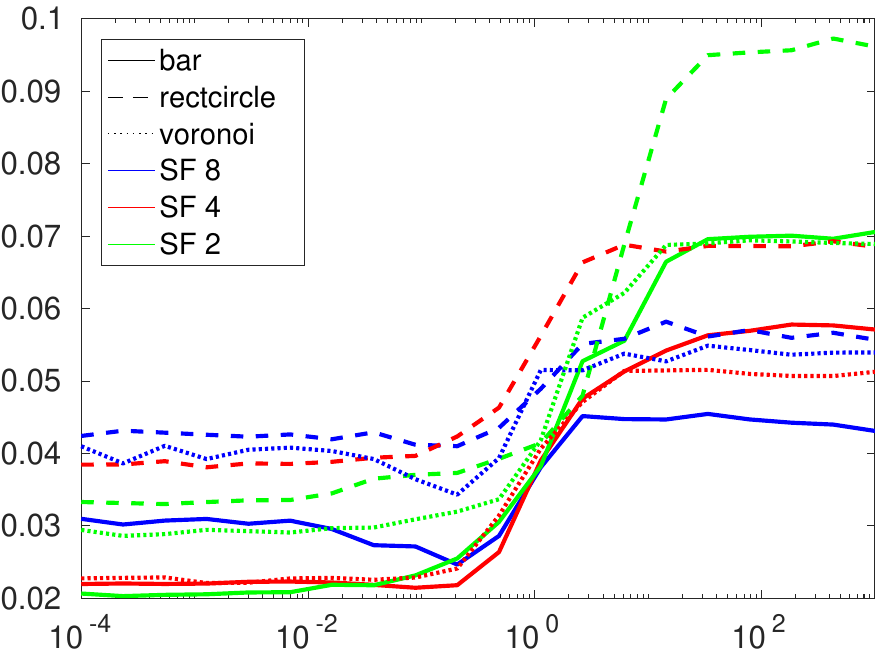}&
    \includegraphics[width=0.225\textwidth]{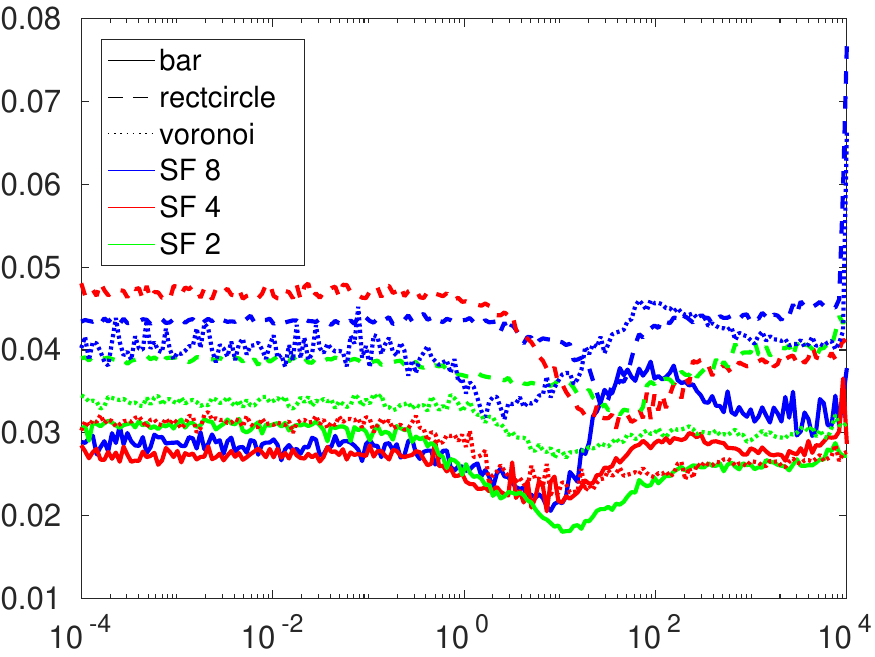}&
    \includegraphics[width=0.225\textwidth]{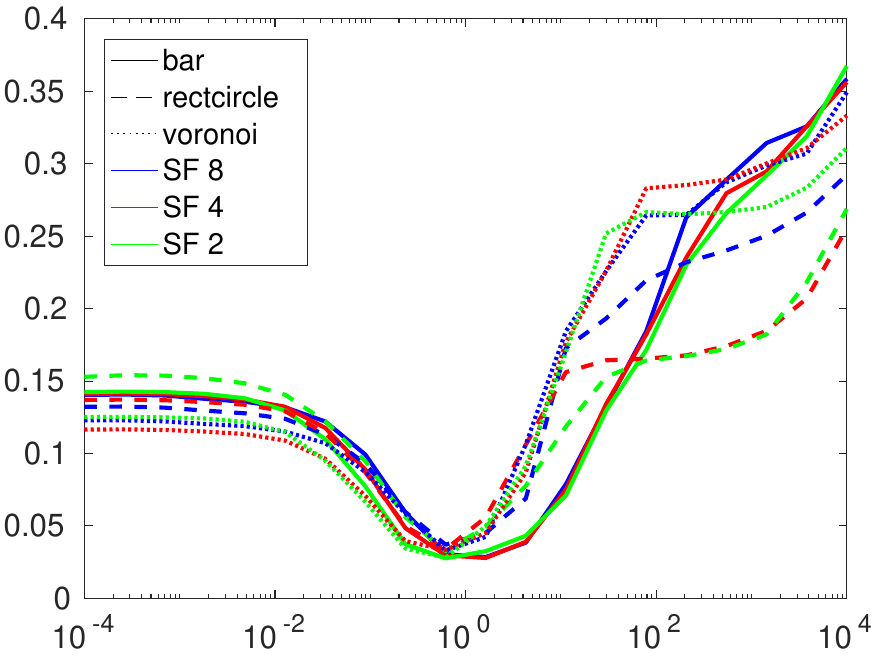}\\
  \end{tabular}
  \caption{Impact of the parameters $(\mu,\nu,\lambda)$ on the accuracy of the albedo and depth estimates. The accuracy of the albedo is evaluated by the root mean square error (RMSE), and that of the depth by the RMSE and the mean angular error (MAE). Based on these experiments, the set of hyper-parameters $(\mu,\nu,\lambda) = (0.1, 0.7, 1)$ is selected.}
  \label{fig:parameter_evaluation_sfs} 
\end{figure*}

In Figure~\ref{fig:parameter_evaluation_sfs}, we use the ``Joyful Yell'' dataset from Figure~\ref{fig:synthetic_data_sfs} in order to determine appropriate values for the hyper-parameters $(\mu,\nu,\lambda)$. For quantitative evaluation, we consider the root mean squared error (RMSE) on the estimated depth and reflectance maps, and the mean angular error (MAE) on surface normals. To select an appropriate set of values for them, we initially set $\mu=0.5$, $\nu=0.01$ and $\lambda=1$. We then evaluate the impact of each parameter by varying it while keeping the remaining two fixed. As could be expected, large values of $\mu$ force the depth map to keep close to the noisy input, while small values make the depth prior less important so not capable of disambiguating shape-from-shading. Inbetween, the range $\mu \in [10^{-1},10]$ seems to provide appropriate results. As for $\nu$, large values produce over-smoothed results and small ones result in slightly noisier depth estimates, although the albedo estimate seems unaffected by this choice. Overall, the range $\nu \in [0.5,10^2]$ seems appropriate. The parameter $\lambda$ strongly impacts both the resulting albedo and depth: too small (resp., high) values for $\lambda$ result in over (resp., under)-segmentation problems, and in both cases shading information gets propagated to the albedo. We found $\lambda\in[10^{-1},10]$ to be a reasonable choice.  Overall, we opted for $(\mu,\nu,\lambda) = (0.1, 0.7, 1)$.

\subsection{Comparison against the State-of-the-art on the Synthetic Dataset}

\begin{figure}[!ht]
  \centering
  \newcommand{\mywidth}{0.065\textwidth}
  \newcommand{\mywidthtwo}{0.1\textwidth}  
  \newcolumntype{C}{ >{\centering\arraybackslash} m{0.02\textwidth} }
  \newcolumntype{X}{ >{\centering\arraybackslash} m{\mywidthtwo} }
  \setlength\tabcolsep{2pt} 
  \begin{tabular}{CXXXX}
    & Lucy ``voronoi'' & Thai Statue ``voronoi'' & Armadillo ``rectcircle'' & \vspace*{.25em}Joyful Yell ``bar'' \\[.25em]
    \cite{Yang2007} &
    \includegraphics[width=\mywidth]{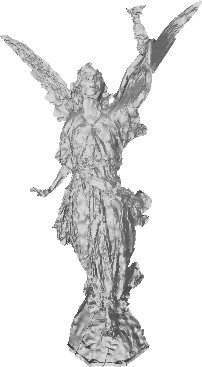}&
    \includegraphics[width=\mywidth]{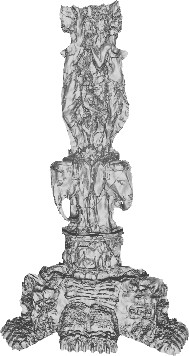}\qquad&\qquad
    \includegraphics[width=\mywidth]{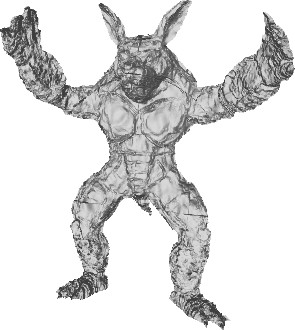}&
    \includegraphics[width=\mywidth]{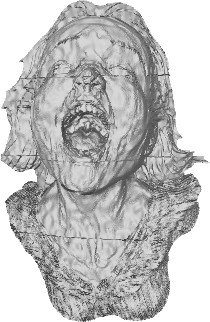} \\
    \cite{Xie2016} &
    \includegraphics[width=\mywidth]{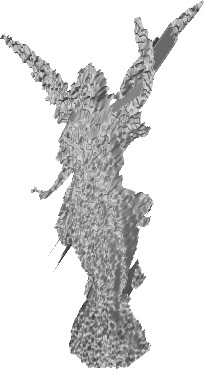}&
    \includegraphics[width=\mywidth]{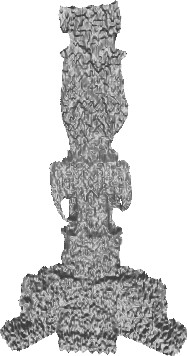}&
    \includegraphics[width=\mywidth]{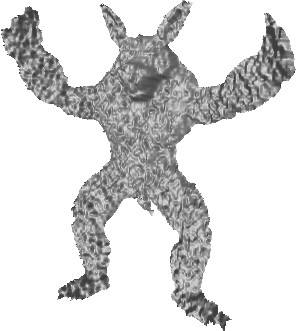}&
    \includegraphics[width=\mywidth]{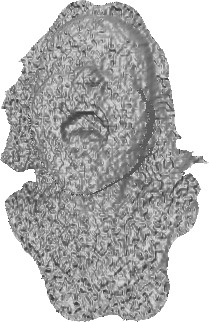} \\
    \cite{Or-El2015} &
    \includegraphics[width=0.07\textwidth]{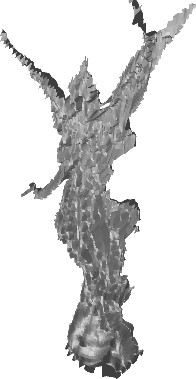}&
    \includegraphics[width=0.07\textwidth]{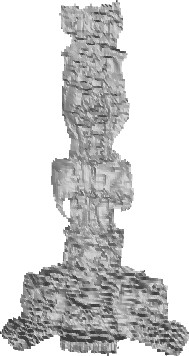}&
    \includegraphics[width=0.07\textwidth]{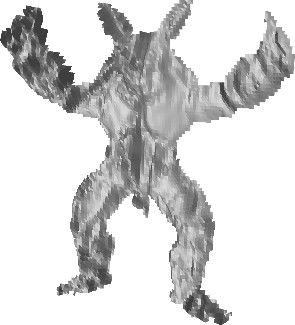}&
    \includegraphics[width=0.07\textwidth]{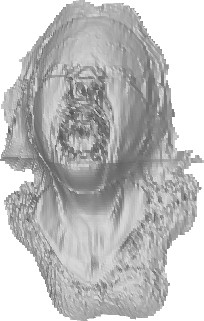} \\
    \rotatebox{90}{Ours} &
    \includegraphics[width=\mywidth]{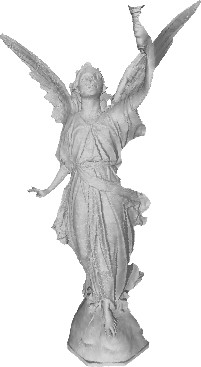}&
    \includegraphics[width=\mywidth]{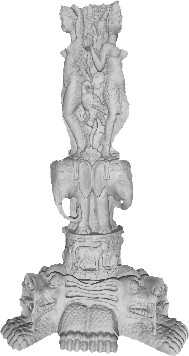}&
    \includegraphics[width=\mywidth]{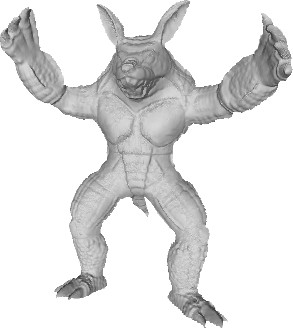}&
    \includegraphics[width=\mywidth]{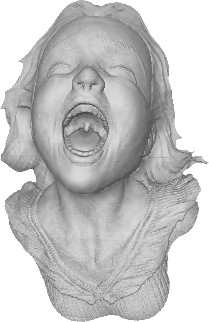}
  \end{tabular}
  \caption{Qualitative comparison between our single-shot results and state-of-the-art's ones (the scaling factor is~$4$).}
  \label{fig:synthetic_comparison_sfs}
\end{figure}

Next, we compare the results obtained by our single-shot approach against the state-of-the-art, on the synthetic dataset from Figure~\ref{fig:synthetic_data_sfs}. We consider two alternative depth super-resolution methods: the image-based one from~\cite{Yang2007}, and the learning-based one from \cite{Xie2016} (since the authors only provide trained data for a factor of $4$, this method was evaluated only for this factor). To emphasise the interest of joint shape-from-shading and depth super-resolution over shading-based depth refinement using downsampled images, we also consider \cite{Or-El2015}. Qualitative results are presented in Figure~\ref{fig:synthetic_comparison_sfs}, and quantitative ones in Table~\ref{tab:synthetic_comparison_sfs}. As can be seen, our method systematically overcomes the competitors in terms of MAE, which indicates that high-frequency geometric details are better recovered. The RMSE on depth rather evaluates the overall (low-frequency) fit to ground truth, and for this metric our results are comparable with~\cite{Yang2007}, which achieves the best results. 

Interestingly, for scaling factors of $4$ and $8$, our approach seems less accurate than~\cite{Yang2007} in terms of RMSE. However, Figure~\ref{fig:synthetic_comparison_sfs} clearly shows that our results are significantly better: we thus believe that only the order of magnitude of the RMSE is meaningful, yet comparison using this metric might not really indicate which method is the best, and MAE should be preferred for this purpose. A more thorough discussion on the relevance of RMSE for evaluation can be found in~\cite{Wang2009}.

\begin{table*}[!ht]                                                                                                                                                                                                                                                                       
\centering                                                                                                                                                                                                                                                                           
\begin{tabular}{|c|c|c|cc|cc|cc|cc|}                                                                                                                                                                                                                                               
\hline                                                                                                                                                                                                                                                                               
\multirow{2}{*}{Albedo} & \multirow{2}{*}{3D-shape} & \multirow{2}{*}{SF} & \multicolumn{2}{c|}{\cite{Yang2007}} & \multicolumn{2}{c|}{\cite{Xie2016}} & \multicolumn{2}{c|}{\cite{Or-El2015}} & \multicolumn{2}{c|}{Ours} \\\cline{4-11}                                                                                                                     
& & & RMSE & MAE & RMSE & MAE & RMSE & MAE & RMSE & MAE \\\hline                                                                                                                     
\hline                                                                                                                                                                                                                  
\multirow{12}{*}{bar}& & 2 & 0.043643 & 38.6274 & -- & -- & 0.41993 & 67.2643 & \textbf{0.034655} & \textbf{16.7496} \\                                                                                                                                                                                    
&Armadillo & 4 & \textbf{0.051558} & 42.2277 & 0.17865 & 45.6972 & 0.45139 & 66.2117 & 0.054679 & \textbf{19.0314} \\                                                                                                                                                             
& & 8 & \textbf{0.072466} & 43.5649 & -- & -- & 0.58837 & 69.3262 & 0.091263 & \textbf{20.8836} \\                                                                                                                                                                                    
\cline{2-11}
& & 2 & 0.05089 & 29.1719 & -- & -- & 0.1721 & 47.4836 & \textbf{0.050694} & \textbf{16.7414} \\                                                                                                                                                                                      
&Joyful Yell & 4 & \textbf{0.066517} & 33.0843 & 0.084094 & 42.611 & 0.22867 & 32.9784 & 0.079271 & \textbf{19.0695} \\                                                                                                                                                                   
& & 8 & \textbf{0.10212} & 36.565 & -- & -- & 0.37923 & 31.2894 & 0.128 & \textbf{21.9886} \\                                                                                                                                                                                         
\cline{2-11}
& & 2 & 0.057987 & 39.4714 & -- & -- & 0.21309 & 66.5525 & \textbf{0.053989} & \textbf{25.0955} \\                                                                                                                                                                                    
&Lucy & 4 & \textbf{0.068502} & 42.7169 & 0.50472 & 47.605 & 0.34091 & 69.2566 & 0.081005 & \textbf{28.3044} \\                                                                                                                                                                   
& & 8 & \textbf{0.098713} & 46.4775 & -- & -- & 0.43619 & 59.5434 & 0.1195 & \textbf{30.1058} \\                                                                                                                                                                                      
\cline{2-11}
& & 2 & 0.040821 & 42.8976 & -- & -- & 0.12948 & 63.06 & \textbf{0.035736} & \textbf{23.9147} \\                                                                                                                                                                                      
&Thai Statue & 4 & \textbf{0.050296} & 47.1017 & 0.22363 & 49.9553 & 0.15489 & 54.6139 & 0.057313 & \textbf{28.492} \\                                                                                                                                                                 
& & 8 & \textbf{0.066515} & 49.8604 & -- & -- & 0.22835 & 56.4247 & 0.087054 & \textbf{31.65} \\                                                                                                                                                                                      
\hline
\multirow{12}{*}{rectcircle} & & 2 & 0.044026 & 39.108 & -- & -- & 0.34323 & 70.8526 & \textbf{0.03494} & \textbf{18.4909} \\                                                                                                                                                                                      
&Armadillo & 4 & \textbf{0.052115} & 43.3175 & 0.17782 & 45.6324 & 0.2338 & 50.6919 & 0.056727 & \textbf{18.8487} \\                                                                                                                                                       
& & 8 & \textbf{0.069467} & 45.4735 & -- & -- & 0.61917 & 70.9363 & 0.09155 & \textbf{21.9959} \\                                                                                                                                                                                     
\cline{2-11}
& & 2 & \textbf{0.051296} & 30.7886 & -- & -- & 0.14841 & 41.5424 & 0.05226 & \textbf{17.134} \\                                                                                                                                                                                      
&Joyful Yell & 4 & \textbf{0.066911} & 33.3 & 0.10328 & 42.7531 & 0.28311 & 51.0665 & 0.080387 & \textbf{19.8717} \\                                                                                                                                                               
& & 8 & \textbf{0.10201} & 36.2961 & -- & -- & 0.39518 & 35.4817 & 0.1281 & \textbf{22.8027} \\                                                                                                                                                                                       
\cline{2-11}
& & 2 & 0.058495 & 39.7374 & -- & -- & 0.19546 & 64.8212 & \textbf{0.054383} & \textbf{24.8427} \\                                                                                                                                                                                    
&Lucy & 4 & \textbf{0.069893} & 43.9016 & 0.50464 & 48.1068 & 0.23235 & 53.2901 & 0.082547 & \textbf{28.7517} \\                                                                                                                                                           
& & 8 & \textbf{0.099402} & 46.3739 & -- & -- & 0.39583 & 64.3269 & 0.12283 & \textbf{29.1531} \\                                                                                                                                                                                     
\cline{2-11}
& & 2 & 0.039821 & 40.6144 & -- & -- & 0.11355 & 58.2254 & \textbf{0.036845} & \textbf{23.9036} \\                                                                                                                                                                                    
&Thai Statue & 4 & \textbf{0.04973} & 46.1154 & 0.20894 & 49.4124 & 0.16749 & 52.9663 & 0.05866 & \textbf{28.155} \\                                                                                                                                                            
& & 8 & \textbf{0.067799} & 50.6515 & -- & -- & 0.21058 & 50.9074 & 0.094688 & \textbf{33.5308} \\                                                                                                                                                                                    
\hline
\multirow{12}{*}{voronoi}& & 2 & 0.043635 & 38.9089 & -- & -- & 0.33005 & 69.3157 & \textbf{0.034751} & \textbf{17.6873} \\                                                                                                                                                                                    
&Armadillo & 4 & \textbf{0.051989} & 41.57 & 0.17182 & 45.5833 & 0.4407 & 65.5811 & 0.056032 & \textbf{20.168} \\                                                                                                                                                             
& & 8 & \textbf{0.07077} & 43.1987 & -- & -- & 0.50548 & 63.8618 & 0.090708 & \textbf{22.2767} \\                                                                                                                                                                                     
\cline{2-11}
& & 2 & \textbf{0.052002} & 28.7903 & -- & -- & 0.16893 & 47.72 & 0.052429 & \textbf{17.0453} \\                                                                                                                                                                                      
&Joyful Yell & 4 & \textbf{0.066557} & 32.3448 & 0.086394 & 43.1744 & 0.24753 & 39.6569 & 0.079888 & \textbf{19.6512} \\                                                                                                                                                              
& & 8 & \textbf{0.10238} & 35.8017 & -- & -- & 0.47694 & 47.4707 & 0.12916 & \textbf{21.6663} \\                                                                                                                                                                                      
\cline{2-11}
& & 2 & 0.058222 & 36.2327 & -- & -- & 0.29164 & 72.9002 & \textbf{0.054442} & \textbf{26.1333} \\                                                                                                                                                                                    
&Lucy & 4 & \textbf{0.068253} & 40.8878 & 0.5066 & 48.0387 & 0.32955 & 71.1042 & 0.079877 & \textbf{28.4506} \\                                                                                                                                                               
& & 8 & \textbf{0.099838} & 43.7671 & -- & -- & 0.37839 & 57.6856 & 0.11877 & \textbf{29.6331} \\                                                                                                                                                                                     
\cline{2-11}
& & 2 & 0.039872 & 39.6508 & -- & -- & 0.13261 & 65.8352 & \textbf{0.037607} & \textbf{25.6126} \\                                                                                                                                                                                    
&Thai Statue & 4 & \textbf{0.049783} & 45.7178 & 0.22688 & 49.4132 & 0.16533 & 58.3933 & 0.058957 & \textbf{28.6314} \\                                                                                                                                                            
& & 8 & \textbf{0.065577} & 48.7962 & -- & -- & 0.21927 & 49.6711 & 0.091959 & \textbf{32.0347} \\                                                                                                                                                                                    
\hline                                                                                                                                                                                                                                                                       
\hline                                                                                                                                                                                                                                                                               
\multicolumn{2}{|c|}{} & 2 & 0.047458 & 39.0085 & -- & -- & 0.18378 & 65.3282 & \textbf{0.044151} & \textbf{21.1973} \\                                                                                                                                                                                    
\multicolumn{2}{|c|}{Median} & 4 & \textbf{0.059316} & 42.4723 & 0.19379 & 46.6511 & 0.24067 & 53.952 & 0.069114 & \textbf{24.1615} \\                                                                                                                                                                     
\multicolumn{2}{|c|}{} & 8 & \textbf{0.085589} & 44.6203 & -- & -- & 0.3955 & 57.0551 & 0.10673 & \textbf{25.9779} \\                                                                                                                                                                                      
\hline                                                                                                                                                                                                                                                                               
\multicolumn{2}{|c|}{} & 2 & 0.048392 & 36.9999 & -- & -- & 0.22154 & 61.2978 & \textbf{0.044394} & \textbf{21.1126} \\                                                                                                                                                                                    
\multicolumn{2}{|c|}{Mean} & 4 & \textbf{0.059342} & 41.0238 & 0.24812 & 46.4986 & 0.27298 & 55.4842 & 0.068779 & \textbf{23.9521} \\                                                                                                                                                                      
\multicolumn{2}{|c|}{} & 8 & \textbf{0.084754} & 43.9022 & -- & -- & 0.40275 & 54.7438 & 0.1078 & \textbf{26.4768} \\                                                                                                                                                                                      
\hline                                                                                                                                                                                                                                                                               
\end{tabular}                                                                                                                                                                                                                                                                        
\caption{Quantitative comparison between our single-shot results and three state-of-the-art methods, on all the synthetic datasets. Our results are always superior in terms of mean angular error (MAE) and in terms of root mean square error (RMSE) when the scaling factor is $2$. For larger synthetic factors our RMSE values are slightly higher than those from~\cite{Yang2007}, but Figure~\ref{fig:synthetic_comparison_sfs} shows that our results are actually of better quality than the latter, so the RMSE values might not be as relevant as the MAE ones.} 
\label{tab:synthetic_comparison_sfs}                                                                                                                                                                                                                                           
\end{table*}

\subsection{Comparison against the State-of-the-art on a Public Real-world Dataset}

\begin{figure*}[!ht]
  \centering
  \newcommand{\mywidth}{0.12\textwidth}
  \newcommand{\mywidthlr}{0.085\textwidth}
  \newcolumntype{C}{ >{\centering\arraybackslash} m{0.02\textwidth} }
  \newcolumntype{X}{ >{\centering\arraybackslash} m{\mywidth} }
  \newcolumntype{Y}{ >{\centering\arraybackslash} m{\mywidthlr} }
  \setlength\tabcolsep{4pt} 
  \begin{tabular}{CXYXXYXX}
    & $\I$ & $\zz$ & \cite{Yang2007} & \cite{Xie2016} & \cite{Or-El2015} & Ours & Ground truth \\
    \rotatebox{90}{bear} &
    \includegraphics[width=\mywidth]{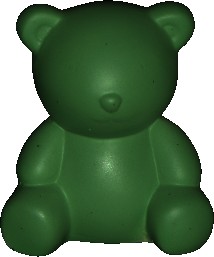}&
    \includegraphics[width=\mywidthlr]{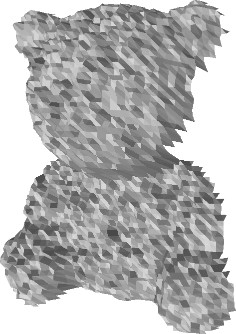}&
    \includegraphics[width=\mywidth]{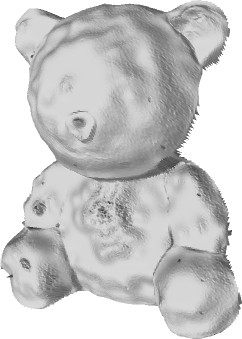}&
    \includegraphics[width=\mywidth]{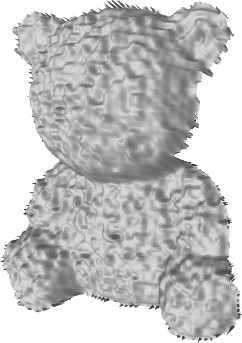}&
    \includegraphics[width=\mywidthlr]{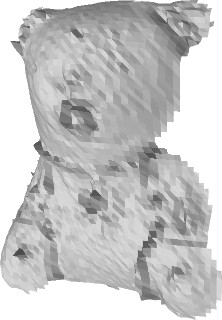}&
    \includegraphics[width=\mywidth]{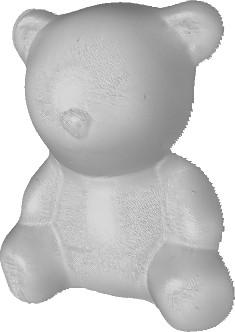}&
    \includegraphics[width=\mywidth]{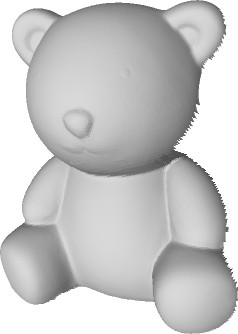} \\
    \rotatebox{90}{buddha} &
    \includegraphics[width=\mywidth]{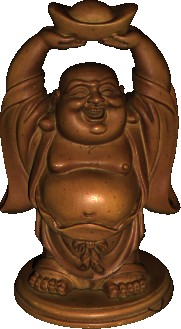}&
    \includegraphics[width=\mywidthlr]{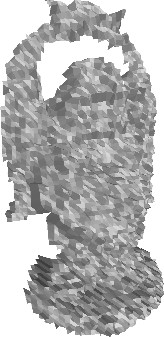}&
    \includegraphics[width=\mywidth]{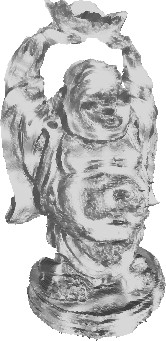}&
    \includegraphics[width=\mywidth]{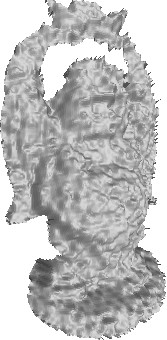}&
    \includegraphics[width=\mywidthlr]{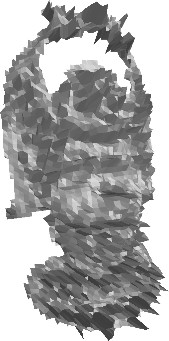}&
    \includegraphics[width=\mywidth]{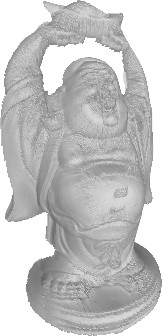}&
    \includegraphics[width=\mywidth]{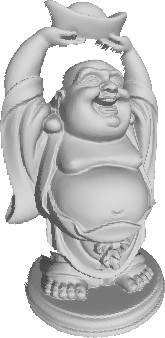} \\
    \rotatebox{90}{cat} &
    \includegraphics[width=\mywidth]{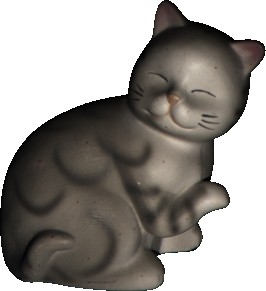}&
    \includegraphics[width=\mywidthlr]{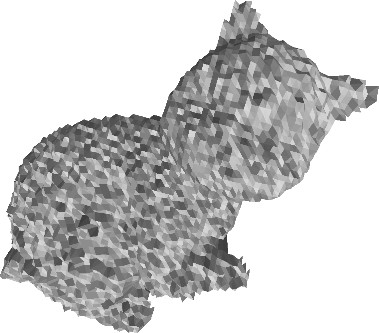}&
    \includegraphics[width=\mywidth]{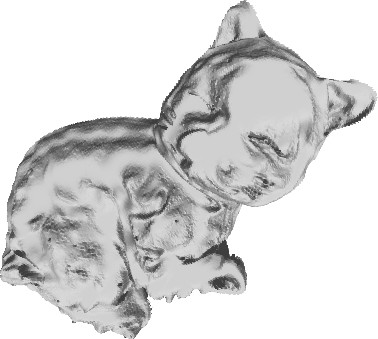}&
    \includegraphics[width=\mywidth]{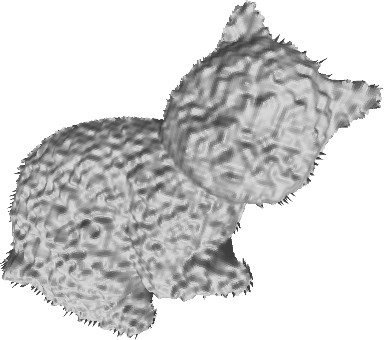}&
    \includegraphics[width=\mywidthlr]{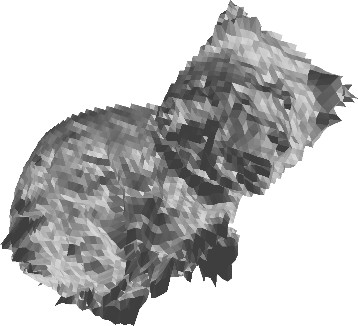}&
    \includegraphics[width=\mywidth]{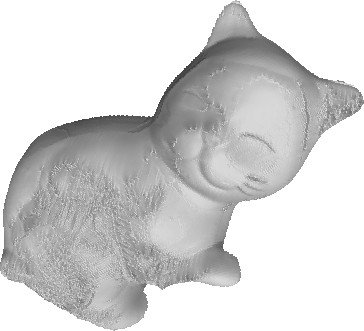}&
    \includegraphics[width=\mywidth]{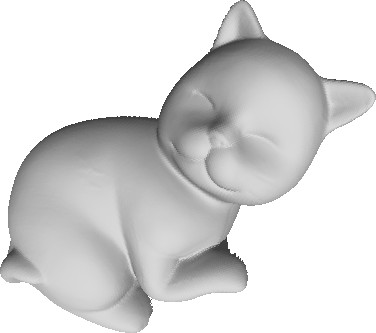} \\
    \rotatebox{90}{cow} &
    \includegraphics[width=\mywidth]{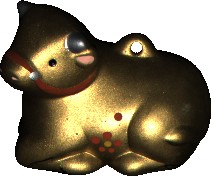}&
    \includegraphics[width=\mywidthlr]{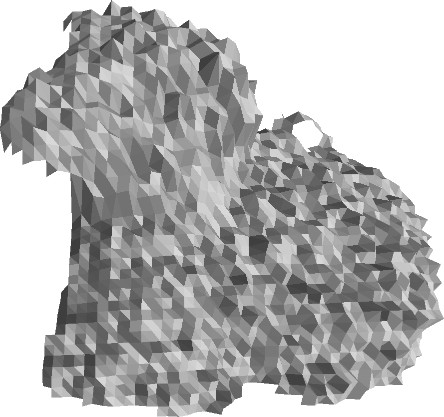}&
    \includegraphics[width=\mywidth]{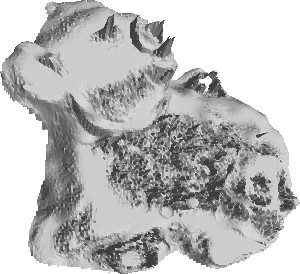}&
    \includegraphics[width=\mywidth]{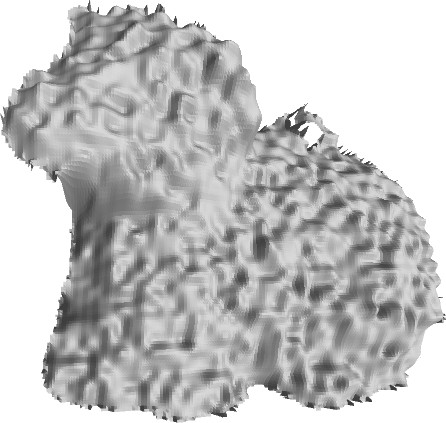}&
    \includegraphics[width=\mywidthlr]{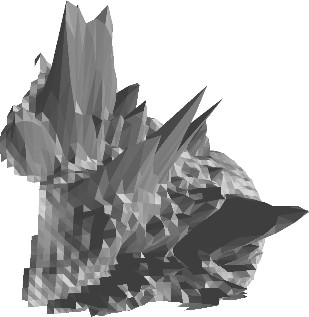}&
    \includegraphics[width=\mywidth]{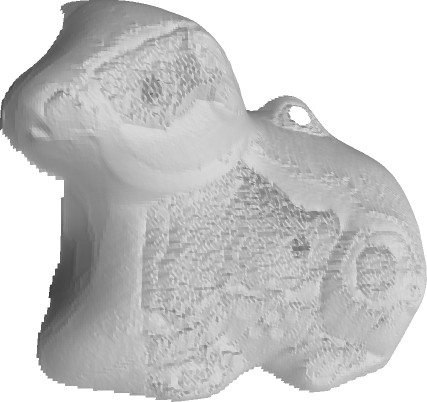}&
    \includegraphics[width=\mywidth]{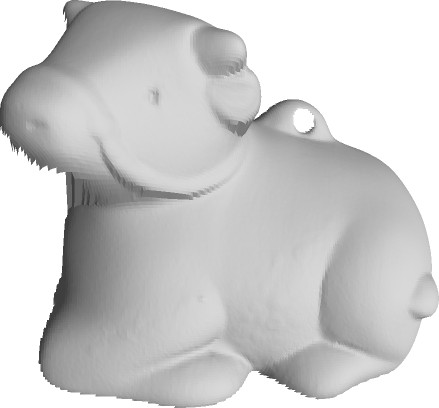} \\
    \rotatebox{90}{goblet} &
    \includegraphics[width=\mywidth]{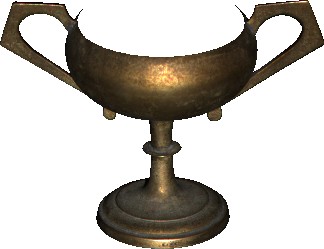}&
    \includegraphics[width=\mywidthlr]{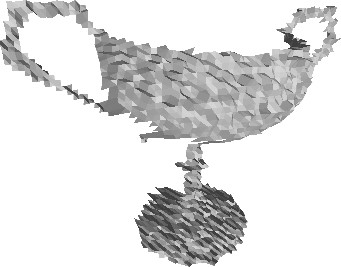}&
    \includegraphics[width=\mywidth]{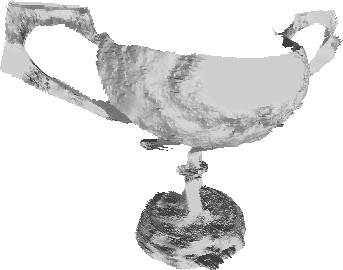}&
    \includegraphics[width=\mywidth]{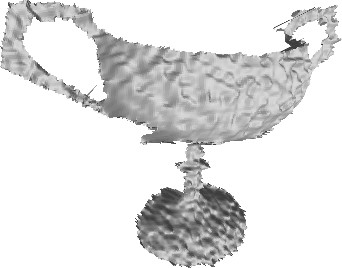}&
    \includegraphics[width=\mywidthlr]{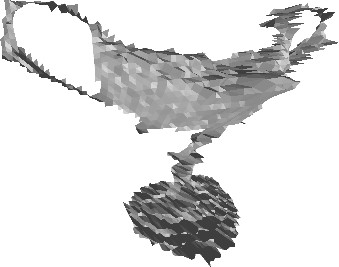}&
    \includegraphics[width=\mywidth]{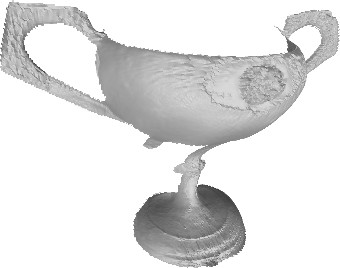}&
    \includegraphics[width=\mywidth]{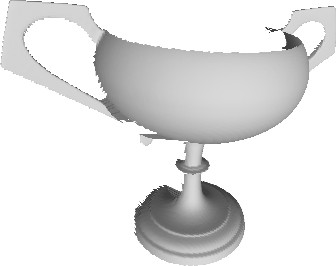} \\
    \rotatebox{90}{harvest} &
    \includegraphics[width=\mywidth]{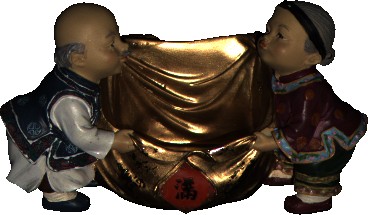}&
    \includegraphics[width=\mywidthlr]{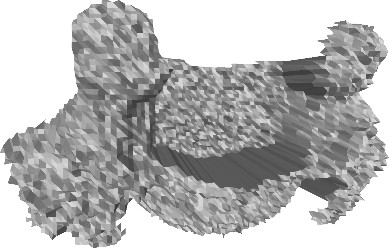}&
    \includegraphics[width=\mywidth]{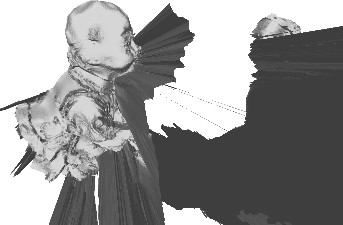}&
    \includegraphics[width=\mywidth]{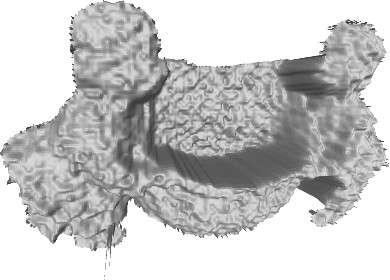}&
    \includegraphics[width=\mywidthlr]{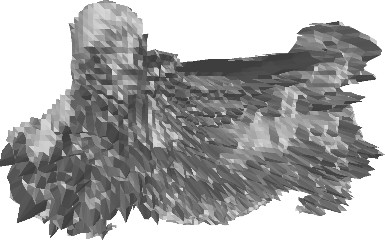}&
    \includegraphics[width=\mywidth]{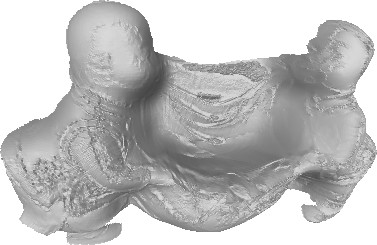}&
    \includegraphics[width=\mywidth]{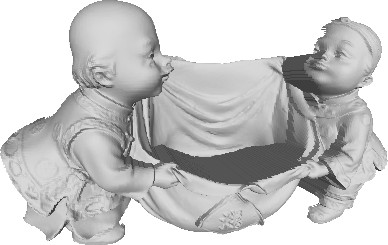} \\
    \rotatebox{90}{pot1} &
    \includegraphics[width=\mywidth]{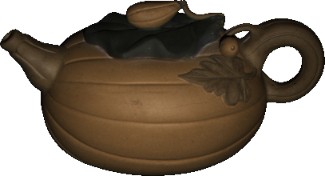}&
    \includegraphics[width=\mywidthlr]{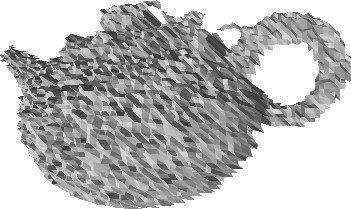}&
    \includegraphics[width=\mywidth]{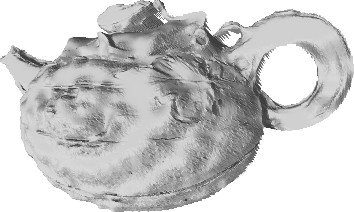}&
    \includegraphics[width=\mywidth]{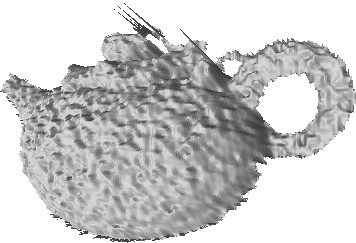}&
    \includegraphics[width=\mywidthlr]{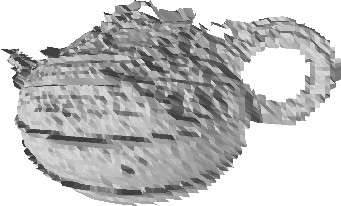}&
    \includegraphics[width=\mywidth]{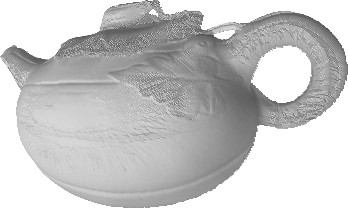}&
    \includegraphics[width=\mywidth]{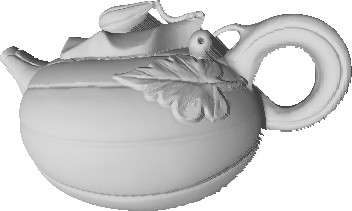} \\
    \rotatebox{90}{pot2} &
    \includegraphics[width=\mywidth]{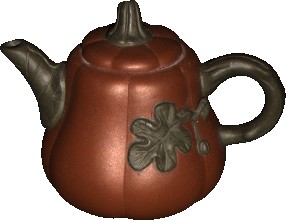}&
    \includegraphics[width=\mywidthlr]{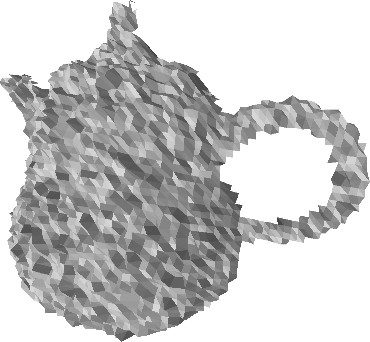}&
    \includegraphics[width=\mywidth]{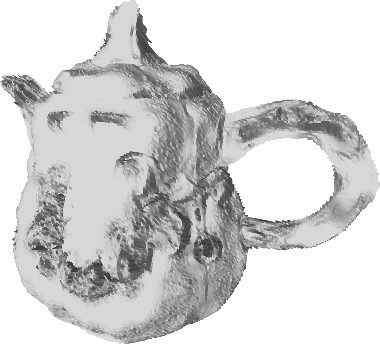}&
    \includegraphics[width=\mywidth]{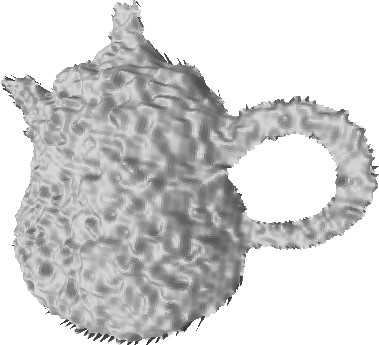}&
    \includegraphics[width=\mywidthlr]{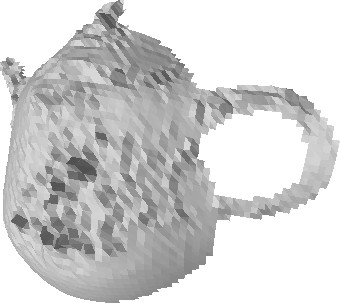}&
    \includegraphics[width=\mywidth]{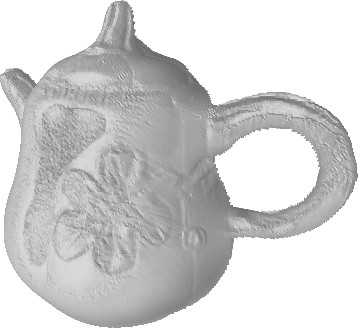}&
    \includegraphics[width=\mywidth]{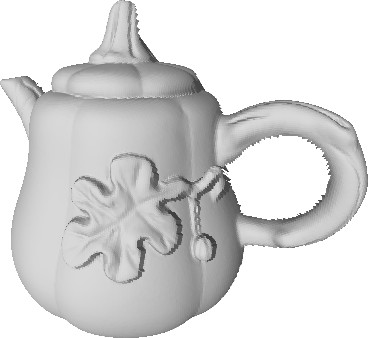} \\
    \rotatebox{90}{reading} &
    \includegraphics[width=\mywidth]{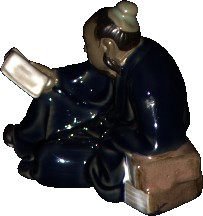}&
    \includegraphics[width=\mywidthlr]{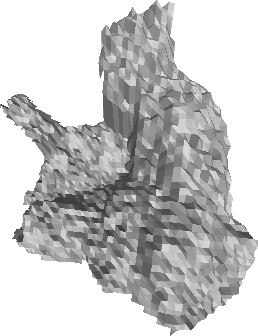}&
    \includegraphics[width=\mywidth]{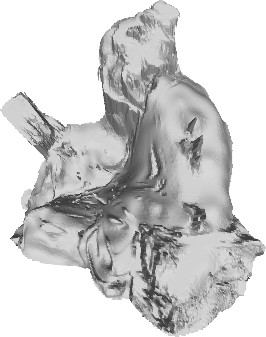}&
    \includegraphics[width=\mywidth]{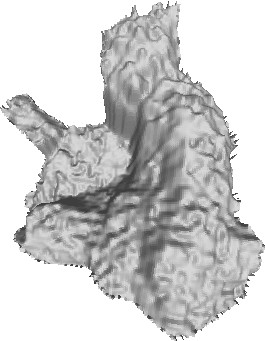}&
    \includegraphics[width=\mywidthlr]{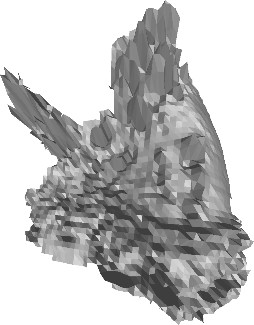}&
    \includegraphics[width=\mywidth]{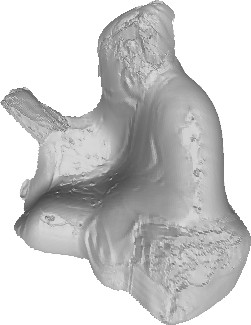}&
    \includegraphics[width=\mywidth]{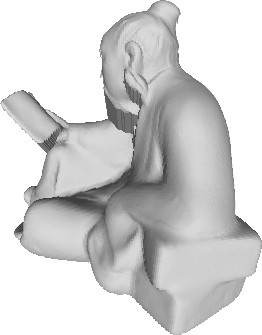}
  \end{tabular}
  \caption{Qualitative comparison between our single-shot results and those from the state-of-the-art, on the DiLiGenT dataset~\cite{Shi2018} (the scaling factor is 4). Our approach outperforms the state-of-the-art in all the experiments.}
  \label{fig:diligent_comparison_sfs}
\end{figure*}

In Figure~\ref{fig:diligent_comparison_sfs}, we qualitatively compare our single-shot results against the state-of-the-art, using the real-world DiLiGenT photometric stereo dataset~\cite{Shi2018} (only one out the 96 images of each object was used). To create noisy low-resolution input depths with a scaling factor of $2$, $4$ and $8$, the ground truth depth is downsampled and Gaussian noise is then added, as in the previous subsection. 

On objects which match our assumption of a Lambertian  surface with piecewise-constant albedo (e.g., ``bear'' and ``pot1''), we obtain very satisfactory results. However, the strong dependency of our approach on the piecewise-constant albedo assumption is clearly visible in the ``cat'' results, which are not as satisfactory: the dark structures in the image are too thin to be appropriately interpreted as piecewise-constant albedo areas and this creates artifacts in the geometry.

Besides, the ``cow'', ``pot2'' and ``reading'' results demonstrate that our approach also strongly depends upon the Lambertian assumption: the specular highlights in the images get propagated into the estimated depth. A natural future extension of our method would thus be to cope with such non-Lambertian effects, either by resorting to robust estimation techniques~\cite{CVPR2017}, or by adapting our approach to a non-Lambertian image formation model\cite{Chen2017}. 

Nevertheless, and despite these important limitations, our results remain qualitatively superior to those of the state-of-the-art in all the experiments. This can also be observed in the quantitative evaluation of Table~\ref{tab:diligent_comparison_sfs}, which confirms the conclusions of the synthetic quantitative evaluation from Table~\ref{tab:synthetic_comparison_sfs}.

\begin{table*}                                                                                                     
\centering                                                                                                         
\begin{tabular}{|c|c|c|c|c|c|c|c|c|c|}                                                                             
\hline                                                                                                             
\multirow{2}{*}{3D-shape} & \multirow{2}{*}{SF} & \multicolumn{2}{c|}{\cite{Yang2007}}  & \multicolumn{2}{c|}{\cite{Xie2016}} & \multicolumn{2}{c|}{\cite{Or-El2015}}  & \multicolumn{2}{c|}{Ours}  \\                          
\cline{3-10}                                                                                                           
 &  & RMSE & MAE & RMSE & MAE & RMSE & MAE & RMSE & MAE \\                                                         
\hline                                                                                                             
\hline                                                                                                             
 & 2 & 0.0066575 & 17.2655 & -- & -- & 0.014616 & 27.9357 & \textbf{0.0047136} & \textbf{12.8781} \\               
bear & 4 & \textbf{0.0065535} & 19.5072 & 0.8825 & 31.5392 & 0.028849 & 31.8918 & 0.0085904 & \textbf{14.8113} \\  
 & 8 & 0.97126 & 76.4581 & -- & -- & 0.055159 & 31.0276 & \textbf{0.018022} & \textbf{20.341} \\                   
\hline                                                                                                             
 & 2 & 0.0099968 & 37.3338 & -- & -- & 0.02972 & 69.0274 & \textbf{0.0080152} & \textbf{26.6017} \\                
buddha & 4 & \textbf{0.0099935} & 39.1319 & 0.86352 & 36.8237 & 0.038584 & 68.6713 & 0.012027 & \textbf{31.0774} \\
 & 8 & 1.3683 & 71.2403 & -- & -- & 0.047353 & 57.6881 & \textbf{0.019676} & \textbf{39.0075} \\                   
\hline                                                                                                             
 & 2 & 0.0085294 & 23.3362 & -- & -- & 0.028382 & 44.6708 & \textbf{0.0084811} & \textbf{18.8204} \\               
cat & 4 & \textbf{0.0096136} & 27.826 & 0.80869 & 30.7428 & 0.042872 & 54.1746 & 0.01353 & \textbf{21.4786} \\     
 & 8 & \textbf{0.015137} & 30.8242 & -- & -- & 0.065853 & 53.4602 & 0.023393 & \textbf{25.3616} \\                 
\hline                                                                                                             
 & 2 & 0.0086552 & 32.7633 & -- & -- & 0.037772 & 59.2638 & \textbf{0.0049385} & \textbf{14.806} \\                
cow & 4 & 0.0090334 & 33.8093 & 0.84557 & 33.7576 & 0.055621 & 55.3108 & \textbf{0.0089681} & \textbf{16.9767} \\  
 & 8 & \textbf{0.010392} & 31.6684 & -- & -- & 0.059261 & 53.5979 & 0.017596 & \textbf{21.03} \\                   
\hline                                                                                                             
 & 2 & \textbf{0.01019} & 30.2473 & -- & -- & 0.032588 & 59.1553 & 0.011007 & \textbf{23.0414} \\                  
goblet & 4 & \textbf{0.011121} & 31.1036 & 1.3435 & 34.0517 & 0.048727 & 56.7471 & 0.017208 & \textbf{24.2692} \\  
 & 8 & \textbf{0.015451} & 36.2801 & -- & -- & 0.084675 & 51.7091 & 0.031125 & \textbf{25.7217} \\                 
\hline                                                                                                             
 & 2 & \textbf{0.014169} & 33.9026 & -- & -- & 0.041792 & 66.3635 & 0.01594 & \textbf{31.1557} \\                  
harvest & 4 & 2.651 & 63.9349 & 0.75973 & 37.0383 & 0.05696 & 66.5893 & \textbf{0.023588} & \textbf{33.6957} \\    
 & 8 & 115.5837 & 79.2204 & -- & -- & 0.074651 & 50.9501 & \textbf{0.037176} & \textbf{35.9762} \\                 
\hline                                                                                                             
 & 2 & 0.0077563 & 22.6961 & -- & -- & 0.020767 & 48.3748 & \textbf{0.007147} & \textbf{16.9523} \\                
pot1 & 4 & \textbf{0.0086358} & 26.2298 & 0.72979 & 31.8426 & 0.03114 & 39.7103 & 0.010863 & \textbf{17.6975} \\   
 & 8 & \textbf{0.013278} & 29.6214 & -- & -- & 0.05537 & 38.9525 & 0.019307 & \textbf{19.9866} \\                  
\hline                                                                                                             
 & 2 & 0.0081729 & 28.8295 & -- & -- & 0.021455 & 50.4214 & \textbf{0.0055283} & \textbf{18.0749} \\               
pot2 & 4 & 0.0088839 & 32.7579 & 0.90388 & 33.4448 & 0.028528 & 28.5455 & \textbf{0.0088442} & \textbf{19.2421} \\ 
 & 8 & \textbf{0.014079} & 35.288 & -- & -- & 0.054661 & 47.9005 & 0.01623 & \textbf{22.4169} \\                   
\hline                                                                                                             
 & 2 & 0.011767 & 28.7648 & -- & -- & 0.030566 & 53.4663 & \textbf{0.0097283} & \textbf{19.2611} \\                
reading & 4 & \textbf{0.011428} & 30.4347 & 0.93384 & 31.764 & 0.047677 & 53.7065 & 0.015536 & \textbf{22.91} \\   
 & 8 & \textbf{0.01607} & 32.2913 & -- & -- & 0.071794 & 52.5448 & 0.028808 & \textbf{29.0107} \\                  
\hline                                                                                                             
\hline                                                                                                             
 & 2 & 0.0086552 & 28.8295 & -- & -- & 0.02972 & 53.4663 & \textbf{0.0080152} & \textbf{18.8204} \\                
Median & 4 & \textbf{0.0096136} & 31.1036 & 0.86352 & 33.4448 & 0.042872 & 54.1746 & 0.012027 & \textbf{21.4786} \\
 & 8 & \textbf{0.015451} & 35.288 & -- & -- & 0.059261 & 51.7091 & 0.019676 & \textbf{25.3616} \\                  
\hline                                                                                                             
 & 2 & 0.0095439 & 28.3488 & -- & -- & 0.028629 & 53.1865 & \textbf{0.0083887} & \textbf{20.1769} \\               
Mean & 4 & 0.30292 & 33.8595 & 0.89678 & 33.4449 & 0.042106 & 50.5941 & \textbf{0.013239} & \textbf{22.4621} \\    
 & 8 & 13.112 & 46.988 & -- & -- & 0.063197 & 48.6479 & \textbf{0.023481} & \textbf{26.5391} \\                    
\hline                                                                                                             
\end{tabular}                                                                                                      
\caption{Quantitative comparison between our single-shot results and those from the state-of-the-art, on the DiliGenT dataset~\cite{Shi2018}. Our approach systematically outperforms the state-of-the-art, consistently with the conclusions from the synthetic experiments drawn in Table~\ref{tab:synthetic_comparison_sfs}.}
\label{tab:diligent_comparison_sfs}
\end{table*}

\subsection{Comparison against State-of-the-art Multi-view Techniques on Publicly Available Real-world Datasets}

\begin{figure}[!ht]
  \centering
  \newcommand{\mywidth}{0.09\textwidth}
  \newcommand{\mywidthlr}{0.07\textwidth}
  \newcolumntype{C}{ >{\centering\arraybackslash} m{0.02\textwidth} }
  \newcolumntype{X}{ >{\centering\arraybackslash} m{\mywidth} }
  \setlength\tabcolsep{0.1pt} 
  \begin{tabular}{CXXXXX}
    & $\I$ & $\mrho$ & $\zz$ & \cite{Zollhoefer2015}/\cite{Maier2017} & Ours \\
    \rotatebox{90}{Augustus} &
    \includegraphics[width=\mywidth]{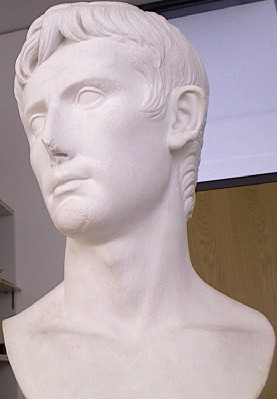}&
    \includegraphics[width=\mywidth]{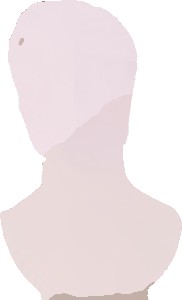}&
    \includegraphics[width=\mywidthlr]{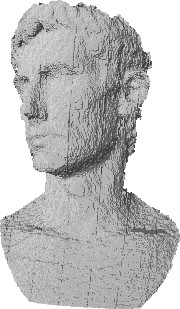}&
    \includegraphics[width=\mywidth]{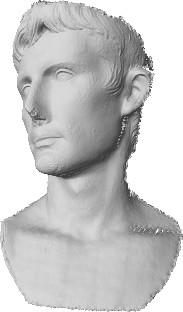}&
    \includegraphics[width=\mywidth]{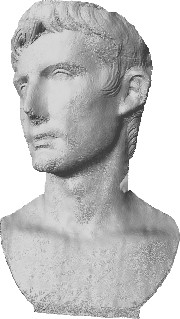}\\
    \rotatebox{90}{Lucy} &
    \includegraphics[width=\mywidth]{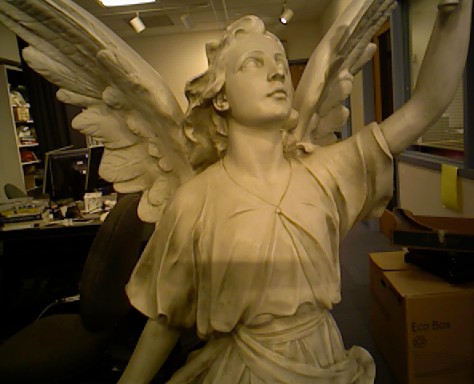}&
    \includegraphics[width=\mywidth]{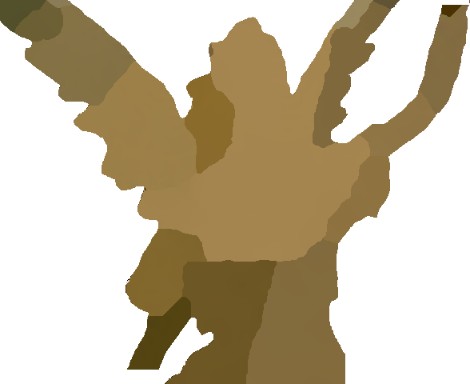}&
    \includegraphics[width=\mywidth]{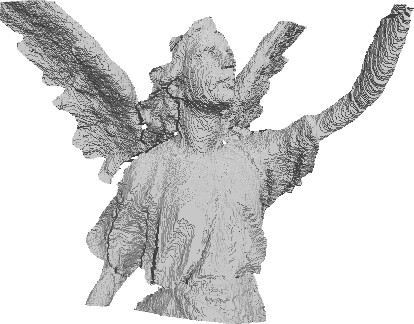}&
    \includegraphics[width=\mywidth]{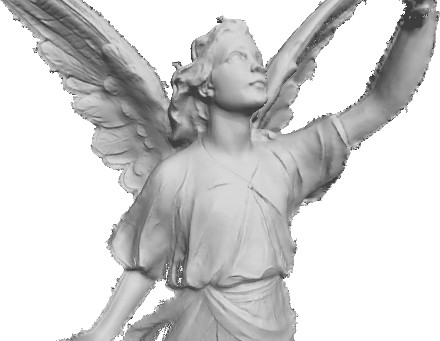}&
    \includegraphics[width=\mywidth]{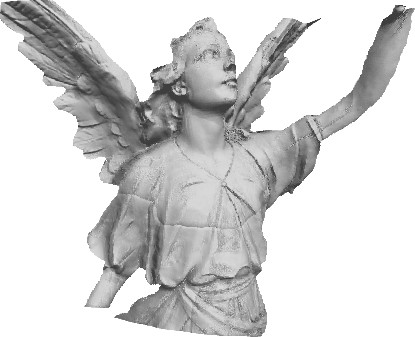}\\
    \rotatebox{90}{Relief} &
    \includegraphics[width=\mywidth]{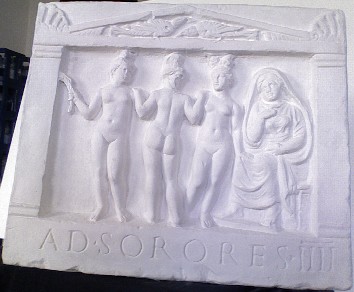}&
    \includegraphics[width=\mywidth]{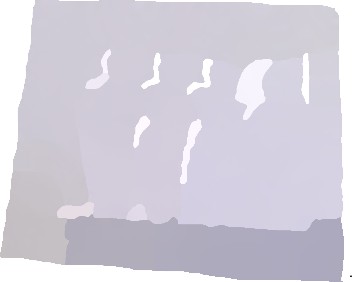}&
    \includegraphics[width=\mywidthlr]{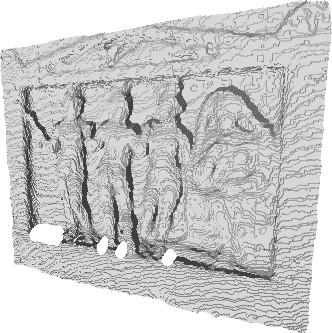}&
    \includegraphics[width=\mywidth]{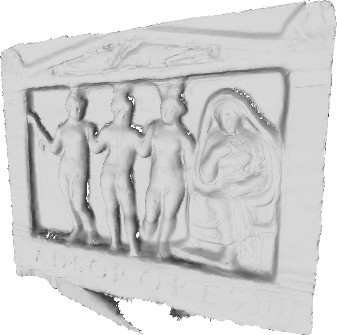}&
    \includegraphics[width=\mywidth]{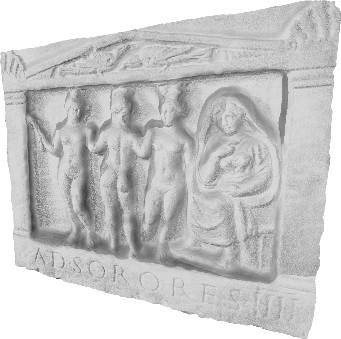}\\
    \rotatebox{90}{Gate} &
    \includegraphics[width=\mywidth]{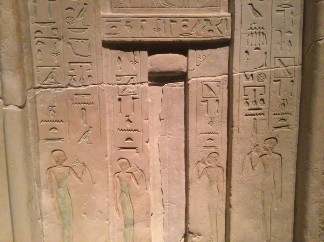}&
    \includegraphics[width=\mywidth]{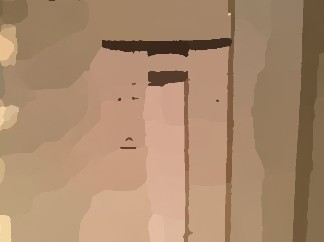}&
    \includegraphics[width=\mywidthlr]{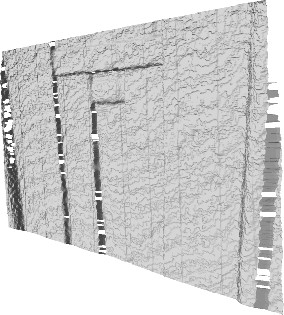}&
    \includegraphics[width=\mywidth]{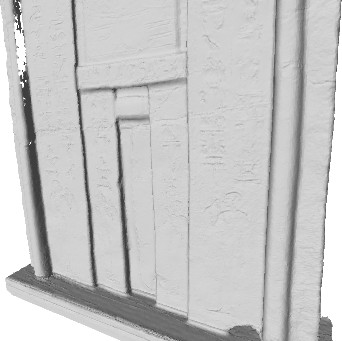}&
    \includegraphics[width=\mywidth]{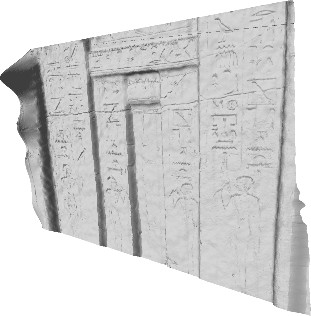}\\
  \end{tabular}
  \caption{Qualitative comparison against state-of-the-art multi-view approaches. Although it uses a single RGB-D frame, our approach results in depth maps whose quality is comparable with those obtained using multi-view data.}
  \label{fig:realworld_mv_comp_sfs}
\end{figure}

Figure \ref{fig:realworld_mv_comp_sfs} shows four qualitative comparisons with state-of-the-art multi-view approaches on publicly available datasets. The ``Augustus'', ``Lucy'' and ``Relief'' datasets~\cite{Zollhoefer2015} were created using a PrimeSense camera, whereas ``Gate''~\cite{Maier2017data} was acquired using a Structure Sensor for the iPad. The scaling factor for ``Augustus'', ``Relief'' and ``Gate'' is $2$, whereas it is $1$ for ``Lucy'' (in this case, our approach only performs shading-based depth refinement without super-resolution). Although our approach needs significantly less data (a single RGB-D image) compared to multi-view approaches, we are still able to recover fine geometry close to the degree of detail of \cite{Zollhoefer2015,Maier2017}. Even under more complex lighting, as for instance in the ``Gate'' experiment, our approach can result in high-resolution depth maps with fine-scale details.

\subsection{Additional Comparison against State-of-the-art Single-shot Techniques on Real-world Datasets we Captured Ourselves}

\begin{figure*}[!ht]
  \centering
  \newcommand{\mywidth}{0.15\textwidth}
  \newcommand{\mywidthlr}{0.1\textwidth}
  \newcolumntype{C}{ >{\centering\arraybackslash} m{0.02\textwidth} }
  \newcolumntype{X}{ >{\centering\arraybackslash} m{\mywidth} }
  \newcolumntype{Y}{ >{\centering\arraybackslash} m{\mywidthlr} }
  \setlength\tabcolsep{0.1pt} 
  \begin{tabular}{CXXYXXYX}
    & $\I$ & $\mrho$ & $\zz$ & \cite{Yang2007} & \cite{Xie2016} & \cite{Or-El2015} & Ours \\
    \rotatebox{90}{Blanket} &
    \includegraphics[width=\mywidth]{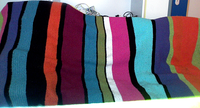}&
    \includegraphics[width=\mywidth]{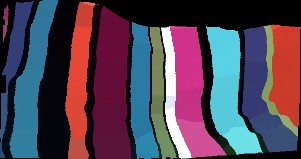}&
    \includegraphics[width=\mywidthlr]{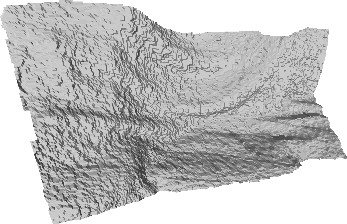}&
    \includegraphics[width=\mywidth]{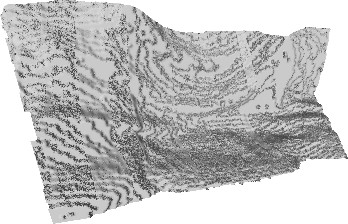}&
    \includegraphics[width=\mywidth]{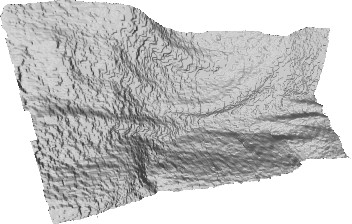}&
    \includegraphics[width=\mywidthlr]{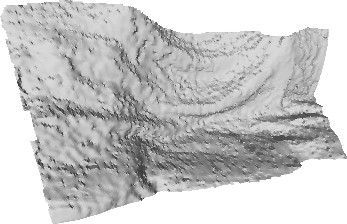}&
    \includegraphics[width=\mywidth]{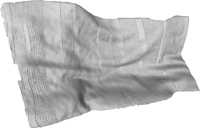}\\
    \rotatebox{90}{Clothes} &
    \includegraphics[width=\mywidth]{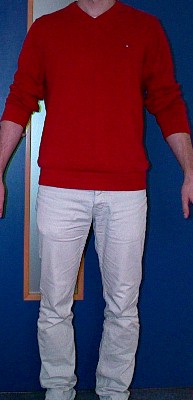}&
    \includegraphics[width=\mywidth]{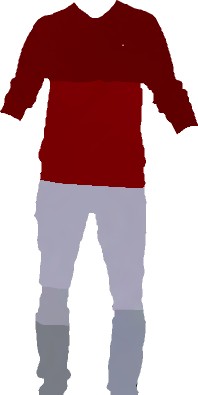}&
    \includegraphics[width=\mywidthlr]{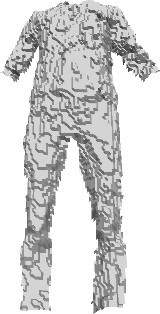}&
    \includegraphics[width=\mywidth]{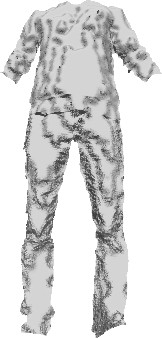}&
    \includegraphics[width=\mywidth]{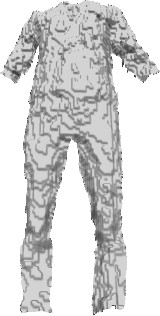}&
    \includegraphics[width=\mywidthlr]{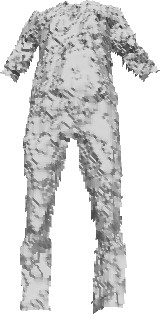}&
    \includegraphics[width=\mywidth]{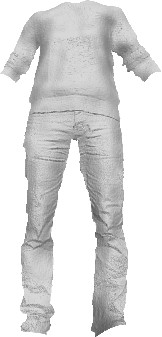}\\
    \rotatebox{90}{Failure} &
    \includegraphics[width=\mywidth]{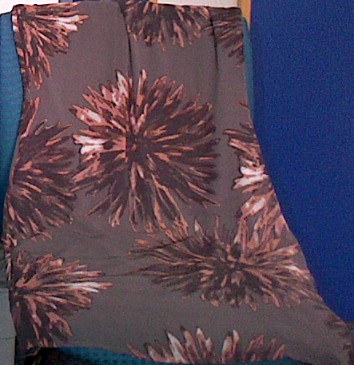}&
    \includegraphics[width=\mywidth]{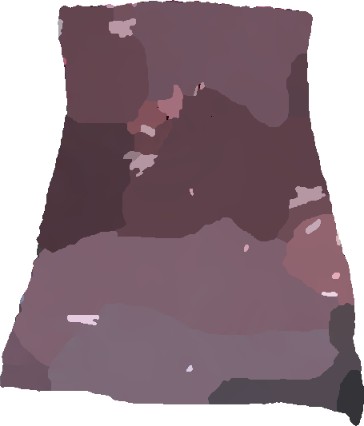}&
    \includegraphics[width=\mywidthlr]{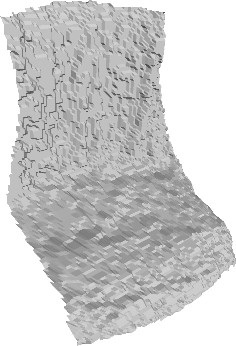}&
    \includegraphics[width=\mywidth]{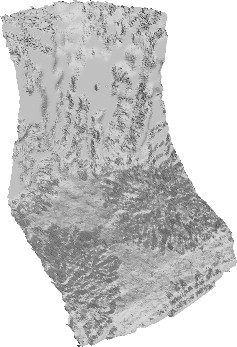}&
    \includegraphics[width=\mywidth]{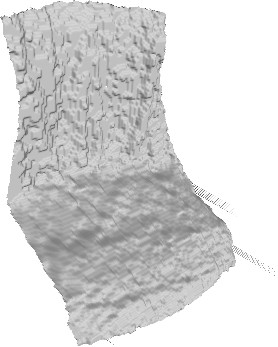}&
    \includegraphics[width=\mywidthlr]{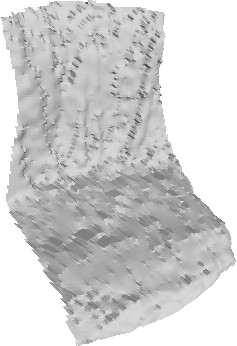}&
    \includegraphics[width=\mywidth]{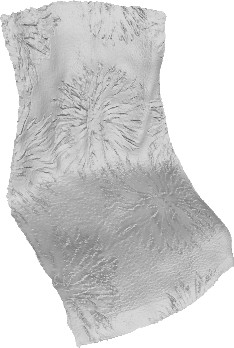}\\
    \rotatebox{90}{Monkey} &
    \includegraphics[width=\mywidth]{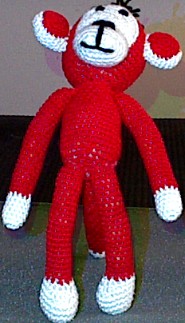}&
    \includegraphics[width=\mywidth]{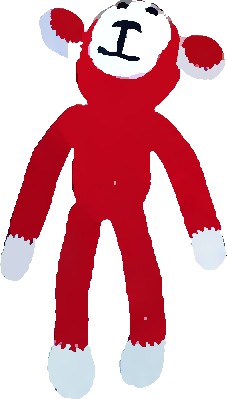}&
    \includegraphics[width=\mywidthlr]{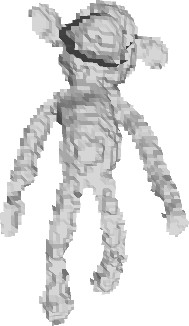}&
    \includegraphics[width=\mywidth]{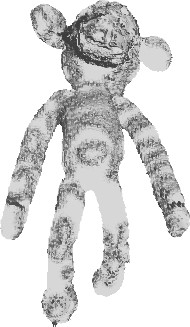}&
    \includegraphics[width=\mywidth]{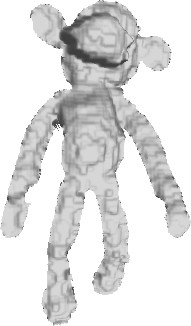}&
    \includegraphics[width=\mywidthlr]{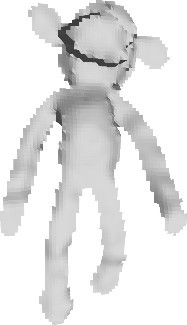}&
    \includegraphics[width=\mywidth]{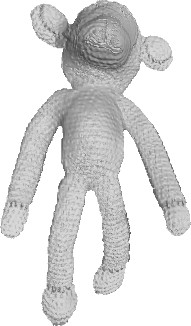}\\
    \rotatebox{90}{Wool} &
    \includegraphics[width=\mywidth]{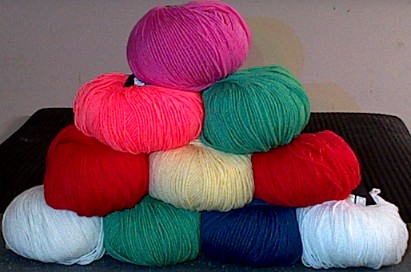}&
    \includegraphics[width=\mywidth]{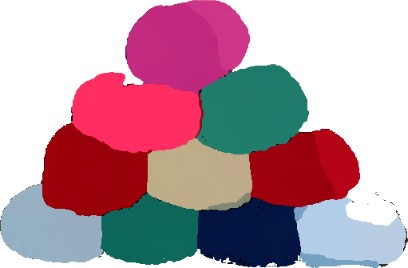}&
    \includegraphics[width=\mywidthlr]{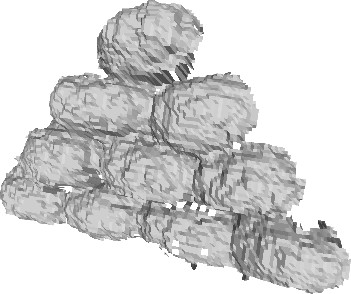}&
    \includegraphics[width=\mywidth]{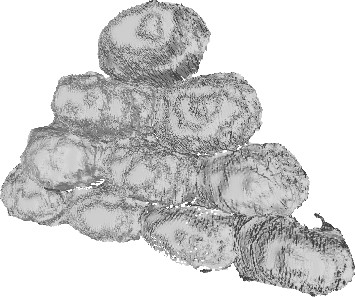}&
    \includegraphics[width=\mywidth]{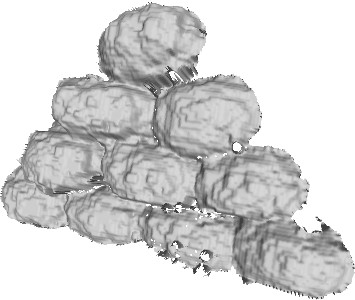}&
    \includegraphics[width=\mywidthlr]{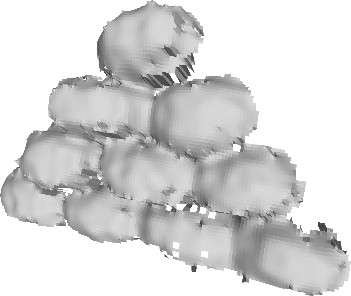}&
    \includegraphics[width=\mywidth]{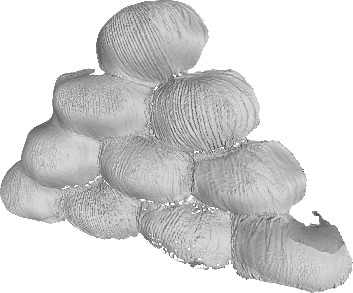}
  \end{tabular}
  \caption{Qualitative comparison of state-of-the-art single-view approaches on five real-world datasets captured with an Asus Xtion Pro Live camera at resolution $1280\times960$ for the RGB images and $320\times240$ for the low-resolution depth.}
  \label{fig:realworld_comp_sfs_kinect_data}
\end{figure*}

Figure \ref{fig:realworld_comp_sfs_kinect_data} shows additional qualitative comparison of single-shot results, on data we captured using an Asus Xtion Pro Live camera (scaling factor of $4$). Once again, our approach outperforms the state-of-the-art, even though under- or over-segmentation of the reflectance may happen.

The ``Clothes'' experiment illustrates a case where over-segmentation of reflectance happens, but interestingly this does not seem to impact depth recovery. Whenever color gets saturated (some of the balls of ``Wool'') or too low (black areas in the ``Blanket''), then minimal surface drives super-resolution: the areas where brightness is not informative are simply smoothed out, which adds robustness. Our method only fails when reflectance does not fit the Potts prior, as shown in the ``Failure'' experiment. In this case of an object with smoothly-varying reflectance, under-segmentation of reflectance happens, and all the thin brightness variations are interpreted in terms of geometry. Two alternative strategies are investigated in this work to cope with this issue: estimate reflectance without a piecewise-constant prior (learning-based strategy), or actively control lighting (photometric stereo-based strategy).

\begin{figure*}[!ht]
  \centering
  \newcommand{\mywidth}{0.15\textwidth}
  \newcommand{\mywidthlr}{0.1\textwidth}
  \newcommand{\mywidthplot}{0.3\textwidth}
  \newcolumntype{C}{>{\centering\arraybackslash} m{0.05\textwidth} }
  \newcolumntype{X}{>{\centering\arraybackslash} m{\mywidth} }
  \newcolumntype{Y}{>{\centering\arraybackslash} m{\mywidthlr} }
  \newcolumntype{Z}{>{\centering\arraybackslash} m{\mywidthplot} }
  \setlength\tabcolsep{0pt} 
  \def\arraystretch{0}
  \begin{tabular}{CXXXYXZ}
    & $\zz$ & \cite{Yang2007} & \cite{Xie2016} & \cite{Or-El2015} & Ours & 1D depth profiles \\
    \rotatebox{90}{Android} &
    \includegraphics[width=\mywidth]{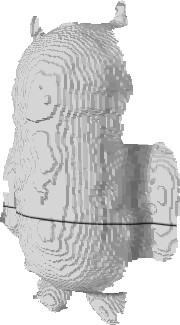} &
    \includegraphics[width=\mywidth]{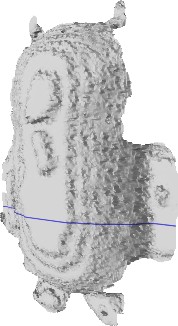} &
    \includegraphics[width=\mywidth]{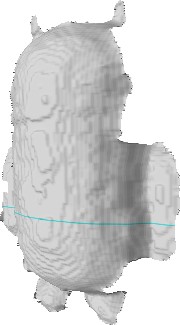} &
    \includegraphics[width=\mywidthlr]{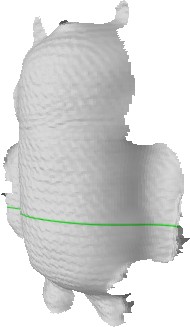} &
    \includegraphics[width=\mywidth]{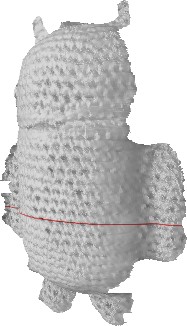} &
    \includegraphics[width=\mywidthplot]{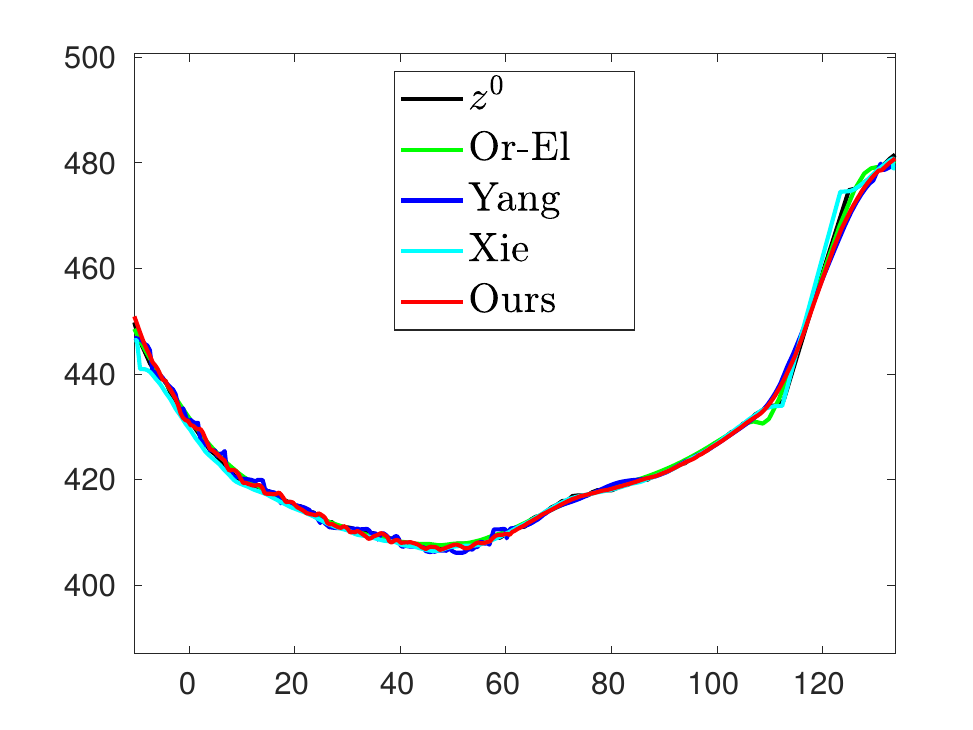} \\
    \rotatebox{90}{Basecap} &
    \includegraphics[width=\mywidth]{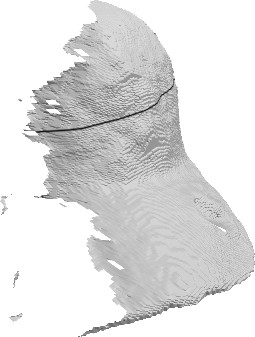} &
    \includegraphics[width=\mywidth]{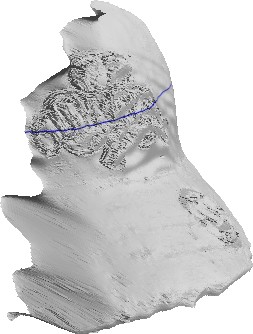} &
    \includegraphics[width=\mywidth]{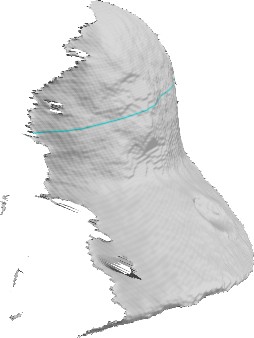} &
    \includegraphics[width=\mywidthlr]{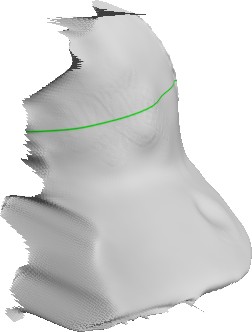} &
    \includegraphics[width=\mywidth]{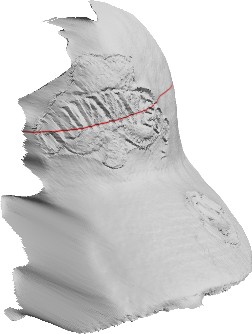} &
    \includegraphics[width=\mywidthplot]{depth_sr_sfs/profiles_basecap_profile.pdf}\\
    \rotatebox{90}{Minion} &
    \includegraphics[width=\mywidth]{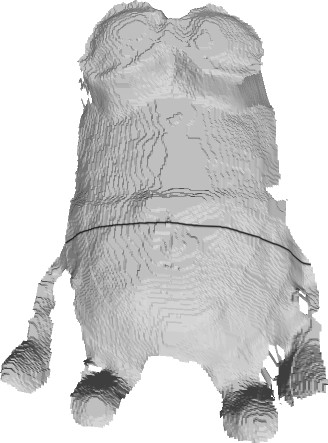} &
    \includegraphics[width=\mywidth]{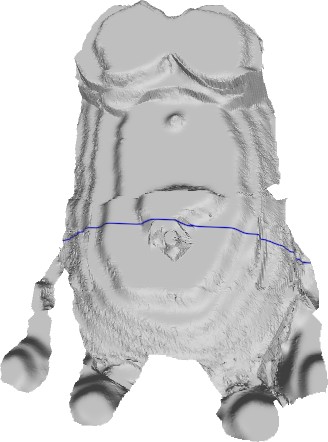} &
    \includegraphics[width=\mywidth]{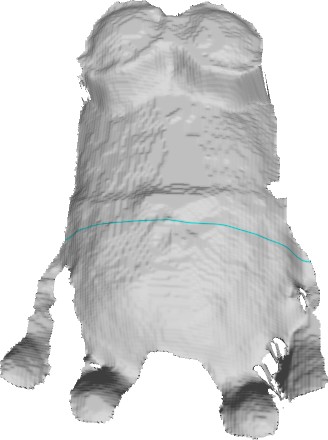} &
    \includegraphics[width=\mywidthlr]{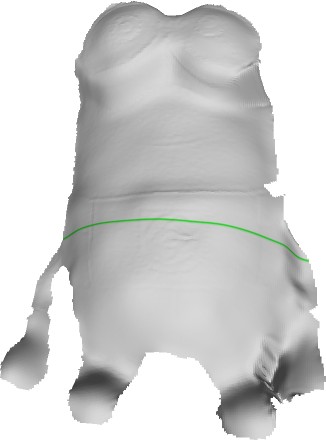} &
    \includegraphics[width=\mywidth]{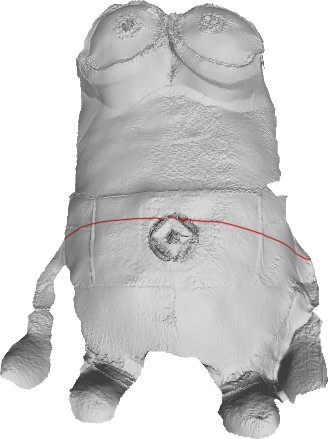} &
    \includegraphics[width=\mywidthplot]{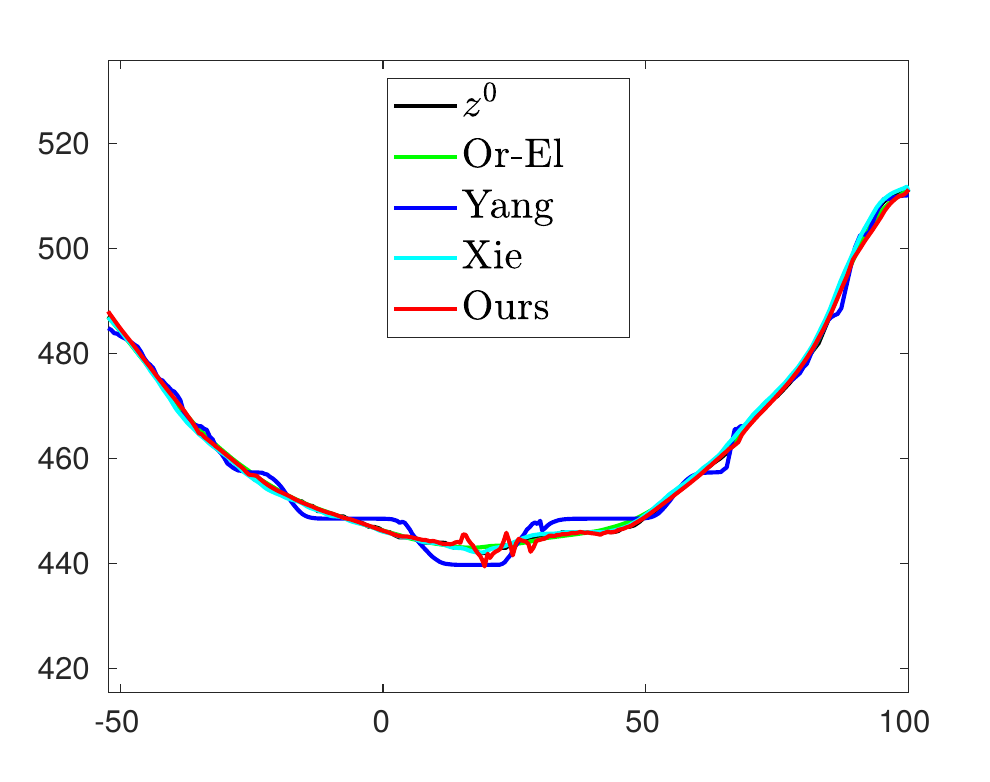}\\
    \rotatebox{90}{Rucksack} &
    \includegraphics[width=\mywidth, height=0.17\textheight]{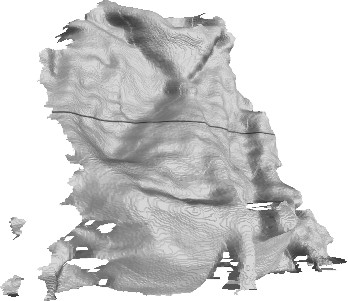} &
    \includegraphics[width=\mywidth, height=0.17\textheight]{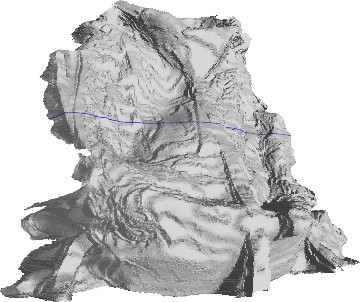} &
    \includegraphics[width=\mywidth, height=0.17\textheight]{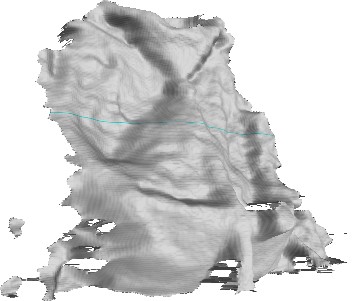} &
    \includegraphics[width=\mywidthlr, height=0.13\textheight]{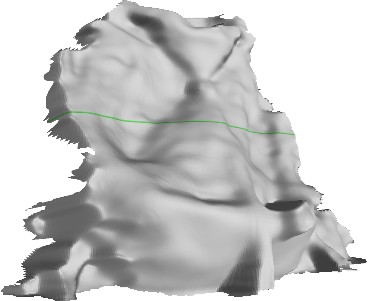} &
    \includegraphics[width=\mywidth, height=0.17\textheight]{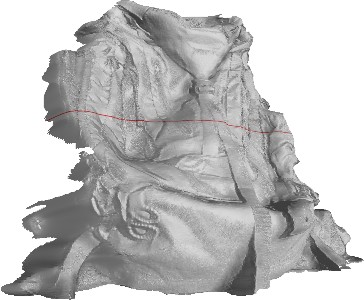} &
    \includegraphics[width=\mywidthplot]{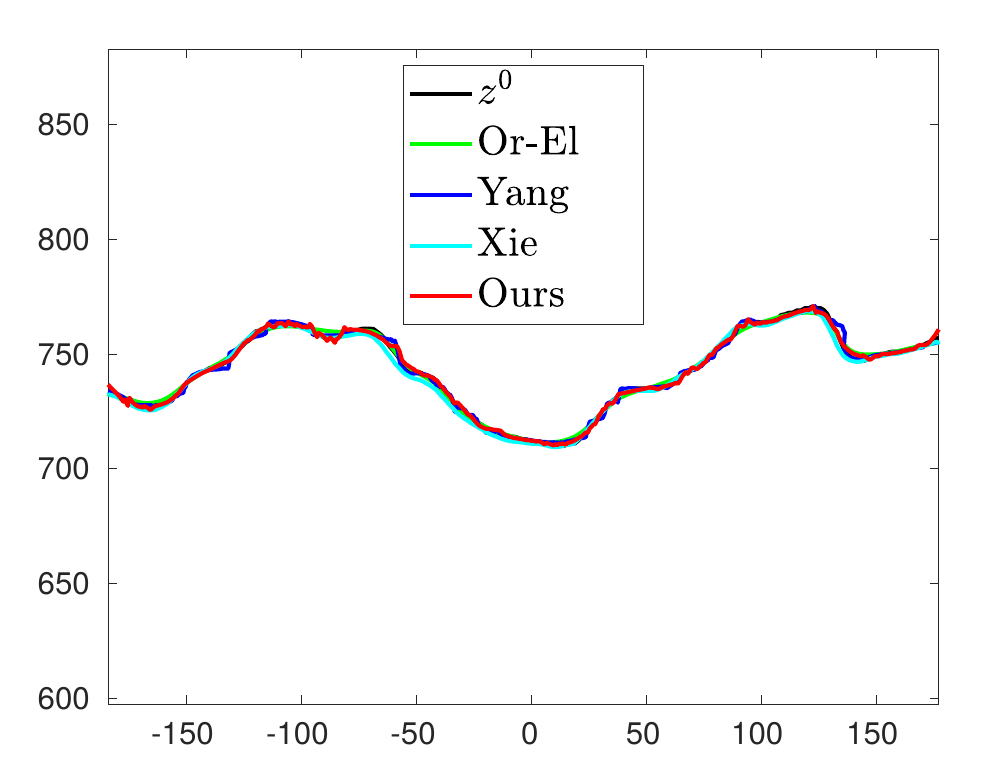}
  \end{tabular}
  \caption{Qualitative comparison between our single-shot results and those from the state-of-the-art, on the datasets from Figure 3 in the main paper. Our approach systematically outperforms the state-of-the-art. The rightest column shows 1D depth profiles corresponding to the lines drawn on the 3D-shapes: although the depth estimated using all the methods overall fit well together, ours is the only which provides reasonable fine-scale details.}
  \label{fig:synthetic_comparison_sfs_1d_profiles} 
\end{figure*}

Eventually, Figure \ref{fig:synthetic_comparison_sfs_1d_profiles} presents another qualitative comparison on the real-world data from Figure 3 in the main paper (captured using a RealSense D415 camera). Note that \cite{Yang2007} seems to give good depth estimates whenever the underlying assumption (an edge in the RGB image coincides with an edge in the depth image) is met, cf. ``Rucksack'' experiment, but it fails to provide detail-preserving depth maps when reflectance is uniform or changes only slightly (``Android'' and ``Minion'' experiments), since it uses only a sparse set of information from the RGB data. Unsurprisingly, the method from \cite{Xie2016} cannot hallucinate surface details, since it does not use the color image. The shading-based depth refinement method of \cite{Or-El2015} does a much better job at improving geometry, but it is largely overcome by the proposed shading-based depth super-resolution approach, because the latter uses information from a higher-resolution RGB image.

\section{Evaluation of the Reflectance Learning-based Approach}  
\label{sec:supp_4}

\subsection{Creation of the Synthetic Data}

Let us first recall that the reflectance learning-based approach was trained on data extracted from the ICT-3DRFE Database~\cite{Database3drfe2011}. In order to evaluate this approach, we considered two subjects from this database as well, each one enacting $10$ different facial expressions. Of course, in order to avoid any bias, these subjects were not used when training the neural network. 

The high-resolution RGB and low-resolution depth images were created in a similar manner as in the previous section: high-resolution RGB images of the faces were rendered at $512\times512$ resolution from the ground truth albedo and depth under first-order spherical harmonics lighting $\l=[0, 0, -1, 0.2]^\top$; and the low-resolution depth maps were created by downsampling the ground truth depth by a scaling factor of $2$, $4$ and $8$. Zero-mean Gaussian noise with standard deviation $1\%$ the maximum RGB intensity was then added to the RGB images, and zero-mean Gaussian noise with standard deviation $10^{-4}$ the squared original depth value (consistently with the real-world measurements from~\cite{Khoshelham2012})
was added to the low-resolution depth maps, before quantisation. 

These synthetic faces were then used for quantitative evaluation of the proposed reflectance learning-based approach against the state-of-the-art and against the proposed fully variational solution, as discussed in the next subsection.

\subsection{Comparison against the State-of-the-art on the Synthetic Dataset}

We next evaluate our method, which first estimates reflectance using deep learning and then achieves variational depth super-resolution using the RGB image, in comparison with end-to-end deep learning solutions for geometry estimation. 

\begin{figure*}[!ht]
	\centering
  \newcommand{\mywidth}{0.16\textwidth}
  \newcommand{\mywidthlr}{0.1\textwidth}
	\newcolumntype{C}{ >{\centering\arraybackslash} m{0.02\textwidth} }
	\newcolumntype{X}{ >{\centering\arraybackslash} m{\mywidth} }
	\newcolumntype{Y}{ >{\centering\arraybackslash} m{\mywidthlr} }
	\setlength\tabcolsep{0.1pt} 
  \def\arrasytrech{1} 
	\begin{tabular}{CXYXYXXX}
	  &$\I$ & $\zz$ & \cite{Hui2016} & \cite{Sfsnet18} & Ours (SfS) & Ours & Ground truth \cite{Database3drfe2011}\\
	  \rotatebox{90}{S4E4}&
		\includegraphics[width=\mywidth]{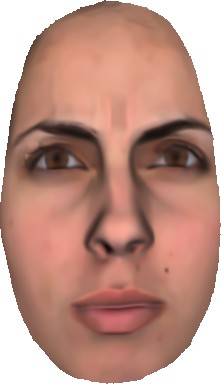}&
		\includegraphics[width=\mywidthlr]{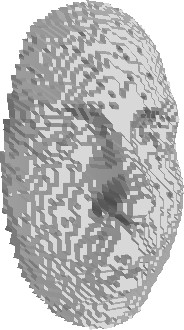}&
		\includegraphics[width=\mywidth]{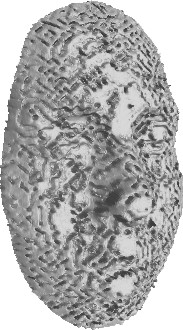}&
		\includegraphics[width=\mywidthlr]{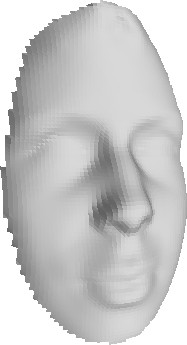}&
		\includegraphics[width=\mywidth]{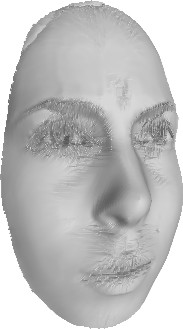}&
		\includegraphics[width=\mywidth]{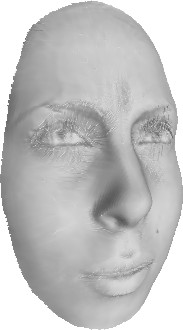}&
		\includegraphics[width=\mywidth]{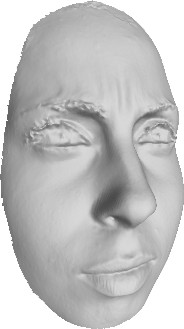}\\
		\rotatebox{90}{S4E9}&
		\includegraphics[width=\mywidth]{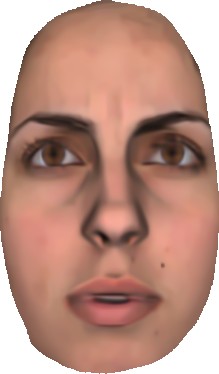}&
		\includegraphics[width=\mywidthlr]{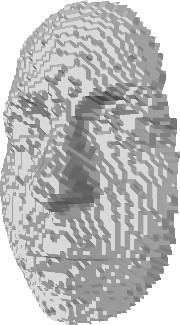}&
		\includegraphics[width=\mywidth]{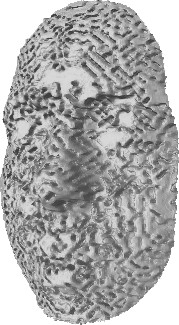}&
		\includegraphics[width=\mywidthlr]{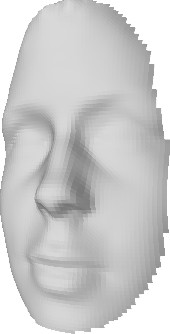}&
		\includegraphics[width=\mywidth]{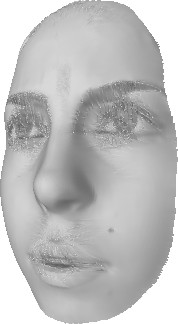}&
		\includegraphics[width=\mywidth]{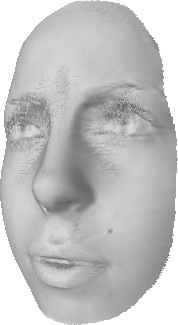}&
		\includegraphics[width=\mywidth]{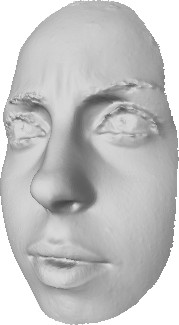}\\
		\rotatebox{90}{S11E5}&
		\includegraphics[width=\mywidth]{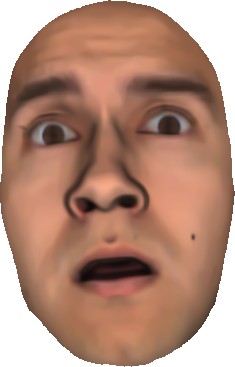}&
		\includegraphics[width=\mywidthlr]{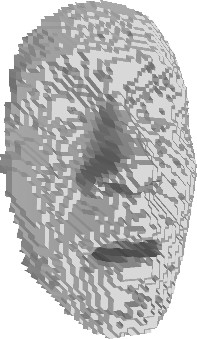}&
		\includegraphics[width=\mywidth]{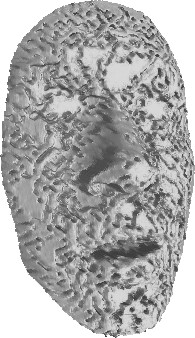}&
		\includegraphics[width=\mywidthlr]{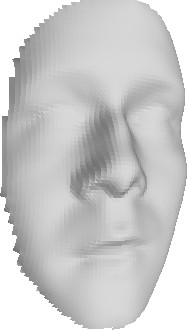}&
		\includegraphics[width=\mywidth]{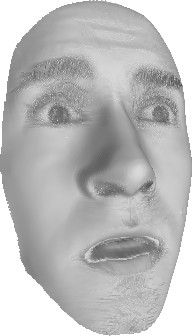}&
		\includegraphics[width=\mywidth]{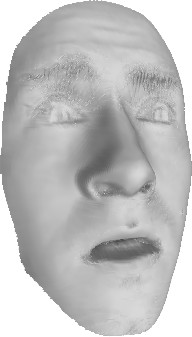}&
		\includegraphics[width=\mywidth]{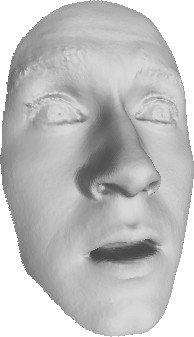}\\
		\rotatebox{90}{S11E6}&
		\includegraphics[width=\mywidth]{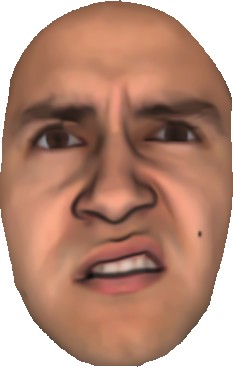}&
		\includegraphics[width=\mywidthlr]{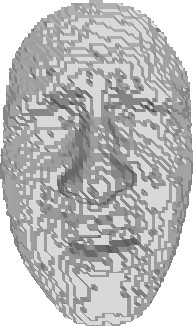}&
		\includegraphics[width=\mywidth]{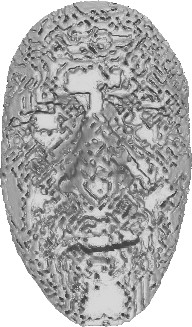}&
		\includegraphics[width=\mywidthlr]{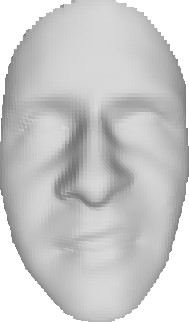}&
		\includegraphics[width=\mywidth]{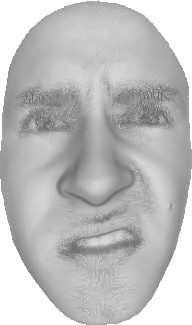}&
		\includegraphics[width=\mywidth]{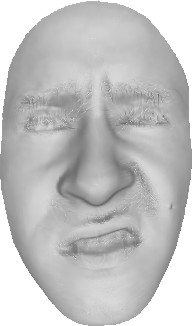}&
		\includegraphics[width=\mywidth]{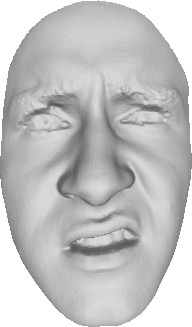}\\
		\end{tabular}
	\caption{Qualitative comparison of the results obtained using the deep learning-based depth super-resolution technique from~\cite{Hui2016}, the deep learning-based shape-from-shading approach from~\cite{Sfsnet18}, the proposed variational approach to shape-from-shading (denoted by SfS), and the proposed combination of deep learning and variational methods. The latter seems the most effective, and this is confirmed by the quantitative evaluation provided in Table~\ref{tab:deepnetTable}. ~\\~\\}
	\label{fig:synthetic_data_deepnet_results}
\end{figure*}

For comparison, we first consider the method introduced in~\cite{Hui2016}, which is an end-to-end depth super-resolution technique based on low-resolution depth data and high-resolution RGB image, i.e. the same inputs as our methods. It can be seen in Figure~\ref{fig:synthetic_data_deepnet_results} that this end-to-end solution fails to recover surface details which are visible in the RGB image.

In order to evaluate the ability of deep networks to reconstruct geometry from a single RGB image, similarly to shape-from-shading techniques, we also show the results of SfSNet~\cite{Sfsnet18}, which is a deep learning-based method estimating albedo and surface normals (which we further integrated into a depth map using the quadratic integation method discussed in~\cite{Queau_variational}) out of a single RGB image. SfSNet is limited to RGB images of size $128\times128$, so it was evaluated only for a scaling factor of $4$ and, since it does not perform depth super-resolution, the ground truth depth was downsampled for the quantitative evaluation of this method. Figure~\ref{fig:synthetic_data_deepnet_results} shows that reasonable results can be expected using SfSNet, yet geometry is slightly oversmoothed in comparison with what can be obtained using the proposed combination of machine learning and variational approaches.

Eventually, we compare this combined approach with the fully variational one from the previous section. The latter does not completely fail at recovering a reasonable geometry, but since the estimated albedo is piecewise-constant and departs significantly from the ground truth, artifacts and noise are propagated to the estimated geometry. This is confirmed by the quantitative evaluation in Table~\ref{tab:deepnetTable}, which clearly indicates that the proposed combination of machine learning and variational methods is more efficient than both end-to-end learning solutions from the state-of-the-art and the proposed fully variational approach.

\begin{table*}[!ht]                                                                                                  
\centering                                                                                                      
\begin{tabular}{|c|c|c|c|c|c|c|c|c|c|c|}                                                                        
\hline                                                                                                          
\multirow{2}{*}{Subject (S)}  & \multirow{2}{*}{Expression (E)}  & \multirow{2}{*}{SF} & \multicolumn{2}{c|}{\cite{Hui2016}} & \multicolumn{2}{c|}{\cite{Sfsnet18}} & \multicolumn{2}{c|}{Ours (SfS)} & \multicolumn{2}{c|}{Ours}\\\cline{4-11}
 &  &  & RMSE & MAE & RMSE & MAE & RMSE & MAE & RMSE & MAE \\                                                   
\hline                                                                                                          
\hline                                                                                                          
 \multirow{30}{*}{4} &   & 2 & 0.1572  & 48.3553 & -- & -- & 0.1613 & 14.1551 & \textbf{0.1355} & \textbf{9.9354} \\                  
                     & 0 & 4 & 0.13284 & 36.9774 & 0.6071 & 13.4647 & 0.10629 & 11.6301 & \textbf{0.086867} & \textbf{8.5443} \\  
                     &   & 8 & 0.63    & 38.0244 & -- & -- & 0.24354 & 14.2278 & \textbf{0.19869} & \textbf{11.0617} \\                 
\cline{2-11}                                                                                                          
                     &   & 2 & 0.15451 & 48.4316 & -- & -- & 0.15824 & 15.2649 & \textbf{0.13413} & \textbf{10.6932} \\              
                     & 1 & 4 & 0.13069 & 36.4169 & 0.7185 & 14.3254 & 0.10972 & 12.8361 & \textbf{0.087058} & \textbf{9.5301} \\    
                     &   & 8 & 0.61295 & 35.7076 & -- & -- & 0.25859 & 15.2818 & \textbf{0.20735} & \textbf{12.0648} \\              
\cline{2-11}                                                                                                          
                     &   & 2 & 0.15358 & 49.0454 & -- & -- & 0.14589 & 14.516 & \textbf{0.14005} & \textbf{14.221} \\                
                     & 2 & 4 & 0.13266 & 37.1973 & 0.8821 & 18.5108 & 0.12322 & 13.9566 & \textbf{0.10993} & \textbf{13.2265} \\   
                     &   & 8 & 0.97825 & 38.8272 & -- & -- & 0.27613 & 17.4243 & \textbf{0.25379} & \textbf{16.0565} \\              
\cline{2-11}                                                                                                          
                     &   & 2 & 0.15657 & 48.4417 & -- & -- & 0.1614 & 14.776 & \textbf{0.14468} & \textbf{11.1379} \\                
                     & 3 & 4 & 0.13335 & 37.0725 & 0.8554 & 16.0271 & 0.11759 & 13.165 & \textbf{0.09558} & \textbf{10.604} \\    
                     &   & 8 & 0.95333 & 38.6755 & -- & -- & 0.26179 & 15.8035 & \textbf{0.21567} & \textbf{13.0665} \\              
\cline{2-11}                                                                                                          
                     &   & 2 & 0.15155 & 48.3665 & -- & -- & 0.14914 & 14.5008 & \textbf{0.13132} & \textbf{11.1096} \\              
                     & 4 & 4 & 0.13093 & 37.2093 & 0.6301 & 15.0882 & 0.1108 & 12.6113 & \textbf{0.091216} & \textbf{10.0432} \\  
                     &   & 8 & 0.81628 & 37.4412 & -- & -- & 0.24872 & 15.1804 & \textbf{0.18053} & \textbf{12.4699} \\              
\cline{2-11}                                                                                                          
                     &   & 2 & \textbf{0.15404} & 47.7565 & -- & -- & 0.17 & 15.3645 & 0.16346 & \textbf{13.0845} \\                 
                     & 5 & 4 & 0.17413          & 37.2071 & 0.9004 & 18.8166 & 0.187 & 15.0677 & \textbf{0.16933} & \textbf{13.3297} \\    
                     &   & 8 & 0.81401          & 37.1801 & -- & -- & 0.35861 & 18.885 & \textbf{0.31589} & \textbf{16.5856} \\               
\cline{2-11}                                                                                                          
                     &   & 2 & 0.15457 & 47.8468 & -- & -- & 0.16725 & 14.4807 & \textbf{0.15373} & \textbf{12.0069} \\              
                     & 6 & 4 & 0.13863 & 36.681  & 0.8684 & 19.1934 & 0.14573 & 13.5136 & \textbf{0.12169} & \textbf{11.3427} \\      
                     &   & 8 & 0.49746 & 36.8001 & -- & -- & 0.31234 & 17.0774 & \textbf{0.26064} & \textbf{14.4585} \\              
\cline{2-11}                                                                                                          
                     &   & 2 & \textbf{0.15476} & 48.4094 & -- & -- & 0.17713 & 15.3454 & 0.1602 & \textbf{12.6357} \\               
                     & 7 & 4 & 0.18215          & 36.0914 & 0.8460 & 19.8673 & 0.18528 & 14.1876 & \textbf{0.15723} & \textbf{11.8129} \\  
                     &   & 8 & 0.82718          & 38.4132 & -- & -- & 0.34986 & 17.1481 & \textbf{0.29932} & \textbf{14.3226} \\              
\cline{2-11}                                                                                                          
                     &   & 2 & 0.15437 & 48.3719 & -- & -- & 0.15093 & 15.9026 & \textbf{0.13533} & \textbf{12.1738} \\              
                     & 8 & 4 & 0.13062 & 37.3586 & 0.4986 & 13.6524 & 0.107 & 13.3844 & \textbf{0.085065} & \textbf{10.5078} \\   
                     &   & 8 & 0.71791 & 37.543  & -- & -- & 0.23366 & 15.5826 & \textbf{0.19305} & \textbf{12.6953} \\               
\cline{2-11}                                                                                                          
                     &   & 2 & 0.15989 & 49.2317 & -- & -- & 0.15939 & 13.879 & \textbf{0.13843} & \textbf{11.097} \\                
                     & 9 & 4 & 0.13373 & 38.022  & 0.6107 & 14.3473 & 0.11108 & 12.4866 & \textbf{0.091548} & \textbf{9.9358} \\   
                     &   & 8 & 0.53732 & 36.8758 & -- & -- & 0.25022 & 15.1722 & \textbf{0.20378} & \textbf{12.7385} \\              
\cline{2-11}                                                                                                          
\hline
 \multirow{30}{*}{11} &   & 2 & 0.16035 & 48.3088 & -- & -- & 0.15248 & 15.1914 & \textbf{0.13971} & \textbf{9.7571} \\               
                      & 0 & 4 & 0.13743 & 36.6588 & 1.0125 & 11.8150 & 0.11775 & 12.5609 & \textbf{0.10129} & \textbf{8.6988} \\   
                      &   & 8 & 0.51743 & 32.4395 & -- & -- & 0.26081 & 14.6577 & \textbf{0.22472} & \textbf{11.6671} \\              
\cline{2-11}                                                                                                          
                      &   & 2 & 0.15231 & 48.2292 & -- & -- & 0.14523 & 15.381 & \textbf{0.13328} & \textbf{9.9593} \\                
                      & 1 & 4 & 0.12957 & 35.6881 & 0.8798 & 10.8757 & 0.11237 & 12.7309 & \textbf{0.097136} & \textbf{8.5511} \\  
                      &   & 8 & 0.52279 & 32.5115 & -- & -- & 0.25173 & 14.7836 & \textbf{0.20387} & \textbf{11.6099} \\              
\cline{2-11}                                                                                                          
                      &   & 2 & 0.15548 & 47.5781 & -- & -- & 0.15421 & 15.7925 & \textbf{0.14821} & \textbf{12.0207} \\              
                      & 2 & 4 & 0.1393  & 36.4907 & 0.9789 & 19.5521 & 0.13943 & 14.875 & \textbf{0.12114} & \textbf{11.8059} \\   
                      &   & 8 & 0.66616 & 36.1001 & -- & -- & 0.30543 & 18.2869 & \textbf{0.26649} & \textbf{15.2768} \\              
\cline{2-11}                                                                                                          
                      &   & 2 & 0.16131 & 48.4766 & -- & -- & 0.15472 & 15.3901 & \textbf{0.14409} & \textbf{10.0102} \\              
                      & 3 & 4 & 0.13652 & 36.0848 & 1.2922 & 13.2403 & 0.12219 & 13.3513 & \textbf{0.10614} & \textbf{8.9464} \\   
                      &   & 8 & 0.94169 & 37.7036 & -- & -- & 0.27622 & 16.4698 & \textbf{0.2266} & \textbf{12.5186} \\               
\cline{2-11}                                                                                                          
                      &   & 2 & 0.15879 & 48.3293 & -- & -- & 0.15457 & 15.0479 & \textbf{0.14001} & \textbf{10.8615} \\              
                      & 4 & 4 & 0.13926 & 36.8105 & 0.8897 & 12.7579 & 0.12404 & 13.557 & \textbf{0.10533} & \textbf{9.5906} \\   
                      &   & 8 & 0.72556 & 35.9876 & -- & -- & 0.27086 & 15.786 & \textbf{0.22974} & \textbf{12.7579} \\               
\cline{2-11}                                                                                                          
                      &   & 2 & 0.16252 & 47.6152 & -- & -- & 0.16964 & 17.0446 & \textbf{0.15787} & \textbf{10.6522} \\              
                      & 5 & 4 & 0.15556 & 36.695  & 1.1557 & 14.7778 & 0.17783 & 15.3102 & \textbf{0.15392} & \textbf{10.6452} \\   
                      &   & 8 & 0.81958 & 35.1608 & -- & -- & 0.32727 & 18.6277 & \textbf{0.28649} & \textbf{14.4657} \\              
\cline{2-11}                                                                                                          
                      &   & 2 & 0.15936 & 48.2603 & -- & -- & 0.15054 & 15.5422 & \textbf{0.14255} & \textbf{10.4559} \\              
                      & 6 & 4 & 0.13906 & 36.2701 & 0.7581 & 13.9221 & 0.13145 & 13.7609 & \textbf{0.1157} & \textbf{9.8813} \\    
                      &   & 8 & 0.68759 & 35.3423 & -- & -- & 0.29362 & 18.0689 & \textbf{0.25192} & \textbf{14.3041} \\              
\cline{2-11}                                                                                                          
                      &   & 2 & \textbf{0.15783} & 46.3708 & -- & -- & 0.19123 & 16.3544 & 0.17274 & \textbf{10.4441} \\              
                      & 7 & 4 & \textbf{0.20118} & 35.1363 & 1.2066 & 18.6458 & 0.23771 & 16.3544 & 0.20955 & \textbf{11.5822} \\  
                      &   & 8 & 0.73165          & 35.5369 & -- & -- & 0.41273 & 19.1395 & \textbf{0.36912} & \textbf{14.8272} \\              
\cline{2-11}                                                                                                          
                      &   & 2 & 0.1601  & 48.3084 & -- & -- & 0.14637 & 18.5782 & \textbf{0.12985} & \textbf{13.3089} \\               
                      & 8 & 4 & 0.13852 & 37.6211 & 0.7112 & 13.2194 & 0.11509 & 15.9898 & \textbf{0.095155} & \textbf{11.6081} \\
                      &   & 8 & 0.78491 & 37.1651 & -- & -- & 0.25296 & 18.0263 & \textbf{0.20633} & \textbf{14.3745} \\              
\cline{2-11}                                                                                                          
                      &   & 2 & 0.15292 & 48.2978 & -- & -- & 0.13997 & 14.5447 & \textbf{0.12648} & \textbf{10.2274} \\              
                      & 9 & 4 & 0.13424 & 36.6469 & 0.9484 & 12.6980 & 0.11997 & 13.0994 & \textbf{0.10137} & \textbf{9.4048} \\  
                      &   & 8 & 0.63803 & 34.7198 & -- & -- & 0.26044 & 15.9719 & \textbf{0.21383} & \textbf{12.7267} \\              
\hline                                                                                                          
\hline                                                                                                          
 \multicolumn{2}{|c|}{\multirow{3}{*}{Median}} & 2 & 0.15693 & 48.3609 & -- & -- & 0.15494 & 15.1423 & \textbf{0.14043} & \textbf{11.0633} \\              
 \multicolumn{2}{|c|}{}                        & 4 & 0.13568 & 36.769  & 0.8741 & 14.3363 & 0.11767 & 13.1322 & \textbf{0.10094} & \textbf{9.9086} \\
 \multicolumn{2}{|c|}{}                        & 8 & 0.72174 & 36.838  & -- & -- & 0.26285 & 15.7971 & \textbf{0.22566} & \textbf{12.7326} \\               
\hline                                                                                                          
 \multicolumn{2}{|c|}{\multirow{3}{*}{Mean}}   & 2 & 0.15694 & 48.2983 & -- & -- & 0.15773 & 15.3515 & \textbf{0.1432} & \textbf{11.1919} \\               
 \multicolumn{2}{|c|}{}                        & 4 & 0.14095 & 36.7644 & 0.8625 & 15.2399 & 0.12908 & 13.4354 & \textbf{0.10935} & \textbf{10.1732} \\
 \multicolumn{2}{|c|}{}                        & 8 & 0.72675 & 36.4789 & -- & -- & 0.27802 & 16.2705 & \textbf{0.23276} & \textbf{13.117} \\               
\hline                                                                                                          
\end{tabular}                                                                                                   
\caption{Quantitative comparison between two state-of-the-art methods, the proposed fully variational approach based on shape-from-shading (denoted by SfS), and the proposed combination of deep learning and variational methods, on the synthetic dataset. The combined solution is the most effective. ~\\~\\ }       
\label{tab:deepnetTable}                                                                          
\end{table*}

\subsection{Qualitative Comparison against the State-of-the-art on Real-world Datasets we Captured Ourselves}

\begin{figure*}[!ht]
	\centering
	\newcommand{\mywidth}{0.135\textwidth}
	\newcommand{\mywidthlr}{0.1\textwidth}
	\newcolumntype{C}{ >{\centering\arraybackslash} m{0.02\textwidth} }
	\newcolumntype{X}{ >{\centering\arraybackslash} m{\mywidth} }
	\newcolumntype{Y}{ >{\centering\arraybackslash} m{\mywidthlr} }
	\setlength\tabcolsep{6pt} 
	\begin{tabular}{YX|X|YY|XX}
		  \multicolumn{2}{c|}{Input} & \cite{Hui2016} & \multicolumn{2}{c|}{\cite{Sfsnet18}}& \multicolumn{2}{|c}{Ours}\\
		 
		 $\zz$ & $\I$ & $z$ & $z$ & $\mrho$ & $z$ & $\mrho$ \\
		 
		\includegraphics[width=\mywidthlr]{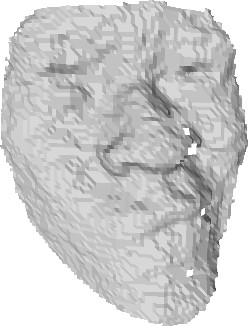}&
		\includegraphics[width=\mywidth]{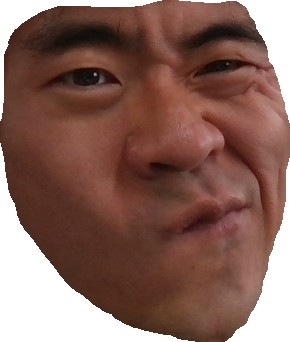}&
		\includegraphics[width=\mywidth]{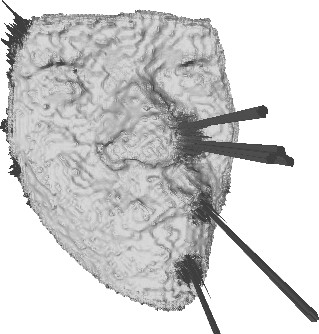}&
		\includegraphics[width=\mywidthlr]{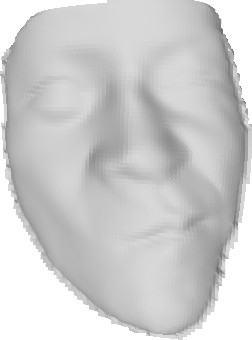}&
		\includegraphics[width=\mywidthlr]{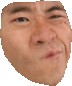}&
		\includegraphics[width=\mywidth]{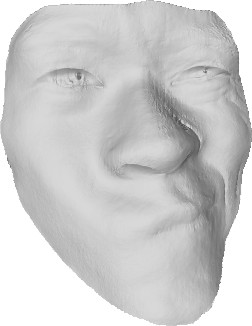}&
		\includegraphics[width=\mywidth]{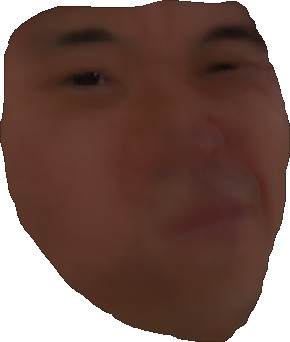}\\
		
		\includegraphics[width=\mywidthlr]{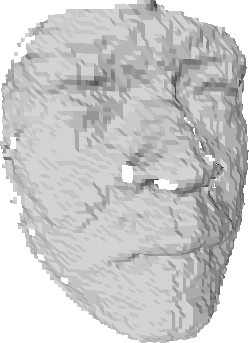}&
		\includegraphics[width=\mywidth]{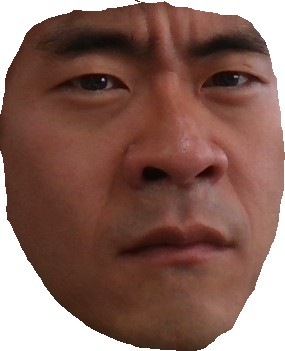}&
		\includegraphics[width=\mywidth]{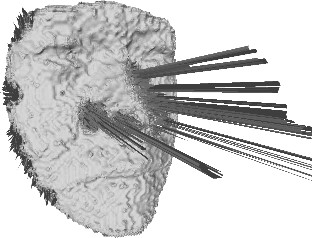}&
		\includegraphics[width=\mywidthlr]{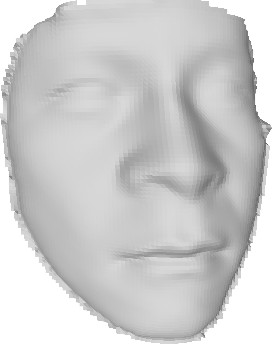}&
		\includegraphics[width=\mywidthlr]{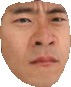}&
		\includegraphics[width=\mywidth]{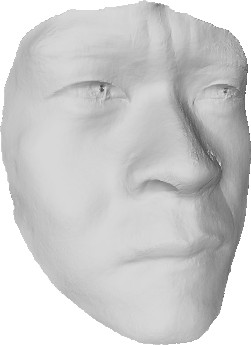}&
		\includegraphics[width=\mywidth]{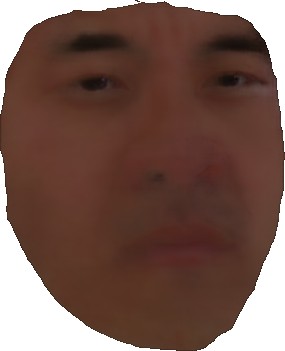}\\
		
		\includegraphics[width=\mywidthlr]{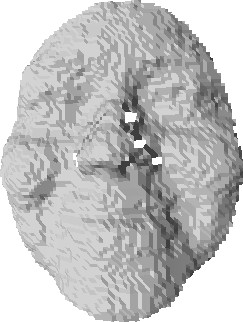}&
		\includegraphics[width=\mywidth]{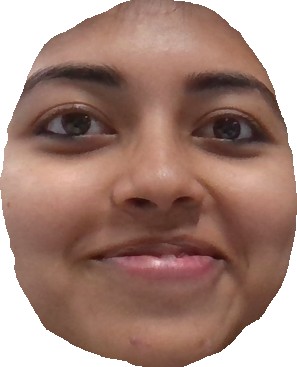}&
		\includegraphics[width=\mywidth]{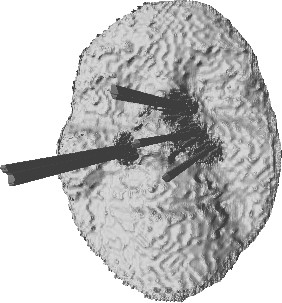}&
		\includegraphics[width=\mywidthlr]{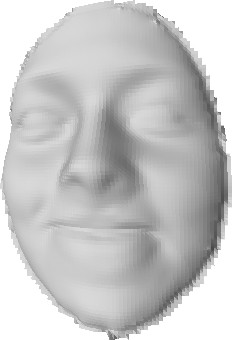}&
		\includegraphics[width=\mywidthlr]{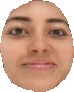}&
		\includegraphics[width=\mywidth]{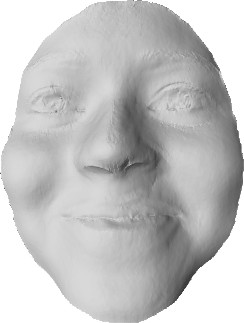}&
		\includegraphics[width=\mywidth]{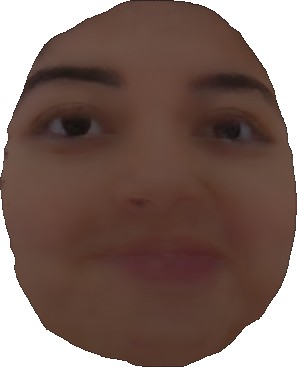}\\
		
		\includegraphics[width=\mywidthlr]{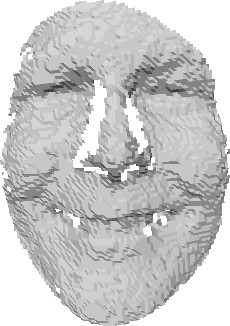}&
		\includegraphics[width=\mywidth]{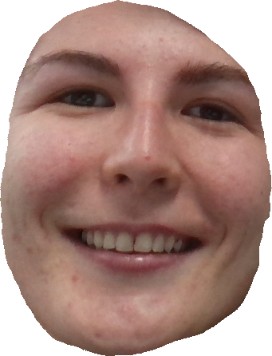}&
		\includegraphics[width=\mywidth]{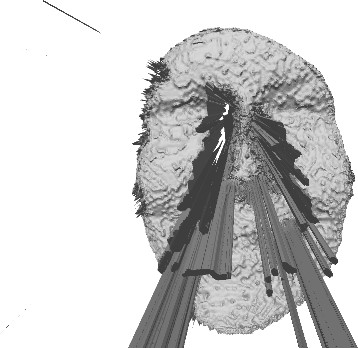}&
		\includegraphics[width=\mywidthlr]{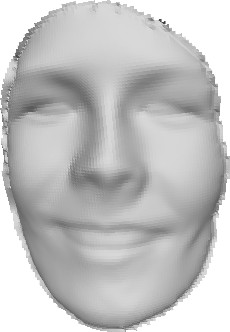}&
		\includegraphics[width=\mywidthlr]{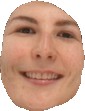}&
		\includegraphics[width=\mywidth]{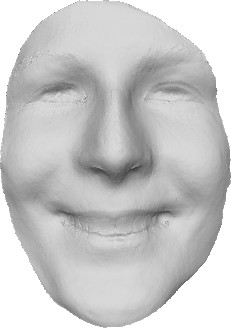}&
		\includegraphics[width=\mywidth]{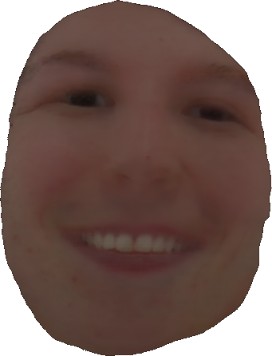}\\
		
		\includegraphics[width=\mywidthlr]{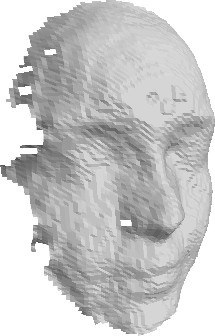}&
		\includegraphics[width=\mywidth]{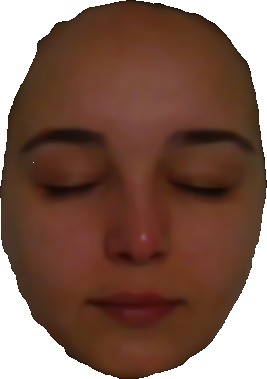}&
		\includegraphics[width=\mywidth]{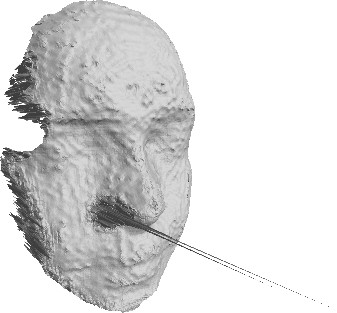}&
		\includegraphics[width=\mywidthlr]{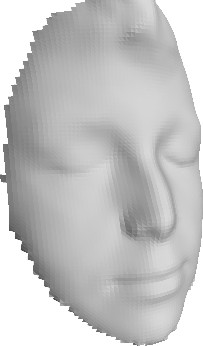}&
		\includegraphics[width=\mywidthlr]{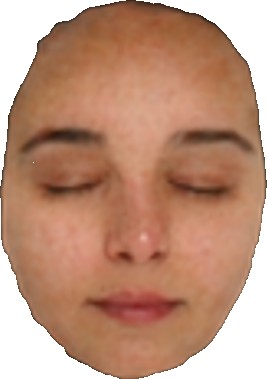}&
		\includegraphics[width=\mywidth]{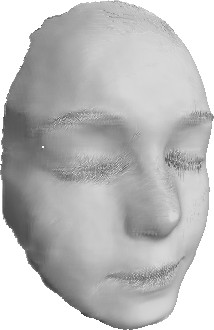}&
		\includegraphics[width=\mywidth]{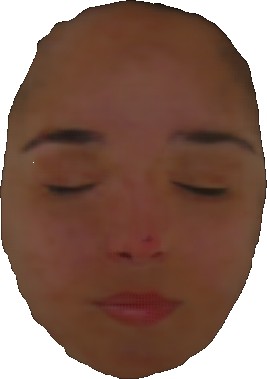}\\
		
		\includegraphics[width=\mywidthlr]{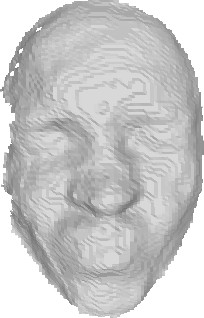}&
		\includegraphics[width=\mywidth]{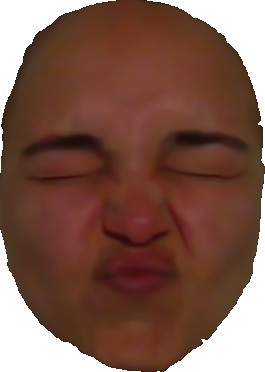}&
		\includegraphics[width=\mywidth]{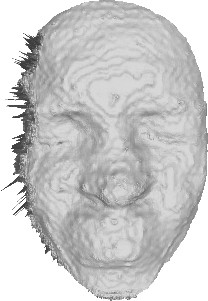}&
		\includegraphics[width=\mywidthlr]{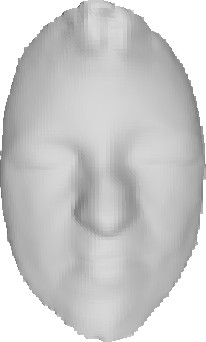}&
		\includegraphics[width=\mywidthlr]{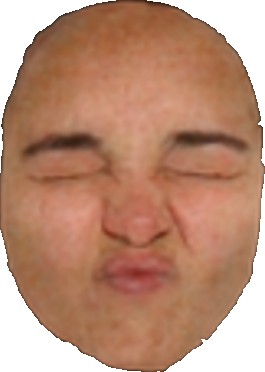}&
		\includegraphics[width=\mywidth]{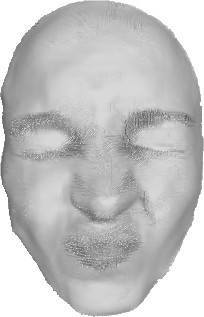}&
		\includegraphics[width=\mywidth]{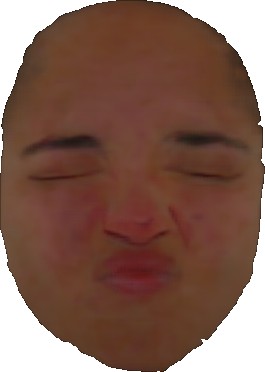}\\
	\end{tabular}
	\caption{Qualitative comparison of our reflectance learning-based results against state-of-the-art methods, on six RGB-D frames of human faces which we captured using an Intel RealSense D415 camera (scaling factor of $4$). Our method provides the most detailed 3D-reconstructions. }
	\label{fig:real_comparison_deepnet}
\end{figure*}

In Figure~\ref{fig:real_comparison_deepnet}, we show additional qualitative comparisons of our results against those from the state-of-the-art, on the dataset from Fig. 5 in the main paper. This dataset consists of RGB-D frames of human faces which we acquired ourselves using an Intel Realsense D415 camera (the scaling factor between the high-resolution RGB image and the low-resolution depth map is $4$). 

This qualitative comparison validates the conclusions from the synthetic experiment in the previous subsection: combining variational and machine learning techniques yields more detailed 3D-reconstructions than end-to-end learning solutions based on neural networks for solving the shape-from-shading~\cite{Sfsnet18} or the depth super-resolution~\cite{Hui2016} problems.

\subsection{Comparison against the State-of-the-art on a Public Real-world Dataset}

\begin{figure*}[!ht]
  \centering
  \newcommand{\mywidth}{0.12\textwidth}
  \newcommand{\mywidthlr}{0.085\textwidth}
  \newcolumntype{C}{ >{\centering\arraybackslash} m{0.02\textwidth} }
  \newcolumntype{X}{ >{\centering\arraybackslash} m{\mywidth} }
  \newcolumntype{Y}{ >{\centering\arraybackslash} m{\mywidthlr} }
  \setlength\tabcolsep{12pt} 
  \begin{tabular}{CXYXYXX}
    & $\I$ & $\zz$ & \cite{Hui2016}  & \cite{Sfsnet18} & Ours & Ground truth \\
    \rotatebox{90}{bear} &
    \includegraphics[width=\mywidth]{diligent_sfs/bear}&
    \includegraphics[width=\mywidthlr]{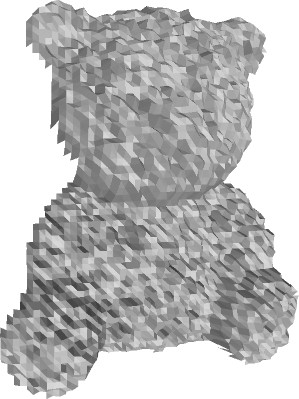}&
    \includegraphics[width=\mywidth]{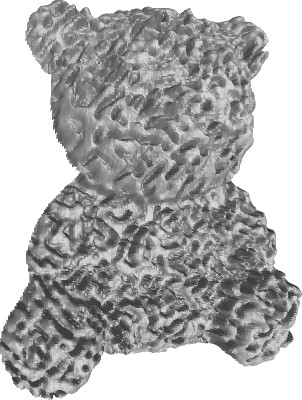}&
    \includegraphics[width=\mywidthlr]{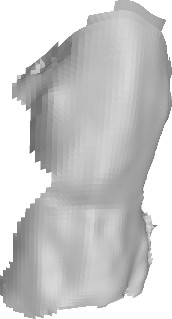}&
    \includegraphics[width=\mywidth]{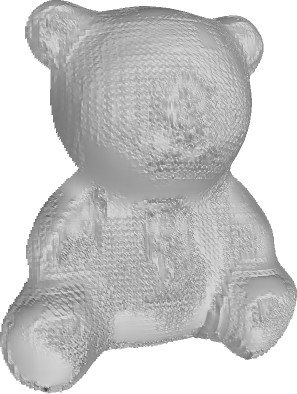}&
    \includegraphics[width=\mywidth]{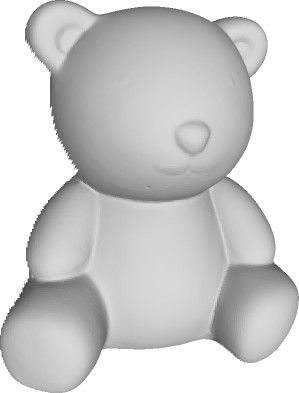} \\
    \rotatebox{90}{buddha} &
    \includegraphics[width=\mywidth]{diligent_sfs/buddha}&
    \includegraphics[width=\mywidthlr]{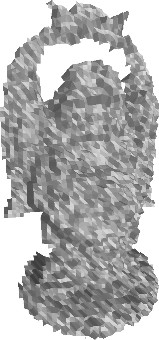}&
    \includegraphics[width=\mywidth]{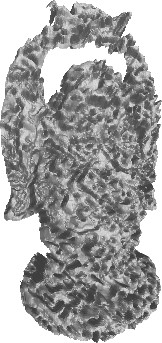}&
    \includegraphics[width=\mywidthlr]{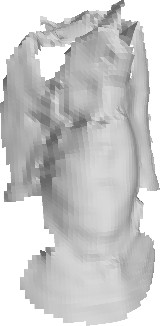}&
    \includegraphics[width=\mywidth]{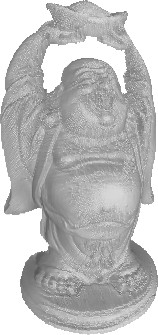}&
    \includegraphics[width=\mywidth]{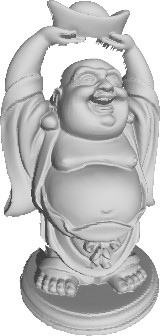} \\
    \rotatebox{90}{cat} &
    \includegraphics[width=\mywidth]{diligent_sfs/cat}&
    \includegraphics[width=\mywidthlr]{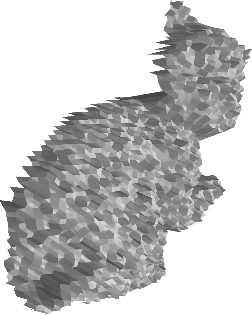}&
    \includegraphics[width=\mywidth]{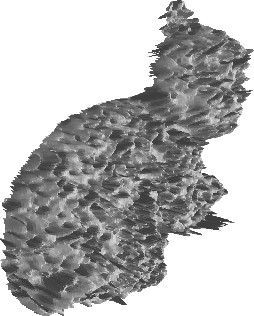}&
    \includegraphics[width=\mywidthlr]{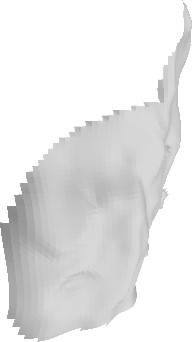}&
    \includegraphics[width=\mywidth]{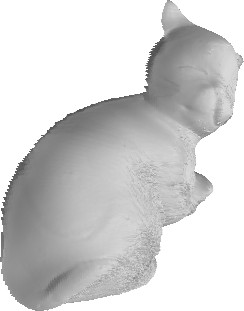}&
    \includegraphics[width=\mywidth]{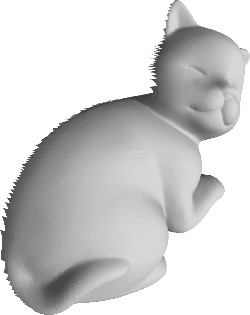} \\
    \rotatebox{90}{cow} &
    \includegraphics[width=\mywidth]{diligent_sfs/cow}&
    \includegraphics[width=\mywidthlr]{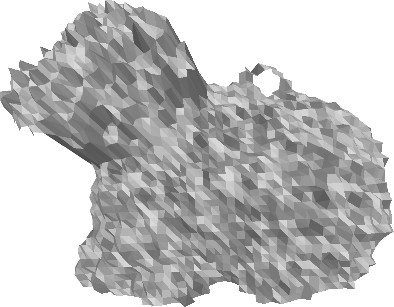}&
    \includegraphics[width=\mywidth]{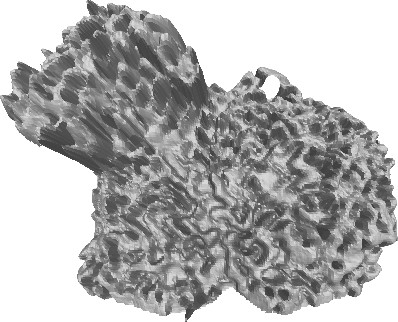}&
    \includegraphics[width=\mywidthlr]{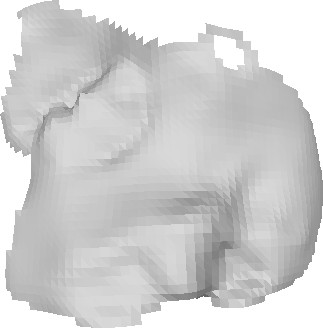}&
    \includegraphics[width=\mywidth]{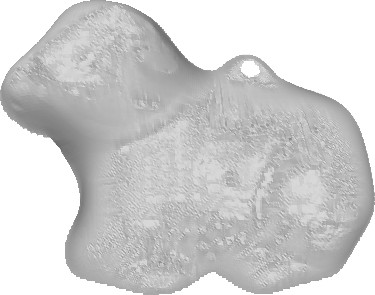}&
    \includegraphics[width=\mywidth]{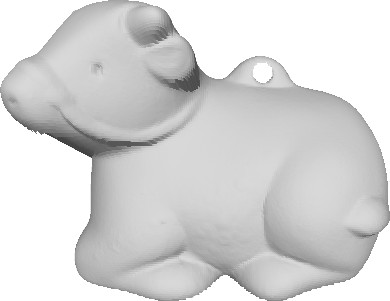} \\
    \rotatebox{90}{goblet} &
    \includegraphics[width=\mywidth]{diligent_sfs/goblet}&
    \includegraphics[width=\mywidthlr]{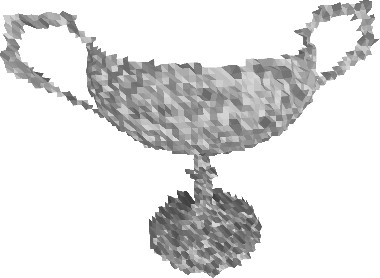}&
    \includegraphics[width=\mywidth]{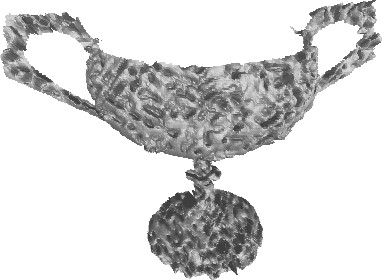}&
    \includegraphics[width=\mywidthlr]{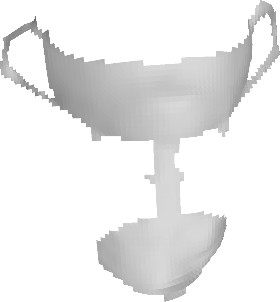}&
    \includegraphics[width=\mywidth]{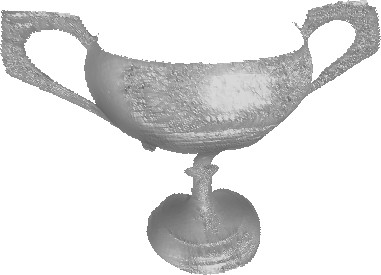}&
    \includegraphics[width=\mywidth]{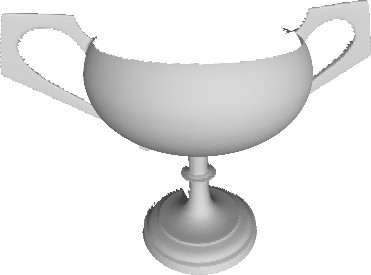} \\
    \rotatebox{90}{harvest} &
    \includegraphics[width=\mywidth]{diligent_sfs/harvest}&
    \includegraphics[width=\mywidthlr]{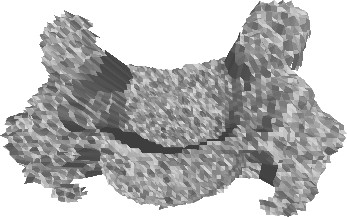}&
    \includegraphics[width=\mywidth]{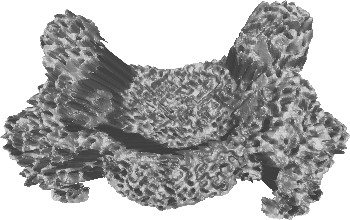}&
    \includegraphics[width=\mywidthlr]{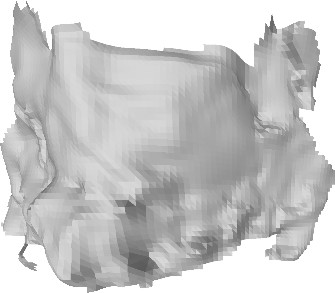}&
    \includegraphics[width=\mywidth]{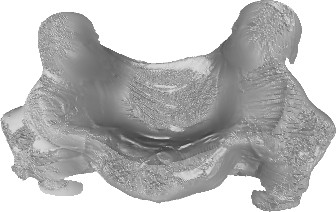}&
    \includegraphics[width=\mywidth]{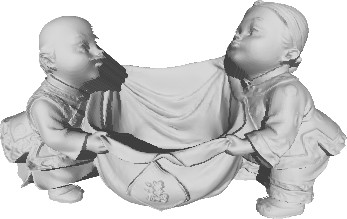} \\
    \rotatebox{90}{pot1} &
    \includegraphics[width=\mywidth]{diligent_sfs/pot1}&
    \includegraphics[width=\mywidthlr]{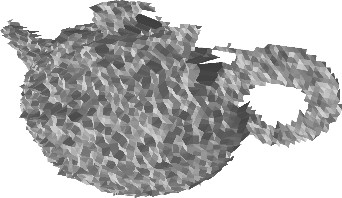}&
    \includegraphics[width=\mywidth]{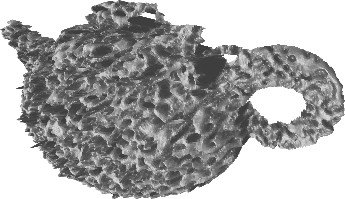}&
    \includegraphics[width=\mywidthlr]{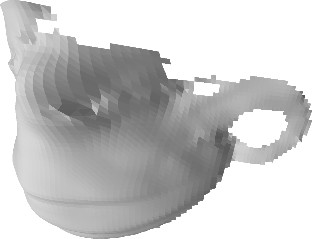}&
    \includegraphics[width=\mywidth]{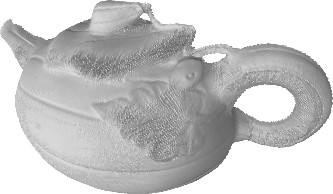}&
    \includegraphics[width=\mywidth]{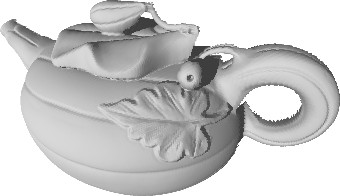} \\
    \rotatebox{90}{pot2} &
    \includegraphics[width=\mywidth]{diligent_sfs/pot2}&
    \includegraphics[width=\mywidthlr]{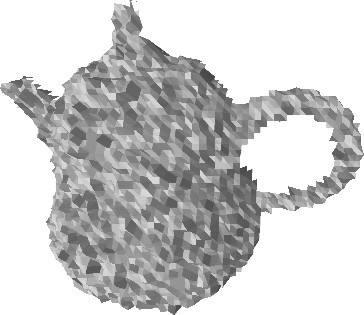}&
    \includegraphics[width=\mywidth]{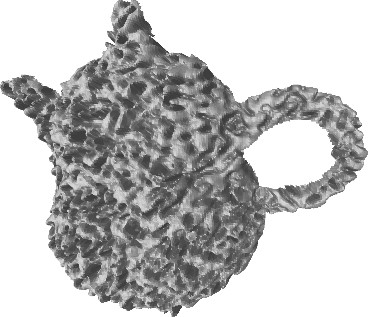}&
    \includegraphics[width=\mywidthlr]{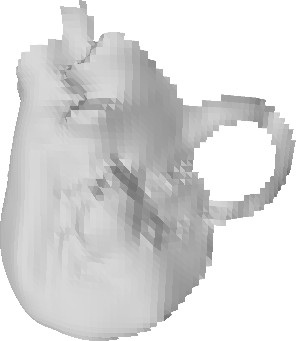}&
    \includegraphics[width=\mywidth]{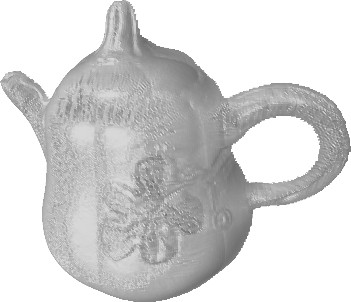}&
    \includegraphics[width=\mywidth]{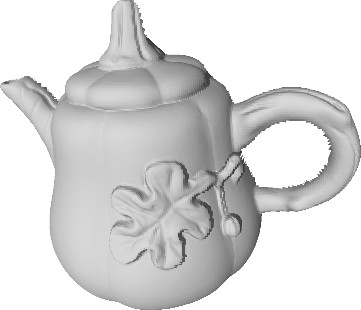} \\
    \rotatebox{90}{reading} &
    \includegraphics[width=\mywidth]{diligent_sfs/reading}&
    \includegraphics[width=\mywidthlr]{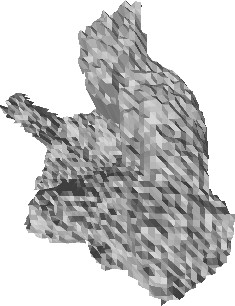}&
    \includegraphics[width=\mywidth]{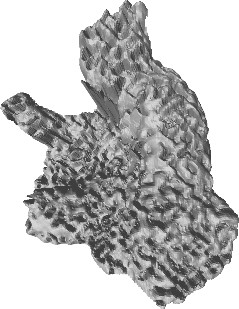}&
    \includegraphics[width=\mywidthlr]{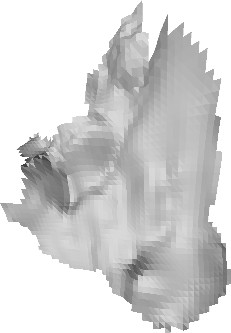}&
    \includegraphics[width=\mywidth]{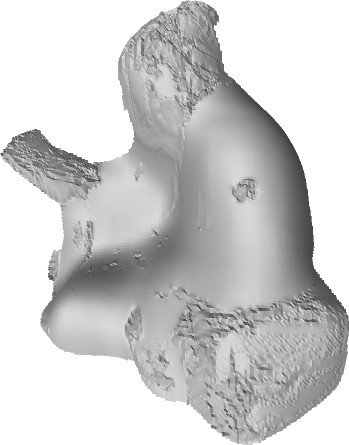}&
    \includegraphics[width=\mywidth]{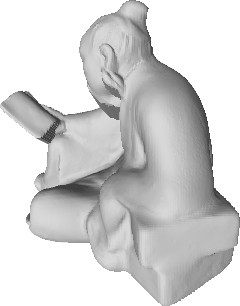} \\
  \end{tabular}
  \caption{Qualitative comparison between our method combining deep learning and variational methods, and state-of-the-art deep learning-based methods, on the DiLiGenT dataset~\cite{Shi2018} (the scaling factor is 4). Our approach outperforms the state-of-the-art in all the experiments.}
  \label{fig:diligent_comparison_deep}
\end{figure*}

Eventually, we compare qualitatively in Figure~\ref{fig:diligent_comparison_deep}, and quantitatively in Table~\ref{tab:table_diligent_deep}, the results of the proposed reflectance learning-based approach against those of the state-of-the-art, on data extracted from the DiLiGenT dataset~\cite{Shi2018}. Note that the datasets are exactly the same as the ones used for the evaluation of the fully variational solution in Figure~\ref{fig:diligent_comparison_sfs} and Table~\ref{tab:diligent_comparison_sfs}, so that the results of the fully variational solution and those of the combined approach can also be compared. 

Let us emphasize that the proposed reflectance learning-based solution was trained on a faces dataset, while none of the objects in this experiment resembles a face. Therefore, this test is rather intended as a test of robustness, and we are not expecting to overcome the results of the fully variational solution.  

Indeed, the results obtained with the combined approach are both qualitatively and quantitatively less satisfactory on this dataset than those obtained with the fully variational solution. However, they remain surprisingly competitive, in comparison with the state-of-the-art. 

Obviously, such a combination of machine learning and variational methods could still be improved by increasing the size of the training database using multiple classes of objects, but the present results already demonstrate its potential.

\begin{table*}[!ht]                                                                                          
\centering                                                                                              
\begin{tabular}{|c|c|c|c|c|c|c|c|}                                                                      
\hline                                                                                                  
\multirow{2}{*}{3D-shape} & \multirow{2}{*}{SF} & \multicolumn{2}{c|}{\cite{Hui2016}}  & \multicolumn{2}{c|}{\cite{Sfsnet18}}  & \multicolumn{2}{c|}{Ours} \\                                    
\cline{3-8}                                                                                                  
 &  & RMSE & MAE & RMSE & MAE & RMSE & MAE \\                                                           
\hline                                                                                                  
\hline                                                                                                  
     & 2 & 0.010946          & 62.708  & -- & -- & \textbf{0.0046569} & \textbf{22.5961} \\                           
bear & 4 & 0.010609          & 49.8753 & 0.1096 & 41.9262 & \textbf{0.0086235} & \textbf{23.2417} \\           
     & 8 & \textbf{0.012821} & 36.8812 & -- & -- & 0.018246 & \textbf{30.7021} \\                           
\hline                                                                                                  
       & 2 & 0.011778          & 63.0557 & -- & -- & \textbf{0.0082909} & \textbf{29.6526} \\                          
buddha & 4 & 0.012539          & 52.2028 & 0.0518 & 40.1120 & \textbf{0.011945} & \textbf{33.5974} \\          
       & 8 & \textbf{0.015423} & 45.132  & -- & -- & 0.019568 & \textbf{41.0636} \\                            
\hline                                                                                                  
    & 2 & 0.013194          & 62.903  & -- & -- & \textbf{0.008389} & \textbf{15.1466} \\                            
cat & 4 & 0.0137            & 50.278  & 0.0647 & 36.9720 & \textbf{0.013534} & \textbf{19.1494} \\                
    & 8 & \textbf{0.015258} & 38.3265 & -- & -- & 0.023363 & \textbf{26.7149} \\                           
\hline                                                                                                  
    & 2 & 0.011679          & 64.5302 & -- & -- & \textbf{0.0053628} & \textbf{17.6086} \\                          
cow & 4 & 0.011237          & 50.6826 & 0.0562 & 39.3336 & \textbf{0.0092811} & \textbf{18.8318} \\            
    & 8 & \textbf{0.014157} & 42.9122 & -- & -- & 0.017689 & \textbf{21.1007} \\                           
\hline                                                                                                  
       & 2 & 0.013153          & 61.8508 & -- & -- & \textbf{0.011713} & \textbf{30.1888} \\                           
goblet & 4 & \textbf{0.01379}  & 48.7097 & 0.1414 & 36.4712 & 0.017615 & \textbf{29.6286} \\             
       & 8 & \textbf{0.016659} & 36.6476 & -- & -- & 0.03133 & \textbf{28.7208} \\                            
\hline                                                                                                  
        & 2 & 0.0167            & 64.113  & -- & -- & \textbf{0.016649} & \textbf{39.602} \\                               
harvest & 4 & \textbf{0.019409} & 53.9958 & 0.1757 & 54.1461 & 0.024208 & \textbf{41.0901} \\         
        & 8 & \textbf{0.028625} & 44.4953 & -- & -- & 0.037441 & \textbf{41.1994} \\                           
\hline                                                                                                  
     & 2 & 0.011218         & 61.9779 & -- & -- & \textbf{0.0070793} & \textbf{18.4819} \\                          
pot1 & 4 & 0.011597         & 50.0199 & 0.1051 & 35.0139 & \textbf{0.010794} & \textbf{18.6248} \\            
     & 8 & \textbf{0.01495} & 40.4749 & -- & -- & 0.019198 & \textbf{20.5408} \\                            
\hline                                                                                                  
     & 2 & 0.010693          & 61.9083 & -- & -- & \textbf{0.0057831} & \textbf{20.0908} \\                          
pot2 & 4 & 0.011123          & 50.5484 & 0.0575 & 32.0884 & \textbf{0.0090011} & \textbf{20.7887} \\           
     & 8 & \textbf{0.014105} & 40.3902 & -- & -- & 0.016243 & \textbf{23.1403} \\                           
\hline                                                                                                  
        & 2 & 0.012058          & 61.2583 & -- & -- & \textbf{0.0098101} & \textbf{20.5263} \\                          
reading & 4 & \textbf{0.012927} & 49.0756 & 0.0817 & 55.4988 & 0.015531 & \textbf{24.2634} \\         
        & 8 & \textbf{0.017714} & 41.0243 & -- & -- & 0.028793 & \textbf{28.8291} \\                           
\hline                                                                                                  
\hline                                                                                                  
       & 2 & 0.011778          & 62.708  & -- & -- & \textbf{0.0082909} & \textbf{20.5263} \\                           
Median & 4 & 0.012539          & 50.278  & 0.0732 & 38.1528 & \textbf{0.011945} & \textbf{23.2417} \\           
       & 8 & \textbf{0.015258} & 40.4749 & -- & -- & 0.019568 & \textbf{28.7208} \\                           
\hline                                                                                                  
     & 2 & 0.01238           & 62.7006 & -- & -- & \textbf{0.0086371} & \textbf{23.766} \\                            
Mean & 4 & \textbf{0.012992} & 50.5987 & 0.0918 & 41.2045 & 0.013392 & \textbf{25.4684} \\            
     & 8 & \textbf{0.016635} & 40.6982 & -- & -- & 0.023541 & \textbf{29.1124} \\                           
\hline                                                                                                  
\end{tabular}                                                                                           
\caption{Quantitative comparison between other state-of-the-art methods and our method combining machine learning and variational methods. Although the results are not as accurate as the fully variational solution (cf. Table~\ref{tab:diligent_comparison_sfs}), since none of the objects here resembles the faces from the training database, they remain superior to the state-of-the-art.}
\label{tab:table_diligent_deep}                                                                   
\end{table*}

\section{Evaluation of the Multi-shot Approach Based on Photometric Stereo}  
\label{sec:supp_5}

\subsection{Creation of the Synthetic Data}

\begin{figure*}[!ht]
  \centering
  \newcolumntype{X}{ >{\centering\arraybackslash} m{0.185\textwidth} }
  \setlength\tabcolsep{14pt} 
  \begin{tabular}{XXXX}
    Lucy~\cite{Levoy2005data} & Thai Statue~\cite{Levoy2005data} & Armadillo~\cite{Levoy2005data} & Joyful Yell~\cite{Bendansie2015} \\
    ``ebsd'' albedo & ``ebsd'' albedo  & ``mandala'' albedo & ``rectcircle'' albedo\\
    \includegraphics[width=0.185\textwidth]{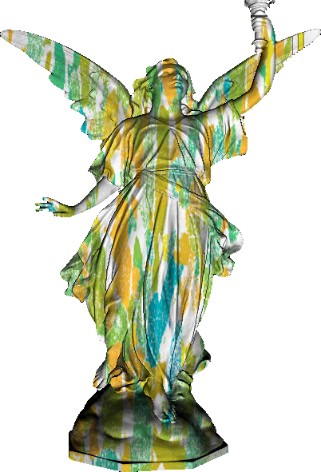}&
    \includegraphics[width=0.185\textwidth]{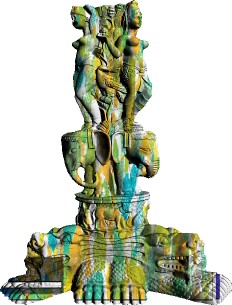}&
    \includegraphics[width=0.185\textwidth]{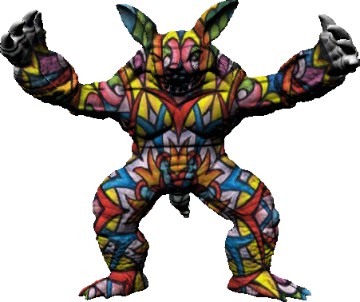}&        
    \includegraphics[width=0.185\textwidth]{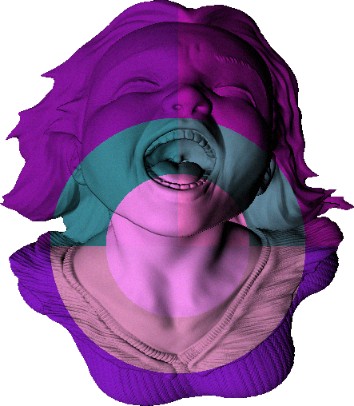}\\
    \includegraphics[width=0.185\textwidth]{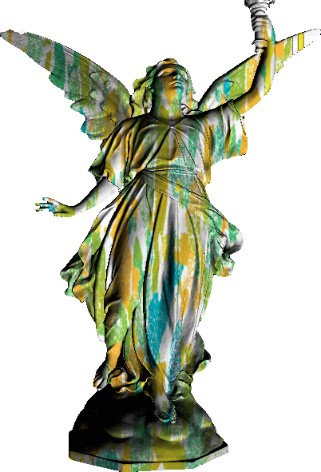}&
    \includegraphics[width=0.185\textwidth]{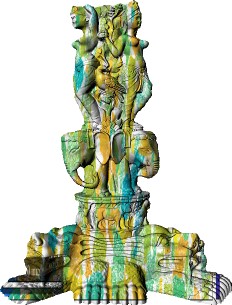}&
    \includegraphics[width=0.185\textwidth]{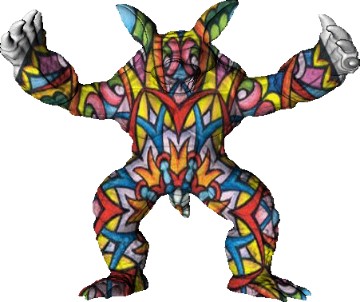}&
    \includegraphics[width=0.185\textwidth]{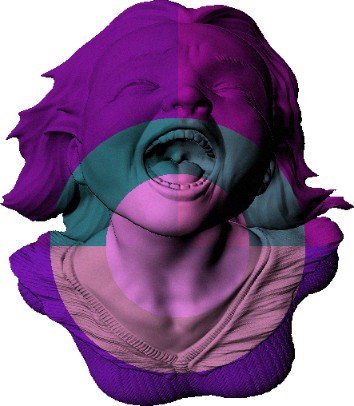}\\
    \includegraphics[width=0.185\textwidth]{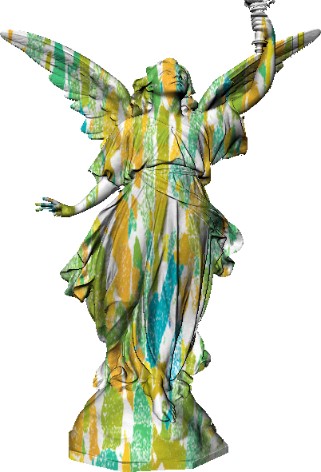}&
    \includegraphics[width=0.185\textwidth]{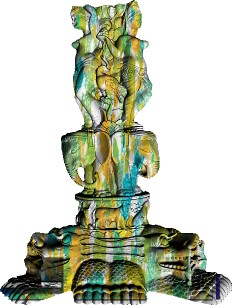}&
    \includegraphics[width=0.185\textwidth]{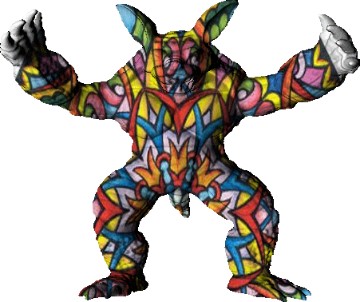}&        
    \includegraphics[width=0.185\textwidth]{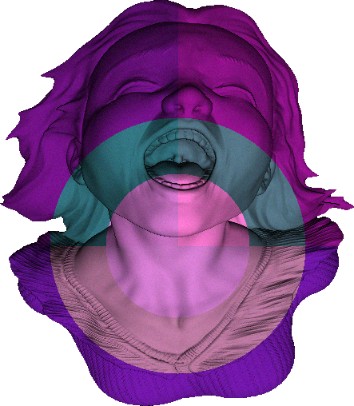}
  \end{tabular}
  \caption{Illustration of the synthetic RGB data used for quantitatively evaluating the multi-shot depth super-resolution approach based on photometric stereo. Each row represents a different illumination condition. Remark that much more complex albedo maps are considered, in comparison with the ones used in the single-shot approach, cf. Figure~\ref{fig:synthetic_data_sfs}.}
  \label{fig:synthetic_data_ups}
\end{figure*}

In order to quantitatively evaluate the proposed photometric stereo-based solution, we consider the same four 3D-shapes as in the shape-from-shading experiments, i.e. ``Lucy'', ``Thai Statue'', ``Armadillo'' and ``Joyful Yell''. However, this time we consider much more complex albedo maps since the multi-shot approach is not limited to piecewise-constant albedos. The albedo maps we consider are ``ebsd''\footnote{\url{https://mtex-toolbox.github.io/files/doc/EBSDSpatialPlots.html}}, ``mandala''\footnote{\url{http://www.cleverpedia.com/mandala-coloring-books-20-coloring-books-with-brilliant-kaleidoscope-designs/}} and ``rectcircle''. The rest of the process for creating the dataset (rendering the high-resolution RGB and low-resolution depth images, and adding noise) is exactly the same as for shape-from-shading, except that multiple RGB images are acquired under randomly varying lighting. Three RGB images of each dataset under three different illumination conditions are presented in Figure~\ref{fig:synthetic_data_ups}, and the corresponding depth maps are those from Figure~\ref{fig:synthetic_data_sfs}.

\subsection{Selecting the Number of Images and Tuning the Hyper-parameters}

\begin{figure*}[!ht]
  \centering
  \newcolumntype{C}{>{\centering\arraybackslash} m{0.0025\textwidth} }
  \newcolumntype{X}{>{\centering\arraybackslash} m{0.31\textwidth} }
  \setlength\tabcolsep{4pt} 
  \begin{tabular}{CXCXCX}
    \rotatebox{90}{RMSE (depth)}&
    \includegraphics[width=0.3\textwidth]{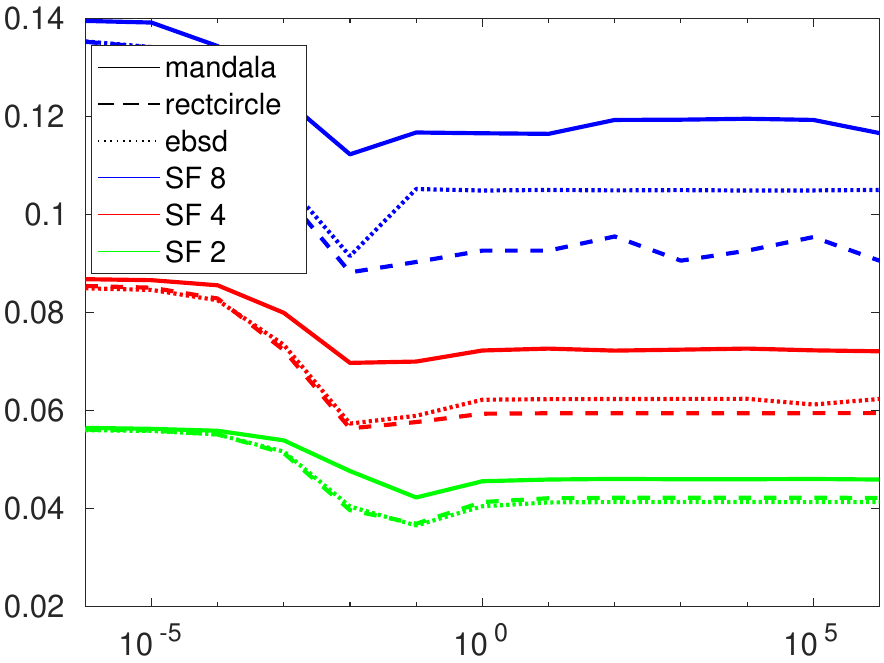} &
    \rotatebox{90}{MAE (depth)}&
    \includegraphics[width=0.3\textwidth]{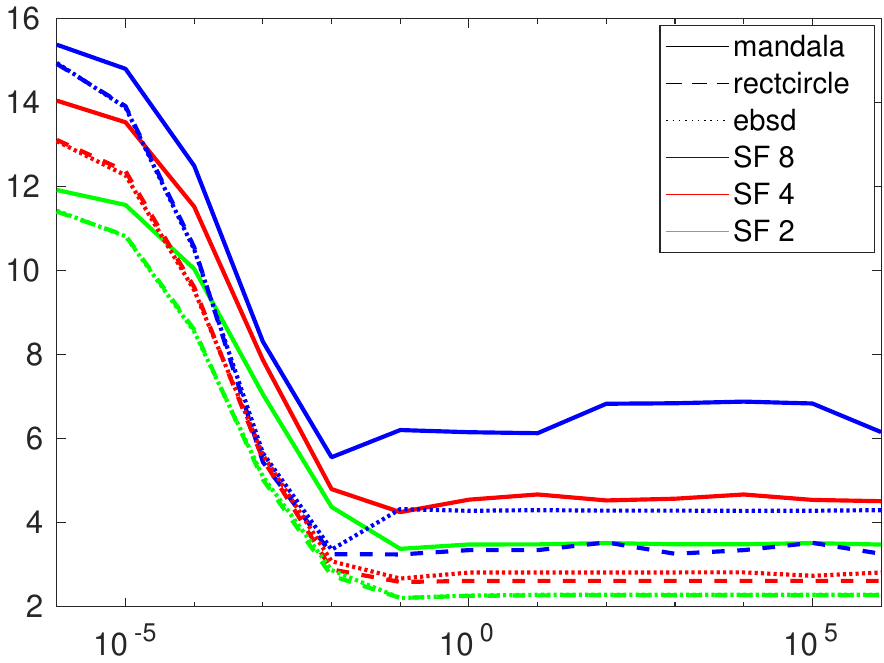} &
    \rotatebox{90}{RMSE (albedo)}&
    \includegraphics[width=0.3\textwidth]{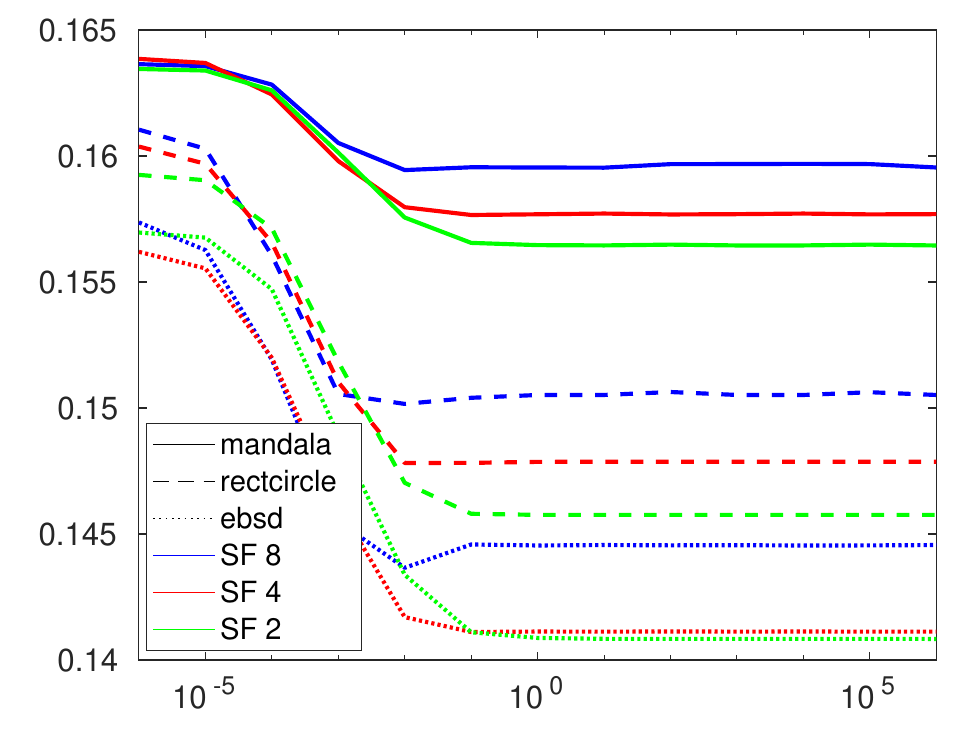}\\
    &$\gamma$&&$\gamma$&&$\gamma$
  \end{tabular}
  \caption{Impact of the parameter $\gamma$ on the accuracy of the albedo and depth estimates using our multi-shot photometric stereo approach ($n = 10$ in this experiment). Based on these results, the value $\gamma = 0.01$ was retained.}
  \label{fig:parameter_evaluation_ups} 
\end{figure*}

Figure~\ref{fig:parameter_evaluation_ups} illustrates the effect of the hyper-parameter $\gamma$ on shape and reflectance estimation. For this purpose, we consider sets of $n=10$ images from the Joyful Yell dataset, and evaluate the RMSE and MAE on depth, as well as the RMSE on albedo, as functions of the number of input images. As can be seen, when $\gamma\to 0$ the estimated depth map sticks to the noisy input, thus results are deceiving. But as soon as $\gamma$ is large enough, photometric stereo drives super-resolution and the accuracy dramatically increases. Interestingly, results remain stable even when $\lambda \to \infty$. This tends to indicate that the ambiguities of uncalibrated photometric stereo vanish as soon as a depth prior is available: it is not necessary to seek a compromise between the depth prior and the photometric 3D-reconstruction, only to plug the information from the former into the latter.

\begin{figure*}[!ht]
  \centering
  \newcolumntype{C}{>{\centering\arraybackslash} m{0.0025\textwidth} }
  \newcolumntype{X}{>{\centering\arraybackslash} m{0.43\textwidth} }
  \setlength\tabcolsep{4pt} 
  \begin{tabular}{CXCX}
    \rotatebox{90}{RMSE (depth)}&
    \includegraphics[width=0.45\textwidth]{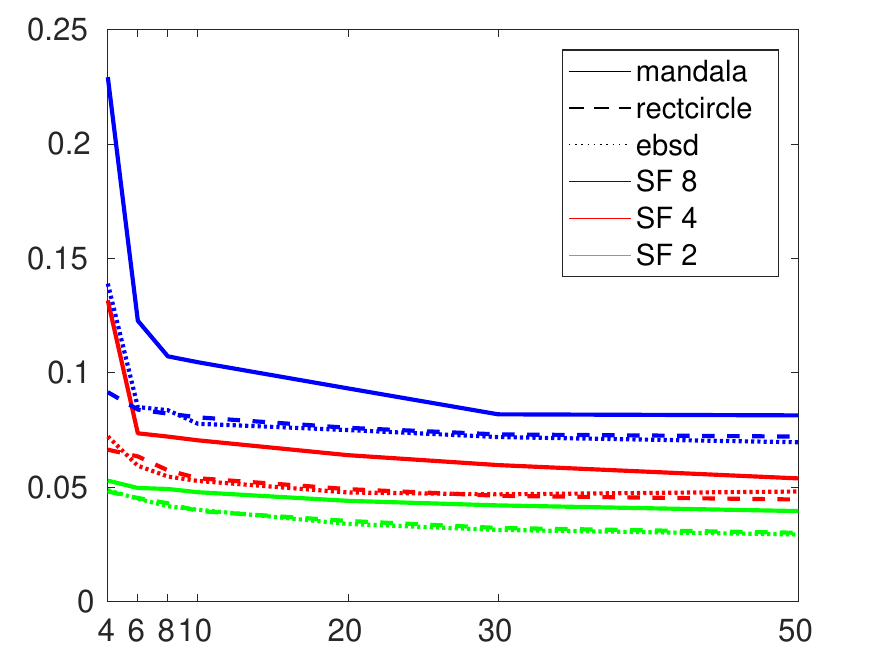} &
    \rotatebox{90}{MAE (depth)}&
    \includegraphics[width=0.45\textwidth]{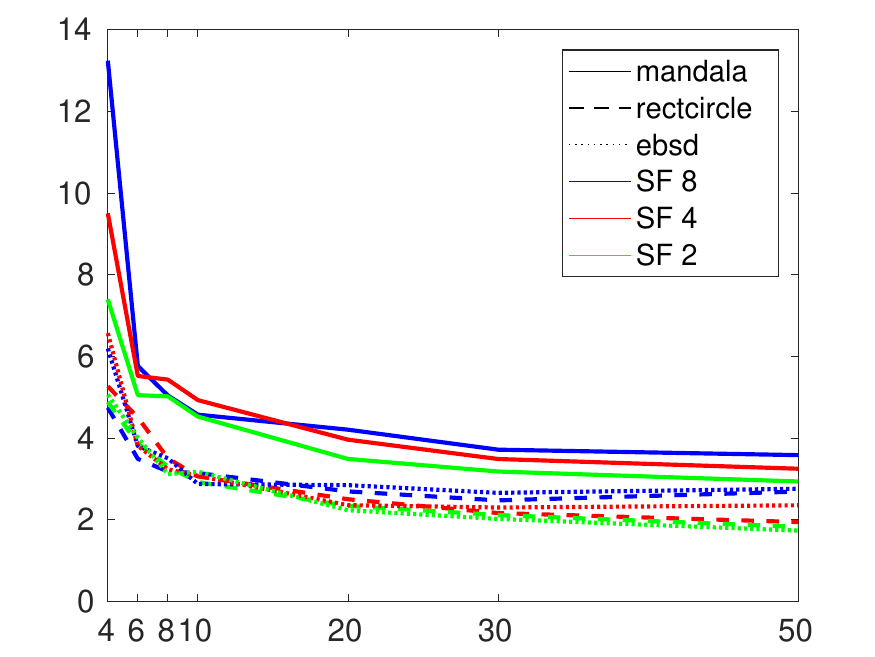}\\
        &$n$&&$n$\\
    \rotatebox{90}{RMSE (albedo)}&
    \includegraphics[width=0.45\textwidth]{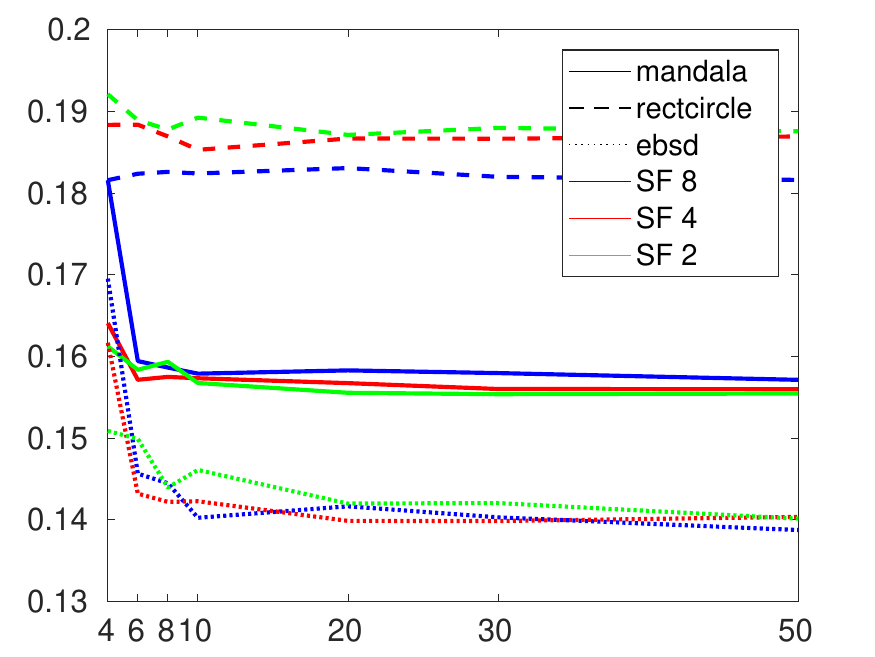}&
    \rotatebox{90}{Runtime ($s$)}&
    \includegraphics[width=0.45\textwidth]{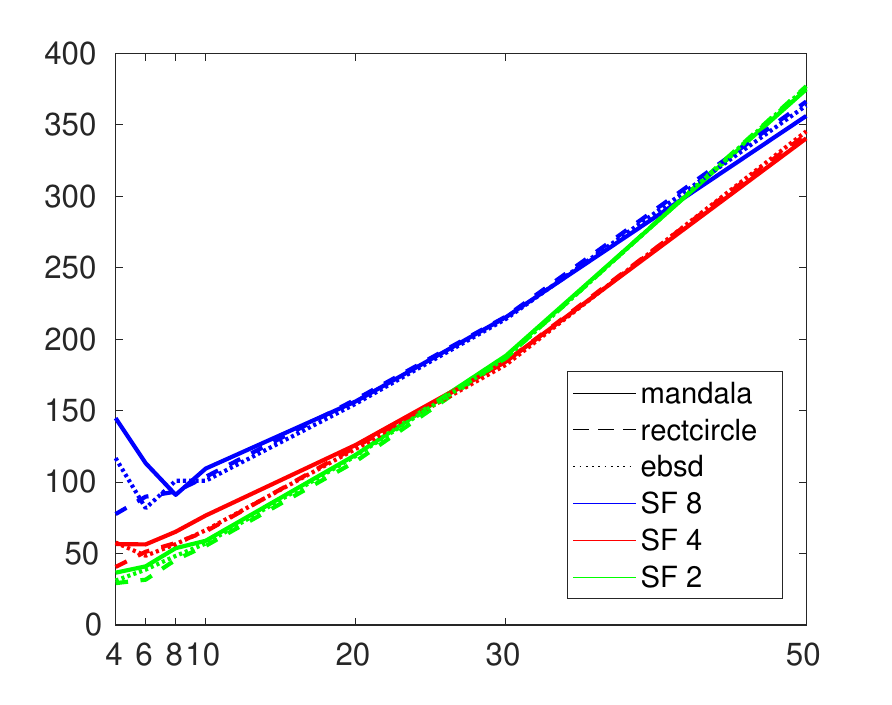}\\
    &$n$&&$n$
  \end{tabular}
  \caption{Impact of the number of images $n$ on the accuracy of the albedo and depth estimates using our multi-shot photometric stereo approach ($\gamma = 0.01$ in this experiment). The range $n \in [10,30]$ represents a reasonable compromise between accuracy and runtime. }
  \label{fig:no_img_evaluation_ups} 
\end{figure*}

Next, we evaluate the number $n$ of input RGB images which would result in the best compromise between accuracy of the 3D-reconstruction and runtime. For this purpose, we consider once again the Joyful Yell synthetic dataset, and evaluate the RMSE and MAE on depth, the RMSE on albedo  and the total runtime required to attain convergence, as functions of $n$. As can be seen in Figure~\ref{fig:no_img_evaluation_ups}, the accuracy of the estimation very quickly increases with $n$, while the runtime increases linearly with $n$. Overall, the choice $n \in [10,30]$ seems to represent a good compromise. 

\subsection{Comparison against the State-of-the-art on the Synthetic Dataset}

Next, we compare our multi-shot approach against the state-of-the-art, on all the synthetic datasets (consistently with the results from the previous subsection, $n=20$ images are considered for each dataset, and $\gamma = 0.01$ in all the experiments). Our results are expected to overcome both pure depth super-resolution and pure uncalibrated photometric stereo, as well as single-shot depth refinement methods acting on low-resolution data. 

To highlight the interest of an explicit photometric model, we first compare our results against an image-based multi-shot depth super-resolution approach adapted from~\cite{Werlberger2009,Unger2010}. It is a personal combination of these papers which achieves variational depth super-resolution by fusing the $n$ low-resolution depth maps, while regularising the gradient of the estimated high-resolution depth map in an anisotropic manner. Here, the anisotropy coefficient is derived from the gradients of the RGB image. This approach is thus a ``pure depth super-resolution'' one, which uses RGB clues but without any explicit photometric model.  

In contrast, we also consider the ``pure'' uncalibrated photometric stereo method from~\cite{Papadhimitri2014b}, which estimates lighting, albedo and high-resolution geometry from the $n$ high-resolution RGB images. In this method, an explicit photometric model is used, as in ours, yet no low-resolution depth clue is considered hence the underlying bas-relief ambiguity may affect the quality of the results.  

As in the evaluation of the shape-from-shading-based approach from Section~\ref{sec:supp_3}, we also show the results of RGB-D refinement~\cite{Or-El2015} applied to the low-resolution RGB-D frame, selecting one image out of $n$. 

\begin{figure*}[!ht]
  \centering
  \newcommand{\mywidth}{0.15\textwidth}
  \newcommand{\mywidthlr}{0.1\textwidth}
  \newcolumntype{C}{ >{\centering\arraybackslash} m{0.03\textwidth} }
  \newcolumntype{X}{ >{\centering\arraybackslash} m{\mywidth} }
  \setlength\tabcolsep{17pt} 
  \begin{tabular}{CXXXX}
    & Lucy& Thai Statue & Armadillo & Joyful Yell \\
    & ``ebsd'' & ``esbd'' & ``mandala'' & ``rectcircle'' \\
    \rotatebox{90}{\cite{Werlberger2009,Unger2010}} &
    \includegraphics[width=\mywidth]{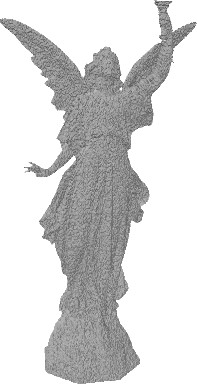}&
    \includegraphics[width=\mywidth]{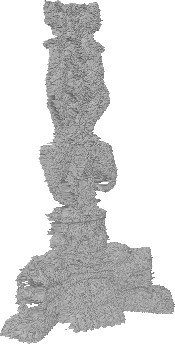}\qquad&\qquad
    \includegraphics[width=\mywidth]{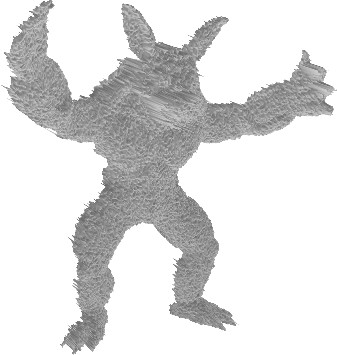}&
    \includegraphics[width=\mywidth]{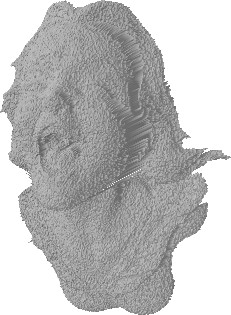} \\
    \rotatebox{90}{\cite{Papadhimitri2014b}} &
    \includegraphics[width=\mywidth]{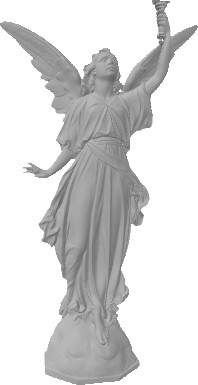}&
    \includegraphics[width=\mywidth]{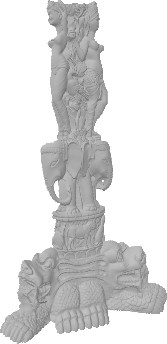}&
    \includegraphics[width=\mywidth]{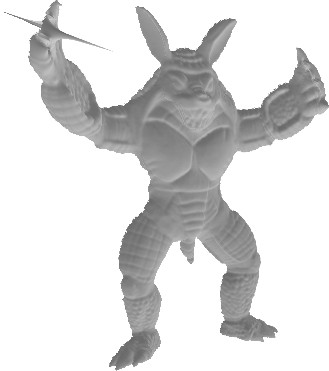}&
    \includegraphics[width=\mywidth]{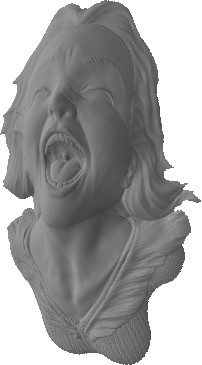} \\
    \rotatebox{90}{\cite{Or-El2015}} &
    \includegraphics[width=\mywidthlr]{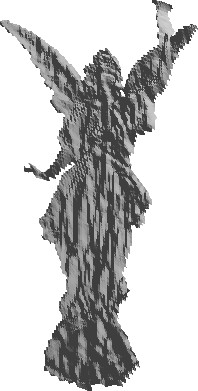}&
    \includegraphics[width=\mywidthlr]{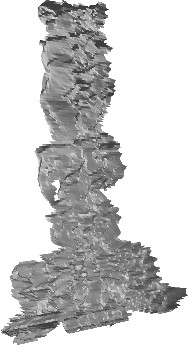}&
    \includegraphics[width=\mywidthlr]{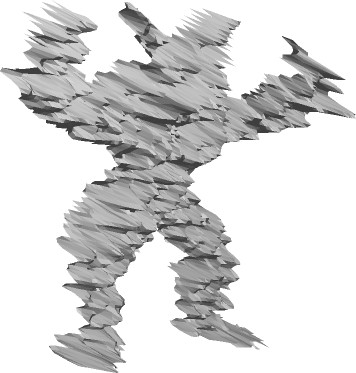}&
    \includegraphics[width=\mywidthlr]{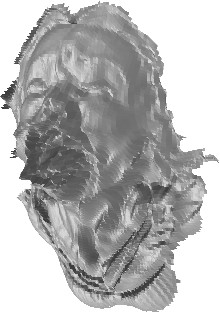} \\
    \rotatebox{90}{Ours} &
    \includegraphics[width=\mywidth]{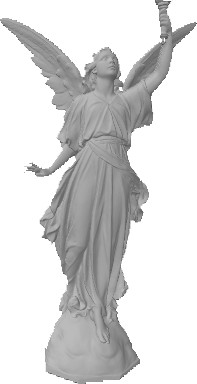}&
    \includegraphics[width=\mywidth]{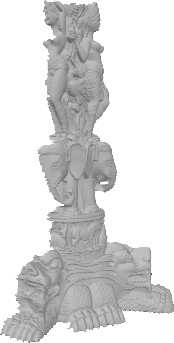}&
    \includegraphics[width=\mywidth]{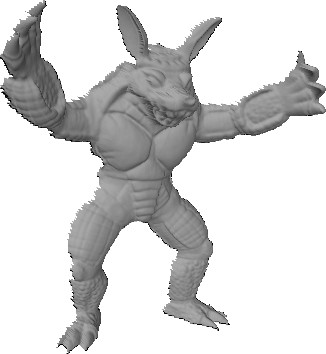}&
    \includegraphics[width=\mywidth]{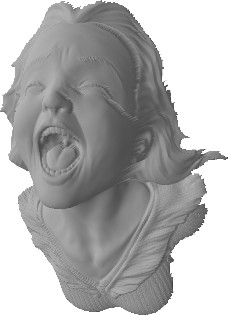}
  \end{tabular}
  \caption{Qualitative comparison of our multi-shot approach against state-of-the-art methods, on four synthetic datasets (scaling factor of 4). Image-based depth super-resolution adapted from~\cite{Werlberger2009,Unger2010} results in noisy geometry, uncalibrated photometric stereo results from~\cite{Papadhimitri2014b} are slightly flattened due to the underlying bas-relief ambiguity, and RGB-D fusion~\cite{Or-El2015} of the low-resolution data is not really successful here. In comparison, the results of the proposed method are extremely satisfactory. }
  \label{fig:synthetic_comparison_ups}
\end{figure*}

The qualitative comparison in Figure~\ref{fig:synthetic_comparison_ups}, and the quantitative ones in Table~\ref{tab:table_ups_comp}, show that our methods result in much more satisfactory high-resolution geometry, in comparison with these methods. This proves that using an explicit model for driving image-based depth super-resolution, and using low-resolution depth clues to disambiguate uncalibrated photometric stereo, both are worthwile. 

\begin{table*} [!ht]                                                                                                            
\centering                                                                                                                   
\begin{tabular}{|c|c|c|cc|cc|cc|cc|}                                                                                       
\hline                                                                                                                       
\multirow{2}{*}{Albedo} & \multirow{2}{*}{3D-shape} & \multirow{2}{*}{SF} & \multicolumn{2}{c|}{Image Based depth SR} & \multicolumn{2}{c|}{\cite{Papadhimitri2014b}$^*$} & \multicolumn{2}{c|}{\cite{Or-El2015}} & \multicolumn{2}{c|}{Ours} \\\cline{4-11}                                                                                                                     
 & & & RMSE & MAE & RMSE & MAE & RMSE & MAE & RMSE & MAE \\\hline                                                                                                                     
\hline                                                                                                                       
\multirow{12}{*}{mandala}&    & 2 & 0.031468 & 46.4149 & 0.51996 & 17.4225 & 0.4320 & 69.7311 & \textbf{0.023266} & \textbf{2.883} \\                    
& Armadillo                   & 4 & 0.042467 & 43.5403 & --  & -- & 0.3948 & 63.7382 & \textbf{0.037789} & \textbf{2.8391} \\   
 &                            & 8 & 0.088849 & 42.6184 & --  & -- & 0.5961 & 83.7853 & \textbf{0.073928} & \textbf{2.9196} \\                    
\cline{2-11}                                                                                                                       
 &                            & 2 & 0.043889 & 46.4903 & 0.37197 & 15.7192 & 0.4755 & 84.6189 & \textbf{0.036334} & \textbf{3.5842} \\                   
 & Lucy                       & 4 & 0.065857 & 44.0677 & --  & -- & 0.4951 & 82.0516 & \textbf{0.051316} & \textbf{3.6142} \\         
 &                            & 8 & 0.12668  & 42.8905 & --  & -- & 0.5231 & 64.7317 & \textbf{0.084713} & \textbf{4.7864} \\                    
\cline{2-11}                                                                                                                       
 &                            & 2 & 0.048887 & 45.0552 & 1.0735  & 14.2243 & 0.3757 & 70.2724 & \textbf{0.044198} & \textbf{3.3143} \\                     
 & Joyful Yell                & 4 & 0.069088 & 42.644  & --  & -- & 0.2985 & 55.6927 & \textbf{0.063392} & \textbf{3.6407} \\          
 &                            & 8 & 0.13103  & 40.0426 & --  & -- & 0.4240 & 44.2549 & \textbf{0.1046}   & \textbf{3.753} \\                        
\cline{2-11}                                                                                                                       
 &                            & 2 & 0.032432 & 47.8575 & 0.37738 & 13.372  & 0.4615 & 70.3271 & \textbf{0.022446} & \textbf{3.579} \\                    
 & Thai Statue                & 4 & 0.053061 & 45.5618 & --  & -- & 0.4211 & 90.2134 & \textbf{0.036245} & \textbf{3.6985} \\     
&                             & 8 & 0.094911 & 43.838  & --  & -- & 0.3371 & 53.2791 & \textbf{0.049733} & \textbf{4.1133} \\                    
\hline                                                                                                                       
\multirow{12}{*}{rectcircle}& & 2 & 0.028459 & 41.506  & 0.52582 & 18.0902 & 0.2844 & 55.3096 & \textbf{0.020885} & \textbf{2.0047} \\                    
& Armadillo                   & 4 & 0.038966 & 38.7345 & --  & -- & 0.3031 & 48.1000 & \textbf{0.035145} & \textbf{1.9458} \\
 &                            & 8 & 0.11182  & 36.3801 & --  & -- & 0.5805 & 80.4625 & \textbf{0.073139} & \textbf{2.1436} \\                    
\cline{2-11}                                                                                                                       
 &                            & 2 & 0.040635 & 42.3051 & 0.32285 & 13.6126 & 0.4868 & 85.9076 & \textbf{0.026858} & \textbf{1.8617} \\                   
 &Lucy                        & 4 & 0.062747 & 39.0783 & --  & -- & 0.4685 & 75.9166 & \textbf{0.041968} & \textbf{2.2851} \\    
 &                            & 8 & 0.12325  & 37.956  & --  & -- & 0.3767 & 56.5020 & \textbf{0.075311} & \textbf{3.8793} \\                     
\cline{2-11}                                                                                                                       
 &                            & 2 & 0.045765 & 39.9946 & 0.84162 & 11.4847 & 0.2012 & 41.3053 & \textbf{0.038698} & \textbf{2.7879} \\                   
 & Joyful Yell                & 4 & 0.064537 & 37.1175 & --  & -- & 0.3189 & 37.2107 & \textbf{0.053871} & \textbf{3.1022} \\      
 &                            & 8 & 0.09492  & 34.7218 & --  & --  & 0.4432 & 36.3990 & \textbf{0.084381} & \textbf{3.2463} \\                     
\cline{2-11}                                                                                                                       
 &                            & 2 & 0.030859 & 44.4276 & 0.38981 & 13.3935 & 0.2625 & 66.0562 & \textbf{0.018374} & \textbf{2.1086} \\                   
  & Thai Statue               & 4 & 0.045516 & 41.7235 & --  & -- & 0.3151 & 85.4734 & \textbf{0.028457} & \textbf{2.2876} \\   
 &                            & 8 & 0.10507  & 39.7697 & --  & -- & 0.2389 & 55.0568 & \textbf{0.041552} & \textbf{3.0519} \\                    
\hline                                                                                                                       
 \multirow{12}{*}{ebsd} &     & 2 & 0.031939 & 46.9515 & 0.49466 & 16.3427 & 0.3473 & 65.4823 & \textbf{0.021037} & \textbf{2.0398} \\                   
 & Armadillo                  & 4 & 0.04424  & 44.2571 & --  & -- & 0.5933 & 58.6932 & \textbf{0.036102} & \textbf{2.0035} \\       
 &                            & 8 & 0.10062  & 42.2539 & --  & -- & 0.6453 & 81.5187 & \textbf{0.073138} & \textbf{1.8159} \\                    
\cline{2-11}                                                                                                                       
 &                            & 2 & 0.04299  & 47.5844 & 0.32989 & 13.0463 & 0.4141 & 84.9623 & \textbf{0.028555} & \textbf{1.9483} \\                    
&Lucy                         & 4 & 0.072388 & 44.5851 & --  & -- & 0.4541 & 75.3771 & \textbf{0.04325}  & \textbf{2.1771} \\            
 &                            & 8 & 0.16385  & 42.4252 & --  & -- & 0.6460 & 74.8618 & \textbf{0.079427} & \textbf{3.6839} \\                    
\cline{2-11}                                                                                                                       
 &                            & 2 & 0.049515 & 46.0065 & 1.0052  & 13.1767 & 0.2645 & 55.3462 & \textbf{0.034162} & \textbf{2.1722} \\                     
& Joyful Yell                 & 4 & 0.069491 & 43.4654 & --  & -- & 0.2770 & 42.4242 & \textbf{0.04818}  & \textbf{2.3335} \\            
 &                            & 8 & 0.11255  & 40.9818 & --  & -- & 0.4589 & 38.8507 & \textbf{0.073515} & \textbf{2.5774} \\                     
\cline{2-11}                                                                                                                       
 &                            & 2 & 0.03307  & 48.7666 & 0.30254 & 12.0112 & 0.2371 & 69.6653 & \textbf{0.019305} & \textbf{2.3639} \\                   
& Thai Statue                 & 4 & 0.046843 & 45.6104 & --  & -- & 0.2792 & 77.7622 & \textbf{0.029185} & \textbf{2.4529} \\        
 &                            & 8 & 0.089646 & 43.7591 & --  & -- & 0.2847 & 64.3520 & \textbf{0.041307} & \textbf{2.9642} \\                   
\hline    \hline                                                                                                                   
\multicolumn{2}{|c|}{}       & 2 & 0.036853 & 46.2107 & 0.44223 & 13.503  & 0.12186 & 45.0229 & \textbf{0.025062} & \textbf{2.2681} \\                    
\multicolumn{2}{|c|}{Median} & 4 & 0.057904 & 43.5029 & --  & -- & 0.18929 & 41.3767 & \textbf{0.039879} & \textbf{2.3932} \\             
\multicolumn{2}{|c|}{}       & 8 & 0.10844  & 41.6178 & --  & -- & 0.31159 & 41.3102 & \textbf{0.073722} & \textbf{3.1491} \\                    
\hline                                                                                                                       
 \multicolumn{2}{|c|}{}      & 2 & 0.038326 & 45.28   & 0.54626 & 14.3246 & 0.11516 & 42.7392 & \textbf{0.027843} & \textbf{2.554} \\                      
\multicolumn{2}{|c|}{Mean}   & 4 & 0.056267 & 42.5321 & --  & -- & 0.18488 & 40.6331 & \textbf{0.042075} & \textbf{2.6984} \\               
 \multicolumn{2}{|c|}{}      & 8 & 0.11193  & 40.6364 & --  & -- & 0.29819 & 40.1205 & \textbf{0.071228} & \textbf{3.2446} \\                    
\hline                                                                                                                       
\end{tabular}                                                                                                                
\caption{Quantitative comparison of the results attained with the proposed multi-shot approach and the state-of-the-art ($^*$: to make the comparison fair, we run the algorithm of \cite{Papadhimitri2014b} on the high resolution RGB images, as it performs uncalibrated photometric stereo on the RGB images without super-resolution -- the scaling factor is thus actually 1 in this case). Our approach overcomes the state-of-the-art in all the experiments.}
\label{tab:table_ups_comp}                                                                                             
\end{table*}

\subsection{Qualitative Comparison against the State-of-the-art on Real-world Datasets we Captured Ourselves}

\begin{figure*}[!ht]
  \newcommand{\mywidth}{0.184\textwidth}
  \newcommand{\mywidthlr}{0.12\textwidth}
  \newcolumntype{C}{ >{\centering\arraybackslash} m{0.02\textwidth} }
  \newcolumntype{X}{ >{\centering\arraybackslash} m{\mywidth} }
  \newcolumntype{Y}{ >{\centering\arraybackslash} m{\mywidthlr} }
  \setlength\tabcolsep{0.1pt} 
  \def\arraystretch{2}
  \begin{tabular}{CXYXXYX}
    &One of $\I_i$ &  $\zz$ & \cite{Werlberger2009,Unger2010} &  \cite{Papadhimitri2014b} &  \cite{Or-El2015} &  Ours \\
    \rotatebox{90}{Tablet Case}&    
    \includegraphics[width = \mywidth]{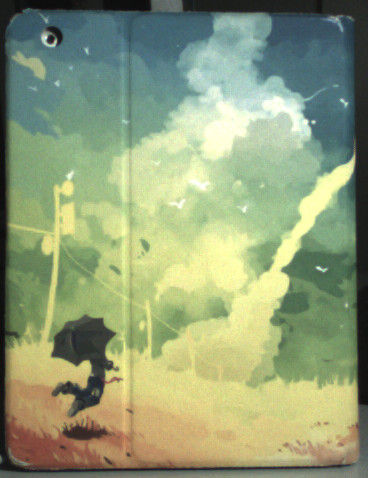}&
    \includegraphics[width = \mywidthlr]{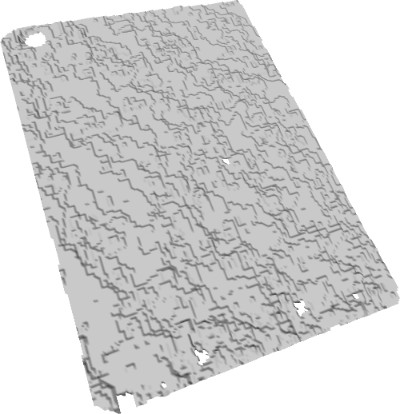}& 
    \includegraphics[width = \mywidth]{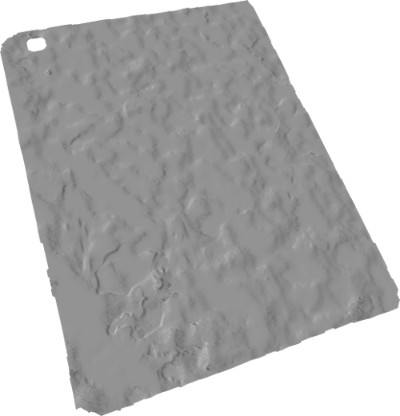}&  
    \includegraphics[width = \mywidth]{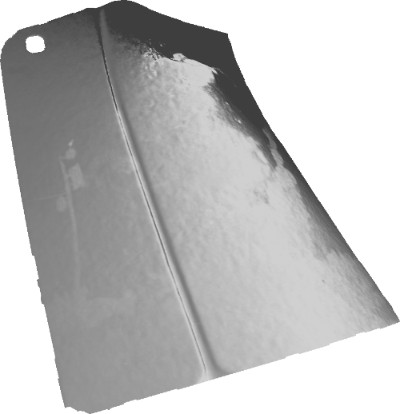}&  
    \includegraphics[width = \mywidthlr]{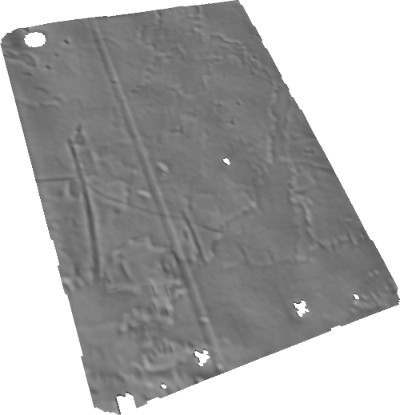}& 
    \includegraphics[width = \mywidth]{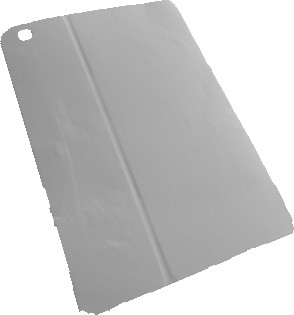}\\
    \rotatebox{90}{Vase}&    
    \includegraphics[width = \mywidth]{depth_sr_ups/vase/vase_rgb_crop.jpg}&
    \includegraphics[width = \mywidthlr]{depth_sr_ups/vase/vase_input}& 
    \includegraphics[width = \mywidth]{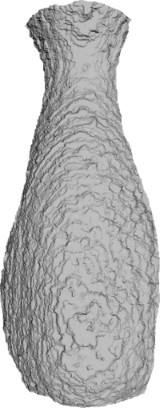}& 
    \includegraphics[width = \mywidth]{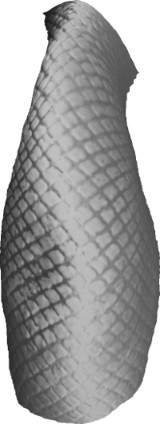}& 
    \includegraphics[width = \mywidthlr]{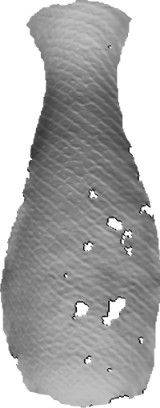}& 
    \includegraphics[width = \mywidth]{depth_sr_ups/vase/vase_shape01_new}\\
    \rotatebox{90}{Shirt}&    
    \includegraphics[width = \mywidth]{depth_sr_ups/shirt/shirt_rgb}&
    \includegraphics[width = \mywidthlr]{depth_sr_ups/shirt/shirt_input}& 
    \includegraphics[width = \mywidth]{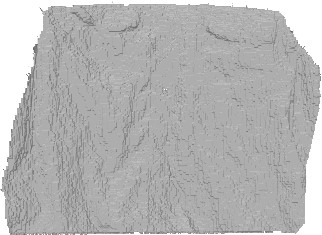}& 
    \includegraphics[width = \mywidth]{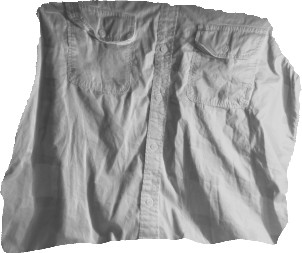}& 
    \includegraphics[width = \mywidthlr]{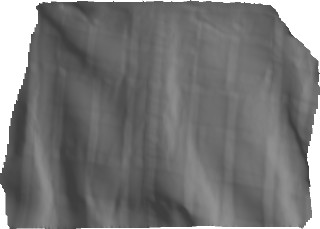}& 
    \includegraphics[width = \mywidth]{depth_sr_ups/shirt/shirt_shape_new}\\
    \rotatebox{90}{Backpack}&    
    \includegraphics[width = \mywidth]{depth_sr_ups/backpack/backpack_rgb}&
    \includegraphics[width = \mywidthlr]{depth_sr_ups/backpack/backpack_input}& 
    \includegraphics[width = \mywidth]{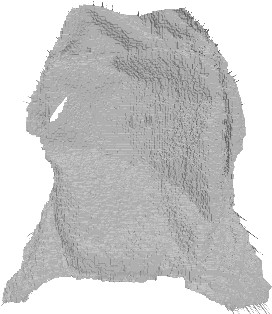}& 
    \includegraphics[width = \mywidth]{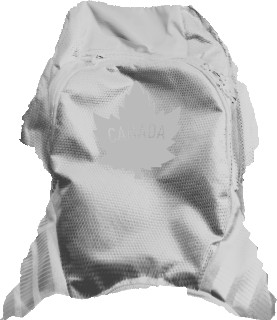}& 
    \includegraphics[width = \mywidthlr]{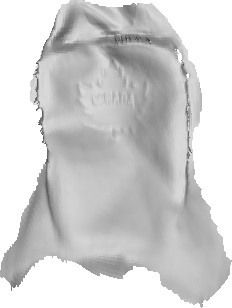}& 
    \includegraphics[width = \mywidth]{depth_sr_ups/backpack/backpack_shape_new} 
  \end{tabular}
\caption{Comparison between the proposed multi-shot method and the state-of-the-art, on real-world datasets captured using an Asus Xtion Pro Live camera. These results confirm the conclusion of the synthetic experiments in Figure~\ref{fig:synthetic_comparison_ups}.~\\~\\}
\label{fig:real_comparison_ups}
\end{figure*}

Figure~\ref{fig:real_comparison_ups} shows four qualitative comparisons against the state-of-the-art, on real-world data from Figures 1 and 6 in the main paper, which was captured with an Asus Xtion Pro Live camera (scaling factor of 4). 

It can be seen that image-based depth super-resolution approach hallucinates reflectance information as geometric information, since the underlying concept allows larger depth variations where strong image gradients are present. The uncalibrated photometric stereo results from~\cite{Papadhimitri2014b} contain much more relevant details, but the approach clearly suffers from a low-frequency bias due to the underlying bas-relief ambiguity, cf. ``Tablet Case'' and ``Vase''. In these experiments the RGB-D fusion results from~\cite{Or-El2015} are reasonable, but not as accurate as the ones obtained with the proposed multi-shot approach. 

\subsection{Comparison against the State-of-the-art on a Public Real-world Dataset}

\begin{figure*}[!ht]
  \centering
  \newcommand{\mywidth}{0.125\textwidth}
  \newcommand{\mywidthlr}{0.085\textwidth}
  \newcolumntype{C}{ >{\centering\arraybackslash} m{0.02\textwidth} }
  \newcolumntype{X}{ >{\centering\arraybackslash} m{\mywidth} }
  \newcolumntype{Y}{ >{\centering\arraybackslash} m{\mywidthlr} }
  \setlength\tabcolsep{10pt} 
  \begin{tabular}{CXYXXXX}
    &One of $\I_i$ &  $\zz$ & \cite{Werlberger2009,Unger2010} &  \cite{Papadhimitri2014b}&  Ours & GT\\
    \rotatebox{90}{bear} &
    \includegraphics[width=\mywidth]{diligent_sfs/bear}&
    \includegraphics[width=\mywidthlr]{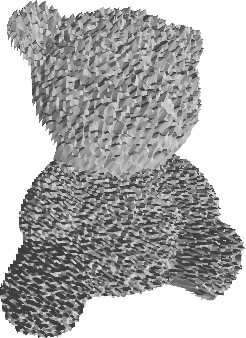}&
    \includegraphics[width=\mywidth]{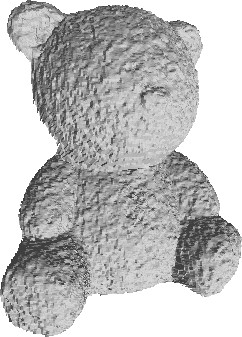}&
    \includegraphics[width=\mywidth]{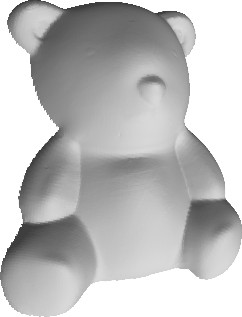}&
    \includegraphics[width=\mywidth]{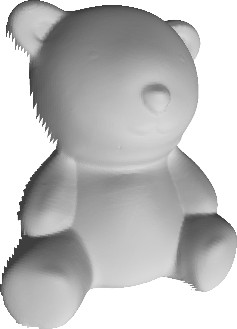}&
    \includegraphics[width=\mywidth]{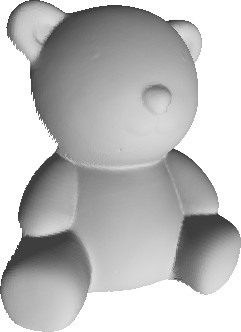} \\
    \rotatebox{90}{buddha} &
    \includegraphics[width=\mywidth]{diligent_sfs/buddha}&
    \includegraphics[width=\mywidthlr]{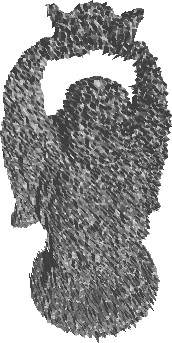}&
    \includegraphics[width=\mywidth]{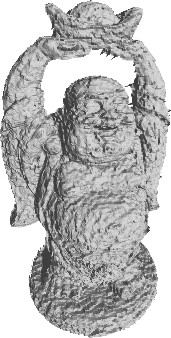}&
    \includegraphics[width=\mywidth]{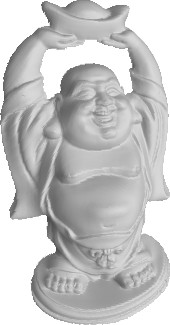}&
    \includegraphics[width=\mywidth]{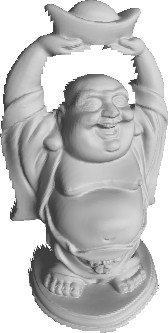}&
    \includegraphics[width=\mywidth]{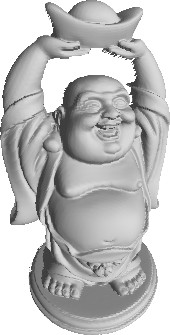} \\
    \rotatebox{90}{cat} &
    \includegraphics[width=\mywidth]{diligent_sfs/cat}&
    \includegraphics[width=\mywidthlr]{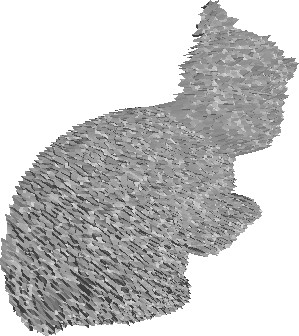}&
    \includegraphics[width=\mywidth]{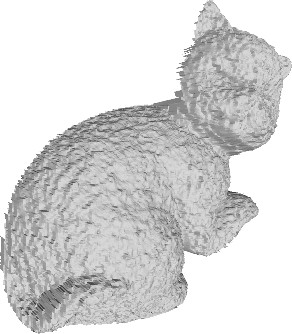}&
    \includegraphics[width=\mywidth]{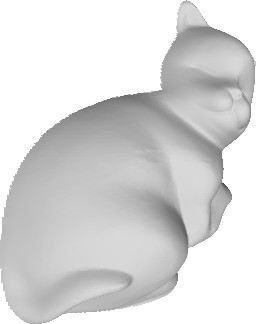}&
    \includegraphics[width=\mywidth]{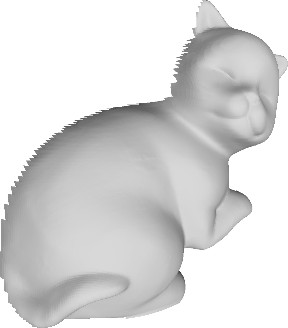}&
    \includegraphics[width=\mywidth]{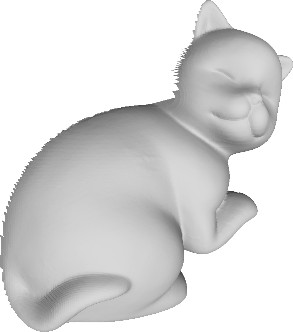} \\
    \rotatebox{90}{cow} &
    \includegraphics[width=\mywidth]{diligent_sfs/cow}&
    \includegraphics[width=\mywidthlr]{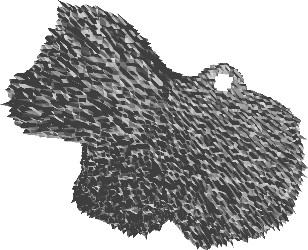}&
    \includegraphics[width=\mywidth]{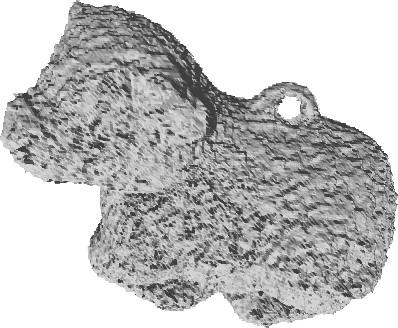}&
    \includegraphics[width=\mywidth]{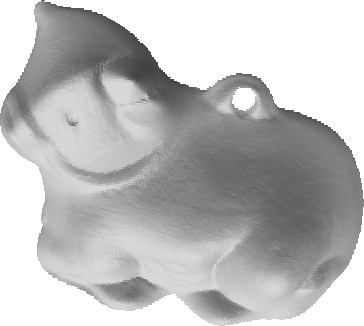}&
    \includegraphics[width=\mywidth]{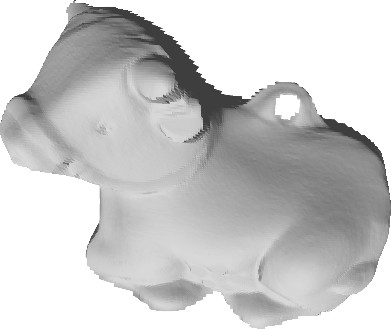}&
    \includegraphics[width=\mywidth]{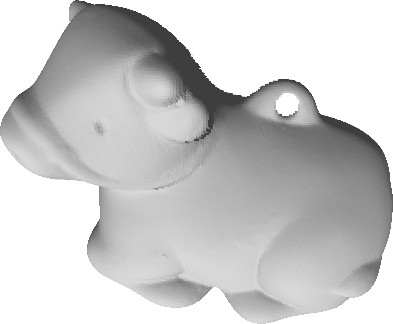} \\
    \rotatebox{90}{goblet} &
    \includegraphics[width=\mywidth]{diligent_sfs/goblet}&
    \includegraphics[width=\mywidthlr]{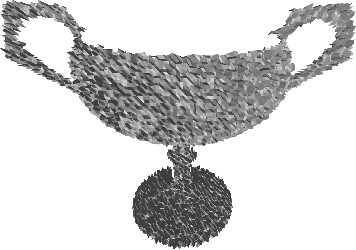}&
    \includegraphics[width=\mywidth]{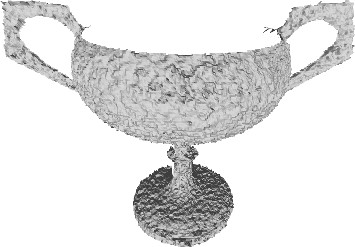}&
    \includegraphics[width=\mywidth]{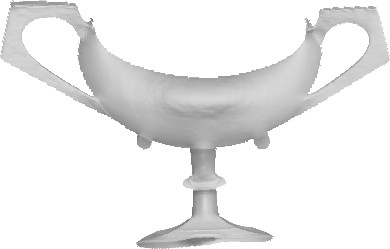}&
    \includegraphics[width=\mywidth]{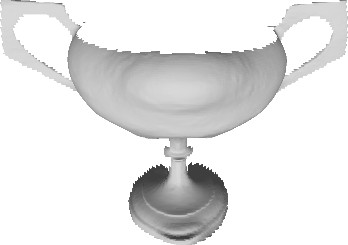}&
    \includegraphics[width=\mywidth]{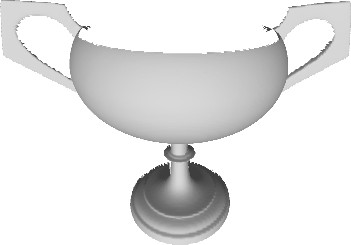} \\
    \rotatebox{90}{harvest} &
    \includegraphics[width=\mywidth]{diligent_sfs/harvest}&
    \includegraphics[width=\mywidthlr]{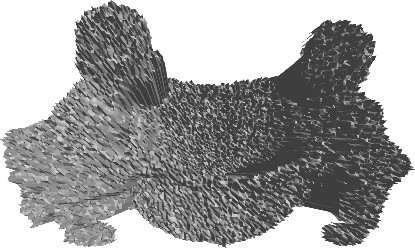}&
    \includegraphics[width=\mywidth]{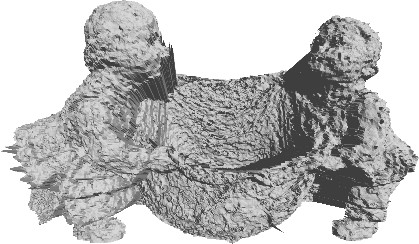}&
    \includegraphics[width=\mywidth]{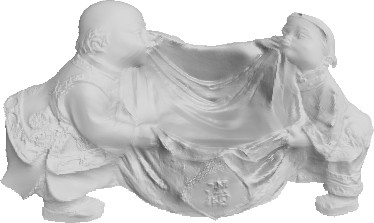}&
    \includegraphics[width=\mywidth]{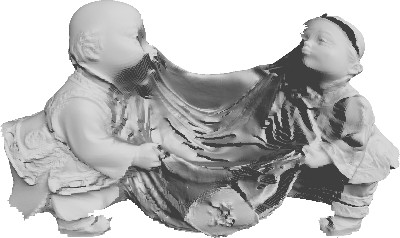}&
    \includegraphics[width=\mywidth]{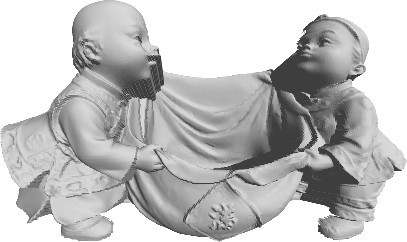} \\
    \rotatebox{90}{pot1} &
    \includegraphics[width=\mywidth]{diligent_sfs/pot1}&
    \includegraphics[width=\mywidthlr]{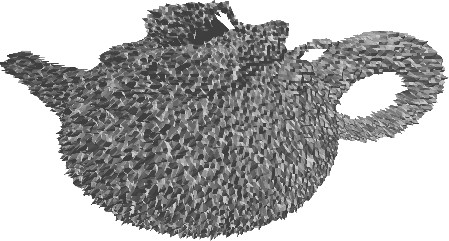}&
    \includegraphics[width=\mywidth]{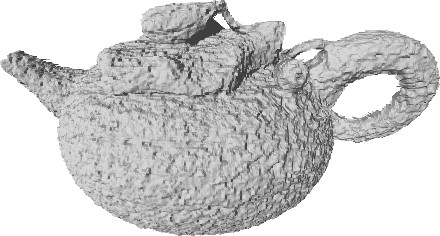}&
    \includegraphics[width=\mywidth]{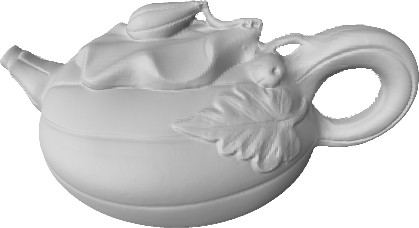}&
    \includegraphics[width=\mywidth]{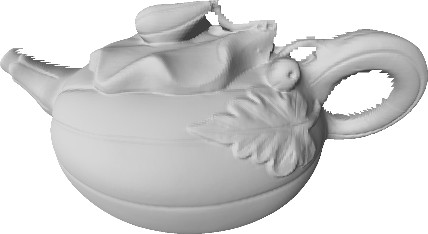}&
    \includegraphics[width=\mywidth]{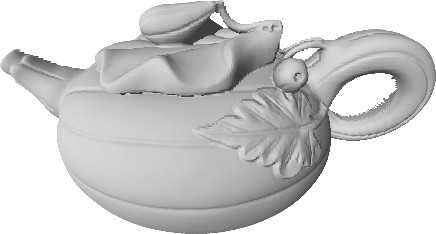} \\
    \rotatebox{90}{pot2} &
    \includegraphics[width=\mywidth]{diligent_sfs/pot2}&
    \includegraphics[width=\mywidthlr]{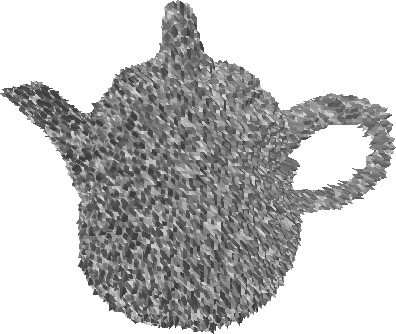}&
    \includegraphics[width=\mywidth]{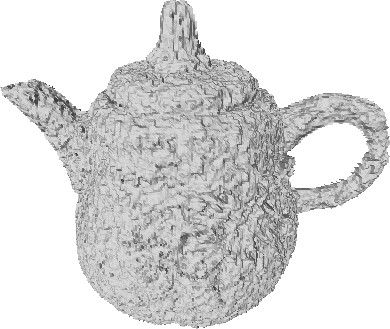}&
    \includegraphics[width=\mywidth]{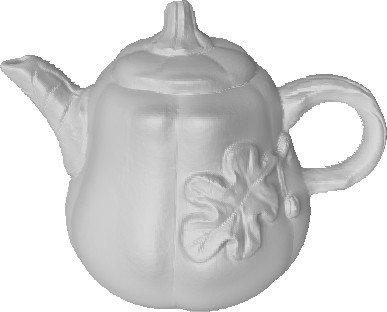}&
    \includegraphics[width=\mywidth]{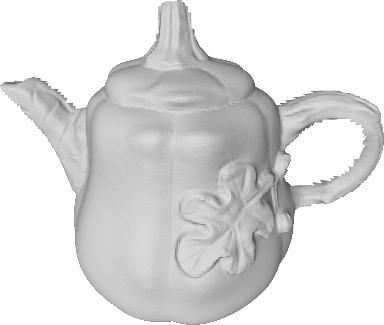}&
    \includegraphics[width=\mywidth]{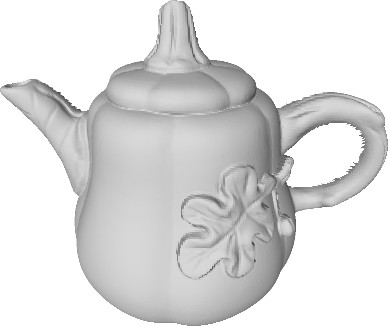} \\
    \rotatebox{90}{reading} &
    \includegraphics[width=\mywidth]{diligent_sfs/reading}&
    \includegraphics[width=\mywidthlr]{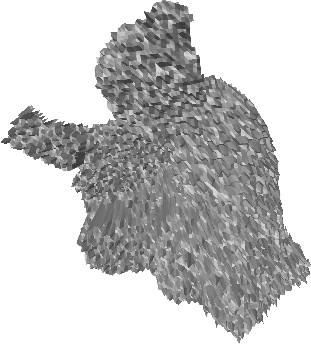}&
    \includegraphics[width=\mywidth]{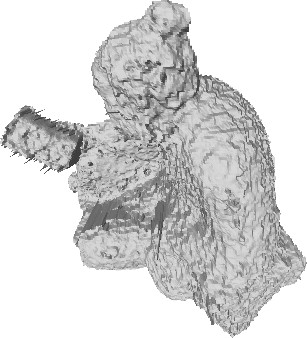}&
    \includegraphics[width=\mywidth]{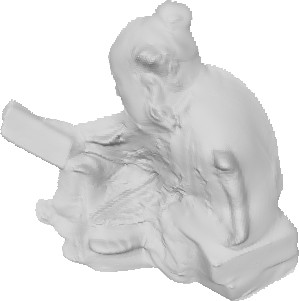}&
    \includegraphics[width=\mywidth]{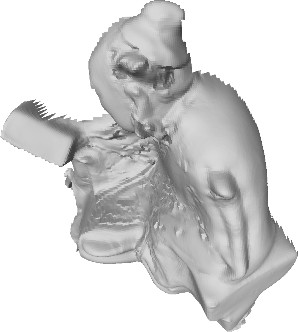}&
    \includegraphics[width=\mywidth]{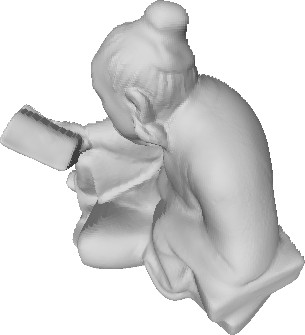} \\
  \end{tabular}
  \caption{Qualitative comparison of our uncalibrated photometric stereo-based approach against state-of-the-art methods, on the DiLiGenT dataset~\cite{Shi2018} (the scaling factor is $2$). Our method overcomes the state-of-the-art in all the experiments.}
  \label{fig:diligent_comparison_ups}
\end{figure*}

Eventually, we compare our results against the state-of-the-art on the DiLiGenT dataset~\cite{Shi2018}. Qualitative results are presented in Figure~\ref{fig:diligent_comparison_ups}, and quantitative ones in Table~\ref{tab:diligent_comparison_ups}. Once again, our method most of the times overcomes the state-of-the art in terms of surface details recovery. It is also interesting to compare these results with the corresponding ones in the previous sections: this comparison clearly shows that resorting to a multi-shot strategy based on photometric stereo is the only way to cope with general reflectance. 

Still, it can be observed that even with redundant data, some results such as the ``harvest''  one remain somewhat disappointing: this is because the proposed method explicitly builds upon the Lambertian assumption, which is not met in this example. Future extensions could thus include coping with non-Lambertian phenomena. 

\begin{table*}[!ht]                                                                                                     
\centering                                                                                                         
\begin{tabular}{|c|c|c|c|c|c|c|c|}                                                                                 
\hline                                                                                                             
\multirow{2}{*}{3D-shape} & \multirow{2}{*}{SF} & \multicolumn{2}{c|}{\cite{Werlberger2009,Unger2010}} & \multicolumn{2}{c|}{\cite{Papadhimitri2014b}$^*$} & \multicolumn{2}{c|}{Ours}\\                                        
\cline{3-8}                                                                                                             
 &  & RMSE & MAE & RMSE & MAE & RMSE & MAE \\                                                                      
\hline                                                                                                             
\hline                                                                                                             
 & 2 & 0.0077882 & 23.2799 & 0.029124 & 8.65 & \textbf{0.0064907} & \textbf{7.056} \\                              
bear & 4 & \textbf{0.0077919} & 19.6628 & -- & -- & 0.0083983 & \textbf{7.2645} \\                                 
 & 8 & \textbf{0.0079796} & 23.8495 & -- & -- & 0.013453 & \textbf{7.0708} \\                                      
\hline                                                                                                             
 & 2 & 0.0077863 & 31.0075 & 0.041827 & 18.0718 & \textbf{0.0066078} & \textbf{12.7816} \\                         
buddha & 4 & 0.0078303 & 28.5663 & -- & -- & \textbf{0.0077671} & \textbf{13.0276} \\                              
 & 8 & \textbf{0.0076309} & 20.1206 & -- & -- & 0.012959 & \textbf{13.6987} \\                                     
\hline                                                                                                             
 & 2 & \textbf{0.0078205} & 24.5162 & 0.039112 & 11.0118 & 0.008108 & \textbf{6.1952} \\                           
cat & 4 & \textbf{0.0076492} & 20.6365 & -- & -- & 0.010542 & \textbf{6.5739} \\                                   
 & 8 & \textbf{0.0078364} & 21.2045 & -- & -- & 0.015403 & \textbf{7.3812} \\                                      
\hline                                                                                                             
 & 2 & 0.0078497 & 31.5175 & 0.030244 & 18.1343 & \textbf{0.0055052} & \textbf{10.4445} \\                         
cow & 4 & \textbf{0.007844} & 26.7532 & -- & -- & 0.0083455 & \textbf{11.3151} \\                                  
 & 8 & \textbf{0.0085472} & 17.225 & -- & -- & 0.015231 & \textbf{12.7818} \\                                      
\hline                                                                                                             
 & 2 & \textbf{0.0078938} & 32.2235 & 0.13005 & 71.5669 & 0.010771 & \textbf{11.16} \\                             
goblet & 4 & \textbf{0.0078725} & 29.261 & -- & -- & 0.015434 & \textbf{11.6484} \\                                
 & 8 & \textbf{0.008322} & 24.9651 & -- & -- & 0.030694 & \textbf{13.9542} \\                                      
\hline                                                                                                             
 & 2 & \textbf{0.0078757} & 32.6288 & 0.06847 & \textbf{29.3081} & 0.024211 & 30.4736 \\                           
harvest & 4 & \textbf{0.0078363} & \textbf{30.6866} & -- & -- & 0.029344 & 31.9109 \\                              
 & 8 & \textbf{0.0077605} & \textbf{33.427} & -- & -- & 0.040837 & 33.5636 \\                                      
\hline                                                                                                             
 & 2 & 0.0078648 & 25.4586 & 0.01869 & 10.3055 & \textbf{0.0063032} & \textbf{7.3048} \\                           
pot1 & 4 & \textbf{0.0078397} & 22.6612 & -- & -- & 0.0080599 & \textbf{7.514} \\                                  
 & 8 & \textbf{0.0079306} & 30.9277 & -- & -- & 0.014455 & \textbf{7.9022} \\                                      
\hline                                                                                                             
 & 2 & 0.0077881 & 29.7433 & 0.022896 & 14.5031 & \textbf{0.0048177} & \textbf{9.4492} \\                          
pot2 & 4 & 0.0080123 & 26.261 & -- & -- & \textbf{0.0066391} & \textbf{9.5829} \\                                  
 & 8 & \textbf{0.0076366} & 21.8009 & -- & -- & 0.012587 & \textbf{10.0768} \\                                     
\hline                                                                                                             
 & 2 & \textbf{0.0077277} & 29.1401 & 0.069057 & 25.0014 & 0.0098433 & \textbf{16.7382} \\                         
reading & 4 & \textbf{0.0076277} & 26.4486 & -- & -- & 0.014885 & \textbf{19.6366} \\                              
 & 8 & \textbf{0.0078612} & \textbf{18.6829} & -- & -- & 0.027963 & 23.2138 \\                                     
\hline                                                                                                             
\hline                                                                                                             
 & 2 & 0.0078205 & 29.7433 & 0.039112 & 18.0718 & \textbf{0.0066078} & \textbf{10.4445} \\                         
Median & 4 & \textbf{0.0078363} & 26.4486 & -- & -- & 0.0083983 & \textbf{11.3151} \\                              
 & 8 & \textbf{0.0078612} & 21.8009 & -- & -- & 0.015231 & \textbf{12.7818} \\                                     
\hline                                                                                                             
 & 2 & \textbf{0.0078216} & 28.835 & 0.049941 & 22.9503 & 0.0091842 & \textbf{12.4004} \\                          
Mean & 4 & \textbf{0.0078115} & 25.6597 & -- & -- & 0.012157 & \textbf{13.1638} \\                                 
 & 8 & \textbf{0.007945} & 23.5781 & -- & -- & 0.020398 & \textbf{14.4048} \\                                      
\hline                                                                                                             
\end{tabular}                                                                                                      
\caption{Quantitative Comparison between other state-of-the-art methods and our multi-shot approach based on photometric stereo ($^*$: to make the comparison fair, we run the algorithm of \cite{Papadhimitri2014b} on the high resolution RGB images, as it performs uncalibrated photometric stereo on the RGB images without super-resolution -- the scaling factor is thus actually equal to~1 in this case). Our approach overcomes the state-of-the-art in terms of the level of geometric details which can be recovered, while being only slightly less accurate in terms of overall RMSE fit.}
\label{tab:diligent_comparison_ups}                                                                          
\end{table*}

\section{Unified Comparison of our Results on a Public Real-world Dataset}
\label{sec:supp_6}

\begin{figure*}[!ht]
	\centering
  \newcommand{\mywidth}{0.10\textwidth}
  \newcommand{\mywidthlr}{0.07\textwidth}
  \newcommand{\mmywidth}{0.12\textwidth}
  \newcommand{\mmywidthlr}{0.08\textwidth}
  \newcommand{\myshift}{25}
	\newcolumntype{C}{ >{\centering\arraybackslash} p{0.02\textwidth} }
	\newcolumntype{X}{ >{\centering\arraybackslash} m{\mywidth} }
	\newcolumntype{Y}{ >{\centering\arraybackslash} m{\mmywidth} }
	\setlength\tabcolsep{0.1pt} 
  \def\arrasytrech{1} 
	\begin{tabular}{CXXXX}
	  &Input $\{\I,\zz\}$ & SfS & SfS + reflectance learning & UPS \\
	  \multirow{2}[\myshift]{*}{\rotatebox{90}{bear}}&
		\includegraphics[width=\mywidth]{diligent_sfs/bear}&
		\includegraphics[width=\mywidth]{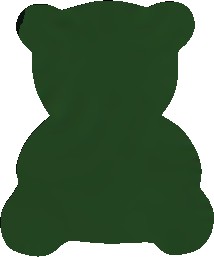}&
		\includegraphics[width=\mywidth]{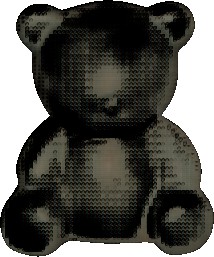}&
		\includegraphics[width=\mywidth]{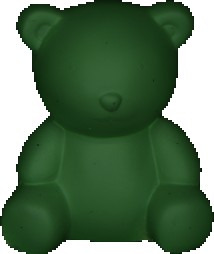}\\
	  &
		\includegraphics[width=\mywidthlr]{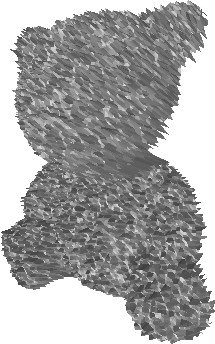}&
		\includegraphics[width=\mywidth]{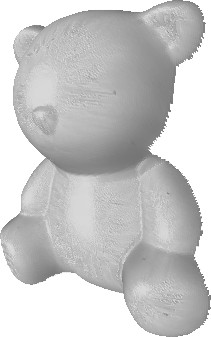}&
		\includegraphics[width=\mywidth]{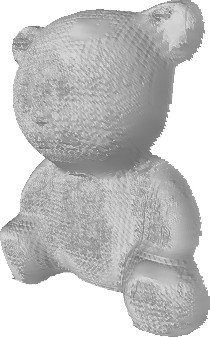}&
		\includegraphics[width=\mywidth]{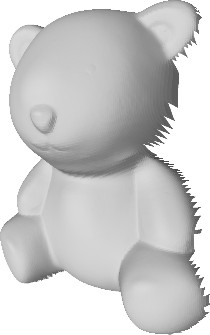}\\
	  \multirow{2}[\myshift]{*}{\rotatebox{90}{buddha}}&
		\includegraphics[width=\mywidth]{diligent_sfs/buddha}&
		\includegraphics[width=\mywidth]{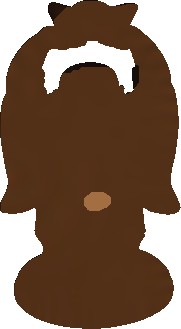}&
		\includegraphics[width=\mywidth]{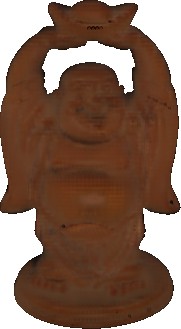}&
		\includegraphics[width=\mywidth]{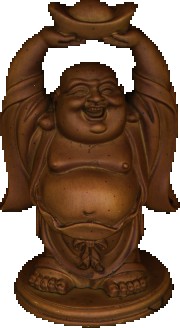}\\
	  &
		\includegraphics[width=\mywidthlr]{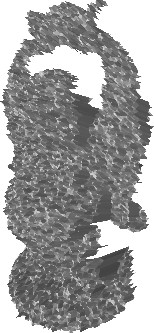}&
		\includegraphics[width=\mywidth]{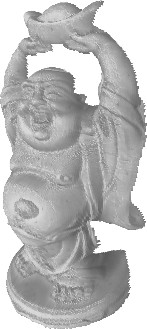}&
		\includegraphics[width=\mywidth]{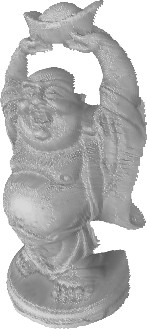}&
		\includegraphics[width=\mywidth]{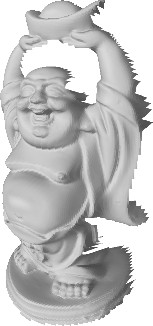}\\
	  \multirow{2}[\myshift]{*}{\rotatebox{90}{cat}}&
		\includegraphics[width=\mywidth]{diligent_sfs/cat}&
		\includegraphics[width=\mywidth]{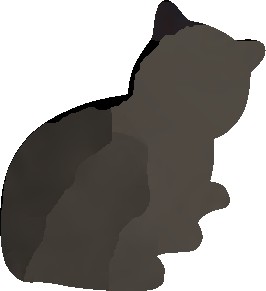}&
		\includegraphics[width=\mywidth]{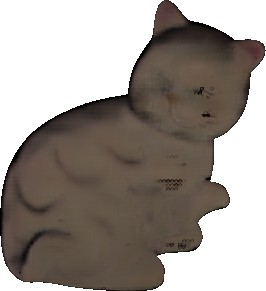}&
		\includegraphics[width=\mywidth]{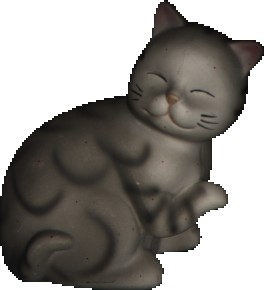}\\
	  &
		\includegraphics[width=\mywidthlr]{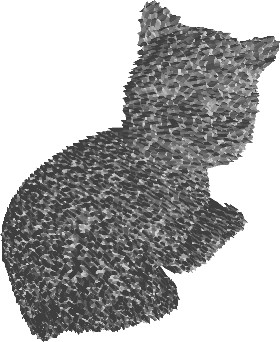}&
		\includegraphics[width=\mywidth]{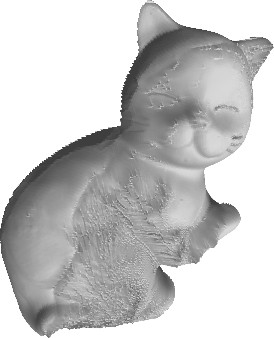}&
		\includegraphics[width=\mywidth]{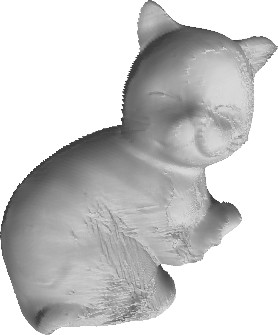}&
		\includegraphics[width=\mywidth]{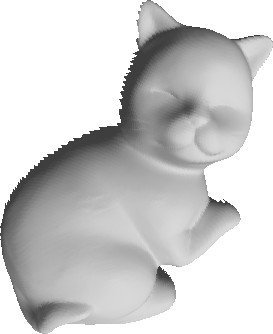}\\
	  \multirow{2}[\myshift]{*}{\rotatebox{90}{cow}}&
		\includegraphics[width=\mywidth]{diligent_sfs/cow}&
		\includegraphics[width=\mywidth]{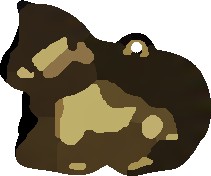}&
		\includegraphics[width=\mywidth]{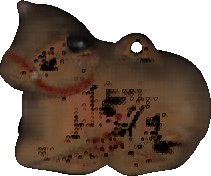}&
		\includegraphics[width=\mywidth]{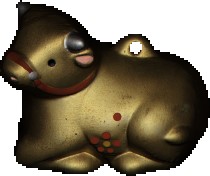}\\
	  &
		\includegraphics[width=\mywidthlr]{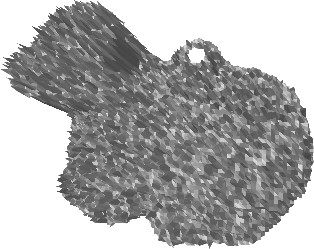}&
		\includegraphics[width=\mywidth]{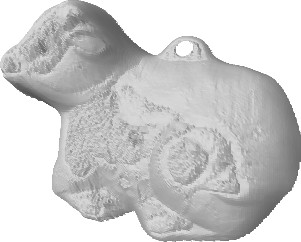}&
		\includegraphics[width=\mywidth]{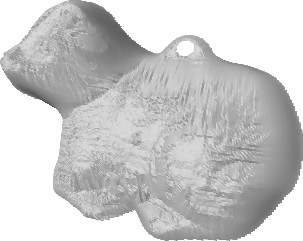}&
		\includegraphics[width=\mywidth]{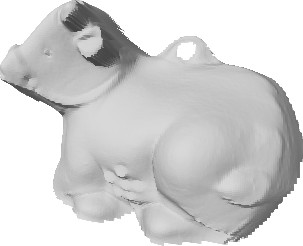}\\
	\end{tabular}\quad\quad
	\begin{tabular}{CYYYY}
    &Input $\{\I,\zz\}$ & SfS & SfS + reflectance learning & UPS \\
    \multirow{2}[\myshift]{*}{\rotatebox{90}{goblet}}&
		\includegraphics[width=\mmywidth]{diligent_sfs/goblet}&
		\includegraphics[width=\mmywidth]{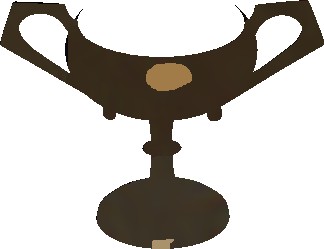}&
		\includegraphics[width=\mmywidth]{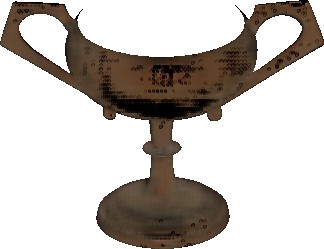}&
		\includegraphics[width=\mmywidth]{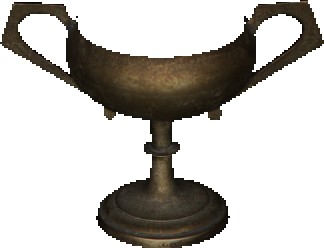}\\
	  &
		\includegraphics[width=\mmywidthlr]{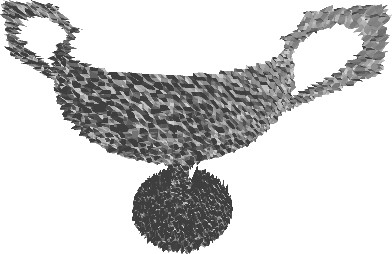}&
		\includegraphics[width=\mmywidth]{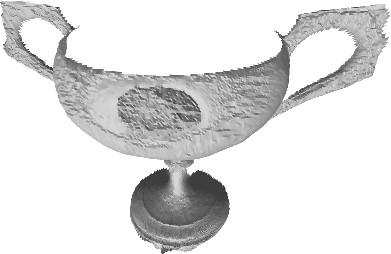}&
		\includegraphics[width=\mmywidth]{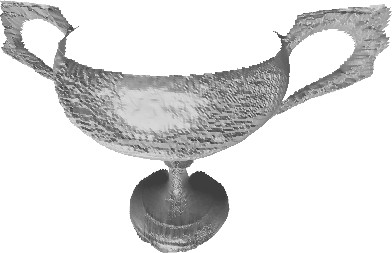}&
		\includegraphics[width=\mmywidth]{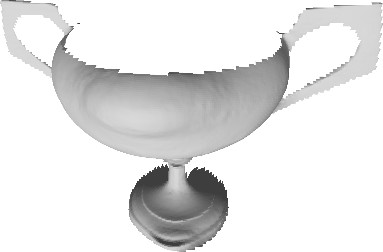}\\
	  \multirow{2}[\myshift]{*}{\rotatebox{90}{harvest}}&
		\includegraphics[width=\mmywidth]{diligent_sfs/harvest}&
		\includegraphics[width=\mmywidth]{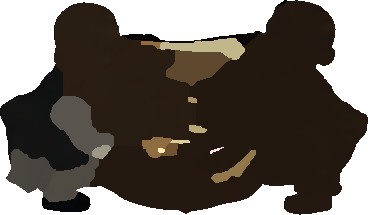}&
		\includegraphics[width=\mmywidth]{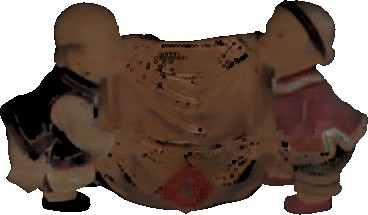}&
		\includegraphics[width=\mmywidth]{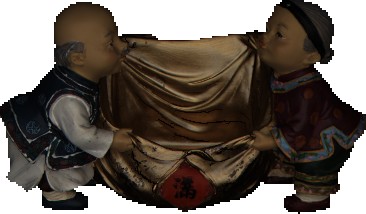}\\
	  &
		\includegraphics[width=\mmywidthlr]{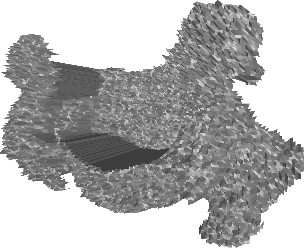}&
		\includegraphics[width=\mmywidth]{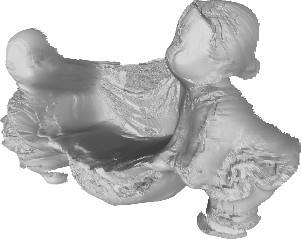}&
		\includegraphics[width=\mmywidth]{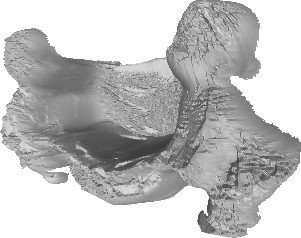}&
		\includegraphics[width=\mmywidth]{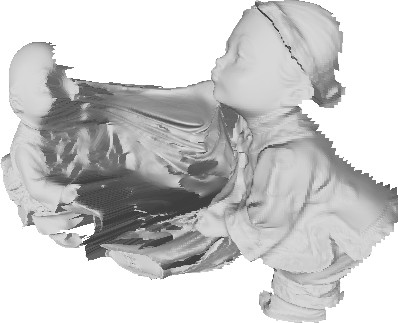}\\
	  \multirow{2}[\myshift]{*}{\rotatebox{90}{pot1}}&
		\includegraphics[width=\mmywidth]{diligent_sfs/pot1}&
		\includegraphics[width=\mmywidth]{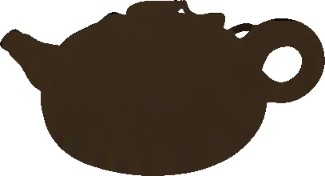}&
		\includegraphics[width=\mmywidth]{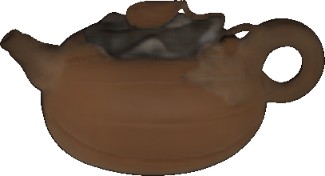}&
		\includegraphics[width=\mmywidth]{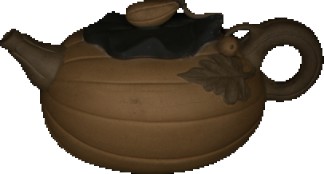}\\
	  &
		\includegraphics[width=\mmywidthlr]{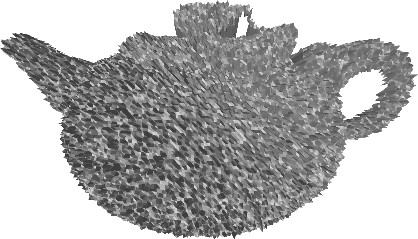}&
		\includegraphics[width=\mmywidth]{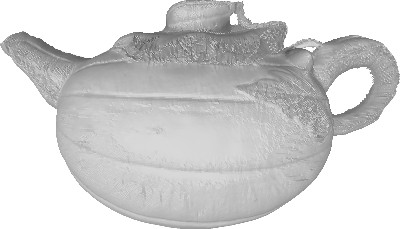}&
		\includegraphics[width=\mmywidth]{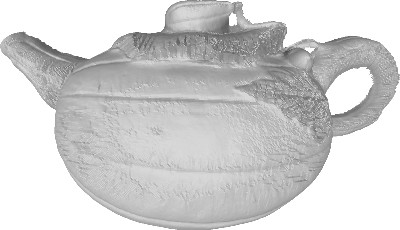}&
		\includegraphics[width=\mmywidth]{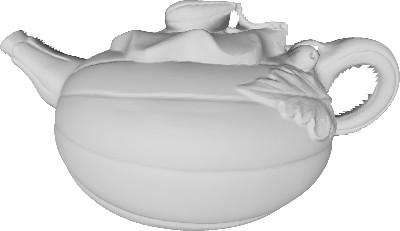}\\
	  \multirow{2}[\myshift]{*}{\rotatebox{90}{pot2}}&
		\includegraphics[width=\mmywidth]{diligent_sfs/pot2}&
		\includegraphics[width=\mmywidth]{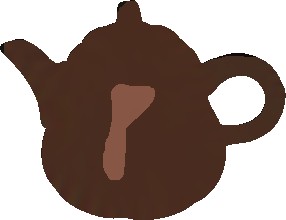}&
		\includegraphics[width=\mmywidth]{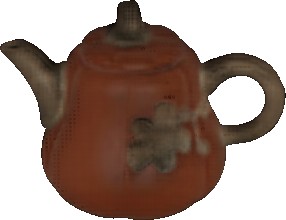}&
		\includegraphics[width=\mmywidth]{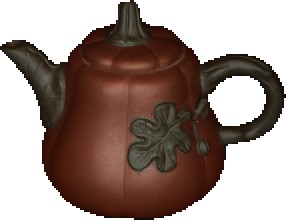}\\
	  &
		\includegraphics[width=\mmywidthlr]{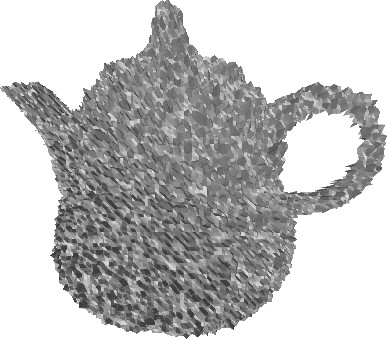}&
		\includegraphics[width=\mmywidth]{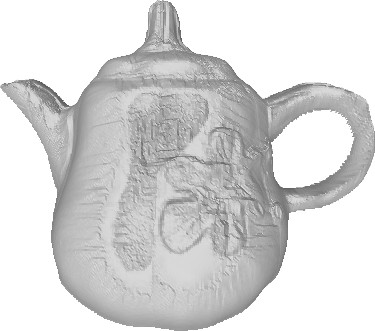}&
		\includegraphics[width=\mmywidth]{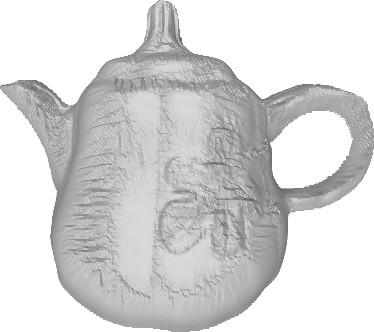}&
		\includegraphics[width=\mmywidth]{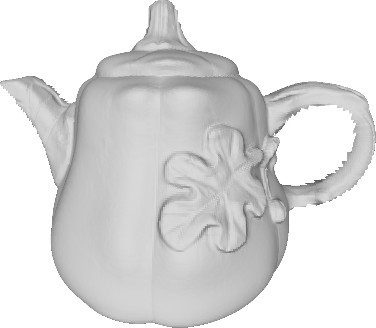}\\
	  \multirow{2}[\myshift]{*}{\rotatebox{90}{reading}}&
		\includegraphics[width=\mmywidth]{diligent_sfs/reading}&
		\includegraphics[width=\mmywidth]{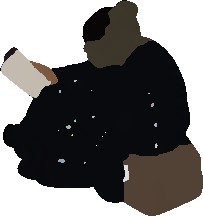}&
		\includegraphics[width=\mmywidth]{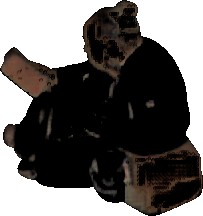}&
		\includegraphics[width=\mmywidth]{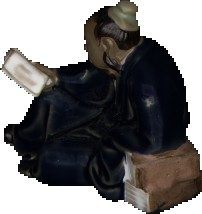}\\
	  &
		\includegraphics[width=\mmywidthlr]{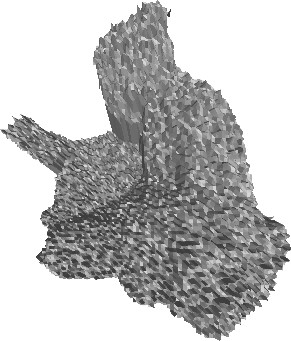}&
		\includegraphics[width=\mmywidth]{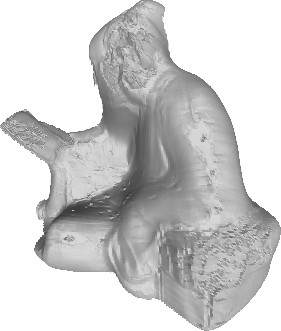}&
		\includegraphics[width=\mmywidth]{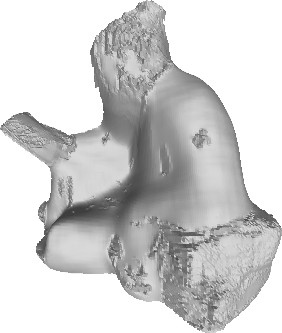}&
		\includegraphics[width=\mmywidth]{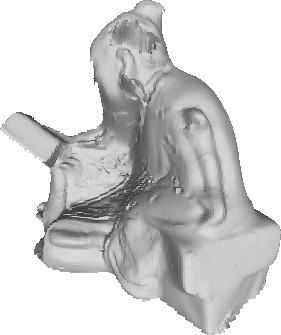}
		\end{tabular}
	\caption{Comparison of the albedo and high-resolution depth maps estimated by the proposed variational approach to shape-from-shading (SfS), the combination of SfS and deep reflectance learning, and the uncalibrated photometric stereo (UPS)-based approach, on the DiLiGenT dataset~\cite{Shi2018}. For quantitative evaluation, we refer the reader to Tables~\ref{tab:diligent_comparison_sfs}, \ref{tab:table_diligent_deep} and \ref{tab:diligent_comparison_ups}. ~\\~\\ }
	\label{fig:diligent_albedo_comparison}
\end{figure*}

Eventually, we present in Figure~\ref{fig:diligent_albedo_comparison} a unified qualitative comparison of the results obtained with the three proposed methods, on the $9$ objects of the DiLiGenT dataset~\cite{Shi2018}. This dataset illustrates well the cases where the single-shot approach can be used (when reflectance is uniform, as for instance in the ``bear'' example) and when it completely fails because the piecewise-constant albedo assumption is not satisfied (e.g., ``Cat''). This method could thus still be improved by designing a more general reflectance prior. The multi-shot approach based on uncalibrated photometric stereo estimates a much more reasonable albedo map, and thus a much more satisfactory depth map, because it does not rely on any assumption regarding piecewise-constantness. Yet, it could still be improved in order to reduce artifacts due to specularirites (e.g., ``reading''). Eventually, the albedo estimated by deep learning is sometimes reasonable (e.g., ``buddha''), but most of the times it is not really satisfactory. This is because the objects do not resemble the training set, which consists only of faces: to cope with a wider variety of objects, the training dataset should contain a broader range of object classes.

\section{Conclusion}
\label{sec:supp_7}
We evaluated in depth the applicability of photometric techniques to resolve depth super-resolution in the context of RGB-D sensing. Multiple self-captured real-world, publicly available real-world and self-generated synthetic datasets were used in order to qualitatively and quantitatively compare the three proposed strageties against state-of-the-art variational, optimization-based and deep learning methods. It appeared that each of the three proposed methods beats the corresponding state-of-the-art ones, which provides an empirical evidence for the soundness of considering photometry as a valuable clue for depth super-resolution in RGB-D sensing. 

In order to have at hand a unified comparison of the three methods presented in this work, we also considered a publicly available real-world photometric stereo benchmark across all experimental sections. This permitted us to clearly highlight the respective strengths and weaknesses of each method. They could still be improved towards, respectively, a more general reflectance prior (single-shot strategy), a broader training dataset (reflectance learning), and the handling of specularities (uncalibrated photometric stereo). 

\bibliographystyle{IEEEtran}
\bibliography{biblio}

\begin{thebibliography}{100}
\providecommand{\url}[1]{#1}
\csname url@samestyle\endcsname
\providecommand{\newblock}{\relax}
\providecommand{\bibinfo}[2]{#2}
\providecommand{\BIBentrySTDinterwordspacing}{\spaceskip=0pt\relax}
\providecommand{\BIBentryALTinterwordstretchfactor}{4}
\providecommand{\BIBentryALTinterwordspacing}{\spaceskip=\fontdimen2\font plus
\BIBentryALTinterwordstretchfactor\fontdimen3\font minus
  \fontdimen4\font\relax}
\providecommand{\BIBforeignlanguage}[2]{{%
\expandafter\ifx\csname l@#1\endcsname\relax
\typeout{** WARNING: IEEEtran.bst: No hyphenation pattern has been}%
\typeout{** loaded for the language `#1'. Using the pattern for}%
\typeout{** the default language instead.}%
\else
\language=\csname l@#1\endcsname
\fi
#2}}
\providecommand{\BIBdecl}{\relax}
\BIBdecl

\bibitem{Unger2010}
M.~Unger, T.~Pock, M.~Werlberger, and H.~Bischof, ``A convex approach for
  variational super-resolution,'' in \emph{Joint Pattern Recognition
  Symposium}, 2010, pp. 313--322.

\bibitem{Park2011}
J.~Park, H.~Kim, Y.-W. Tai, M.~S. Brown, and I.~S. Kweon, ``{High quality depth
  map upsampling for 3F-TOF cameras},'' in \emph{Proceedings of the IEEE
  International Conference on Computer Vision (ICCV)}, 2011, pp. 1623--1630.

\bibitem{Queau_Survey}
Y.~Qu\'{e}au, J.-D. Durou, and J.-F. Aujol, ``{Normal Integration: A Survey},''
  \emph{{Journal of Mathematical Imaging and Vision}}, vol.~60, no.~4, pp.
  576--593, 2018.

\bibitem{Basri2003}
R.~Basri and D.~P. Jacobs, ``{Lambertian reflectances and linear subspaces},''
  \emph{IEEE Transactions on Pattern Analysis and Machine Intelligence},
  vol.~25, no.~2, pp. 218--233, 2003.

\bibitem{Ramamoorthi2001}
R.~Ramamoorthi and P.~Hanrahan, ``{An Efficient Representation for Irradiance
  Environment Maps},'' in \emph{Proceedings of the Annual Conference on
  Computer Graphics and Interactive Techniques}, 2001, pp. 497--500.

\bibitem{Goldlucke2014}
B.~Goldl{\"u}cke, M.~Aubry, K.~Kolev, and D.~Cremers, ``A super-resolution
  framework for high-accuracy multiview reconstruction,'' \emph{International
  Journal of Computer Vision}, vol. 106, no.~2, pp. 172--191, 2014.

\bibitem{Maier2015}
R.~Maier, J.~St{\"u}ckler, and D.~Cremers, ``{Super-resolution keyframe fusion
  for 3D modeling with high-quality textures},'' in \emph{{Proceedings of the
  International Conference on 3D Vision (3DV)}}, 2015, pp. 536--544.

\bibitem{Schuon2009}
S.~Schuon, C.~Theobalt, J.~Davis, and S.~Thrun, ``{Lidarboost: Depth
  superresolution for TOF 3D shape scanning},'' in \emph{Proceedings of the
  IEEE Conference on Computer Vision and Pattern Recognition (CVPR)}, 2009, pp.
  343--350.

\bibitem{Macaodha2012}
O.~Mac~Aodha, N.~D.~F. Campbell, A.~Nair, and G.~J. Brostow, ``Patch based
  synthesis for single depth image super-resolution,'' in \emph{Proceedings of
  the European Conference on Computer Vision (ECCV)}, 2012, pp. 71--84.

\bibitem{Xie2016}
J.~Xie, R.~S. Feris, and M.-T. Sun, ``Edge-guided single depth image super
  resolution,'' \emph{IEEE Transactions on Image Processing}, vol.~25, no.~1,
  pp. 428--438, 2016.

\bibitem{Hornacek2013}
M.~Horn{\'a}cek, C.~Rhemann, M.~Gelautz, and C.~Rother, ``{Depth super
  resolution by rigid body self-similarity in 3D},'' in \emph{Proceedings of
  the IEEE Conference on Computer Vision and Pattern Recognition (CVPR)}, 2013,
  pp. 1123--1130.

\bibitem{Li2014}
J.~Li, Z.~Lu, G.~Zeng, R.~Gan, and H.~Zha, ``Similarity-aware patchwork
  assembly for depth image super-resolution,'' in \emph{Proceedings of the IEEE
  Conference on Computer Vision and Pattern Recognition (CVPR)}, 2014, pp.
  3374--3381.

\bibitem{Xie2015}
J.~Xie, R.~S. Feris, S.-S. Yu, and M.-T. Sun, ``Joint super resolution and
  denoising from a single depth image,'' \emph{IEEE Transactions on
  Multimedia}, vol.~17, no.~9, pp. 1525--1537, 2015.

\bibitem{Ferstl2015}
D.~Ferstl, M.~R{\"u}ther, and H.~Bischof, ``Variational depth superresolution
  using example-based edge representations,'' in \emph{Proceedings of the IEEE
  International Conference on Computer Vision (ICCV)}, 2015, pp. 513--521.

\bibitem{Riegler2016}
G.~Riegler, M.~R{\"u}ther, and H.~Bischof, ``{ATGV-net: accurate depth
  super-resolution},'' in \emph{Proceedings of the European Conference on
  Computer Vision (ECCV)}, 2016, pp. 268--284.

\bibitem{Horn1970}
B.~K.~P. Horn, ``{Shape From Shading: A Method for Obtaining the Shape of a
  Smooth Opaque Object From One View},'' Ph.D. dissertation, Department of
  Electrical Engineering and Computer Science, Massachusetts Institute of
  Technology, 1970.

\bibitem{Breuss2012}
M.~Breu{\ss}, E.~Cristiani, J.-D. Durou, M.~Falcone, and O.~Vogel,
  ``Perspective shape from shading: Ambiguity analysis and numerical
  approximations,'' \emph{SIAM Journal on Imaging Sciences}, vol.~5, no.~1, pp.
  311--342, 2012.

\bibitem{Durou2008}
J.-D. Durou, M.~Falcone, and M.~Sagona, ``{Numerical Methods for
  Shape-from-shading: A New Survey with Benchmarks},'' \emph{Computer Vision
  and Image Understanding}, vol. 109, no.~1, pp. 22--43, 2008.

\bibitem{Zhang1999}
R.~Zhang, P.-S. Tsai, J.~E. Cryer, and M.~Shah, ``Shape-from-shading: a
  survey,'' \emph{IEEE Transactions on Pattern Analysis and Machine
  Intelligence}, vol.~21, no.~8, pp. 690--706, 1999.

\bibitem{Horn1986}
B.~K.~P. Horn and M.~J. Brooks, ``The variational approach to shape from
  shading,'' \emph{Computer Vision, Graphics, and Image Processing}, vol.~33,
  no.~2, pp. 174--208, 1986.

\bibitem{Ikeuchi1981}
K.~Ikeuchi and B.~K. Horn, ``Numerical shape from shading and occluding
  boundaries,'' \emph{Artificial intelligence}, vol.~17, no. 1-3, pp. 141--184,
  1981.

\bibitem{Cristiani2007}
E.~Cristiani and M.~Falcone, ``Fast semi-lagrangian schemes for the eikonal
  equation and applications,'' \emph{SIAM Journal on Numerical Analysis},
  vol.~45, no.~5, pp. 1979--2011, 2007.

\bibitem{Falcone1997}
M.~Falcone and M.~Sagona, ``An algorithm for the global solution of the
  shape-from-shading model,'' in \emph{Proceedings of the International
  Conference on Image Analysis and Processing (ICIAP)}, 1997, pp. 596--603.

\bibitem{Lions1993}
P.-L. Lions, E.~Rouy, and A.~Tourin, ``Shape-from-shading, viscosity solutions
  and edges,'' \emph{Numerische Mathematik}, vol.~64, no.~1, pp. 323--353,
  1993.

\bibitem{Rouy1992}
E.~Rouy and A.~Tourin, ``A viscosity solutions approach to
  shape-from-shading,'' \emph{SIAM Journal on Numerical Analysis}, vol.~29,
  no.~3, pp. 867--884, 1992.

\bibitem{Adelson1996}
E.~H. Adelson and A.~P. Pentland, \emph{Perception as Bayesian
  inference}.\hskip 1em plus 0.5em minus 0.4em\relax Cambridge University
  Press, 1996, ch. {The perception of shading and reflectance}, pp. 409--423.

\bibitem{Huang2011}
R.~Huang and W.~A.~P. Smith, ``Shape-from-shading under complex natural
  illumination,'' in \emph{Proceedings of the IEEE International Conference on
  Image Processing (ICIP)}, 2011, pp. 13--16.

\bibitem{Johnson2011}
M.~K. Johnson and E.~H. Adelson, ``Shape estimation in natural illumination,''
  in \emph{Proceedings of the IEEE Conference on Computer Vision and Pattern
  Recognition (CVPR)}, 2011, pp. 2553--2560.

\bibitem{Richter2015}
S.~R. Richter and S.~Roth, ``Discriminative shape from shading in uncalibrated
  illumination,'' in \emph{Proceedings of the IEEE Conference on Computer
  Vision and Pattern Recognition (CVPR)}, 2015, pp. 1128--1136.

\bibitem{Queau2017}
Y.~Qu\'{e}au, J.~M\'{e}lou, F.~Castan, D.~Cremers, and J.-D. Durou, ``{A
  Variational Approach to Shape-from-shading Under Natural Illumination},'' in
  \emph{{Energy Minimization Methods for Computer Vision and Pattern
  Recognition (EMMCVPR)}}, 2017, pp. 342--357.

\bibitem{Barron2015}
J.~Barron and J.~Malik, ``Shape, illumination, and reflectance from shading,''
  \emph{IEEE Transactions on Pattern Analysis and Machine Intelligence},
  vol.~37, no.~8, pp. 1670--1687, 2015.

\bibitem{Woodham1980}
R.~J. Woodham, ``{Photometric Method for Determining Surface Orientation from
  Multiple Images},'' \emph{Optical Engineering}, vol.~19, no.~1, pp. 139--144,
  1980.

\bibitem{Hayakawa1994}
H.~Hayakawa, ``{Photometric stereo under a light source with arbitrary
  motion},'' \emph{Journal of the Optical Society of America A}, vol.~11,
  no.~11, pp. 3079--3089, 1994.

\bibitem{Belhumeur1999}
P.~N. Belhumeur, D.~J. Kriegman, and A.~L. Yuille, ``{The bas-relief
  ambiguity},'' \emph{International Journal of Computer Vision}, vol.~35,
  no.~1, pp. 33--44, 1999.

\bibitem{Basri2007}
R.~Basri, D.~W. Jacobs, and I.~Kemelmacher, ``{Photometric stereo with general,
  unknown lighting},'' \emph{International Journal of Computer Vision},
  vol.~72, no.~3, pp. 239--257, 2007.

\bibitem{Alldrin2007}
N.~G. Alldrin, S.~P. Mallick, and D.~J. Kriegman, ``{Resolving the generalized
  bas-relief ambiguity by entropy minimization},'' in \emph{Proceedings of the
  IEEE Conference on Computer Vision and Pattern Recognition (CVPR)}, 2007.

\bibitem{Papadhimitri2014b}
T.~Papadhimitri and P.~Favaro, ``{A closed-form, consistent and robust solution
  to uncalibrated photometric stereo via local diffuse reflectance maxima},''
  \emph{International Journal of Computer Vision}, vol. 107, no.~2, pp.
  139--154, 2014.

\bibitem{JMIV2015}
Y.~Qu\'{e}au, F.~Lauze, and J.-D. Durou, ``{Solving Uncalibrated Photometric
  Stereo using Total Variation},'' \emph{{Journal of Mathematical Imaging and
  Vision}}, vol.~52, no.~1, pp. 87--107, 2015.

\bibitem{Lu2018}
F.~Lu, X.~Chen, I.~Sato, and Y.~Sato, ``Symps: Brdf symmetry guided photometric
  stereo for shape and light source estimation,'' \emph{IEEE Transactions on
  Pattern Analysis and Machine Intelligence}, vol.~40, no.~1, pp. 221--234,
  2018.

\bibitem{Mo2018}
Z.~Mo, B.~Shi, F.~Lu, S.-K. Yeung, and Y.~Matsushita, ``Uncalibrated
  photometric stereo under natural illumination,'' in \emph{Proceedings of the
  IEEE Conference on Computer Vision and Pattern Recognition (CVPR)}, 2018, pp.
  2936--2945.

\bibitem{Shi2018}
B.~Shi, Z.~Mo, Z.~Wu, D.~Duan, S.~K. Yeung, and P.~Tan, ``A benchmark dataset
  and evaluation for non-lambertian and uncalibrated photometric stereo,''
  \emph{IEEE Transactions on Pattern Analysis and Machine Intelligence},
  vol.~41, no.~2, pp. 271--284, 2019.

\bibitem{CVPR2017}
Y.~Qu\'{e}au, T.~Wu, F.~Lauze, J.-D. Durou, and D.~Cremers, ``{A Non-Convex
  Variational Approach to Photometric Stereo under Inaccurate Lighting},'' in
  \emph{{Proceedings of the IEEE Conference on Computer Vision and Pattern
  Recognition (CVPR)}}, 2017, pp. 350--359.

\bibitem{Ikehata2018}
S.~Ikehata, ``{CNN-PS}: {CNN}-based photometric stereo for general non-convex
  surfaces,'' in \emph{Proceedings of the European Conference on Computer
  Vision (ECCV)}, 2018, pp. 3--18.

\bibitem{Chen2019}
G.~Chen, K.~Han, B.~Shi, Y.~Matsushita, and K.-Y.~K. Wong, ``Self-calibrating
  deep photometric stereo networks,'' in \emph{Proceedings of the IEEE
  Conference on Computer Vision and Pattern Recognition (CVPR)}, 2019.

\bibitem{Choe2017}
G.~Choe, J.~Park, Y.-W. Tai, and I.~S. Kweon, ``Refining geometry from depth
  sensors using {IR} shading images,'' \emph{International Journal of Computer
  Vision}, vol. 122, no.~1, pp. 1--16, 2017.

\bibitem{Maier2017}
R.~Maier, K.~Kim, D.~Cremers, J.~Kautz, and M.~Nie{\ss}ner, ``Intrinsic3d:
  High-quality {3D} reconstruction by joint appearance and geometry
  optimization with spatially-varying lighting,'' in \emph{Proceedings of the
  IEEE International Conference on Computer Vision (ICCV)}, 2017, pp.
  3114--3122.

\bibitem{Zollhoefer2015}
M.~Zollh{\"o}fer, A.~Dai, M.~Innman, C.~Wu, M.~Stamminger, C.~Theobalt, and
  M.~Nie{\ss}ner, ``Shading-based refinement on volumetric signed distance
  functions,'' \emph{ACM Transactions on Graphics}, vol.~34, no.~4, pp.
  96:1--96:14, 2015.

\bibitem{Han2013}
Y.~Han, J.-Y. Lee, and I.~S. Kweon, ``{High Quality Shape from a Single RGB-D
  Image under Uncalibrated Natural Illumination},'' in \emph{Proceedings of the
  IEEE International Conference on Computer Vision (ICCV)}, 2013, pp.
  1617--1624.

\bibitem{Kim2015}
K.~Kim, A.~Torii, and M.~Okutomi, ``Joint estimation of depth, reflectance and
  illumination for depth refinement,'' in \emph{Proceedings of the IEEE
  International Conference on Computer Vision (ICCV)}, 2015, pp. 199--207.

\bibitem{Or-El2016}
R.~Or-El, R.~Hershkovitz, A.~Wetzler, G.~Rosman, A.~M. Bruckstein, and
  R.~Kimmel, ``Real-time depth refinement for specular objects,'' in
  \emph{Proceedings of the IEEE Conference on Computer Vision and Pattern
  Recognition (CVPR)}, 2016, pp. 4378--4386.

\bibitem{Or-El2015}
R.~Or-El, G.~Rosman, A.~Wetzler, R.~Kimmel, and A.~Bruckstein, ``{RGBD-Fusion:
  Real-Time High Precision Depth Recovery},'' in \emph{Proceedings of the IEEE
  Conference on Computer Vision and Pattern Recognition (CVPR)}, 2015, pp.
  5407--5416.

\bibitem{Wu2014}
C.~Wu, M.~Zollh\"{o}fer, M.~Nie{\ss}ner, M.~Stamminger, S.~Izadi, and
  C.~Theobalt, ``Real-time shading-based refinement for consumer depth
  cameras,'' \emph{ACM Transactions on Graphics}, vol.~33, no.~6, pp.
  200:1--200:10, 2014.

\bibitem{Yu2013}
L.-F. Yu, S.-K. Yeung, Y.-W. Tai, and S.~Lin, ``{Shading-based shape refinement
  of RGB-D images},'' in \emph{Proceedings of the IEEE Conference on Computer
  Vision and Pattern Recognition (CVPR)}, 2013, pp. 1415--1422.

\bibitem{Anderson2011}
R.~Anderson, B.~Stenger, and R.~Cipolla, ``Augmenting depth camera output using
  photometric stereo,'' in \emph{Proceedings of the IAPR Conference on Machine
  Vision Applications (MVA)}, 2011, pp. 369--372.

\bibitem{Chatterjee2015}
A.~Chatterjee and V.~Madhav~Govindu, ``Photometric refinement of depth maps for
  multi-albedo objects,'' in \emph{Proceedings of the IEEE Conference on
  Computer Vision and Pattern Recognition (CVPR)}, 2015, pp. 933--941.

\bibitem{Xie2018}
L.~Xie, Y.~Xu, X.~Zhang, W.~Bao, C.~Tong, and B.~Shi, ``A self-calibrated
  photo-geometric depth camera,'' \emph{The Visual Computer}, 2018.

\bibitem{Zhang2018}
Y.~Zhang, Q.~Zhang, and W.~Feng, ``{High-Resolution Depth Refinement by
  Photometric and Multi-shading Constraints},'' in \emph{PRICAI 2018: Trends in
  Artificial Intelligence}, 2018, pp. 201--209.

\bibitem{Diebel2006}
J.~Diebel and S.~Thrun, ``{An application of Markov random fields to range
  sensing},'' in \emph{{Advances in Neural Information Processing Systems}},
  2006, pp. 291--298.

\bibitem{Ferstl2013}
D.~Ferstl, C.~Reinbacher, R.~Ranftl, M.~R{\"u}ther, and H.~Bischof, ``Image
  guided depth upsampling using anisotropic total generalized variation,'' in
  \emph{Proceedings of the IEEE International Conference on Computer Vision
  (ICCV)}, 2013, pp. 993--1000.

\bibitem{Yang2007}
Q.~Yang, R.~Yang, J.~Davis, and D.~Nist{\'e}r, ``Spatial-depth super resolution
  for range images,'' in \emph{Proceedings of the IEEE Conference on Computer
  Vision and Pattern Recognition (CVPR)}, 2007.

\bibitem{Li2018}
B.~Li, Y.~Zhou, Y.~Zhang, and A.~Wang, ``Depth image super-resolution based on
  joint sparse coding,'' \emph{Pattern Recognition Letters}, 2019, (in press).

\bibitem{Hui2016}
T.-W. Hui, C.~C. Loy, and X.~Tang, ``Depth map super-resolution by deep
  multi-scale guidance,'' in \emph{Proceedings of European Conference on
  Computer Vision (ECCV)}, 2016, pp. 353--369.

\bibitem{Tan2008}
P.~Tan, S.~Lin, and L.~Quan, ``Subpixel photometric stereo,'' \emph{IEEE
  Transactions on Pattern Analysis and Machine Intelligence}, vol.~30, no.~8,
  pp. 1460--1471, 2008.

\bibitem{Chaudhuri2005}
S.~Chaudhuri and M.~V. Joshi, \emph{Motion-free super-resolution}.\hskip 1em
  plus 0.5em minus 0.4em\relax Springer Verlag, 2005.

\bibitem{Lu2013}
Z.~Lu, Y.-W. Tai, F.~Deng, M.~Ben-Ezra, and M.~S. Brown, ``{A 3D imaging
  framework based on high-resolution photometric-stereo and low-resolution
  depth},'' \emph{International Journal of Computer Vision}, vol. 102, no. 1-3,
  pp. 18--32, 2013.

\bibitem{Haefner2018}
B.~Haefner, Y.~Quéau, T.~Möllenhoff, and D.~Cremers, ``Fight ill-posedness
  with ill-posedness: Single-shot variational depth super-resolution from
  shading,'' in \emph{Proceedings of the I{EEE} {C}onference on {C}omputer
  {V}ision and {P}attern {R}ecognition (CVPR)}, 2018, pp. 164--174.

\bibitem{Peng2017}
S.~Peng, B.~Haefner, Y.~Qu{\'e}au, and D.~Cremers, ``Depth super-resolution
  meets uncalibrated photometric stereo,'' in \emph{Proceedings of the IEEE
  International Conference on Computer Vision (ICCV) Workshops}, 2017, pp.
  2961--2968.

\bibitem{Mumford1994}
D.~Mumford, ``Bayesian rationale for the variational formulation,'' in
  \emph{Geometry-driven diffusion in computer vision}, 1994, pp. 135--146.

\bibitem{Graber2015}
G.~Graber, J.~Balzer, S.~Soatto, and T.~Pock, ``Efficient minimal-surface
  regularization of perspective depth maps in variational stereo,'' in
  \emph{Proceedings of the IEEE Conference on Computer Vision and Pattern
  Recognition (CVPR)}, 2015, pp. 511--520.

\bibitem{Land1977}
E.~H. Land, ``The retinex theory of color vision,'' \emph{Scientific American},
  vol. 237, no.~6, pp. 108--120, 1977.

\bibitem{Boyd2011}
S.~Boyd, N.~Parikh, E.~Chu, B.~Peleato, and J.~Eckstein, ``{Distributed
  Optimization and Statistical Learning via the Alternating Direction Method of
  Multipliers},'' \emph{Foundations and Trends in Machine Learning}, vol.~3,
  no.~1, pp. 1--122, 2011.

\bibitem{Eckstein1992}
J.~Eckstein and D.~P. Bertsekas, ``{On the Douglas--Rachford splitting method
  and the proximal point algorithm for maximal monotone operators},''
  \emph{{Mathematical Programming}}, vol.~55, no.~1, pp. 293--318, 1992.

\bibitem{Glowinski1975}
R.~Glowinski and A.~Marroco, ``{Sur l'approximation, par {\'e}l{\'e}ments finis
  d'ordre un, et la r{\'e}solution, par p{\'e}nalisation-dualit{\'e} d'une
  classe de probl{\`e}mes de Dirichlet non lin{\'e}aires},'' \emph{{Revue
  fran{\c{c}}aise d'automatique, informatique, recherche op{\'e}rationnelle.
  Analyse num{\'e}rique}}, vol.~9, no.~R2, pp. 41--76, 1975.

\bibitem{Strekalovskiy2014}
E.~Strekalovskiy and D.~Cremers, ``{Real-time minimization of the piecewise
  smooth Mumford-Shah functional},'' in \emph{Proceedings of the European
  Conference on Computer Vision (ECCV)}, 2014, pp. 127--141.

\bibitem{Schmidt}
M.~Schmidt, ``{minFunc: unconstrained differentiable multivariate optimization
  in Matlab},'' 2005,
  \url{http://www.cs.ubc.ca/~schmidtm/Software/minFunc.html}.

\bibitem{Liu1989}
D.~C. Liu and J.~Nocedal, ``{On the limited memory BFGS method for large scale
  optimization},'' \emph{Mathematical programming}, vol.~45, no.~1, pp.
  503--528, 1989.

\bibitem{InpaintNans}
``inpaint\_nans,'' 2012,
  \url{https://fr.mathworks.com/matlabcentral/fileexchange/4551-inpaint_nans}.

\bibitem{He2013}
K.~He, J.~Sun, and X.~Tang, ``Guided image filtering,'' \emph{IEEE Transactions
  on Pattern Analysis and Machine Intelligence}, no.~6, pp. 1397--1409, 2013.

\bibitem{IntrisicImageUsingOpt2011}
J.~Shen, X.~Yang, Y.~Jia, and X.~Li, ``Intrinsic images using optimization,''
  in \emph{Proceedings of the IEEE Conference on Computer Vision and Pattern
  Recognition (CVPR)}, 2011, pp. 3481--3487.

\bibitem{Fan2018}
Q.~Fan, J.~Yang, G.~Hua, B.~Chen, and D.~Wipf, ``Revisiting deep intrinsic
  image decompositions,'' in \emph{Proceedings of The IEEE Conference on
  Computer Vision and Pattern Recognition (CVPR)}, 2018, pp. 8944--8952.

\bibitem{FaceDecomWithPriors2014}
C.~Li, K.~Zhou, and S.~Lin, ``Intrinsic face image decomposition with human
  face priors,'' in \emph{Proceedings of the European Conference on Computer
  Vision (ECCV)}, 2014, pp. 218--233.

\bibitem{Eigen2014}
D.~Eigen, C.~Puhrsch, and R.~Fergus, ``Depth map prediction from a single image
  using a multi-scale deep network,'' in \emph{Advances in Neural Information
  Processing Systems}, 2014, pp. 2366--2374.

\bibitem{FaceNormals2017}
G.~Trigeorgis, P.~Snape, I.~Kokkinos, and S.~Zafeiriou, ``Face normals
  "in-the-wild" using fully convolutional networks,'' in \emph{Proceedings of
  the IEEE Conference on Computer Vision and Pattern Recognition (CVPR)}, 2017,
  pp. 38--47.

\bibitem{NeuralFace2017}
Z.~Shu, E.~Yumer, S.~Hadap, K.~Sunkavalli, E.~Shechtman, and D.~Samaras,
  ``Neural face editing with intrinsic image disentangling,'' in
  \emph{Proceedings of the IEEE Conference on Computer Vision and Pattern
  Recognition (CVPR)}, 2017, pp. 5444--5453.

\bibitem{Pixel2mesh2018}
N.~Wang, Y.~Zhang, Z.~Li, Y.~Fu, W.~Liu, and Y.-G. Jiang, ``Pixel2mesh:
  Generating 3d mesh models from single rgb images,'' in \emph{Proceedings of
  the European Conference on Computer Vision (ECCV)}, 2018.

\bibitem{Sfsnet18}
S.~Sengupta, A.~Kanazawa, C.~D. Castillo, and D.~W. Jacobs, ``{SfSNet: Learning
  Shape, Refectance and Illuminance of Faces in the Wild},'' in
  \emph{Proceedings of the IEEE Conference on Computer Vision and Pattern
  Recognition (CVPR)}, 2018, pp. 6296--6305.

\bibitem{Jian2017}
J.~Shi, Y.~Dong, H.~Su, and S.~X. Yu, ``Learning non-lambertian object
  intrinsics across shapenet categories,'' in \emph{Proceedings of the IEEE
  Conference on Computer Vision and Pattern Recognition (CVPR)}, 2017, pp.
  5844--5853.

\bibitem{Database3drfe2007}
W.-C. Ma, T.~Hawkins, P.~Peers, C.-F. Chabert, M.~Weiss, and P.~Debevec,
  ``Rapid acquisition of specular and diffuse normal maps from polarized
  spherical gradient illumination,'' in \emph{Proceedings of the 18th
  Eurographics Conference on Rendering Techniques}, 2007, pp. 183--194.

\bibitem{Database3drfe2011}
G.~Stratou, A.~Ghosh, P.~Debevec, and L.~Morency, ``Effect of illumination on
  automatic expression recognition: A novel 3d relightable facial database,''
  in \emph{Face and Gesture}, 2011, pp. 611--618.

\bibitem{unet2015}
O.~Ronneberger, P.~Fischer, and T.~Brox, ``U-net: Convolutional networks for
  biomedical image segmentation,'' in \emph{Medical Image Computing and
  Computer-Assisted Intervention (MICCAI)}, 2015, pp. 234--241.

\bibitem{Werlberger2009}
M.~Werlberger, W.~Trobin, T.~Pock, A.~Wedel, D.~Cremers, and H.~Bischof,
  ``{Anisotropic Huber-L1 Optical Flow},'' in \emph{Proceedings of the British
  Machine Vision Conference}, 2009, pp. 108.1--108.11.

\bibitem{Chen2017}
L.~Chen, Y.~Zheng, B.~Shi, A.~Subpa-Asa, and I.~Sato, ``A microfacet-based
  reflectance model for photometric stereo with highly specular surfaces,'' in
  \emph{Proceedings of the IEEE International Conference on Computer Vision
  (ICCV)}, 2017, pp. 3162--3170.

\bibitem{Gardner2017}
M.-A. Gardner, K.~Sunkavalli, E.~Yumer, X.~Shen, E.~Gambaretto, C.~Gagn{\'e},
  and J.-F. Lalonde, ``Learning to predict indoor illumination from a single
  image,'' \emph{ACM Transactions on Graphics}, vol.~36, no.~6, pp.
  176:1--176:14, 2017.

\bibitem{LEDS}
Y.~Qu\'{e}au, B.~Durix, T.~Wu, D.~Cremers, F.~Lauze, and J.-D. Durou,
  ``{LED-based Photometric Stereo: Modeling, Calibration and Numerical
  Solution},'' \emph{{Journal of Mathematical Imaging and Vision}}, vol.~60,
  no.~3, pp. 313--340, 2018.

\bibitem{Frolova2004}
D.~Frolova, D.~Simakov, and R.~Basri, ``{Accuracy of spherical harmonic
  approximations for images of Lambertian objects under far and near
  lighting},'' in \emph{Proceedings of the European Conference on Computer
  Vision (ECCV)}, 2004, pp. 574--587.

\bibitem{Takuya2015}
M.~M. Takuya~Narihira and S.~X. Yu, ``Direct intrinsics: Learning
  albedo-shading decomposition by convolutional regression,'' in
  \emph{Proceedings of the IEEE International Conference on Computer Vision
  (ICCV)}, 2015.

\bibitem{Sintel2012}
D.~J. Butler, J.~Wulff, G.~B. Stanley, and M.~J. Black, ``A naturalistic open
  source movie for optical flow evaluation,'' in \emph{Proceedings of the
  European Conference on Computer Vision (ECCV)}, 2012, pp. 611--625.

\bibitem{MIT2009}
R.~Grosse, M.~K. Johnson, E.~H. Adelson, and W.~T. Freeman, ``Ground truth
  dataset and baseline evaluations for intrinsic image algorithms,'' in
  \emph{Proceedings of the IEEE International Conference on Computer Vision
  (ICCV)}, 2009, pp. 2335--2342.

\bibitem{Shapenet2015}
A.~X. Chang, T.~Funkhouser, L.~Guibas, P.~Hanrahan, Q.~Huang, Z.~Li,
  S.~Savarese, M.~Savva, S.~Song, H.~Su, J.~Xiao, L.~Yi, and F.~Yu,
  ``{ShapeNet: An Information-Rich 3D Model Repository},'' Stanford University
  --- Princeton University --- Toyota Technological Institute at Chicago, Tech.
  Rep. arXiv:1512.03012 [cs.GR], 2015.

\bibitem{Levoy2005data}
M.~Levoy, J.~Gerth, B.~Curless, and K.~Pull, ``The stanford 3d scanning
  repository,'' 2005, \url{http://www-graphics. stanford. edu/data/3dscanrep}.

\bibitem{Bendansie2015}
``The joyful yell,'' 2015, \url{https://www.thingiverse.com/thing:897412}.

\bibitem{Khoshelham2012}
K.~Khoshelham and S.~O. Elberink, ``{Accuracy and resolution of Kinect depth
  data for indoor mapping applications},'' \emph{Sensors}, vol.~12, no.~2, pp.
  1437--1454, 2012.

\bibitem{Wang2009}
Z.~Wang and A.~C. Bovik, ``Mean squared error: Love it or leave it? a new look
  at signal fidelity measures,'' \emph{IEEE signal processing magazine},
  vol.~26, no.~1, pp. 98--117, 2009.

\bibitem{Maier2017data}
R.~Maier, K.~Kim, D.~Cremers, J.~Kautz, and M.~Nie{\ss}ner, ``{Intrinsic3D
  Dataset},'' 2017, \url{http://vision.in.tum.de/data/datasets/intrinsic3d}.

\bibitem{Queau_variational}
Y.~Qu{\'e}au, J.-D. Durou, and J.-F. Aujol, ``Variational methods for normal
  integration,'' \emph{Journal of Mathematical Imaging and Vision}, vol.~60,
  no.~4, pp. 609--632, 2018.

\end{thebibliography}

\end{document}